\documentclass[12pt,a4paper,onecolumn,twoside,openright,final]{book}
\usepackage[includehead,top=20mm, bottom=20mm, bindingoffset=7mm, marginparwidth=2.3cm, marginparsep=2mm]{geometry}
\usepackage{svg}
\usepackage{times}
\usepackage{epsfig}
\usepackage{graphicx}
\usepackage{amsmath}
\usepackage{amssymb}
\usepackage[ruled,vlined,algo2e]{algorithm2e}
\usepackage{subcaption}
\usepackage{afterpage}

\newcommand{\ov}[1]{\overline{#1}}

\newcommand{\cad}{{CompAD}}

\newcommand{\rev}[1]{\textcolor{black}{#1}}

\newcommand{\imagepath}{../imgs} 
\usepackage{preamble}    % formatting, definitions and settings style file
\usepackage[thesis]{optional}

\begin{document}

    \onehalfspacing

%----FRONTMATTER--------------------------------------------------------------
    \frontmatter

    \pagestyle{plain}
    \begin{titlepage}

	\begin{minipage}{0.9\textwidth}
	%\vspace{2cm}
	\end{minipage}
	
	\begin{center}
		\Large
		\textbf{Spatiotemporal Event Graphs for Dynamic Scene Understanding} 
	\end{center}	
		
	\vspace{0.5cm}

	\begin{center}		
		\Large
		{Salman Khan}
	\end{center}
	
	\vspace{0.05cm}
	
	\begin{center}		
            A Thesis Submitted in Partial Fulfilment\\
            of the Requirements for the Degree of \\
            Doctor of Philosophy    \\
            in \\
	Computer Science and Mathematics
	\end{center}

	\vspace{0.3in}

	\begin{center}
	Visual Artificial Intelligence Laboratory (VAIL)\\
	School of Engineering, Computing and Mathematics\\
	Oxford Brookes University
	\end{center}
	
	\vspace{1.0in}

	\begin{center}
	Supervised by :\\
	Prof. Fabio Cuzzolin and Dr. Tjeerd Olde Scheper \\
	\end{center}

	\vspace{0.1in}
	
	\begin{center}		
		\small
		January 2023
	\end{center}
	
\end{titlepage}

        % --- copyright ---
    \cleardoublepage
    \phantomsection
%     \addcontentsline{toc}{chapter}{}
    \newpage
\begin{minipage}{0.9\textwidth}
\vspace{5cm}

\begin{center}
\emph{The author holds the \textbf{\emph{copyright}} of this thesis. 
Any person(s) intending to use a part or whole of the materials in the thesis in a proposed publication must seek copyright release from the author.}
\end{center}

\end{minipage}

    % --- examiners ---
    \cleardoublepage
    \phantomsection
%     \addcontentsline{toc}{chapter}{}
    \newpage

\begin{minipage}{0.9\textwidth}
\vspace{5cm}
\end{minipage}

\begin{center}
    \large
    \textbf{Thesis/Assessment Committee}
\end{center}

\vspace{0.3in}
\begin{center}
Prof. Dima Damen (External Examiner)\\
Professor of Computer Vision\\
Lead of Machine Learning and Computer Vision Group\\
Department of Computer Science\\
University of Bristol\\
\end{center}

\vspace{0.3in}
\begin{center}
Dr. Matthias Rolf (Internal Examiner)\\
\rev{Reader}\\
School of Engineering, Computing and Mathematics\\
Oxford Brookes University\\
\end{center}

     % --- Preface ---
    \cleardoublepage
    \phantomsection
    \addcontentsline{toc}{chapter}{Preface}
    \newpage
\begin{center}	
{\large
\textbf{Preface}
}
\end{center}
\vspace{1cm}

This dissertation is submitted for the degree of Doctor of Philosophy at  Oxford Brookes University, United Kingdom.
The research presented herein was undertaken under the kind supervision of 
Prof. Fabio Cuzzolin and Dr. Tjeerd Olde Scheper between February 2020 and February 2023.
To the best of my knowledge, this work is original, excluding those previous works for which acknowledgment and reference have been made.
Neither this nor any substantially similar dissertation has been or is being submitted for any other degree, diploma, or qualification at any other university. 
Part of this work has been published as follows:

%FAB: I added the Machine Learning paper with Eleonora and the LNAI proceedings

\begin{itemize}
 \item \textbf{Salman Khan}, Izzeddin Teeti, Andrew Bradley, Mohamed Elhoseiny, and Fabio Cuzzolin, 
 ``A Hybrid Graph Network for Complex Activity Detection in Video'', 
    Pre-print arXiv, (2023).

 \item Gurkirt Singh, ..., \textbf{Salman Khan}, ..., and Fabio Cuzzolin, 
 ``ROAD: The ROad event Awareness Dataset for autonomous Driving'', in IEEE Transactions on Pattern Analysis and Machine Intelligence (IEEE TPAMI) 45, pp. 1036-1054, January 2023.

 \item
 E. Giunchiglia, \textbf{Salman Khan}, ..., F. Cuzzolin and T. Lukasiewicz,
``ROAD-R: the Autonomous Driving Dataset for Learning with Requirements'',
Machine Learning, Springer, September 2022 (accepted with minor revision)

\item
 A. Shahbaz, \textbf{Salman Khan}, ... and F. Cuzzolin,
``International Workshop on Continual Semi-Supervised Learning: Introduction, Benchmarks and Baselines'',
in CSSL 2021, Lecture Notes in Artificial Intelligence Vol. 13418, pages 1-14, Springer, 2022
 
 \item \textbf{Salman Khan} and Fabio Cuzzolin, 
 ``Spatiotemporal Deformable Scene Graphs for Complex Activity Detection'', 
 in Proceedings of The 32nd British Machine Vision Conference (BMVC), 2021.
 
 \item Eleonora Giunchiglia, Mihaela Catalina Stoian, \textbf{Salman Khan}, Fabio Cuzzolin and Thomas Lukasiewicz,
 ``ROAD-R: The Autonomous Driving Dataset with Logical Requirements'', 
 in Proceedings of IJCAI 2022 Workshop on Artificial Intelligence for Autonomous Driving (AI4AD 2022) and IJCLR 2022.

 \item Ajmal Shahbaz, \textbf{Salman Khan}, Mohammad Asiful Hossain, Vincenzo Lomonaco, Kevin Cannons, and Fabio Cuzzolin,
 International Workshop on Continual Semi-Supervised Learning: Introduction, Benchmarks and Baselines,
 in Proceedings of IJCAI Workshop, International Workshop on Continual Semi-Supervised Learning (2021).
 
  \item Fabio Cuzzolin, \textbf{Salman Khan}, Ajmal Shahbaz
 ``Continual Semi-Supervised Learning for Classification via a Time-Varying
Hidden Markov Model'', 
 Pre-print arXiv 2023.

\end{itemize}

    % --- Dedication ---
    \cleardoublepage
    \phantomsection
    \addcontentsline{toc}{chapter}{Dedication}
    \newpage
\begin{minipage}{0.9\textwidth}

% \begin{center}
% \large
% \emph{\textbf{Dedication}}
% \end{center}

\vspace{50mm}

\begin{center}

{\Large \textit{Dedication}}

\vspace{20mm}

    In memory of my father\\
    Late \emph{Mushtaq Ahmad}\\
    With love and eternal appreciation
    
\vspace{10mm}
\&
\vspace{10mm}
\end{center}

\textit{
I would like to dedicate this work to my parents, siblings, and wife. Without whom none of my success would be possible
}

\vspace{50mm}

\emph{“The best things that a man can leave behind are three: A righteous son who will pray for him, ongoing charity whose reward will reach him, and knowledge which is acted upon after his death.”}

\vspace{4mm}
\begin{flushright}
\textnormal{[The Messenger of Allah: Prophet Muhammad SAW \\- Sunan Ibn Majah Vol. 1, Book 1, Hadith 241 ]}
\end{flushright}

\end{minipage}
\vfill

    % --- Acknowledgements ---
    \cleardoublepage    
    \phantomsection
    \addcontentsline{toc}{chapter}{Acknowledgements}
    
\newpage
\heading{Acknowledgements}
\noindent

First and foremost, I would like to praise Allah the Almighty, the Most
Gracious, and the Most Merciful for His blessing given to me during my study
and in completing this thesis. May Allah’s blessing goes to His final Prophet
Muhammad (peace be upon him), his family, and his companions.

%------------------------------------

I would like to express my heartfelt thanks to 
my PhD supervisor Prof. Fabio Cuzzolin, and co-supervisor Dr. Tjeerd Olde Scheper 
for providing such an exceptional opportunity 
for me to pursue a doctorate in such an interesting topic as Computer Vision! 
I also thank my supervisors for their invaluable time, guidance and support.
%------------------------------------
In particular, 
Fabio's strive for excellence, openness, and positive criticism
always inspired me to push myself beyond my limits
and allowed me to gradually improve my scientific, technical and academic skills.
%------------------------------------
Moreover, 
human values like elegance and generosity that I found in Prof. Cuzzolin
have inspired me to work in the Visual Artificial Intelligence Lab (VAIL),
School of Engineering, Computing and Mathematics, Oxford Brookes University.

%------------------------------------

I would also like to thank my co-authors and collaborators Dr. Andrew Bradley, Dr. Gurkirt Singh, Dr. Eleonora Giunchiglia, and Izzeddin Teeti for their time and effort spent during our long technical discussions in our various projects.

%------------------------------------

I want to thank: Tjeerd Olde Scheper and Matthias Rolf for their invaluable guidance and support as research tutors;
Jill Organ, Catherine Joyejob, and Claire Ferone for helping me with all administrative tasks; all the members of VAIL with whom I had a chance to work, especially Inna Skarga-Bandurova, Vivek Singh, Neha Bhargava,  Dinesh Jackson Samuel, Mohamed Mohamed, Bogdan Cirstea, Ajmal Shahbaz, Shireen Kudukkil, Aytac Kanaci, and Amirul Islam.

I am grateful to all external collaborators I worked with: Reza Javanmard, Kevin Cannons, Naeemullah Khan, Mihaela Catalina Stoian, and Valentina Musat.
%------------------------------------

My special thanks go to Prof. Mohamed H. Elhoseiny for offering me an internship at his lab Vision-CAIR, King Abdullah University of Science and Technology (KAUST), Saudi Arabia. I feel fortunate to have had the chance to work in a highly stimulating environment and thankful to the Vision-CAIR members for their invaluable time and guidance during my stay at KAUST.
%------------------------------------

Due acknowledgement goes to my MS supervisors Prof. Sung Wook Baik and Prof. Khan Muhammad
for their time, advice and support during my MS dissertation. I would also like to thank my BS supervisor Prof. Muhammad Sajjad for his guidance and support.
%------------------------------------

I am thankful to my friends, who have greatly supported me for many years, Dr. Amin Ullah, Dr. Ijaz Ul Haq, Dr. Tanveer Husaain, Dr. Fath U Min Ullah, Muhammad Irfan, Imran Ullah, Hayat Ullah, Abid Ali, Hamza Ikram, Yahya Shah, Malik Yasir, and Uzair Shah.

%------------------------------------
Last but not least, I am thankful for my \emph{parents} and my \emph{family} in Pakistan for their endless 
patience, encouragement, and generosity. Especially, my late Father \emph{Mushtaq Ahmad} who did not only raise and mature me but also struggled over the years for my education to make me what I am today, and my elder brother \emph{Dr. Ishtiaq Ahmad} to be my mentor since my childhood. Without their perseverance, I would not be writing this today. My heartfelt thanks go to my wife \emph{Malaika} for her unbounded patience and pushing me to be my best self.

    % --- Abstract ---
    \cleardoublepage
    \phantomsection
    \addcontentsline{toc}{chapter}{Abstract}
    \newpage
\heading{Abstract}

Dynamic scene understanding is the ability of a computer system to interpret and
make sense of the visual information present in a video of a real-world scene. In this thesis, we present a series of frameworks for dynamic scene understanding starting from road event detection from an autonomous driving perspective to complex video activity detection, followed by continual learning approaches for the life-long learning of the models.

Firstly, we introduce the ROad event Awareness Dataset (ROAD) for Autonomous
Driving, to our knowledge the first of its kind. ROAD is designed to test an autonomous vehicle’s ability to detect road events, defined as triplets composed by
an active agent, the action(s) it performs and the corresponding scene locations. Due to the lack of datasets equipped with formally specified logical requirements,
we also introduce the ROad event Awareness Dataset with logical Requirements
(ROAD-R), the first publicly available dataset for autonomous driving with requirements expressed as logical constraints, as a tool for driving neurosymbolic research in the area. 

Next, we extend event detection to holistic scene understanding by proposing
two complex activity detection methods. In the first method, we present a deformable, spatiotemporal scene graph approach, consisting of three main building
blocks: action tube detection, a 3D deformable RoI pooling layer designed for learning the flexible, deformable geometry of the constituent action tubes, and a scene graph constructed by considering all parts as nodes and connecting them based on different semantics. In a second approach evolving from the first, we propose a hybrid graph neural network that combines attention applied to a graph encoding of the local (short-term) dynamic scene with a temporal graph modelling the overall long-duration activity. Our contribution is threefold: i) a feature extraction technique; ii) a method for constructing a local scene graph followed by graph attention, and iii) a graph for temporally connecting all the local dynamic scene graphs.

Finally, the last part of the thesis is about presenting a new continual semi-supervised learning (CSSL) paradigm, proposed to the attention of the machine learning community. We also propose to formulate the continual semi-supervised learning problem as a latent-variable.

    \cleardoublepage
    \phantomsection
    \tableofcontents

%--- List of Figures ---
    \cleardoublepage
    \phantomsection
    \addcontentsline{toc}{chapter}{List of Figures}
    \listoffigures

    %--- List of Tables ---
    \cleardoublepage
    \phantomsection
    \addcontentsline{toc}{chapter}{List of Tables}
    \listoftables

    %--- Nomenclature ---
    \cleardoublepage
    \phantomsection
    \addcontentsline{toc}{chapter}{Nomenclature}    
    \printnomenclature  

%----FRONTMATTER--------------------------------------------------------------

%----MAINMATTER--------------------------------------------------------------
    % Mainmatter
    \mainmatter
    \pagestyle{fancy}

\Nomenclature[A]{ROAD: }{The ROad Awareness Dataset for Autonomous Driving}
\Nomenclature[A]{AV: }{Autonomous Vehicle}
\Nomenclature[A]{HMM: }{Hidden Markov Model}
\Nomenclature[A]{T-HMM: }{Time-varying Hidden Markov Model}
\Nomenclature[A]{CNN: }{Convolutional Neural Network}
\Nomenclature[A]{GCN: }{Graph Convolutional Network}
\Nomenclature[A]{SSL: }{Semi-Supervised Learning}
\Nomenclature[A]{ESAD: }{the Endoscopic Surgeon Action Detection Dataset}
\Nomenclature[A]{RE: }{ROAD Event}
\Nomenclature[A]{BB: }{Bounding Box}
\Nomenclature[A]{RGB: }{ Red Green Blue, 3-Channel Camera output}
\Nomenclature[A]{RNN: }{Recurrent Neural Network}
\Nomenclature[A]{LSTM: }{Long Short-Term Memory recurrent neural network}
\Nomenclature[A]{FPN: }{Feature Pyramid Network}
\Nomenclature[A]{fps: }{frames per second}
\Nomenclature[A]{1D: }{One Dimensional}
\Nomenclature[A]{2D: }{Two Dimensional (spatially for image)}
\Nomenclature[A]{3D: }{Three Dimensional (spatiotemporal for video)}
\Nomenclature[A]{C3D: }{ Convolutional 3D Network}
\Nomenclature[A]{I3D: }{ Inflated Convolutional 3D Network}
\Nomenclature[A]{IoU: }{Intersection over Union}
\Nomenclature[A]{ROAD-R: }{The ROad Awareness Dataset with logical Requirements}
\Nomenclature[A]{HMC: }{Hierarchical Multi-label Classification}
\Nomenclature[A]{MC: }{Multi-label Classification}
\Nomenclature[A]{CL: }{Constrained Loss}
\Nomenclature[A]{CO: }{Constrained Output}
\Nomenclature[A]{AUC: }{Area Under the Curve}
\Nomenclature[A]{mAP: }{ mean Average Precision}
\Nomenclature[A]{TAL: }{ Temporal Action Localisation}
\Nomenclature[A]{CompAD: }{Complex Activity Detection}
\Nomenclature[A]{GAT: }{Graph Attention Network}
\Nomenclature[A]{CSSL: }{Continual Semi-Supervised Leanring}
\Nomenclature[A]{CAR: }{Continual Activity Recognition}
\Nomenclature[A]{CCC: }{Continual Crowd Counting}
\Nomenclature[A]{MAE: }{Mean Absolute Error}
\Nomenclature[A]{EM: }{Expectation Maximisation}

\chapter{Introduction}
\label{chapter:intro}

%FAB: 1.1, 1.5 and 1.6 have relatively few or little references, maybe add a few

Scene understanding from videos has a significant role
in the computer vision community due to its vast real-world applications such as autonomous driving \cite{caesar2020nuscenes,camara2020pedestrian}, surveillance \cite{liang2019peeking}, medical robotics \cite{zia2018surgical} or team sports analysis \cite{hu2020progressive}. The scene understanding problem can be addressed by a variety of techniques starting from simple video action recognition to complex activity detection. In this thesis, we address the problem in a comprehensive fashion, starting from individual event detection to holistic complex scene analysis (complex activity detection).

\section{Video Action Recognition and Detection}

\emph{Video action recognition} is the task of identifying and classifying human actions in video sequences. 
Video \emph{action detection}, on the other hand, is the process of identifying and locating the actions of an individual within a video, in both space and time. In the current literature, a number of convolutional neural networks (CNNs) frameworks are used to learn spatial features from individual video frames, 
which are then fed to
recurrent neural networks (RNNs) to model the temporal dynamics across frames \cite{ullah2017action}. Another approach is to use 3D CNNs, which can learn spatiotemporal features directly from the video data \cite{singh2019recurrent}.

One of the key challenges in video action recognition and detection is the need to deal with the large variability in the appearance and motion of the actions across different videos. To address this, researchers often use techniques such as data augmentation and transfer learning to improve the robustness of the models \cite{yun2020videomix,song2020richly}. Additionally, some researchers have proposed using attention mechanisms to allow the model to focus on the most relevant parts of the video when detecting and recognising actions \cite{meng2019interpretable}. Video action recognition and detection have applications to a wide range of areas such as surveillance, human-computer interaction and sports analytics, to name a few. 

\section{Event Detection}

\emph{Video event detection} refers to the task of identifying and localising specific events or activities in a video. Such activities may be performed by a single agent (object) or by multiple agents at the same time. Examples include activities such as sports events  \cite{yu2020spatiotemporal}, traffic incidents \cite{li2022real}, disaster management \cite{khan2019energy,khan2021deepsmoke} or criminal activities. The most popular deep learning-based method for video event detection is the two-stream CNN, which uses two streams of CNNs, one for processing the RGB frames and the other for processing the optical flow frames \cite{saha2020two}. Another popular method is the 3D CNN, which uses 3D convolutional kernels that can learn spatiotemporal features from the video. These features can then be used to detect events over time, making 3D CNNs useful for tasks such as road events for decision-making in autonomous driving.

\subsection{Road Event Detection}

In this thesis, we define \emph{road events} to be the collection of three labels (agent, action, and location) associated with a single ``tube" (the spatiotemporal representation of a series of bounding boxes). The agent is an active road user (object); the event triple also contains the action(s) the latter performs (possibly more than one at the same time), and the location(s) in which this takes place (which may vary from the start to the end of the event itself), as seen from the point of the view of an autonomous vehicle.

\section{Temporal Action Localisation}

The ability to recognise and localise activities temporally from untrimmed videos is a research problem which has been attracting significant attention, due in part to the rapid growth of online video generation. The current state-of-the-art in action detection is \emph{Temporal Action Localisation} (TAL) - whose approaches not only recognise the action label(s) present in a video, but also identify the start and end time of each action instance occurring.

TAL can be instrumental in generating sports highlights \cite{zeng2021graph}, understanding road scenes in autonomous driving \cite{skhan2021comp}, video summarisation in surveillance videos \cite{xiao2020convolutional} and video captioning \cite{mun2019streamlined}.
A number of TAL methods \cite{lin2018bsn,zeng2019graph,lin2019bmn,long2019gaussian,liu2019multi,xu2020g,xia2022learning,liu2022empirical,hsieh2022contextual,bao2022opental} have recently been proposed. 
%competing to achieve state-of-the-art performance on selected benchmark datasets.
Whereas various new datasets have been recently proposed, %but 
the two most common relevant benchmarks remain ActivityNet 1.3 \cite{caba2015activitynet}, and Thumos-14 \cite{idrees2017thumos}. State-of-the-art performance on Thumos-14 has improved in four years by some 19\% \cite{zhu2022learning}.

Almost all TAL approaches contemplate two major aspects: \emph{features/scene representation} and \emph{temporal localisation}.
In the scene representation stage, snippets (continuous sequences of frames) are processed to understand the local scene in the video.
Methods \cite{liu2022empirical,hsieh2022contextual} typically employ pre-extracted features obtained using a  sequential learning model (e.g., I3D \cite{carreira2017quo}), often pre-trained on the Kinetics \cite{kay2017kinetics} dataset. 
Such features are then processed, e.g., via a temporal or semantic graph neural network, by applying appropriate feature encoding techniques or by generating temporal proposals, in an object detection style \cite{chao2018rethinking}.
In the second step of TAL, actions need to be temporally localised.
Existing methods vary in the way this is done, e.g. through temporal graphs~\cite{xu2020g,zeng2019graph}, boundary regression and proposals generation \cite{lin2018bsn,lin2019bmn,hsieh2022contextual} or encoder-decoder approaches \cite{zhu2022learning}. All such methods, however, are only applicable to short- or mid-duration actions that last for a few seconds ({e.g., a person jumping or pitching a baseball}).

\begin{figure}
    \centering
    \includegraphics[width=0.96\textwidth]{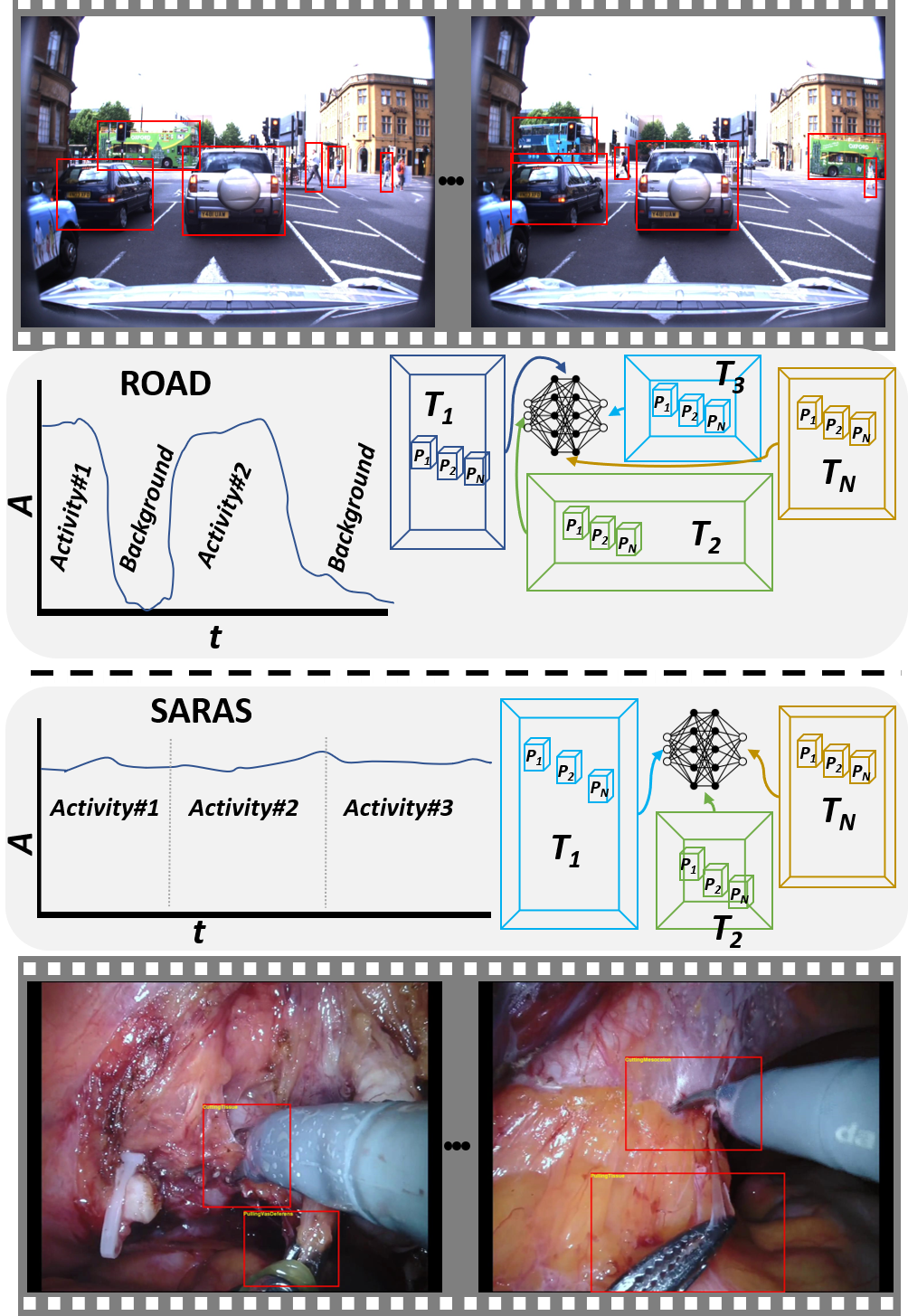}
    \caption{Our complex activity recognition framework workflow for two videos from the ROAD (autonomous driving) and SARAS (surgical robotics) datasets. The overall concept of detecting complex activities via a part-based deformation mechanism is the same in both cases. However, in ROAD background frames exist which do not contain any complex road activity. In contrast, the SARAS complex activities are actually the phases of a surgical procedure, which are contiguous without the need for a background label. 
    }
    \label{fig:workflow} 
\end{figure}

\section{Complex Activity Detection}

{\emph{Complex activities}, in opposition, can be defined as longer-term events, typically comprising a series of shorter actions. For example, an autonomous vehicle (AV) ``negotiating a pedestrian crossing'' presents a complex activity - first the AV drives along the road, then the traffic lights switches to red, the vehicle stops and several pedestrians cross the road. Eventually, the lights change back to green, and the AV drives off again. Whereas a number of studies have attempted to address the localisation of longer-term activities \cite{khan2021spatiotemporal,cheng2022tallformer,shou2016temporal}, 
these approaches mostly focus on activities performed by single actor. To the author's knowledge, \cite{khan2021spatiotemporal} is the only existing study which attempts to model truly complex activities, performed by multiple actors over an extended period of time.
The work, however, relies upon the availability of heavily-annotated datasets which include granular labels for the individual actions which make up a complex activity, and the corresponding frame-level bounding boxes. A general overview of a complex activity framework is visualised in Figure \ref{fig:workflow}. 

\section{Scene Graphs}

A \emph{scene graph} is a graph-based representation of a scene, where the nodes of the graph represent entities in the scene (such as objects, people, and locations) and the edges represent relationships between those entities (such as proximity, or interaction) \cite{gu2019scene}. Scene graphs are used in computer vision \cite{ji2020action} and natural language processing \cite{koner2021graphhopper} to represent the semantic structure of a scene and to facilitate tasks such as object detection, scene understanding, and natural language generation.

Scene graphs are typically constructed by analysing images or videos, using techniques such as object detection, instance segmentation, and scene understanding \cite{fan2022vision}. The nodes of the graph are typically labeled with object categories, and the edges are labeled with the type of relationship between the entities.

Due to the vast importance of scene graphs in scene understanding, in this thesis, we also adopt a scene graph representation for modelling complex activities.

\section{Continual Learning}

\emph{Continual learning} refers to the ability of a machine learning model to learn and adapt to new information over time, without forgetting previously acquired knowledge \cite{lomonaco2021avalanche}. In other words, it allows a model to continuously improve its performance by learning from new data, even after it has been deployed in the real world. This is an important aspect of developing AI systems that can function in dynamic, real-world environments and adapt to changing conditions \cite{lesort2020continual}. Continual learning is an active area of research in machine learning, with the goal of developing models that can learn and adapt throughout their lifetime. A typical example is that of an object detector which needs to be extended to include classes not originally in its list (e.g., ‘donkey’) while retaining its ability to correctly detect, say, a ‘horse’. The hidden assumption there is that we are quite satisfied with the model we have, and we merely wish to extend its capabilities to new settings and classes.

In the last part of the thesis, we also proposed two continual learning paradigms in Chapter \ref{chapter:continual} for updating the machine learning model continually in a semi-supervised manner.

\section{Contributions of the Thesis}

The main contributions of this thesis are the following:

\begin{enumerate}
    \item A ROad event Awareness Dataset for Autonomous Driving (ROAD), the first of its kind, designed to support a paradigm shift toward the detection of semantically meaningful road events, and allow the testing of a range of tasks related to situation awareness for autonomous driving: agent and/or action detection, event detection, ego-action classification.

    \item The ROad event Awareness Dataset with logical Requirements (ROAD-R), the first publicly available dataset whose commonsense requirements are expressed in full propositional logic, equipped with baseline models.

    \item A spatiotemporal scene graph generation and processing mechanism for complex activity detection able to cope with a variable number of parts while learning the overall semantics of an activity class.

    \item An \rev{extension of 2D deformable RoI pooling to 3D} for flexibly pooling features from the various components of the detected tubes, to create an overall representation for activity parts.

    \item A hybrid graph network approach for general complex activity detection, comprising a local scene graph as well as a global temporal graph, capable of localising both long and short-term activities.

    \item A feature extraction strategy for fine-grained temporal activity localisation without the need for additional annotation, consisting in detecting and tracking the relevant scene agents to construct tubes, followed by feature extraction from both the whole scene and each agent tube individually, as opposed to existing methods which only rely on snippet-level features. 

    \item A formal definition of the continual semi-supervised learning (CSSL) problem, and the associated training and testing protocols, with a suitable baseline. The first two benchmark datasets for the validation of semi-supervised continual learning approaches, one for classification (continual activity recognition) and one for regression (continual crowd counting), which we propose as the foundations for the first challenge in this new domain area.

    \item A principled two-stage approach to the CSSL problem, based on a latent model formulation which reduces the task to continual supervised learning.
    
\end{enumerate}

\section{Organisation of the Thesis}

Firstly, we review all the relevant related work, including the benchmark datasets and evaluation metrics, in Chapter \ref{chapter:related_work}. Next, the main body of the thesis (Chapters 3 to 7) is divided into three parts; Part I, Part II, and Part III. Finally, the thesis is concluded in Chapter \ref{chapter:conclusion} by a summary and a discussion of possible future extensions.

The theme of Part I \ref{part:part1} is ROAD (The ROad event Awareness Dataset for Autonomous Driving) which includes two chapters \ref{chapter:road}. In Chapter \ref{chapter:road}, we present the baseline ROAD dataset, which is designed to test an autonomous vehicle’s ability to detect road events, defined as triplets composed by an active agent, the action(s) it performs, and the corresponding scene locations. We also introduce the ROad event Awareness Dataset with logical Requirements (ROAD-R) in Part I, the first publicly available dataset for autonomous driving with requirements expressed as logical constraints, as a neurosymbolic extension to ROAD.

Next, in Part II \ref{part:part2} we present the concept of Complex Activity Detection (CompAD), over two Chapters (Chapter \ref{chapter:deformsgraph} and Chapter \ref{chapter:hgraph}). Chapter \ref{chapter:deformsgraph} presents a novel deformable, spatiotemporal scene graph approach for CompAD, consisting of three main building blocks: (i) action tube detection, (ii) the modelling of the deformable geometry of parts, and (iii) a graph convolutional network. The latter is further improved in Chapter \ref{chapter:hgraph} by addressing the CompAD problem using a hybrid graph neural network that combines attention applied to a graph encoding the local (short-term) dynamic scene with a temporal graph modelling the overall long-duration activity.

Finally, Part III \ref{part:part3} is about continual learning and is composed of a single Chapter\ref{chapter:continual}. 
%In Chapter \ref{chapter:continual}, 
There, we present a new continual semi-supervised learning (CSSL) paradigm, which we proposed to the attention of the machine learning community via the IJCAI 2021 International Workshop on Continual Semi-Supervised Learning. In addition to the baseline made available for the challenge, we also propose to formulate the continual semi-supervised learning problem as a latent-variable one, in which the labels of a series of unlabelled, streaming data points play the role of the hidden states, while the
instances play that of the observations. 

 \section{Publications}

\paragraph{Part of thesis:} the following is the list of publications that form the core of this PhD thesis.

%FAB: please update this list as in the preface above

\begin{enumerate}
 \item \textbf{Salman Khan}, Izzeddin Teeti, Andrew Bradley, Mohamed Elhoseiny, and Fabio Cuzzolin, 
 ``A Hybrid Graph Network for Complex Activity Detection in Video'', 
    Pre-print arXiv, (2023).

 \item Gurkirt Singh, ..., \textbf{Salman Khan}, ..., and Fabio Cuzzolin, 
 ``ROAD: The ROad event Awareness Dataset for autonomous Driving'', in IEEE Transactions on Pattern Analysis and Machine Intelligence (IEEE TPAMI) 45, pp. 1036-1054, January 2023.

 \item
 E. Giunchiglia, \textbf{Salman Khan}, ..., F. Cuzzolin and T. Lukasiewicz,
``ROAD-R: the Autonomous Driving Dataset for Learning with Requirements'',
Machine Learning, Springer, September 2022 (accepted with minor revision)

\item
 A. Shahbaz, \textbf{Salman Khan}, ... and F. Cuzzolin,
``International Workshop on Continual Semi-Supervised Learning: Introduction, Benchmarks and Baselines'',
in CSSL 2021, Lecture Notes in Artificial Intelligence Vol. 13418, pages 1-14, Springer, 2022
 
 \item \textbf{Salman Khan} and Fabio Cuzzolin, 
 ``Spatiotemporal Deformable Scene Graphs for Complex Activity Detection'', 
 in Proceedings of The 32nd British Machine Vision Conference (BMVC), 2021.
 
 \item Eleonora Giunchiglia, Mihaela Catalina Stoian, \textbf{Salman Khan}, Fabio Cuzzolin and Thomas Lukasiewicz,
 ``ROAD-R: The Autonomous Driving Dataset with Logical Requirements'', 
 in Proceedings of IJCAI 2022 Workshop on Artificial Intelligence for Autonomous Driving (AI4AD 2022) and IJCLR 2022.

 \item Ajmal Shahbaz, \textbf{Salman Khan}, Mohammad Asiful Hossain, Vincenzo Lomonaco, Kevin Cannons, and Fabio Cuzzolin,
 International Workshop on Continual Semi-Supervised Learning: Introduction, Benchmarks and Baselines,
 in Proceedings of IJCAI Workshop, International Workshop on Continual Semi-Supervised Learning (2021).
 
  \item Fabio Cuzzolin, \textbf{Salman Khan}, Ajmal Shahbaz
 ``Continual Semi-Supervised Learning for Classification via a Time-Varying
Hidden Markov Model'', 
 Pre-print arXiv 2023.

\end{enumerate}

\paragraph{Other works:} Below is the list of other works which I published in the course of my PhD term, but are not directly connected to the topic of this thesis.

\begin{enumerate}

\item Muhammad, Khan, Hayat Ullah, \textbf{Salman Khan}, Mohammad Hijji, and Jaime Lloret. "Efficient Fire Segmentation for Internet-of-Things-Assisted Intelligent Transportation Systems." IEEE Transactions on Intelligent Transportation Systems, September 2022.

\item \textbf{Salman Khan}, Khan Muhammad, Tanveer Hussain, Javier Del Ser, Fabio Cuzzolin, Siddhartha Bhattacharyya, Zahid Akhtar, and Victor Hugo C. de Albuquerque. "Deepsmoke: Deep learning model for smoke detection and segmentation in outdoor environments." Expert Systems with Applications 182 (2021): 115125.

\item Teeti, Izzeddin, \textbf{Salman Khan}, Ajmal Shahbaz, Andrew Bradley, and Fabio Cuzzolin. "Vision-based Intention and Trajectory Prediction in Autonomous Vehicles: A Survey." Proceedings of IJCAI 2022 (Survey track), pages 5630-5637.

\item Teeti, Izzeddin, Valentina Musat, \textbf{Salman Khan}, Alexander Rast, Fabio Cuzzolin, and Andrew Bradley. "Vision in adverse weather: Augmentation using CycleGANs with various object detectors for robust perception in autonomous racing." ICML 2022 - Workshop on Safe Learning for Autonomous Driving (SL4AD 2022).

\item Onyeogulu, Tochukwu, Amirul Islam, \textbf{Salman Khan}, Izzeddin Teeti, and Fabio Cuzzolin. "Situation Awareness for Automated Surgical Check-listing in AI-Assisted Operating Room." arXiv preprint arXiv:2209.05056 (2022).

\item Jin, Kaizhe, Adrian Rubio-Solis, Ravi Naik, Tochukwu Onyeogulu, Amirul Islam, \textbf{Salman Khan}, Izzeddin Teeti et al. "Identification of Cognitive Workload during Surgical Tasks with Multimodal Deep Learning." arXiv preprint arXiv:2209.06208 (2022).

\end{enumerate}

\section{List of Packages and Demos}

\begin{itemize}

\item All resources of our ROAD project (IEEE TPAMI  2022 \cite{singh2022road} and The ROAD Workshop and Challenge: Event Detection for Situation Awareness in Autonomous Driving, ICCV 2021) are publicly available at:

\subitem \emph{Workshop and Challenge link}: \url{https://sites.google.com/view/roadchallangeiccv2021/}
\subitem \emph{ROAD dataset}: \url{https://github.com/gurkirt/road-dataset}
\subitem \emph{3D RetinaNet baseline}: \url{https://github.com/gurkirt/3D-RetinaNet}
\subitem \emph{Inference demo}: \url{https://www.youtube.com/watch?v=CmxPjHhiarA}
\subitem \emph{Challenge evalAI} \url{https://eval.ai/web/challenges/challenge-page/1059/overview.}

\item The resources of our ROAD-R project \cite{giunchiglia2022road} are publicly available at:
\subitem \emph{code}: \url{https://github.com/EGiunchiglia/ROAD-R}
\subitem \emph{video}: \url{https://www.youtube.com/watch?v=_-Ll5d1VQXY}.

\item The source code of our BMVC 2021 ~\cite{skhan2021comp} is publicly available at: \url{https://github.com/salmank255/SDSG_Complex_Activity}.

\item The source code of our CompAD paper Pre-print is publicly available at: \url{https://anonymous.4open.science/r/HG_CompAD-7697/README.md}.

\item All resources of the continual learning project (Continual Semi-Supervised Learning Workshop and Challenge at IJCAI 2021 \cite{shahbaz2022international}) are publicly available at:

\subitem \emph{Workshop link}: \url{https://sites.google.com/view/sscl-workshop-ijcai-2021/}
\subitem \emph{Code}: \url{https://github.com/salmank255/cssl}
\subitem \emph{CAR challenge baseline}: \url{https://github.com/salmank255/IJCAI-2021-Continual-Activity-Recognition-Challenge}
\subitem \emph{CAR evalAI}: \url{https://eval.ai/web/challenges/challenge-page/984/overview}
\subitem \emph{CCC challenge baseline}: \url{https://github.com/Ajmal70/IJCAI_2021_Continual_Crowd_Counting_Challenge}
\subitem \emph{CCC evalAI}: \url{https://eval.ai/web/challenges/challenge-page/986/overview}.

\end{itemize}

    \chapter{Related Work}
\label{chapter:related_work}

This chapter presents a literature review of the most relevant methods in the areas (discussed in Chapter \ref{chapter:intro}) which most contribute to scene understanding. 
%In the following sections, we started
We start
from a comparison of the existing major autonomous driving datasets (Section \ref{subsec:av-datasets}) with our ROAD dataset, which is presented in Chapter \ref{chapter:road}. Next, we provide a literature review of temporal action localisation, complex activity detection, graph convolutional networks, and continual learning in Sections \ref{rel:sec:tal}, \ref{rel:sec:cad}, \ref{rel:sec:gnn}, and \ref{rel:sec:cl}, respectively. We also discuss the benchmark datasets used for the evaluation of our proposed methods in Section \ref{related_work:sec:datasets}. Lastly, we introduce all the evaluation metrics (Section \ref{related_work:sec:evalmat}) used for the evaluation of the different tasks targeted in this thesis.

\section{Autonomous Driving Datasets} \label{subsec:av-datasets}

In recent years a multitude of AV datasets have been released, mostly focusing on object detection, scene segmentation, trajectory prediction, and action detection. We can categorise them into two main bins: (1) RGB without range data (single modality) which is relatively less time-consuming and expensive than building multimodal datasets including range data from LiDAR or radar and (2) RGB with range data (multimodal) that contain multiple types of data, such as lidar, camera, and radar data. 

To compare ROAD (proposed in Chapter \ref{chapter:road}) with autonomous driving perception datasets, we provide a comparative analysis of the recent state-of-the-art datasets in Table \ref{table:ROADtubecomparison}. As it can be noted in the table, the unique feature of ROAD is its diversity in terms of the types of actions and events portrayed, by all types of road agents in the scene. With 12 agent classes, 30 action classes and 15 location classes ROAD provides (through a combination of these three elements) a much more refined description of road scenes.

\begin{table}[t]
	\centering
	\caption{ \rev{Comparison of ROAD with similar datasets for perception in autonomous driving in terms of diversity of labels. The comparison is based on the number of classes portrayed and the availability of action annotations and tube tracks for both pedestrians and vehicles, as well as location information. Most competitor datasets do not provide action annotation for either pedestrians or vehicles.}}
	\label{tab:datasets}
	\setlength{\tabcolsep}{4pt}
	    %{\footnotesize
		\scalebox{0.72}{
\rev{\begin{tabular}{ccccccccccccc}
	\toprule
	\multicolumn{1}{c}{\multirow{2}{*}{\textbf{Dataset}}} & \multicolumn{1}{c}{\multirow{2}{*}{\textbf{Class Num.}}} &\multicolumn{1}{c}{\multirow{2}{*}{\textbf{Ann. frames}}}&\multicolumn{1}{c}{\multirow{2}{*}{\textbf{Countries}}} & \multicolumn{1}{c}{\multirow{2}{*}{\textbf{Loc. lab.}}} & \multicolumn{1}{c}{\multirow{2}{*}{\textbf{Avg. obj./f}}}& \multicolumn{2}{c}{\textbf{Action Ann}}            & \multicolumn{2}{c}{\textbf{Tube Ann}}              \\ \cline{7-10} 
	\multicolumn{1}{c}{} & \multicolumn{1}{c}{} & \multicolumn{1}{c}{}& \multicolumn{1}{c}{}& \multicolumn{1}{c}{} & \multicolumn{1}{c}{}& \multicolumn{1}{c}{Ped.} & \multicolumn{1}{c}{Veh.} & \multicolumn{1}{c}{Ped.} & \multicolumn{1}{c}{Veh.}  \\ \hline
	\multicolumn{1}{c}{SYNTHIA\cite{ros2016synthia}} & \multicolumn{1}{c}{13}& \multicolumn{1}{c}{13400}& \multicolumn{1}{c}{1}& \multicolumn{1}{c}{pixelwise ann.} & \multicolumn{1}{c}{-}& \multicolumn{1}{c}{-}& \multicolumn{1}{c}{-}& \multicolumn{1}{c}{-}& \multicolumn{1}{c}{-} \\ %\hline
	\multicolumn{1}{c}{Cityscapes \cite{cordts2016cityscapes}} & \multicolumn{1}{c}{30}& \multicolumn{1}{c}{5000}& \multicolumn{1}{c}{2}& \multicolumn{1}{c}{pixel level sem.} & \multicolumn{1}{c}{-}& \multicolumn{1}{c}{-}& \multicolumn{1}{c}{-}& \multicolumn{1}{c}{-}& \multicolumn{1}{c}{-}\\ %\hline	
	\multicolumn{1}{c}{ Waymo \cite{sun2019scalability}} & \multicolumn{1}{c}{4}& \multicolumn{1}{c}{1,200,000}& \multicolumn{1}{c}{1}& \multicolumn{1}{c}{-}& \multicolumn{1}{c}{20} & \multicolumn{1}{c}{-}& \multicolumn{1}{c}{-}& \multicolumn{1}{c}{\checkmark}& \multicolumn{1}{c}{\checkmark}\\ %\hline	
	\multicolumn{1}{c}{ Apolloscape \cite{wang2019apolloscape}} & \multicolumn{1}{c}{25}& \multicolumn{1}{c}{140,000}& \multicolumn{1}{c}{1}& \multicolumn{1}{c}{pixel level sem.} & \multicolumn{1}{c}{-}& \multicolumn{1}{c}{-}& \multicolumn{1}{c}{-}& \multicolumn{1}{c}{\checkmark}& \multicolumn{1}{c}{\checkmark}\\ %\hline
	\multicolumn{1}{c}{ PIE \cite{rasouli2019pie}} & \multicolumn{1}{c}{6}& \multicolumn{1}{c}{293,437}& \multicolumn{1}{c}{1}& \multicolumn{1}{c}{-}& \multicolumn{1}{c}{2.5} & \multicolumn{1}{c}{\checkmark}& \multicolumn{1}{c}{-}& \multicolumn{1}{c}{\checkmark}& \multicolumn{1}{c}{-}\\ %\hline		
	\multicolumn{1}{c}{ TITAN \cite{malla2020titan}} & \multicolumn{1}{c}{50}& \multicolumn{1}{c}{75,262}& \multicolumn{1}{c}{1}& \multicolumn{1}{c}{-}& \multicolumn{1}{c}{5.2} & \multicolumn{1}{c}{\checkmark}& \multicolumn{1}{c}{\checkmark}& \multicolumn{1}{c}{\checkmark}& \multicolumn{1}{c}{\checkmark}\\ %\hline	
	\multicolumn{1}{c}{ KITTI360 \cite{Liao2021ARXIV}} & \multicolumn{1}{c}{19}& \multicolumn{1}{c}{83,000}& \multicolumn{1}{c}{1}& \multicolumn{1}{c}{sem. ann.} & \multicolumn{1}{c}{-}& \multicolumn{1}{c}{-}& \multicolumn{1}{c}{-}& \multicolumn{1}{c}{-}& \multicolumn{1}{c}{-}\\ 
	\multicolumn{1}{c}{  A*3D \cite{pham20193d}} & \multicolumn{1}{c}{7}& \multicolumn{1}{c}{39,000}& \multicolumn{1}{c}{1}& \multicolumn{1}{c}{-}& \multicolumn{1}{c}{-} & \multicolumn{1}{c}{4.8}& \multicolumn{1}{c}{-}& \multicolumn{1}{c}{-}& \multicolumn{1}{c}{-}\\ %\hline	
	\multicolumn{1}{c}{  H3D \cite{patil2019h3d}} & \multicolumn{1}{c}{8}& \multicolumn{1}{c}{27,721}& \multicolumn{1}{c}{1}& \multicolumn{1}{c}{-}& \multicolumn{1}{c}{16} & \multicolumn{1}{c}{-}& \multicolumn{1}{c}{-}& \multicolumn{1}{c}{\checkmark}& \multicolumn{1}{c}{\checkmark}\\ %\hline
	\multicolumn{1}{c}{  Argoverse \cite{chang2019argoverse}} & \multicolumn{1}{c}{15}& \multicolumn{1}{c}{70,557}& \multicolumn{1}{c}{1}& \multicolumn{1}{c}{-} & \multicolumn{1}{c}{18}& \multicolumn{1}{c}{-}& \multicolumn{1}{c}{-}& \multicolumn{1}{c}{\checkmark}& \multicolumn{1}{c}{\checkmark}\\ %\hline	
	\multicolumn{1}{c}{ NuScense \cite{caesar2020nuscenes}} & \multicolumn{1}{c}{23}& \multicolumn{1}{c}{40,000}& \multicolumn{1}{c}{2}& \multicolumn{1}{c}{3D sem. seg.} & \multicolumn{1}{c}{20}& \multicolumn{1}{c}{-}& \multicolumn{1}{c}{-}& \multicolumn{1}{c}{\checkmark}& \multicolumn{1}{c}{\checkmark}\\ %\hline
	\multicolumn{1}{c}{ DriveSeg \cite{mmke-dv03-20}} & \multicolumn{1}{c}{12}& \multicolumn{1}{c}{5,000}& \multicolumn{1}{c}{1}& \multicolumn{1}{c}{sem. ann.} & \multicolumn{1}{c}{-}& \multicolumn{1}{c}{-}& \multicolumn{1}{c}{-}& \multicolumn{1}{c}{-}& \multicolumn{1}{c}{-}\\ %\hline
	\midrule
	\multicolumn{5}{c}{Spatiotemporal action detection datasets} & \\
	\midrule
	\multicolumn{1}{c}{ UCF24 \cite{jiang2014thumos}} & \multicolumn{1}{c}{24}& \multicolumn{1}{c}{80175}& \multicolumn{1}{c}{-}& \multicolumn{1}{c}{-}& \multicolumn{1}{c}{1.5} & \multicolumn{1}{c}{\checkmark}& \multicolumn{1}{c}{-}& \multicolumn{1}{c}{\checkmark}& \multicolumn{1}{c}{-}\\ 
	\multicolumn{1}{c}{ AVA \cite{ava2017gu}} & \multicolumn{1}{c}{80}& \multicolumn{1}{c}{1,600,000}& \multicolumn{1}{c}{-}& \multicolumn{1}{c}{-}& \multicolumn{1}{c}{3} & \multicolumn{1}{c}{\checkmark}& \multicolumn{1}{c}{-}& \multicolumn{1}{c}{\checkmark}& \multicolumn{1}{c}{-}\\ 
	\multicolumn{1}{c}{ Multisports \cite{li2021multisports}} & \multicolumn{1}{c}{66}& \multicolumn{1}{c}{37701}& \multicolumn{1}{c}{-} & \multicolumn{1}{c}{-} & \multicolumn{1}{c}{2.5}& \multicolumn{1}{c}{\checkmark}& \multicolumn{1}{c}{-}& \multicolumn{1}{c}{\checkmark}& \multicolumn{1}{c}{-}\\ 
	\midrule
	\multicolumn{1}{c}{ \textbf{ROAD \cite{singh2022road}} } & \multicolumn{1}{c}{$43$}& \multicolumn{1}{c}{122,000}& \multicolumn{1}{c}{1}& \multicolumn{1}{c}{\checkmark} & \multicolumn{1}{c}{5}& \multicolumn{1}{c}{\checkmark}& \multicolumn{1}{c}{\checkmark}& \multicolumn{1}{c}{\checkmark}& \multicolumn{1}{c}{\checkmark}\\ 
	\bottomrule 
	\multicolumn{7}{l}{Ped. Pedestrian, Veh. Vehicle, ann. annotation, sem. seg. semantic segmentation} \\
\end{tabular}
}
}
%}
\label{table:ROADtubecomparison}
\end{table}

\section{Temporal Action Localisation} \label{rel:sec:tal}

Significant progress has been made in recent years toward temporal action localisation (TAL) in untrimmed videos. The main goal of TAL is to remove the irrelevant background from the input videos and identify the start and end instants of the action of interest. Temporal localisation methods can be divided into supervised \cite{bai2020boundary,huang2020improving} versus weakly/unsupervised ones \cite{shou2018autoloc,paul2018w,zhang2020metal}. While the former requires temporal annotations to learn a model able to identify the start and end of an action, in weakly-supervised settings the model is learned from partial annotation. 

In Part II of the thesis, our goal is the same as temporal activity localisation, with the difference that we wish to target \emph{complex, long duration events articulated into a possibly large number of parts} (complex activity detection), whose internal structure cannot be leveraged by naive temporal segmentation methods.

\section{Complex Activity Detection} \label{rel:sec:cad}

 Most recent work on complex activity recognition concerns scalar sensors \cite{bharti2018human,thakur2019improved,zhou2020deep} or a combination of both scalar and vision sensors \cite{arzani2020switching,kwon2020imutube}. Recently, though, several vision-based complex activity recognition methods have been proposed \cite{huang2020improving,wu2019long,zhao2019hacs,sudhakaran2019lsta,luo2019grouped,ji2020action} with the goal of understanding an overall scene by recognising and segmenting atomic actions. 
These methods can be further divided into (i) sliding windows approaches \cite{wang2014action,shou2016temporal}, in which an activity classifier is applied to each snippet, and (ii) boundaries analyses \cite{xu2017r,gao2017turn}, in which a model is trained to identify the start and end time of each action.  
Overall, current activity recognition methods are geared to recognise short-term activities via a combination of small atomic actions.

Unlike existing approaches, the objective of our complex activity model (proposed in Chapter \ref{chapter:deformsgraph}) is to understand \emph{long-term} activities in dynamic scenes, such as the \emph{phases} a surgical procedure is broken into, whose detection is crucial to inform the decision making of automated robotic assistants.

\subsection{One-Stage vs Two-Stage Approaches}

Complex activity methods can been broadly divided into \emph{one-stage} and \emph{two-stage} approaches. 
The former
\cite{buch2019end,huang2019decoupling,long2019gaussian,lin2021learning} detect actions/activities in a single shot,
and can be easily trained in an end-to-end manner. For example, Wang et al. \cite{wang2021oadtr} detect actions using transformers which, unlike RNNs, do not suffer from nonparallelism and vanishing gradients. 
Most one-stage methods, however, only perform action classification, rather than spatiotemporal localisation. In contrast, Lin et al. \cite{lin2021learning} have recently proposed an anchor-free, one-stage light model which generates proposals, locates actions within them, and classifies them end-to-end.

The latter group of methods
\cite{lin2018bsn,lin2019bmn,bai2020boundary, Bao_2022_CVPR,Zhu_Wang_Tang_Liu_Zheng_Hua_2022}, instead, \rev{consists} of two stages, similarly to region-proposal object detectors. The first stage generates suitable proposals for predicting the start and end time of an activity,
while the second stage extracts features and processes the proposals before passing them to both a classification head and a regression head (for temporal localisation). Some work, including \cite{lin2019bmn}, focus on the first stage to improve the quality of the proposals, while others focus on the processing or refining of the proposals. \cite{Zhu_Wang_Tang_Liu_Zheng_Hua_2022}, for instance, uses an off-the-shelf method for proposal generation. The second stage consists of two networks, a `disentanglement' network to separate the classification and regression representations, and a `context aggregation' network to add them together. Such methods are not trainable end-to-end and limited to short-or mid-duration actions.
In contrast, we propose a \emph{complex activity detection} method in Chapter \ref{chapter:hgraph},
which exhibits the advantages of both classes of methods, thanks to our hybrid graph approach capable of
recognising and localising both short- and long-duration activities.

\section{Graph Networks} \label{rel:sec:gnn}

Recently, Graph Convolutional Networks (GCNs) have been widely used for action and activity detection and recognition, building on their success in different areas of computer vision such as point cloud segmentation \cite{wang2019dynamic,xie2020point} and 3D object detection \cite{gkioxari2019mesh}. Relevant GCN approaches have been broadly focussing on either action recognition \cite{wang2018videos,liu2019learning,chen2019graph} or TAL \cite{zeng2019graph,xu2020g}. 
In the former, videos are represented in different spatiotemporal formats such as 3D point clouds and time-space region graphs, and methods focus on recognising atomic actions only.
In contrast, Zeng et al. \cite{zeng2019graph} use GCN for temporal activity localisation by considering action proposals as nodes and a relation between two proposals as an edge. In opposition, in our model nodes are action tubes and their connections are based on an array of semantics.
In another recent study, Xu et al. \cite{xu2020g} generate graphs by considering temporal snippets as nodes and drawing connections between them based on temporal appearance and semantic similarity. 

Most graph-based activity detection methods \cite{xu2020g,li2020graph,zeng2019graph} construct a graph for a whole video by taking snippets as nodes and their temporal linkage as edges, not paying much attention to the constituent atomic actions within each snippet, and are typically limited to shorter videos or memory dependent \rev{in case of larger videos due to adjacency matrices. For a large number of snippets (nodes) in the video required memory can be substantial. The adjacency matrix of the graph with n nodes (snippets) requires $O(n^2)$ memory, which can be impractical for large videos.}

In contrast, our proposed framework presented in Chapter \ref{chapter:deformsgraph} is designed to construct a \rev{scene} graph for each snippet \rev{rather than the whole video} which reflects the structure of a dynamic scene in terms of atomic action tubes (nodes) and the different types of relationships between them.

\subsection{GCN-Based Temporal Activity Localisation}
GCNs have been extensively investigated for TAL \cite{zeng2019graph,bai2020boundary,xu2020g,nawhal2021activity,yang2022acgnet}. 
GCN-based TAL methods can also be further divided into two-stage and one-stage methods. 
The former, once again, perform localisation after generating suitable proposals.
For instance, in %Zeng et al. 
\cite{zeng2019graph} %proposed
two different types of boundary proposals are generated and then individually passed to the same graph,  resulting in both an action label and a temporal boundary. 

One-stage GCN-based TAL methods, instead, solve the detection problem without proposals in one go by learning 
spatiotemporal features in an end-to-end manner. For examples, in \cite{xu2020g} a graph is first generated by connecting the snippets both temporally and by virtue of meaningful semantics.
This graph is then divided into sub-graphs (anchors), where each anchor represents the activity in an untrimmed video. In contrast, Khan et al. \cite{skhan2021comp} proposed a spatiotemporal scene graph-based long-term TAL method where each of the snippets is considered as a separate graph, which is heavily dependent on the particular actions present in the scene, and is only applicable to datasets providing (label and bounding box) annotations for each individual action.

In Chapter \ref{chapter:hgraph}, we propose a new framework leveraging GCNs, by incorporating them in an overall hybrid graph capable of modelling both the local scenes, via a Graph Attention Network (GAT) \cite{velivckovic2017graph}, and the overall global activity via a temporal graph. GATs build on the transformer concept by applying attention to graphs, and were originally proposed in \cite{velivckovic2017graph} for node classification. The idea is to update the representation of the current node with respect to its neighbours by applying attention to learn the importance of the various connections.

In Chapter \ref{chapter:hgraph}, GAT is used at the local scene level to learn the features of each node (active agent tube), to generate a more robust local scene representation.

\section{Continual Learning} \label{rel:sec:cl}

Current continual learning methods can be categorised into three major families based on how the information of previous data is stored and used. \emph{Prior-focused} methods \cite{Farquhar2018TowardsRE} use a penalty term to regularise the parameters rather than a hard constraint. \emph{Parameter isolation} methods \cite{Mallya2018PackNetAM} dedicate different parameters for different tasks to prevent interference. Finally, \emph{replay-based} approaches store the examples’ information in a ‘replay buffer’ or a generative model, which is used for rehearsal/retraining or to provide constraints on the current learning step \cite{Aljundi2019online}.

\subsection{Continual Unsupervised Learning} 
Continual unsupervised learning is a relatively unexplored territory due to the heavy dependency of existing works on the availability of ground truth information. Another challenge lies in the clear distinction of class boundaries for controlling model updates \cite{maha}. Clustering approaches that provide structure to the unlabelled stream are a popular theme of existing continual unsupervised works \cite{maha,he,zhan}. A self-adaptive clustering approach \cite{maha} is proposed which does not rely on labelled data for model updates. The labelled data is initially leveraged to learn the cluster-to-class associations. Similarly, in \cite{he} the pseudo labels for new data are obtained by a K-means global clustering approach, whereas the updated model without fully connected layers is employed as a feature extractor.

\subsection{Time-Varying Hidden Markov Models} 
A number of competing formulations exist for HMMs whose parameters depend on time. They have been employed recently for useful life prediction in engineering \cite{Li2019time-varying}, analysing time-varying networks in EEG data \cite{Williams2018markov}, and incorporating periodic variability in animal movement \cite{li2017incorporating}. 
Hidden \emph{semi-Markov} models \cite{marhasev2006hsmm}, in particular, have been proposed, which augment HMMs with time-varying parameters, so that the transition probability to a new state varies in accordance with the time spent in the current state. In opposition, the TVHMM model formulated in \cite{wang2009event} encodes non-stationarity into a finite sequence of time varying transition densities and can simulate state duration of infinite length. A common approach \cite{chung2007latent,li2017incorporating} is to model the transition probabilities as functions of a covariate process $\{z_t\}$, which is observed in parallel to the main observation process \cite{diebold1993regime}. 
As the covariates $z_t$ are simultaneously correlated with the hidden variables $x_t$, it may be preferable to use lagged informative variables or artificial variables. 
In fact, in \cite{Otranto2008thmm} a time-varying hidden Markov model with latent information was proposed in which an additional hidden variable (`signal') with AR(1) dynamics drives the transition probabilities in the main hidden variable, $x_t$.

In \cite{chib2004non-markovian}, the authors observed that the states respond more to
their own past values than  any other variable. They thus proposed a model in which the state depends on the sign of an autoregressive latent variable, which induces time-varying probabilities in a non-Markovian process, as the probability of a certain state at time $t$ depends on the lagged value of the latent variable and not the state at the previous time.

\section{Benchmark Datasets}\label{related_work:sec:datasets}

In the literature, several datasets have been proposed to evaluate the performance of scene understanding from simple action detection to event detection and complex activity recognition. Here, in this chapter, we describe all the datasets used in this thesis.

\subsection{ROAD}

ROAD (The ROad event Awareness Dataset for Autonomous Driving) \cite{singh2022road} is a multi-labeled dataset proposed for road agent, action, and location detection. The combination of these three labels is referred to in \cite{singh2022road} as a `road event'. The ROAD dataset consists of a total of 22 videos with an average duration of 8 minutes, captured by the Oxford RobotCar \cite{RobotCarDatasetIJRR} under diverse lighting and weather conditions. The dataset was further extended as a testbed for complex activity detection in \cite{skhan2021comp}. Complex activities in the ROAD dataset belong to six different classes, including: negotiating an intersection, negotiating a pedestrian crossing, waiting in a queue, merging into the (ego) vehicle lane, sudden appearance (of other vehicles/agents), and (people) walking in the middle of the road. Activities can span up to two minutes and involve a large number of road agents. We introduce ROAD in Chapter \ref{chapter:road} and evaluate our 3D-RetinaNet baseline in the same chapter. We also use ROAD extension for complex activity detection in Chapter \ref{chapter:deformsgraph} and Chapter \ref{chapter:hgraph}

% \subsection{UCF-101}
% UCF-101 \cite{soomro2012ucf101} is a dataset of human action videos collected from YouTube, which is often used for evaluating action recognition algorithms. UCF-101 contains 101 different action classes, such as running, jumping, playing basketball, and so on.

% It contains 13,320 video clips, each of which is around 6 seconds long. The videos were collected from YouTube, and the action categories were chosen to be as diverse and challenging as possible. Each video is labeled with one of the 101 action classes, making it a useful resource for training and evaluating action recognition algorithms.

% For action detection, 24 categories are annotated spatiotemporally as a subset of UCF-101 known as UCF101-24 released for the THUMOS-2013 challenge\footnote{\url{https://www.crcv.ucf.edu/ICCV13-Action-Workshop/download.html}}. UCF101-24 has been used in Chapter \ref{chapter:road} for the evaluation 3D-RetinaNet baseline.

\subsection{SARAS-ESAD}

ESAD (the Endoscopic Surgeon Action Detection Dataset) \cite{bawa2021saras} is a benchmark devised for surgeon action detection in real-world endoscopic surgery videos. The goal of ESAD is contributing to increasing the potential and dependability of surgical assistant robots by realistically testing their awareness of the actions performed by a surgeon. In ESAD, surgeon actions are classed into 21 different categories and annotated with the help of professional surgeons. This dataset is used in Chapter \ref{chapter:deformsgraph} for complex activity detection.

\subsection{Thumos-14} 
Thumos-14 \cite{idrees2017thumos} is one of the benchmark datasets for temporal action localisation. Thumos-14 contains 413 untrimmed temporally annotated videos categorised into 20 actions. Videos are characterised by a large variance in duration, from one second to 26 minutes. On average, each video contains 16 action instances. To compare our performance with the state-of-the-art, we also adopt the standard practice of using the validation set (200 videos) for training while evaluating our model on the test set (213 videos). We use this dataset for the evaluation of our complex activity framework in Chapter \ref{chapter:hgraph}.

\subsection{ActivityNet-1.3} 
ActivityNet-1.3 \cite{caba2015activitynet} is one of the largest action localisation datasets with around 20K untrimmed videos comprising 200 action categories. The videos are divided into training, validation, and testing folds according to a ratio of 2:1:1, respectively. The number of action instances per video is 1.65, which is quite low compared to Thumos-14. Following the previous art, we train our model on the training set and test it on the validation set. We use this dataset in Chapter \ref{chapter:deformsgraph} for the evaluation of our complex activity framework.

\subsection{50-Salad}

50-Salad \cite{stein2013combining} is an activity segmentation dataset with the aim of stimulate research on recognizing manipulative gestures. It consists of 50 videos of 25 cooks (actors) preparing salads, where each cook recorded two videos of themselves preparing salads by following different sequence of steps. The dataset contains two levels of annotations: ``high-level" labels only span three classes, whereas ``low-level" annotation involves a total of 51 classes. 50-Salad is a multimodel dataset contains other modality besides RGB videos including RGB-D (depth information), 3-axis accelerometer devices attached to a knife, a mixing spoon, a small spoon, a peeler, a glass, an oil bottle, and a pepper dispenser. This dataset is used in Chapter \ref{chapter:continual} for the evaluation of our latent model.

\section{Evaluation Metrics}\label{related_work:sec:evalmat}

We also provide the standard evaluation metrics used for the evaluation of our methods proposed in this thesis.

\subsection{Classification Accuracy}
Classification accuracy is a measure of how well a classification model is able to correctly predict the class labels of a given dataset. It is defined as the proportion of correctly classified instances to the total number of instances in the dataset. Classification accuracy is used in Chapter \ref{chapter:deformsgraph} for recognition of activities.

\subsection{Precision}
Precision is a measure of the accuracy of a classification model's positive predictions. It is defined as the proportion of true positive predictions (correctly predicted positive instances) to the total number of positive predictions (true positives plus false positives). 

A high precision value means that the model has a low number of false positives, or that it is good at not classifying negative instances as positive. In other words, it means that the model is good at identifying only the relevant instances among the positive predictions. Precision is used in Chapter \ref{chapter:deformsgraph} and Chapter \ref{chapter:continual} for recognition of activities.

\subsection{Recall}
Recall, also known as Sensitivity or the True Positive Rate (TPR), is a measure of a classification model's ability to correctly identify all relevant instances. It is defined as the proportion of true positive predictions (correctly predicted positive instances) to the total number of actual positive instances (true positives plus false negatives).

A high recall value means that the model has a low number of false negatives, or that it is good at identifying all relevant instances. In other words, it means that the model is good at detecting all the positive instances among all the actual positive instances. Recall is used in Chapter \ref{chapter:deformsgraph} and Chapter \ref{chapter:continual} for recognition of activities.

\subsection{F1-score}
The F1 Score is a measure of a classification model's accuracy that combines precision and recall. It is defined as the harmonic mean of precision and recall.

The F1 Score is a single value that represents the balance between precision and recall. A higher F1 Score indicates that the model has a better balance between precision and recall.

It is particularly useful when the classes are imbalanced, or when the cost of false positives and false negatives is different. In these cases, it is possible to have a high accuracy score but a low F1 score if precision and recall are not in balance. F1 score is used in Chapter \ref{chapter:deformsgraph} and Chapter \ref{chapter:continual} for recognition of activities.

\subsection{Mean Average Precision}
Mean Average Precision (mAP) is a common evaluation metric used in action detection, which is a subfield of computer vision that deals with recognizing and locating instances of specific actions within video sequences.
MAP calculates the average precision across all classes of actions in a video dataset, by taking into account both the precision and recall of the action detection algorithm.

Precision measures the proportion of true positive detections (correctly detected actions) to the total number of positive detections (both true positives and false positives). Recall measures the proportion of true positive detections to the total number of instances of the action that should have been detected.

mAP is calculated by first computing the average precision for each action class, and then taking the mean of those values. A higher MAP value indicates a better performance of the action detection algorithm. This evaluation metric is used for two purposes; firstly for the spatiotemporal detection of events in Chapter \ref{chapter:road}. Secondly, used for the temporal evaluation of complex activities in Chapter \ref{chapter:deformsgraph} and Chapter \ref{chapter:hgraph}.

\subsection{Mean Absolute Error}
Mean Absolute Error (MAE) is a measure of the average magnitude of the errors in a set of predictions, without considering their direction. It measures the difference between the predicted values and the true values.

MAE is a commonly used regression evaluation metric and it is relatively easy to understand and interpret. It is expressed in the same unit as the target variable and it gives an idea of the magnitude of the error, regardless of the direction of the error. MAE is used in Chapter \ref{chapter:continual} for the evaluation of continual crowd counting (regression problem).

    \pagestyle{plain}
    \cleardoublepage
\phantomsection
\newcommand{\pname}{Part I : ROAD (The ROad event Awareness Dataset for Autonomous Driving)} \label{part:part1}
\addcontentsline{toc}{chapter}{\pname}\label{partI}

\pagebreak
\hspace{14pt}
\vfill
\begin{center}
\textbf{\large{\pname}}
\end{center}
\vfill
\hspace{0pt}
\pagebreak
    \pagestyle{fancy}
    
\chapter{The ROad event Awareness Dataset for Autonomous Driving}
\renewcommand{\imagepath}{figures/offline/} 
\label{chapter:road}
\section{Introduction} 
\label{road:intro}

As anticipated,
the first part of this thesis concerns individual event detection, specifically the detection of road events from an autonomous driving perspective. Injecting capabilities such as the understanding of dynamic road events and their evolution into autonomous vehicles has the potential to take situational awareness and decision-making closer to human-level performance. 
To this purpose, in this chapter, we introduce the ROad event Awareness Dataset (ROAD) for Autonomous Driving. ROAD is designed to test an autonomous vehicle’s ability to detect road events.
\rev{To the best of our knowledge, ROAD is the first multi-label multi-action autonomous driving dataset where each of the bounding box is associated with three labels (agent, action, and location). This unique labeling schema makes ROAD more demanding and challenging in comparison to its competitors.} ROAD comprises videos originally from the Oxford RobotCar Dataset, annotated with bounding boxes showing the location in the image plane of each road event. We benchmark various detection tasks, proposing as a baseline a new incremental algorithm for online road event awareness termed 3D-RetinaNet.
We also report the performance on the ROAD tasks of Slowfast and YOLOv5 detectors, as well as that of the winners of the ICCV2021 ROAD challenge, which highlights the challenges faced by situation awareness in autonomous driving. ROAD is designed to allow scholars to investigate exciting tasks such as complex (road) activity detection, future event anticipation and continual learning.

 Further down the line, we also introduce logical constraints to ROAD. The main intuition behind these logical constraints is that neural networks have proven to be very powerful at computer vision tasks. However, they often exhibit unexpected behaviours, violating known requirements and expressing background knowledge.  This calls for models (i) able to learn from the requirements, and (ii) be guaranteed to be compliant with the requirements themselves. Unfortunately, the development of such models is hampered by the lack of datasets equipped with formally specified requirements. Therefore, we introduce the ROad event Awareness Dataset with logical Requirements (ROAD-R), the first publicly available dataset for autonomous driving with requirements expressed as logical constraints. 

\subsection{Motivation}

\rev{ The history of \textit{autonomous driving} has witnessed significant milestones and advancements since the Defense Advanced Research Projects Agency (DARPA\footnote{\url{https://www.darpa.mil/}}) Grand Challenge 2004, which marked a pivotal moment in the field. The DARPA Grand Challenge was a groundbreaking competition organized by the U.S. Department of Defense's DARPA agency. Its aim was to accelerate the development of autonomous vehicle technology. The challenge required participating vehicles to navigate a designated route in a desert environment, relying solely on their autonomous capabilities. The main purpose of the Challenge was to push the boundaries of autonomous vehicle technology. The goal was to foster innovation and encourage advancements that could potentially benefit military applications, such as unmanned ground vehicles for reconnaissance or supply missions \cite{behringer2004darpa}. The DARPA Grand Challenge helped establish autonomous driving as a viable research area and laid the foundation for subsequent advancements in the field.}

In recent years, \textit{autonomous driving} has emerged as a fast-growing research area. The race towards fully autonomous vehicles pushed many large companies, such as Google, Toyota, and Ford, to develop their own concept of \emph{robot-car}~\cite{winn2006layout,KirstenNov2017,gpandey-2011a}. 
While self-driving cars are widely considered to be a  major development and testing ground for the real-world application of artificial intelligence, 
major reasons for concern remain 
in terms of safety, ethics, cost, and reliability~\cite{maurer2016autonomous}.
From a safety standpoint, in particular, smart cars need to robustly interpret the behaviour of the humans (drivers, pedestrians or cyclists) they share the environment with, in order to cope with their decisions.
\emph{Situation awareness} and the ability to %predict intent and 
understand the behaviour of other road users are thus crucial for the safe deployment of autonomous vehicles (AVs). \rev{Therefore, in this chapter, we only consider the context of \emph{vision-based} autonomous driving~\cite{BERTOZZI20001} from video sequences captured by cameras mounted on the vehicle in a streaming, online fashion.}

While detector networks~\cite{YOLO9000} are routinely trained to facilitate object and actor recognition in road scenes, this simply allows the vehicle to 'see' what is around it. The philosophy of this work is that robust self-driving capabilities require a deeper, more human-like understanding of dynamic road environments (and of the evolving behaviour of other road users over time) in the form of semantically meaningful concepts, as a stepping stone for intention prediction and automated decision making. One advantage of this approach is that it allows the autonomous vehicle to focus on a much smaller amount of relevant information when learning how to make its decisions, in a way arguably closer to how decision-making takes place in humans.

\begin{figure*}[ht!] 
  \centering
  \includegraphics[width=0.99\textwidth]{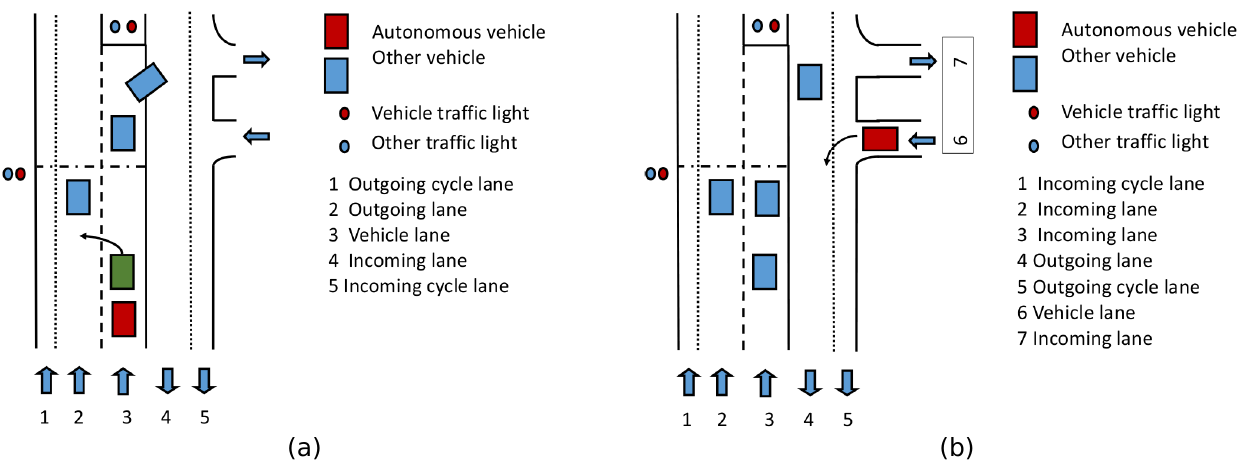}
  \caption{Use of labels in ROAD to describe typical road scenarios. (a) A green car is in front of the AV while changing lanes, as depicted by the arrow symbol. The associated event will then carry the following labels: \textit{in vehicle lane} (location), \textit{moving left} (action). Once the event is completed, the location label will change to: \textit{in outgoing lane}. (b) Autonomous vehicle turning left from lane 6 into lane 4: lane 4 will be the \textit{outgoing lane} as the traffic is moving in the same direction as the AV. However, if the AV turns right from lane 6 into lane 4 (a wrong turn), then lane 4 will become the \textit{incoming lane} as the vehicle will be moving into the incoming traffic. The overall philosophy of ROAD is to use suitable combinations of multiple label types to fully describe a road situation, and allow a machine learning algorithm to learn from this information.}
  \label{fig:typical} 
\end{figure*}

On the opposite side of the spectrum lies end-to-end reinforcement learning. There, the behaviour of a human driver in response to road situations is used to train, in an {imitation learning} setting~\cite{codevilla2018end}, an autonomous car to respond in a more ‘human-like’ manner to road scenarios. This, however, requires an astonishing amount of data from 
a myriad of road situations. For highway driving only, a relatively simple task when compared to city driving, Fridman et al. in~\cite{Fridman2017arxiv} had to use a whole fleet of vehicles to collect 45 million frames. 
Perhaps more importantly, in this approach, the network learns a mapping from the scene to control inputs, without attempting to model the significant facts taking place in the scene or the reasoning of the agents therein. As discussed in~\cite{Cuzzolin2020tom}, many authors~\cite{Rasouli2020,Rudenko2019} have recently highlighted the insufficiency
of models which directly map observations to actions \cite{Armstrong2018}, specifically in the scenario of the self-driving car.

\subsection{ROAD: a multi-label, multi-task dataset} \label{road:subsec:intro-road}

\textbf{Concept}. This work aims to propose a new framework for situation awareness and perception, departing from the disorganised collection of object detection, semantic segmentation or pedestrian intention tasks which is the focus of much current work. We propose to do so in a “holistic”, multi-label approach in which agents, actions and their locations are all ingredients in the fundamental concept of \emph{road event} (RE). Road events are defined as triplets $E = (Ag, Ac, Loc)$ composed by an active road agent $Ag$, the action(s) $Ac$ it performs (possibly more than one at the same time), and \rev{the location(s) $Loc$\footnote{\rev{Note: The label location $Loc$ in our case does not refer to the spatiotemporal coordinates (x, y) of the bounding box.}} in which this takes place (which may vary from the start to the end of the event itself), as seen from the point of the view of an autonomous vehicle. The complete list of all three label types is provided in Table \ref{tab:labels}.} This takes the problem to a higher conceptual level, in which AVs are tested on their \emph{understanding of what is going on} in a dynamic scene rather than their ability to describe what the scene \emph{looks like}, putting them in a position to use that information to make decisions and a plot course of action. Modelling dynamic road scenes in terms of road events can also allow us to model the causal relationships between what happens; these causality links can then be exploited to predict further future consequences.

To transfer this conceptual paradigm into practice, this chapter introduces ROAD, the first \emph{ROad event Awareness in Autonomous Driving Dataset}, as an entirely new type of dataset designed to allow researchers in autonomous vehicles to test the situation awareness capabilities of their stacks in a manner impossible until now. Unlike all existing benchmarks, ROAD provides ground truth for the action performed by all road agents, not just humans.
In this sense ROAD is unique in the richness and sophistication of its annotation, designed to support the proposed conceptual shift. We are confident this contribution will be very useful moving forward for both the autonomous driving and the computer vision community. \rev{However, at the same time, it is also worth mentioning that the quantisation of the action and activities in the context of autonomous driving with several limitation and challenges including computational complexity, robustness to uncertainty, and control and decision-making. }

\textbf{Features}.
ROAD is built upon (a fraction of) the Oxford RobotCar Dataset \cite{maddern20171}, by carefully annotating 22 carefully selected, relatively long-duration videos. Road events are represented as '{tubes}', i.e., time series of frame-wise bounding box detections. ROAD is a dataset of significant size, most notably in terms of the richness and complexity of its annotation rather than the raw number of video frames. A total of $122K$ video frames are labelled for a total of $560K$ detection bounding boxes in turn associated with $1.7M$ unique individual labels,  broken down into $560K$ agent labels, $640K$ action labels and $499K$ location labels.

The dataset was designed according to the following principles.
\begin{itemize}
    \item A \emph{multi-label} benchmark: each road event is composed by the label of the (moving) agent responsible, the label(s) of the type of action(s) being performed, and labels describing where the action is located. 
    \item Each event can be assigned \emph{multiple instances} of the same label type whenever relevant (e.g., an RE can be an instance of both \textit{moving away} and \textit{turning left}).
    \item The labelling is done \emph{from the point of view of the AV}: the final goal is for the autonomous vehicle to use this information to make the appropriate decisions.
    \item The meta-data is intended to contain all the information required to fully describe a road scenario: an illustration of this concept is given in Figure \ref{fig:typical}. After closing one's eyes, the set of labels associated with the current video frame should be sufficient to recreate the road situation in one's head (or, equivalently, sufficient for the AV to be able to make a decision).
\end{itemize}

In an effort to take action detection into the real world, ROAD moves away from human body actions almost entirely, to consider (besides pedestrian behaviour) actions performed by humans as drivers of various types of vehicles, shifting the paradigm from \emph{actions performed by human bodies} to \emph{events caused by agents}.
As shown in our experiments, ROAD is more challenging than current action detection benchmarks due to the complexity of road events happening in real, non-choreographed driving conditions, the number of active agents present and the variety of weather conditions encompassed.

\textbf{Tasks}.
ROAD allows one to validate manifold tasks associated with situation awareness for self-driving, each associated with a label type (agent, action, location) or a combination thereof: 
\begin{enumerate} 
\item \emph{Agent detection}, the output is in the form of 'agent tubes' collecting the BBs associated with an active road agent in consecutive frames. 
\item \emph{Action detection}, the output is in the form of 'action tubes' (series of bounding box (BB) detections with an attached action label).
\item \emph{Location detection}, the output is in the form of 'agent-location' tubes collecting the BBs associated with an agent and its location in consecutive frames.
\item \emph{Agent-action detection}, the detector is trained on pairs of agent-action labels. The output is again in the form of a tube. 
\item \emph{Road event detection}, as well as the AV's decisions make use of all the three types of information provided by ROAD, detecting RE triplets is crucial for autonomous driving applications. 
\item \emph{Temporal \rev{AV actions detection}}, the output of this \rev{task} is the class of action the AV performs in each video frame, from a list suitably compiled. 

\end{enumerate}
For each task, one can assess both \emph{frame-level} detection, which outputs independently for each video frame the bounding box(es) of the instances there present and the relevant class labels, and \emph{video-level} detection, which consists in regressing the whole series of temporally-linked bounding boxes (i.e., in current terminology, a ’tube’) associated with an instance, together with the relevant class label. In this chapter, we conduct tests on both. All tasks come with both the necessary annotation and a shared baseline, which is described in Section \ref{road:sec:baseline-method}.

\begin{table}[h]

    \centering
    \caption{\rev{ROAD labels.}}
    \label{tab:labels}
    \begin{tabular}{l l l}
    \toprule
         \textbf{Agents} & \textbf{Actions} & \textbf{Locations} \\
    \midrule
    Pedestrian & Move away & AV lane \\
    Car & Move towards & Outgoing lane\\
    Cyclist & Move & Outgoing cycle lane \\
    Motorbike & Brake &  Incoming lane\\
    Medium vehicle & Stop & Incoming cycle lane \\ 
    Large vehicle & Indicating left & Pavement\\
    Bus & Indicating right & Left pavement \\
    Emergency vehicle & Hazards lights on & Right pavement\\
    AV traffic light & Turn left & Junction\\
    Other traffic light & Turn right & Crossing location \\
                        & Overtake & Bus stop \\
                        & Wait to cross & Parking\\
                        & Cross from left \\
                        & Cross from right \\
                        & Crossing \\
                        & Push object \\
                        & Red traffic light \\
                        & Amber traffic light \\
                        & Green traffic light \\
    \bottomrule
    \end{tabular}
\end{table}

\subsection{ROAD-R}
In this chapter, we also generalize hierarchical multi-label classification (HMC) problems by introducing {\sl multi-label classification problems with (full) propositional logic requirements}. Thus,   given a multi-label classification problem with labels $A$, $B$, and $C$, we can, for example, write the requirement: 
$$
(\neg A \wedge B) \vee C,
$$
stating that for each data point in the dataset either the label $C$ is predicted, or $B$ but not $A$ are predicted. 

Then, we present the ROad event Awareness Dataset with logical Requirements (ROAD-R), the first 
publicly available dataset for autonomous driving with requirements expressed as logical constraints.  ROAD-R extends the ROAD dataset by manually annotated ROAD-R with 243 constraints, each verified to hold for each bounding box. A typical constraint is thus ``a traffic light cannot be red and green at the same time'', while there are no constraints like ``pedestrians should cross at crossings”, which should always be satisfied in theory, but which might not be in real-world scenarios.

Given ROAD-R, we considered 6 current state-of-the-art (SOTA) models, and we showed that they are not able to learn the requirements just from the data points, as more than 90\% of the times, they produce predictions that violate the constraints. 

% Then, we
% faced the problem of how to leverage the additional knowledge provided by constraints with the goal of (i)~improving their performance, measured by the frame mean average precision (f-mAP) at intersection over union (IoU) thresholds 0.5 and 0.75; see, e.g.,~\cite{kalogeiton2017a,li2018map}, and (ii) guaranteeing that they are compliant with the constraints. 
% To achieve the above two goals, three new models are proposed including CL models (with a {\sl constrained loss}),  CO models (with a {\sl constrained output}), and CLCO models (with both a constrained loss and a constrained output).

% In particular, we consider three different ways to build CL (resp.,  CO, CLCO) models. More specifically, we run the $9 \times 6$ models obtained by equipping the 6 current SOTA models with a constrained loss and/or a constrained output, and we show that it is always possible to
% \begin{enumerate}
%     \item improve the performance of each SOTA model, and 
%     \item be compliant with (i.e., strictly satisfy) the constraints. 
% \end{enumerate}
% Overall, the best performing model (for IoU = 0.5 and also IoU = 0.75) is CLCO-RCGRU, i.e., the SOTA model RCGRU equipped with both constrained loss and constrained output: CLCO-RCGRU (i)~always satisfies the requirements  and (ii)~has f-mAP = 31.81 for IoU = 0.5, and f-mAP = 17.27 for IoU = 0.75.
% On the other hand, the standard RCGRU model 
% (i)~produces predictions that violate the constraints at least 92\% of the times,
% and (ii) has f-mAP = 30.78 for IoU = 0.5 and f-mAP = 15.98 for IoU = 0.75. 

\subsection{Main Contributions}

The major contributions of this work are thus the following.
\begin{itemize}
\item
A conceptual shift in situation awareness centred on a formal definition of the notion of road event, as a triplet composed by a road agent, the action(s) it performs and the location(s) of the event, seen from the point of view of the AV.
\item A new ROad event Awareness Dataset for Autonomous Driving (ROAD), the first of its kind, designed to support this paradigm shift and allow the testing of a range of tasks related to situation awareness for autonomous driving: agent and/or action detection, event detection, ego-action classification.

\item A multi-label classification problem with propositional logic requirements.
\item Introduction of ROAD-R, which is the first publicly available dataset whose requirements are expressed in full propositional logic.
\item Considered 6 SOTA models and show that on ROAD-R,  they produce predictions violating the requirements more than $\sim$90\% of the times, 

\end{itemize}
Instrumental to the introduction of ROAD as the benchmark of choice for semantic situation awareness, we propose a robust baseline for online action/agent/event detection (termed \emph{3D-RetinaNet}) which combines state-of-the-art single-stage object detector technology with an online tube construction method~\cite{singh2017online}, with the aim of linking detections over time to create \emph{event tubes}~\cite{saha2016deep,Georgia-2015a}. Results for two additional baselines based on a Slowfast detector architecture \cite{feichtenhofer2019slowfast} and YOLOv5\footnote{\url{https://github.com/ultralytics/yolov5}.} (for agent detection only) are also reported and critically assessed.

\paragraph{Related publications:}
The work of this chapter appeared in IEEE transactions on pattern analysis and machine intelligence (IEEE TPAMI 2022) \cite{singh2022road} and The ROAD Workshop and Challenge: Event Detection for Situation Awareness in Autonomous Driving at ICCV 2021\footnote{\url{https://sites.google.com/view/roadchallangeiccv2021/}}. The dissertation's author was the co-author of the TPAMI paper and co-organiser of the workshop and challenge. 
The main contributions of the dissertation's author to this project are: adding a new backbone (SlowFast) to the 3D RetinaNet architecture (baseline for ROAD), implementing a YOLOv5 for agents detection, and setting up the evaluation site for the challenge. All the resources of this project are publically available and are accessed from \emph{ROAD dataset}\footnote{\url{https://github.com/gurkirt/road-dataset}}, \emph{3D RetinaNet baseline}\footnote{\url{https://github.com/gurkirt/3D-RetinaNet}}, \emph{Inference demo}\footnote{\url{https://www.youtube.com/watch?v=CmxPjHhiarA}}, and \emph{Challenge evalAI}\footnote{\url{https://eval.ai/web/challenges/challenge-page/1059/overview}}.

The work of ROAD-R appeared in IJCAI 2022 Workshop on Artificial Intelligence for Autonomous Driving (AI4AD 2022)\footnote{\url{https://learn-to-race.org/workshop-ai4ad-ijcai2022/}} and IJCLR 2022 the International Joint Conference on Learning and Reasoning \cite{giunchiglia2022road}. This work received \textbf{the best paper award}  at IJCAI 2022 AI4AD and won \textbf{the best student paper award} at IJCLR 2022. The dissertation's author was the co-author in both papers. The main contribution of the dissertation's author to this project is providing the vision baselines for on ROAD dataset for logical constraint. All the resources of this project are publically available and are accessed from \emph{code}\footnote{\url{https://github.com/EGiunchiglia/ROAD-R}}, video\footnote{\url{https://www.youtube.com/watch?v=_-Ll5d1VQXY}}.

\paragraph{Outline:}
The remainder of the chapter is organised as follows. Section \ref{road:sec:road-dataset} presents our ROAD dataset in full detail, including: its multi-label nature (Sec. \ref{road:sec:dataset-multilabel}), data collection (Sec. \ref{road:sec:dataset-collection}), annotation (Sec. \ref{road:subsec:dataset-annotation}), the tasks it is designed to validate (Sec. \ref{road:sec:dataset-tasks}), and a quantitative summary (Sec. \ref{road:sec:dataset-summary}). 
Section \ref{road:sec:baseline-method} presents an overview of the proposed 3D-RetinaNet baseline, and recalls the ROAD challenge organised by some of us at ICCV 2021 to disseminate this new approach to situation awareness within the autonomous driving and computer vision communities, using ROAD as the benchmark. Section \ref{road_r:sec:roadr} describes ROAD-R. Experiments are described in Section \ref{road:sec:exp}, where a number of ablation studies are reported and critically analysed in detail, together with the results of the ROAD challenge's top participants. Section \ref{road:sec:extensions} outlines additional exciting tasks the dataset can be used as a benchmark for in the near future, such as future event anticipation, decision making and machine theory of mind \cite{Cuzzolin2020tom}. Conclusions and future work are outlined in Section \ref{road:sec:conclusion}.

\section{The Dataset} \label{road:sec:road-dataset}

\subsection{A Multi-Label Benchmark} \label{road:sec:dataset-multilabel}

The ROAD dataset is specially designed from the perspective of self-driving cars, and thus includes actions performed not just by humans but by all road agents in specific locations, to form \emph{road events} (REs). 
REs are annotated by drawing a bounding box around each active road agent present in the scene, and linking these bounding boxes over time to form 'tubes'. 
As explained, to this purpose three different types of labels are introduced, namely: 
(i) the category of \emph{road agent} involved (e.g. \textit{Pedestrian}, \textit{Car}, \textit{Bus}, \textit{Cyclist}); 
(ii) the \emph{type of action} being performed by the agent (e.g. \textit{Moving away}, \textit{Moving towards}, \textit{Crossing} and so on), and
(iii) the \emph{location} of the road user relative the autonomous vehicle perceiving the scene (e.g. \textit{In vehicle lane}, \textit{On right pavement}, \textit{In incoming lane}).
In addition, ROAD labels the actions performed by the vehicle itself. Multiple agents might be present at any given time, and each of them may perform multiple actions simultaneously (e.g. a \textit{Car} may be \textit{Indicating right} while \textit{Turning right}). Each agent is always associated with at least one action label. The complete list of all three labels types are given in Table \ref{tab:labels}.

\emph{Agent labels}. %\label{subsec:agent-labels}
Within a road scene, the objects or people able to perform actions which can influence the decision made by the autonomous vehicle are termed \textit{agents}.
We only annotate \emph{active} agents (i.e., a parked vehicle or a bike or a person visible to the AV but located away from the road are not considered to be 'active' agents).
Three types of agent are considered to be of interest, in the sense defined above, to the autonomous vehicle: people, vehicles and traffic lights.
For simplicity, the AV itself is considered just like another agent: this is done by labelling the vehicle's bonnet.
People are further subdivided into two sub-classes: pedestrians and cyclists.
The vehicle category is subdivided into six sub--classes: car, small--size motorised vehicle, medium--size motorised vehicle, large--size motorised vehicle, bus, motorbike, emergency vehicle.
Finally, the `traffic lights' category is divided into two sub--classes: \textit{Vehicle traffic light}
(if they apply to the AV) and \textit{Other traffic light} (if they apply to other road users). Only one agent label can be assigned to each active agent present in the scene at any given time.

\emph{Action labels}. %\label{subsec:action-labels}
Each {agent} can perform one or more \textit{actions} at any given time instant. For example, a traffic light can only carry out a single action: it can be either red, amber, green or `black'. A car, instead, can be associated with two action labels simultaneously, e.g., \textit{Turning right} and \textit{Indicating right}. Although some road agents are inherently multitasking, some action combinations can be suitably described by a single label: for example, pushing an object (e.g. a pushchair or a trolley-bag) while walking can be simply labelled as \textit{Pushing object}. The latter was our choice. 

\emph{AV own actions}. %~\label{subsec:av-actions}
Each video frame is also labelled with the action label associated with what the AV is doing. To this end, a bounding box is drawn on the bonnet of the AV. The AV can be assigned one of the following seven action labels: \textit{AV-move}, \textit{AV-stop}, \textit{AV-turn-left}, \textit{AV-turn-right}, \textit{AV-overtake}, \textit{AV-move-left} and \textit{AV-move-right}.

\emph{Location labels}. %\label{subsec:location-labels}
Agent \textit{location} is crucial for deciding what action the AV should take next. 
As the final, long-term objective of this project is to assist autonomous decision making, we propose to label the location of each agent from the perspective of the autonomous vehicle. For example, a pedestrian can be found on the right or the left pavement, in the vehicle's own lane, while crossing or at a bus stop. The same applies to other agents and vehicles as well. 
There is no location label for the traffic lights as they are not movable objects, but agents of a static nature and well-defined location.
To understand this concept, Fig. \ref{fig:typical} illustrates two scenarios in which the location of the other vehicles sharing the road is depicted from the point of view of the AV. 
\emph{Traffic light} is the only agent type missing location labels, all the other agent classes are associated with at least one location label.

\subsection{Data Collection} \label{road:sec:dataset-collection}

ROAD is composed of 22 videos from the publicly available Oxford Robot Car Dataset ~\cite{maddern20171} (OxRD) released in 2017 by the Oxford Robotics Institute\footnote{\url{http://robotcar-dataset.robots.ox.ac.uk/}}, covering diverse road scenes under various weather conditions. The OxRD dataset, collected from the narrow streets of the historic city of Oxford, was selected because it presents challenging scenarios for an autonomous vehicle due to the diversity and density of various road users and road events. 

Note, however, that our labelling process (described below) is not limited to OxRD. In principle, other autonomous vehicle datasets (e.g.~\cite{yu2018bdd100k,geiger2013vision}) may be labelled in the same manner to further enrich the ROAD benchmark, we plan to do exactly so in the near future.

\emph{Video selection}. %~\label{subsubsec:videoselection}
Within OxRD, videos were selected with the objective of ensuring diversity in terms of weather conditions, times of the day and types of scenes recorded. 
Specifically, the 22 videos have been recorded both during the day (in strong sunshine, rain or overcast conditions, sometimes with snow present on the surface) and at night. 

\emph{Preprocessing}. First, the original sets of video frames were downloaded and demosaiced, in order to convert them to red, green, and blue (RGB) image sequences. Then, they were encoded into proper video sequences using \texttt{ffmpeg}\footnote{\href{https://www.ffmpeg.org/}{https://www.ffmpeg.org/}} at the rate of 12 frames per second (fps). 
Although the original frame rate in the considered frame sequences varies from 11 fps to 16 fps, we uniformised it to keep the annotation process consistent.

\subsection{Annotation Process} \label{road:subsec:dataset-annotation}

\emph{Annotation tool}.
Annotating tens of thousands of frames rich in content is a very intensive process; therefore, a tool is required which can make this process both fast and intuitive. For this work, we adopted Microsoft's VoTT\footnote{\url{https://github.com/Microsoft/VoTT/}}. The most useful feature of this annotation tool is that it can copy annotations (bounding boxes and their labels) from one frame to the next, while maintaining a unique identification for each box, so that boxes across frames are automatically linked together. 
Moreover, VoTT also allows for multiple labels, thus lending itself well to ROAD's multi-label annotation concept.

\emph{Annotation protocol}. All salient objects and actors within the frame were labelled, with the exception of inactive participants (mostly parked cars) and objects/actors at large distances from the ego vehicle, as the latter were judged to be irrelevant to the AV’s decision-making. This can be seen in the attached 30-minute video\footnote{\url{https://www.youtube.com/watch?v=CmxPjHhiarA}.} portraying ground truth and predictions.
As a result, pedestrians, cyclists, and traffic lights were always labelled. Vehicles, on the other hand, were only labelled when active (i.e., moving, indicating, being stopped at lights or stopping with hazard lights on the side of the road). As mentioned, only parked vehicles were not considered active (as they do not arguably influence the AV's decision-making) and were thus not labelled.

\emph{Event label generation}. 
Using the annotations manually generated for actions and agents in the multi-label scenario as discussed above it is possible to generate \emph{event-level} labels about agents, e.g. \textit{Pedestrian / Moving towards} the AV \textit{On right pavement} or \textit{Cyclist / Overtaking / In vehicle lane}. 
Any combinations of location, action and agent labels are admissible.
If location labels are ignored, the resulting event labels become location-invariant. 
\\
Namely, given an agent tube and the continuous temporal sequence of action labels attached to its constituent bounding box detections, we can generate action tubes by looking for changes in the action label series associated with each agent tube. For instance, a \textit{Car} appearing in a video might be first \textit{Moving away} before \textit{Turning left}. The agent tube for the car will then be formed by two contiguous agent-action tubes: a first tube with label pair \textit{Car / Moving away} and a second one with pair \textit{Car / Turning left}.

\begin{table}[ht!]
     \centering
     \caption{ROAD tasks and attributes.}     
     \label{tab:task_types}
     \begin{tabular}{lccc}
         Task type & Problem type & Output & Multiple labels \\
         \midrule
         Active agent & Detection & Box\&Tube & No \\
         Action & Detection & Box\&Tube & Yes \\
         Location & Detection & Box\&Tube & Yes \\
         Event & Detection & Box\&Tube & Yes \\
         AV-action & \rev{Temporal detection} & Start/End & No \\
         \bottomrule
     \end{tabular}
 \end{table}

\subsection{Tasks} \label{road:sec:dataset-tasks}

ROAD is designed as a sandbox for validating the six tasks relevant to situation awareness in autonomous driving outlined in Sec. \ref{road:subsec:intro-road}.
Five of these tasks are detection tasks, while the last one is a frame-level action recognition task sometimes referred to as 'temporal action segmentation'~\cite{sigurdsson2018charadesego}, Table~\ref{tab:task_types} shows the main attributes of these tasks.

All detection tasks are evaluated both at frame-level and at video- (tube-)level. \emph{Frame-level detection} refers to the problem of identifying in each video frame the bounding box(es) of the instances there present, together with the relevant class labels. \emph{Video-level detection} consists in regressing a whole series of temporally-linked bounding boxes (i.e., in current terminology, a '{tube}') together with the relevant class label. 
In our case, the bounding boxes will mark a specific active agent in the road scene. The labels may issue (depending on the specific task) either from one of the individual label types described above (i.e., agent, action or location) or from one of the meaningful combinations described in~\ref{road:subsec:dataset-annotation} (i.e., event). 

Below we list all the tasks for which we currently provide a baseline, with a short description.

\begin{enumerate}
\item{\emph{Active agent detection} (or \emph{agent detection}) %is the task which 
aims at localising an active agent using a bounding box (frame-level) or a tube (video-level) and assigning a class label to it.}  
\item{\emph{Action detection} 
seeks to localise an active agent occupied in performing a specific action from the list of action classes.} 
\item
In \emph{agent location detection} (or \emph{location detection}) a label from the relevant list of locations (as seen from the AV) is sought and attached to the relevant bounding box or tube.
\item{\emph{Road event detection} (or \emph{event detection}) consists in assigning to each box or tube a triplet of class labels.}
\item{\emph{Autonomous vehicle temporal action segmentation} is a frame-level action classification task in which each video frame is assigned a label from the list of possible AV own actions. 
We refer to this task as 'AV-action segmentation', similarly to \cite{sigurdsson2018charadesego}.}
\end{enumerate}

\begin{figure*}[t]
  \centering{
      \includegraphics[width=0.99\textwidth]{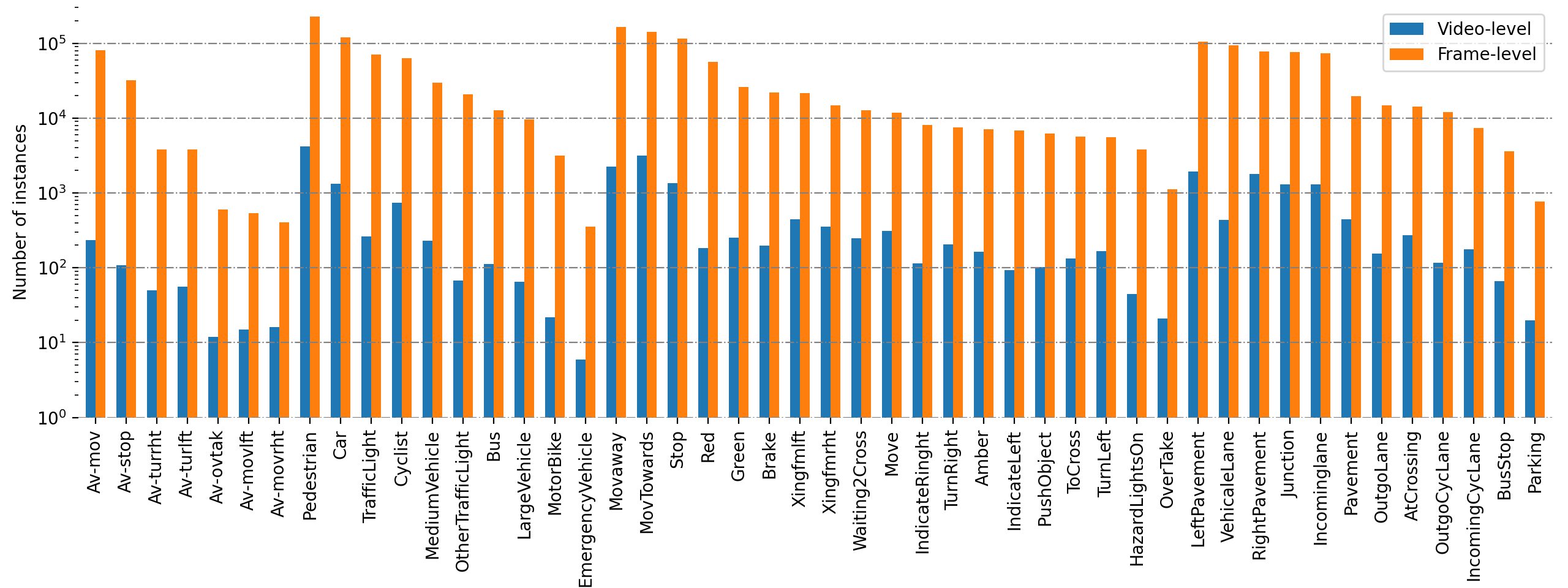}
  \caption{Number of instances of each class of individual label-types, in logarithmic scale. \label{fig:stats}}
  }
\end{figure*}
 
\subsection{Quantitative Summary} \label{road:sec:dataset-summary}

Overall, $122K$ frames extracted from 22 videos were labelled, in terms of both AV own actions (attached to the entire frame) and bounding boxes with attached one or more labels of each of the three types: agent, action, location.
In total, ROAD includes $560K$ bounding boxes with $1.7M$ instances of individual labels. The latter figure can be broken down into $560K$ instances of agent labels, $640K$ instances of action labels, and $499K$ instances of location labels.
Based on the manually assigned individual labels, we could identify $454K$ instances of triplets (event labels).

The number of instances for each individual class from the three lists is shown in Fig.~\ref{fig:stats} (frame-level, in orange). 
The $560K$ bounding boxes make up $7,029$, $9,815$, $8,040$, and $8,394$ tubes for the label types agent, action, location, and event, respectively.
Figure~\ref{fig:stats} also shows the number of tube instances for each class of individual label types as number of video-level instances (in blue). \rev{The figure also highlights the presence of an imbalance in the dataset distribution among the classes. Specifically, certain classes exhibit a significant disparity in sample sizes, for example, "car" has a considerably larger number of samples compared to "emergency vehicle."}

\section{Baseline and Challenge} \label{road:sec:baseline-method}

Inspired by the success of recent 3D CNN architectures~\cite{carreira2017quo} for video recognition and of feature-pyramid networks (FPN)~\cite{lin2017feature} with focal loss~\cite{lin2017focal}, we propose a simple yet effective 3D feature pyramid network (3D-FPN) with focal loss as a baseline method for ROAD's detection tasks. We call this architecture \emph{3D-RetinaNet}.

\begin{figure*}[t!]
    \centering{
        \includegraphics[width=\textwidth]{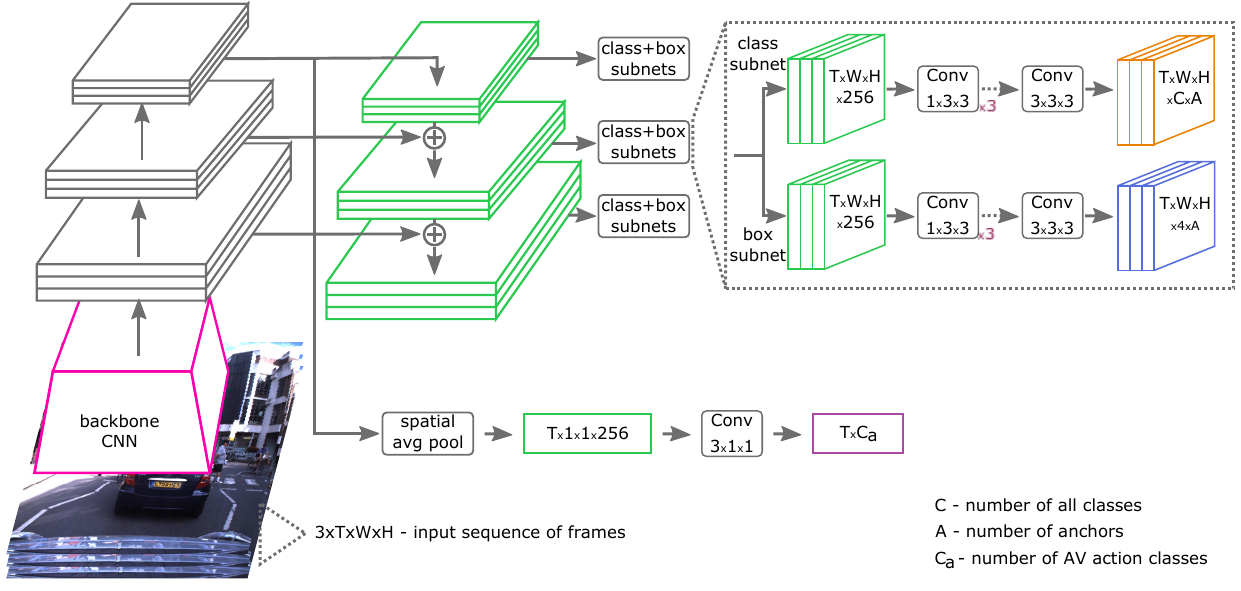}
    \caption{Proposed 3D-RetinaNet architecture for online video processing.}
    \label{fig:3dfpn}
    }
\end{figure*}

\subsection{3D-RetinaNet Architecture}

The data flow of 3D-RetinaNet is shown in Figure~\ref{fig:3dfpn}.
The input is a sequence of $T$ video frames. As in classical FPNs~\cite{lin2017feature}, the initial block of 3D-RetinaNet consists of a backbone network outputting a series of forward feature pyramid maps, and of lateral layers producing the final feature pyramid composed by $T$ feature maps. 
The second block is composed by two sub-networks which process these features maps to produce both bounding boxes ($4$ coordinates) and $C$ classification scores for each anchor location (over $A$ possible locations). 
In the case of ROAD, the integer $C$ is the sum of the numbers of agent, action, location, and agent-action-location (event) classes, plus one reserved for an \emph{agentness} score. The extra class agentness is used to describe the presence or absence of an active agent.
As in FPN~\cite{lin2017feature}, we adopt ResNet50~\cite{he2016deep} as the backbone network. 

\emph{2D versus 3D backbones}. 
In our experiments we show results obtained using three different backbones: frame-based ResNet50 (2D), inflated 3D (I3D)~\cite{carreira2017quo} and Slowfast~\cite{feichtenhofer2019slowfast}, in the manner also explained in ~\cite{nonlocal2018wang,feichtenhofer2019slowfast}. 
Choosing a 2D backbone makes the detector completely online~\cite{singh2017online}, with a delay of a single frame. Choosing an I3D or a Slowfast backbone, instead, causes a 4-frame delay at detection time.
Note that, as Slowfast and I3D networks makes use of a max-pool layer with stride 2, the initial feature pyramid in the second case contains $T/2$ feature maps. Nevertheless, in this case we can simply linearly upscale the output to $T$ feature maps.

\emph{AV action prediction heads}. 
In order for the method to also address the prediction of the AV's own actions (e.g. whether the AV is stopping, moving, turning left etc.),
we branch out the last feature map of the pyramid (see Fig. \ref{fig:3dfpn}, bottom)
and apply spatial average pooling, followed by a temporal convolution layer. The output is a score for each of the $C_a$ classes of AV actions, for each of the $T$ input frames. 

\emph{Loss function}. 
As for the choice of the loss function, we adopt a binary cross-entropy-based focal loss~\cite{lin2017focal}. We chose a binary cross entropy because our dataset is multi-label in nature \rev{ for which we only train a single binary cross entropy function. The rationale behind selecting a focal-type loss function stems from the anticipation that it could potentially help the neural network in addressing the challenge of long tail distribution of data (see Figure~\ref{fig:stats}), where some classes have significantly more samples than others. The focal-type loss accomplishes this by giving more weight to samples from minority classes during training, enabling the model to better learn their distinctive characteristics.}

\subsection{Online Tube Generation via Agentness Score} \label{road:subsec:tube-generation}

The autonomous driving scenario requires any suitable method for agent, action or event tube generation to work in an \emph{online} fashion, by incrementally updating the existing tubes as soon as a new video frame is captured.
For this reason, this work adopts a recent algorithm proposed by Singh \etal \cite{singh2017online}, which incrementally builds action tubes in an online fashion and at real-time speed. To be best of our knowledge, \cite{singh2017online} was the first online multiple action detection approach to appear in the literature, and was later adopted by almost all subsequent works~\cite{kalogeiton2017action,li2020actionsas,singh2018tramnet} on action tube detection. 

\emph{Linking of detections}. 
We now briefly review the tube-linking method of Singh~\etal~\cite{singh2017online}, and show how it can be adapted to build agent tubes based on an 'agentness' score, rather than build a tube separately for each class as proposed in the original paper.
This makes the whole detection process faster, since the total number of classes is much larger than in the original work~\cite{singh2017online}.
The proposed 3D-RetinaNet is used to regress and classify detection boxes in each video frame potentially containing an active agent of interest. Subsequently, detections whose score is lower than $0.025$ are removed and non-maximal suppression is applied based on the agentness score. 

At video start, each detection initialises an agentness tube. From that moment on, at any time instance $t$ the highest scoring tubes in terms of mean agentness score up to $t-1$ are linked to the detections with the highest agentness score in frame $t$ which display an Intersection-over-Union (IoU) overlap with the latest detection in the tube above a minimum threshold $\lambda$. The chosen detection is then removed from the pool of frame-$t$ detections. This \rev{continues} until the tubes are either assigned or not assigned a detection from current frame. Remaining detections at time $t$ are used to initiate new tubes.  A tube is terminated after no suitable detection is found for $n$ consecutive frames. 
As the linking process takes place, each tube carries scores for all the classes of interest for the task at hand (e.g., action detection rather than event detection), as produced by the classification subnet of 3D-RetinaNet. We can then label each agentness tube using the $k$ classes that show the highest mean score over the duration of the tube.

% \emph{Temporal trimming}. %~\label{subsec:temporal-trimming}
% Most tubelet based methods~\cite{kalogeiton2017action,li2020actionsas,simonyan2014two} do not perform any temporal trimming of the action tubes generated in such a way (i.e., they avoid deciding when they should start or end). Singh~\etal~\cite{singh2017online} proposed to pose the problem in a label consistency formulation solved via dynamic programming. However, as it turns out, temporal trimming~\cite{singh2017online} does not actually improve performance, as shown in \cite{singh2018tramnet}, except in some settings, for instance in the DALY ~\cite{daly2016weinzaepfel} dataset. 

% The situation is similar for our ROAD dataset as opposed to what happens on UCF-101-24, for which temporal trimming based on solving the label consistency formulation in terms of the actionness score, rather than the class score, does help improve localisation performance. Therefore, in our experiments we only use temporal trimming on the UCF-101-24 dataset but not on ROAD.

\subsection{The ROAD Challenge} \label{sec:challenge}

To introduce the concept of road event, our new approach to situation awareness and the ROAD dataset to the computer vision and AV communities, some of us have organised in October 2021 the workshop "The ROAD challenge: Event Detection for Situation Awareness in Autonomous Driving"\footnote{\url{https://sites.google.com/view/roadchallangeiccv2021/}.}.
For the challenge, we selected (among the tasks described in Sec. \ref{road:sec:dataset-tasks}) only three tasks: agent detection, action detection and event detection, which we identified as the most relevant to autonomous driving.

% protocol
As standard in action detection, evaluation was done in terms of video mean average precision (video-mAP).
3D-RetinaNet was proposed as the baseline for all three tasks.
Challenge participants had 18 videos available for training and validation. The remaining 4 videos were to be used to test the final performance of their model. 
This split was applied to all the three challenges (split 3 of the ROAD evaluation protocol, see Section \ref{road:subsec:exp:road}).

% timeline
The challenge opened for registration on April 1 2021, with the training and validation folds released on April 30, the test fold released on July 20 and the deadline for submission of results set to September 25. 
For each stage and each Task the maximum number of submissions was capped at 50, with an additional constraint of 5 submissions per day.
The workshop, co-located with ICCV 2021, took place on October 16 2021.

In the validation phase we had between three and five teams submit between 15 and 17 entries to each of three challenges. In the test phase, which took place after the summer, we noticed a much higher participation with 138 submissions from 9 teams to the agent challenge, 98 submissions from 8 teams to the action challenge, and 93 submission from 6 teams to the event detection challenge.

The methods proposed by the winners of each challenge are briefly recalled in Section \ref{road:sec:experiments-challenge}.

\emph{Benchmark maintenance}. After the conclusion of the ROAD @ ICCV 2021 workshop, the challenge has been re-activated to allow for submissions indefinitely. The ROAD benchmark will be maintained by withholding the test set from the public on the \url{eval.ai} platform\footnote{\url{https://eval.ai/web/challenges/challenge-page/1059/overview}}, where teams can submit their predictions for evaluation. Training and validation sets can be downloaded from \url{https://github.com/gurkirt/road-dataset}.

\section{ROAD-R}\label{road_r:sec:roadr}

ROAD-R extends the ROAD dataset by introducing a set  $\Pi$ of 243 constraints that specify the space of admissible outputs. In order to
improve the usability of our dataset, we write the constraints in a way that allows us to easily express $\Pi$ as a single formula in conjunctive normal form (CNF). The above can be done without any loss in generality, as any propositional formula can be expressed in CNF, and is important because many solvers expects formulas in CNF as input. Thus, each requirement in $\Pi$ has form: 
\begin{equation}\label{eq:req_lit}
l_1 \vee l_2 \vee \cdots \vee l_n\,, 
\end{equation}
where $n \geq 1$, and each $l_i$ is either a negative label $\neg A$ or a positive label~$A$. 
The requirements have been manually specified  following three steps: 
\begin{enumerate}
     \item 
an initial set of constraints $\Pi_1$ was manually created, 
     \item 
a subset $\Pi_2 \subset \Pi_1$ was retained by eliminating all those constraints that were entailed by the others,
     \item 
the final subset $\Pi \subset \Pi_2$ was retained by keeping only those requirements that were always satisfied by the ground-truth labels of the entire ROAD-R dataset.
\end{enumerate}

\begin{table}[t]
    \centering
    \caption{Constraint statistics. $\Pi_n$ is the set of constraints $r$ in $\Pi$ with $|r| = n$, i.e., with $n$ positive and negative labels. $\ov{\mathcal{C}} = \{\neg {A} : A \in \mathcal{C}\}$. Each row shows the number of rules $r$ with $|r|=n$, and the average number of negative and positive labels in such rules.}    
    \label{tab:req_stats1}
    \begin{tabular}{rrrr}
    \toprule
%\# labels per rule	&	 \# rules &	avg. neg. labels 	&	avg. pos. labels \\
$n$	&	$|\Pi_n|$ &	$\text{avg}_{r \in \Pi_n} (|r \cap \ov{\mathcal{C}}|)$ 	&	$\text{avg}_{r \in \Pi_n} (|r \cap \mathcal{C}|)$ \\
    \midrule
2	&	215	&	1.995	&	0.005 \\
3	&	5	&	1	&	2 \\
7	&	1	&	1	&	6\\
8	&	6	&	1	&	7\\
9	&	6	&	1	&	8\\
10	&	1	&	0	&	10\\
12	&	1	&	1	&	11\\
14	&	1	&	0	&	14\\
15	&	7	&	1	&	14\\
    \midrule
Total	&	243	&	1.87	&	0.96\\
    \bottomrule
    \end{tabular}

\end{table}

Finally, redundancy in the constraints has been automatically checked with {\sc relsat}\footnote{https://github.com/roberto-bayardo/relsat/}. 
Note that our process of gathering and further selecting the logical requirements follows more closely the software engineering paradigm rather than the machine learning view.
To this end, we ensured that the constraints were consistent with the provided labels from the ROAD dataset in the sense that they were acting as strict conditions to be absolutely satisfied by the ground-truth labels, as emphasized in the third step of the annotation pipeline above.
Table \ref{tab:req_stats1} gives a high-level description of the properties of the set $\Pi$ of constraints. Notice that, with a slight abuse of notation, in the tables we use a set based notation for the requirements. Each requirement of form (\ref{eq:req_lit}) thus becomes 
$$
\{l_1, l_2, \ldots, l_n\}.
$$ Such notation allows us to express the properties of the requirements in a more succinct way.
In addition to the information in the tables, we report that of the 243 constraints, there are two in which all the labels are positive (expressing that there must be at least one agent and that every agent but traffic lights has at least one location), and 214 in which all the labels are negative (expressing mutual exclusion between two labels).  All the constraints with more than two labels have at most one negative label, as they express a one-to-many relation
 between actions and agents (like \rev{in our dataset} ``if something is crossing, then it is a pedestrian or a cyclist'').
Constraints like ``pedestrians should cross at crossings'', which might not be satisfied in practice, are not included.
Additionally embedding such logical constraints would require, e.g., using modal operators and, while it would be an interesting study to see the impact on the model's predictions when adding more complex layers to the expressivity of our logic, we opted for using a simpler logic in this first instance.
This also provides more transparency to the wider research community, as the full propositional logic covers a vast range of applications that do not require extra logical operators. 
% The list with all the 243 requirements, with their natural language explanations, is  in Appendix~\ref{app:req_list},
% Tables~\ref{tab:req_list1}, \ref{tab:req_list2}, and \ref{tab:req_list3}.
Notice that the 243 requirements restrict the number of admissible prediction to $4985868 \sim 5 \times 10^6$,
thus ruling out $(2^{41} - 4985868) \sim 10^{12}$ non-admissible predictions.\footnote{The number of admissible predictions has been computed with relsat: https://github.com/roberto-bayardo/relsat/.}
In principle, the set of admissible predictions can be further reduced by adding other constraints. Indeed, the 243 requirements are not guaranteed to be complete from every possible point of view: as standard in the software development cycle,
the requirement specification process deeply involves the stakeholders of the system (see, e.g., \cite{sommerville}). For example, we decided not to include constraints like ``it is not possible to both move towards and move away", which were not satisfied by all the data points because of errors in the ground truth labels. In these cases, we decided to dismiss the constraint in order to maintain (i) consistency between the knowledge provided by the constraints and by the data points, and (ii) backward compatibility. 

As an additional point, we underline that, even though the annotation of the requirements introduces some overhead in the annotation process, it is also the case that the effort of manually writing 243 constraints (i) is negligible when compared to the effort of manually annotating the 22 videos, and (ii) can improve such last process, e.g., allowing to prevent errors in the annotation of the data points.
\section{Experimental Validation} \label{road:sec:exp}

In this section we present results on the various tasks the ROAD dataset is designed to benchmark (see Sec. \ref{road:sec:dataset-tasks}).

\subsection{Implementation Details} \label{road:subsec:details}

The results are evaluated in terms of both frame-level bounding box detection and tube detection. In the first case, the evaluation measure of choice is \emph{frame mean average precision} (f-mAP). We set the Intersection over Union (IoU) detection threshold to $0.5$ (signifying a 50\% overlap between the predicted and true bounding box). For the second set of results, we use \emph{video mean average precision} (video-mAP), as information on how the ground-truth BBs are temporally connected is available. These evaluation metrics are standard in action detection~\cite{Saha2016,Weinzaepfel-2015,kalogeiton2017action,singh2017online,li2018recurrent}.
\\
We also evaluate actions performed by AV, 
as described in \ref{road:sec:dataset-multilabel}. Since this is a temporal segmentation problem, we adopt the mean average precision metric computed at frame-level, as standard on the Charades~\cite{sigurdsson2018charadesego} dataset.

\begin{table}
    \centering
    \setlength{\tabcolsep}{4pt}
    \caption{Splits of training, validation and test sets for the ROAD dataset with respect to weather conditions. The table shows the number of videos in each set or split. For splits, the first figure is the number of training videos, the second number that of validation videos.}     \label{tab:splits} 
    {\footnotesize
    \scalebox{0.99}{
    \begin{tabular}{lcccc}
    \toprule
    Condition & sunny & overcast & snow & night \\ \midrule
    Training and validation & 7 & 7 & 1 & 3 \\
    \textit{Split-1} & 7/0 & 4/3 & 1/0 & 3/0 \\
    \textit{Split-2} & 7/0 & 7/0 & 1/0 & 0/3 \\
    \textit{Split-3} & 4/3 & 7/0 & 1/0 & 3/0 \\ \midrule
    Testing & 1 & 1 & 1 & 1 \\ \bottomrule
    \end{tabular}
    }
    }

\end{table}

We use sequences of $T = 8$ frames as input to 3D-RetinaNet. 
Input image size is set to $512\times682$. This choice of %the length of the input frame sequences 
$T$ is the result of GPU memory constraints; however, at test time, we unroll our convolutional 3D-RetinaNet for sequences of 32 frames, showing that it can be deployed in a streaming fashion.
We initialise the backbone network with weights pretrained on Kinetics~\cite{kay2017kinetics}. For training we use an SGD optimiser with step learning rate. The initial learning rate is set to $0.01$ and drops by a factor of $10$ after $18$ and $25$ epochs, up to an overall $30$ epochs. 

The parameters of the tube-building algorithm (Sec. \ref{road:subsec:tube-generation}) are set by cross validation. 
For ROAD we obtain $\lambda = 0.5$ and $k = 4$. 

\subsection{Experimental Results on ROAD} \label{road:subsec:exp:road}

\subsubsection{Three splits: modelling weather variability}~\label{road:subsec:splits}
For the benchmarking of the ROAD tasks, we divided the dataset into two sets. The first set contains 18 videos for training and validation purposes, while the second set contains 4 videos for testing, equally representing the four types of weather conditions encountered.

The group of training and validation videos is further subdivided into three different ways ('splits'). In each split, 15 videos are selected for training and 3 for validation. Details on the number of videos for each set and split are shown in Table~\ref{tab:splits}. 
All 3 validation videos for Split-1 are overcast; 4 overcast videos are also present in the training set. As such, Split-1 is designed to assess the effect of different overcast conditions.
Split-2 has all 3 night videos in the validation subset, and none in the training set. It is thus designed to test model robustness to day/night variations. Finally, Split-3 contains 4 training and 3 validation videos for sunny weather: it is thus designed to evaluate the effect of different sunny conditions, as camera glare can be an issue when the vehicle is turning or facing the sun directly.

Note that there is no split to simulate a bias towards snowy conditions, as the dataset only contains one video of that kind.
The test set (bottom row) is more uniform, as it contains one video from each environmental condition.

\subsubsection{Results on the various tasks} \label{road:sec:experiments-tasks}

Results are reported for the tasks discussed in Section~\ref{road:sec:dataset-tasks}. 

\emph{{Frame-level} results across the five detection tasks} are summarised in Table~\ref{tab:frame-avg-results} using the frame-mAP (f-mAp) metric, for a detection threshold of $\delta = 0.5$. 
The reported figures are averaged across the three splits described above, {in order to assess the overall robustness of the detectors to domain variations.}
Performance within each split is evaluated on both the corresponding validation subset and test set. 
Each row in the Table shows the result of a particular combination of backbone network (2D, I3D, or Slowfast) and test-time sequence length (in a number of frames, 8 and 32).
Frame-level results vary between 16.8\% (events) and 65.4\% (agentness) for I3D, and between 23.9\% and 69.2\% for Slowfast.

\emph{Video-level results} 
are reported in terms of video-mAP in Table~\ref{tab:video-avg-results}. As for the frame-level results, tube detection performance  (see Sec.~\ref{road:subsec:tube-generation}) is averaged across the three splits. One can appreciate the similarities between frame- and video-level results, which follow a similar trend albeit at a much lower absolute level. Again, results are reported for different backbone networks and sequence lengths.

\begin{table}[t!]
    %\vskip -3mm
    \centering
    \setlength{\tabcolsep}{4pt}
    \caption{Frame-level results (mAP $\%$) averaged across the three splits of ROAD. 
    The considered models differ in terms of backbone network (2D, I3D, and Slowfast) and clip length (08 vs 32). The performance of YOLOv5 on agent detection is also reported. Detection threshold $\delta = 0.5$.
    Both validation and test performance are reported for each entry. }    \label{tab:frame-avg-results} 
    {\footnotesize
    \scalebox{0.99}{
    \begin{tabular}{lccccc}
    \toprule
    Model & {Agentness}  & Agents & Actions & Locations &  {Events} \\ \midrule
    2D-08     & 51.8/63.4 & 30.9/39.5 & 15.9/22.0 & 23.2/30.8  & 10.6/12.8\\
    2D-32    & 52.4/64.2 & 31.5/39.8 & 16.3/22.6 & 23.6/31.4 &  10.8/13.0\\
    I3D-08    & 52.3/65.1 & 32.2/39.5 & 19.3/25.4 & 24.5/34.9 &  12.3/16.5\\
    I3D-32    & 52.7/65.4 & 32.3/39.2 & 19.7/25.9 & 24.7/35.3  & 12.6/16.8\\
    Slowfast-08 &   68.8/\textbf{69.2} & 41.9/47.5 & 26.9/31.1     & 34.6/\textbf{37.3} &  \textbf{18.1}/23.7\\
    Slowfast-32 &   \textbf{69.3}/68.7 &  42.6/43.7 & \textbf{27.3}/\textbf{31.7} & \textbf{34.8}/36.4 & 18.0/\textbf{23.9}\\
    YOLOv5 &  - & \textbf{57.9}/\textbf{56.9} &   - &  - & -\\
   \bottomrule
    \end{tabular}
    }
    }

\end{table}

\begin{table}[t!]
  \centering
  \setlength{\tabcolsep}{4pt}
  \caption{Video-level results (mAP $\%$) averaged across the three ROAD splits. 
  The models differ in terms of backbone network (2D, I3D and Slowfast) and test time clip length (08 vs 32). The performance of YOLOv5 on agent detection is also reported.
  Both validation and test performance are reported for each entry.}  \label{tab:video-avg-results} 
  % \vspace{-2mm}
  {\footnotesize
  \scalebox{0.99}{
  \begin{tabular}{lcccc}
  \toprule
  Model & Agents & Actions & Locations &  {Events}\\ \midrule
  \midrule
  \multicolumn{5}{l}{Detection threshold $\delta = 0.2$} \\
  \midrule
  2D-08     & 22.2/25.1 & 10.3/13.9 & 18.2/24.8 &  12.8/14.7\\
  2D-32    & 22.6/25.0 & 11.2/14.5 & 18.5/25.9 &  13.0/15.3\\
  I3D-08    & 23.2/26.5 & 14.1/15.8 & 20.8/25.8 &  14.9/17.4\\
  I3D-32    & 24.4/26.9 & 14.3/17.5 & 21.3/27.1 &  15.9/17.9\\
  Slowfast-08 & 24.1/{29.0} & \textbf{16.0}/\textbf{20.5} & 28.3/\textbf{33.0}  & 18.9/22.4\\
  Slowfast-32 & 24.2/28.6 & \textbf{16.0}/19.55 & \textbf{29.0}/29.7 & \textbf{19.1}/\textbf{22.5}\\
  YOLOv5 & \textbf{38.8}/\textbf{43.3} &  - & - & - \\
  \midrule
  \multicolumn{5}{l}{Detection threshold $\delta = 0.5$} \\ 
  \midrule
  2D-08     & 8.9/7.5 & 2.3/3.0 & 5.2/6.1 & 5.1/5.3\\ 
  2D-32    & 8.3/8.0 & 2.7/3.3 & 5.6/7.1 & 5.0/5.7\\ 
  I3D-08    & 9.2/9.6 & \textbf{4.0}/4.3 & 5.8/6.9 &  4.6/5.4\\ 
  I3D-32    & 9.7/10.2 & \textbf{4.0}/4.6 & 6.4/7.7 &  4.8/6.1\\ 
  Slowfast-08    & 7.1/8.9 & 3.9/\textbf{4.7} & 7.1/\textbf{11.0} & \textbf{6.5}/6.6\\ 
  Slowfast-32    & 8.3/9.8 & 3.7/4.4 & \textbf{8.4}/10.0 & 5.3/\textbf{7.3}\\ 
  YOLOv5 &  \textbf{18.7}/\textbf{13.9} &  - & - & - \\
  % \midrule
  \bottomrule
  \end{tabular}
  }
  }

\end{table}

\begin{table*}[t]
  \centering
  \setlength{\tabcolsep}{4pt}
  \caption{Number of video- and frame-level instances for each label (individual or composite), left. %of the three types: agent, action and location (left). 
  Corresponding frame-/video-level results (mAP@$\%$) for each of the three ROAD splits (right). {Val-$n$ denotes the validation set for Split $n$}. Results produced by an I3D backbone.}\label{tab:classwise-splitwise-primary-labels}  %[{\color{red}AB: Not clear what Val-1 Val-2 Val-3 are?}]
  % \vspace{-2mm}
  {\footnotesize
  \scalebox{0.65}{
  \begin{tabular}{lccccc | cccccc}
  \toprule
  & \multicolumn{5}{c}{Number of instance}  & \multicolumn{6}{c}{Frame-mAP@$0.5$/Video-mAP@$0.2$} \\ 
  \midrule
  Train subset & \multicolumn{5}{c}{\#Boxes/\#Tubes} & \multicolumn{2}{c}{Train-1}  & \multicolumn{2}{c}{Train-2}  & \multicolumn{2}{c}{Train-3} \\ 
  \midrule 
  Eval subset & All & Val-1 & Val-2  & Val-3 & Test  & Val-1 & Test  & Val-2 & Test  & Val-3 & Test \\ 
  \midrule
Agent & 559142/7029  & 60103/781    & 79119/761    & 83750/809    & 82465/1138   & 44.5/30.1    & 34.0/25.7    & 17.2/16.0    & 40.9/27.4    & 35.3/27.1    & 42.6/27.5    \\ 
 Action & 639740/9815  & 69523/1054   & 89142/1065   & 95760/1111   & 94669/1548   & 26.2/17.0    & 26.6/17.4    & 11.7/11.4    & 25.3/17.3    & 21.2/14.6    & 25.7/17.9    \\ 
 Location & 498566/8040  & 56594/851    & 67116/864    & 77084/914    & 70473/1295   & 34.9/28.6    & 35.2/26.4    & 13.7/12.1    & 33.9/26.3    & 25.4/23.2    & 36.7/28.6    \\ 
%  \midrule
%  \multicolumn{11}{l}{Agent-action detection (average across classes)} \\ 
%  \midrule
%  Total/Mean 
%  \midrule
%  \multicolumn{11}{l}{Event (triplet) detection (average across classes)} \\ 
%  \midrule
%  Total/Mean  
 Event & 453626/8394  & 43569/883    & 65965/963    & 72152/967    & 64545/1301   & 17.7/18.6    & 15.9/15.8    & 6.4/11.8     & 16.4/18.9    & 13.7/17.2    & 18.1/18.9    \\ 
 \midrule
 \multicolumn{6}{c}{Number of instances
 %AV-action segmentation
 } & \multicolumn{6}{c}{Frame-AP} \\ 
 \midrule
% Total/Mean  
 AV-action
 & 122154/490   & 17929/67     & 18001/56     & 16700/85     & 20374/82     & 57.9     & 45.7     & 33.5     & 43.6     & 43.7    & 48.2 \\
 \bottomrule
  \end{tabular}
  }
  }

\end{table*}

\emph{Streaming deployment}. Increasing test sequence length from 8 to 32 does not much impact performance. 
This indicates that, even though the network is trained on 8-frame clips, being fully convolutional (including the heads in the temporal direction), it can be easily unrolled to process longer sequences at test time, making it easy to deploy in a streaming fashion. 
Being deployable in an incremental fashion is a must for autonomous driving applications; this is a quality that other tubelet-based online action detection methods~\cite{kalogeiton2017action,singh2018tramnet,li2020actionsas} fail to exhibit, as they can only be deployed in a sliding window fashion. Interestingly, the latest work on streaming object detection~\cite{li2020towards} 
proposes an approach that integrates latency and accuracy into a single metric for real-time online perception, termed `streaming accuracy'. We will consider adopting this metric in the future evolution of ROAD.

\emph{Impact of the backbone}. Broadly speaking, the Slowfast~\cite{feichtenhofer2019slowfast} and I3D~\cite{carreira2017quo} versions of the backbone perform as expected, much better than the 2D version.
A Slowfast backbone can particularly help with tasks that require the system to `understand' movement, e.g. when detecting actions and road events, at least at 0.2 IoU. Under more stringent localisation requirements ($\delta = 0.5$), it is interesting to notice how Slowfast's advantage is quite limited, with the I3D version often outperforming it. \rev{This is due to the fact that the 3D-RetinaNet baseline model's choice of fixed-length input that may not fully leverage the strengths of the Slowfast backbone, which excels in scenarios requiring motion information and temporal context, such as video analysis. Therefore, any critique of the model should consider this limitation and its potential impact on capturing temporal dependencies and motion-related features.} Finally, even stronger backbones using transformers~\cite{liu2021video,fan2021multiscale} can be plugged in our 3D-RetinaNet baseline as backbone.

\emph{Level of task challenge}. The overall results on event detection (last column in both Table~\ref{tab:frame-avg-results} and Table~\ref{tab:video-avg-results}) are encouraging, but they remain in the low 20s at best, showing how challenging situation awareness is in road scenarios.

\emph{Comparison across tasks}. The headline figures are not really comparable since, as we know, the number of \rev{classes per task} varies. More importantly, within-class variability is often lower for composite labels. For example, the score for \textit{Indicating right} 
is really low, whereas \textit{Car / Indicating-right} 
has much better performance. This is because the within-class variability of the pair \textit{Car / Indicating-right} is much lower than that of \textit{Indicating right}, which puts together instances of differently-looking types of vehicles (e.g. buses, cars and vans) all indicating right.
Interestingly, results on agents are comparable among the four baseline models (especially for f-mAP and v-mAP at 0.2, see Tables \ref{tab:frame-avg-results} and \ref{tab:video-avg-results}).

\emph{YOLOv5 for Agent detection}. For completeness, we also trained YOLOv5\footnote{https://github.com/ultralytics/yolov5} for the detection of active agents. The results are shown in the last row of both Table~\ref{tab:frame-avg-results} and Table~\ref{tab:video-avg-results}. Keeping is mind that YOLOv5 is trained only on single input frames, it shows a remarkable improvement over the other baseline methods for active agent detection. We believe that is because YOLOv5 is better at the regression part of the detection problem -- namely, Slowfast has a recall of 71\% compared to the 94\% of YOLOv5, so that Slowfast has a 10\% lower mAP for active agent detection. We leave the combination of YOLOv5 for bounding box proposal generation and Slowfast for proposal classification as a promising future extension, which could lead to a general improvement across all tasks.

\emph{Validation vs test results}. Results on the test set are, on average, superior to those on the validation set. This is because the test set includes data from all weather/visibility conditions (see Table \ref{tab:splits}), whereas for each split the validation set only contains videos from a single weather condition. E.g., in Split 2 all validation videos are nighttime ones.

\subsubsection{Results under different weather conditions} \label{road:sec:experiments-weather}

Table~\ref{tab:classwise-splitwise-primary-labels} shows, instead, the results obtained under the three different splits we created on the basis of the weather/environmental conditions of the ROAD videos, discussed in Section~\ref{road:subsec:splits} and summarised in Table~\ref{tab:splits}. 
Note that the total number of instances (boxes for frame-level results or tubes for video-level ones) of the five detection tasks is comparable for all the three splits.

We can see how Split-2 (for which all three validation videos are taken at night and no nighttime videos are used for training, see Table~\ref{tab:splits})
has the lowest validation results, as seen in Table~\ref{tab:classwise-splitwise-primary-labels} (Train-2, Val-2). 
When the network trained on Split-2's training data is evaluated on the (common) test set, instead, its performance is similar to that of the networks trained on the other splits (see Test columns). 
Split-1 has three overcast videos in the validation set, but also four overcast videos in the training set. The resulting network has the best performance across the three validation splits. Also, under overcast conditions one does not have the typical problems with night-time vision, nor glares issues as in sunny days, as it has sunny videos in both train and validation sets. 

These results seem to attest a certain robustness of the baseline to weather variations, for no matter the choice of the validation set used to train the network parameters (represented by the three splits), the performance on test data (as long as the latter fairly represents a spectrum of weather conditions) is rather stable.

\subsubsection{Results of AV \rev{temporal action detection}} \label{road:sec:experiments-av}

Table~\ref{tab:classwise-av-actions} shows the results of using 3D-RetinaNet to \rev{temporally detect AV-action} classes, averaged across all three splits on both validation and test set. As we can see, the results for classes \textit{AV-move} and \textit{AV-stop} are very good, we think because these two classes are predominately present in the dataset. The performance of the 'turning' classes is reasonable, but the results for the bottom three classes 
are really disappointing. We believe this is mainly due the fact that the dataset is very heavily biased (in terms of number of instances) towards the other classes. 
As we do intend to further expand this dataset in the future by including more and more videos, we hope the class imbalance issue can be mitigated over time. A measure of performance weighing mAP using the number of instances per class could be considered, but this is not quite standard in the action detection literature. At the same time, ROAD provides an opportunity for testing methods designed to address class imbalance. 

\begin{figure*}[ht!]
\centering{
    \includegraphics[width=1.0 \textwidth]{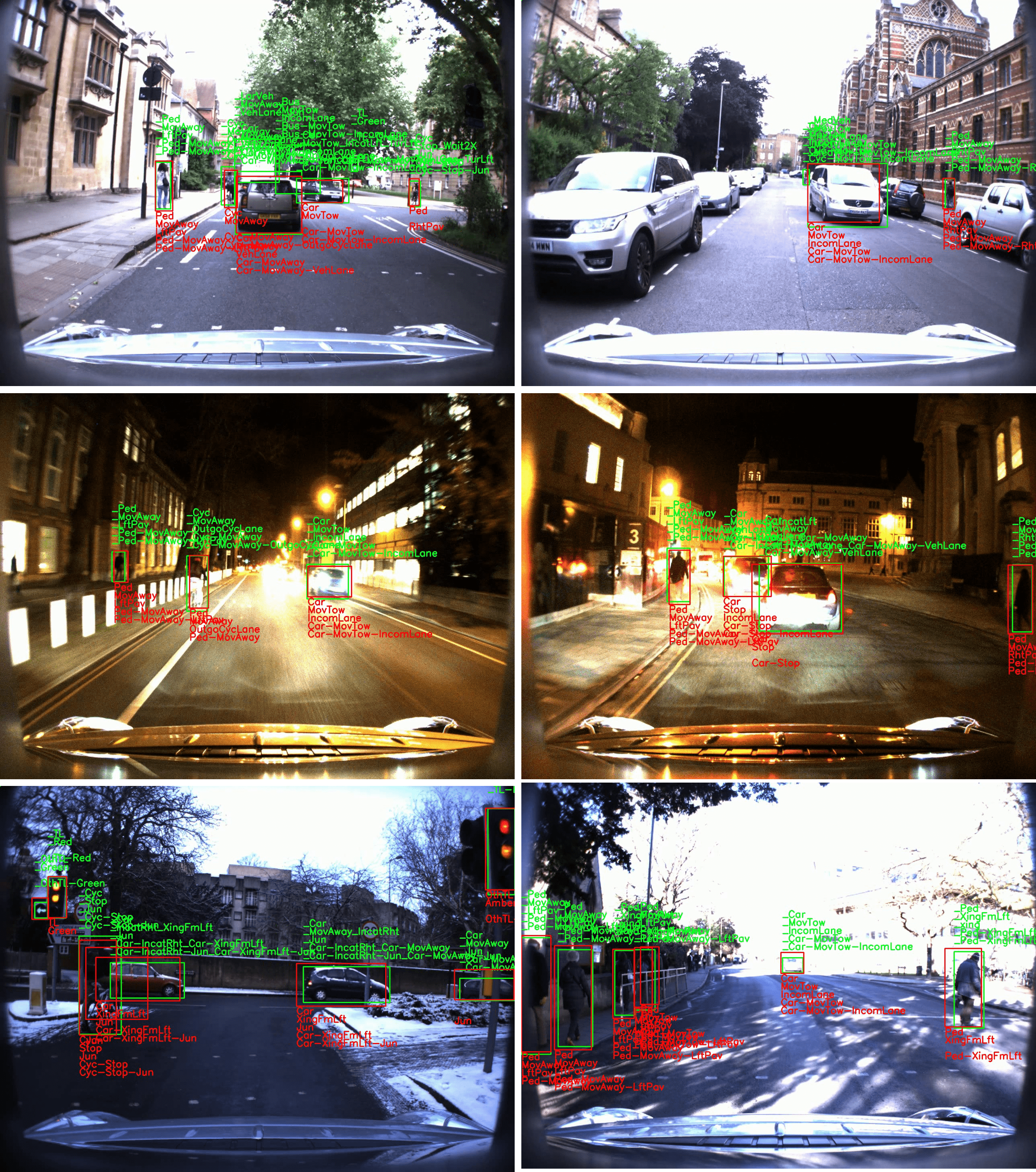}
\caption{Success cases in which our model detects the actions and locations correctly, and only for those agents which are active. Ground truth bounding boxes and labels are shown in green, while the predictions of our model are shown in red. 
}
\label{fig:Success}
}
\end{figure*}

\begin{figure*}[ht!]
\centering{
    \includegraphics[width=1.0 \textwidth]{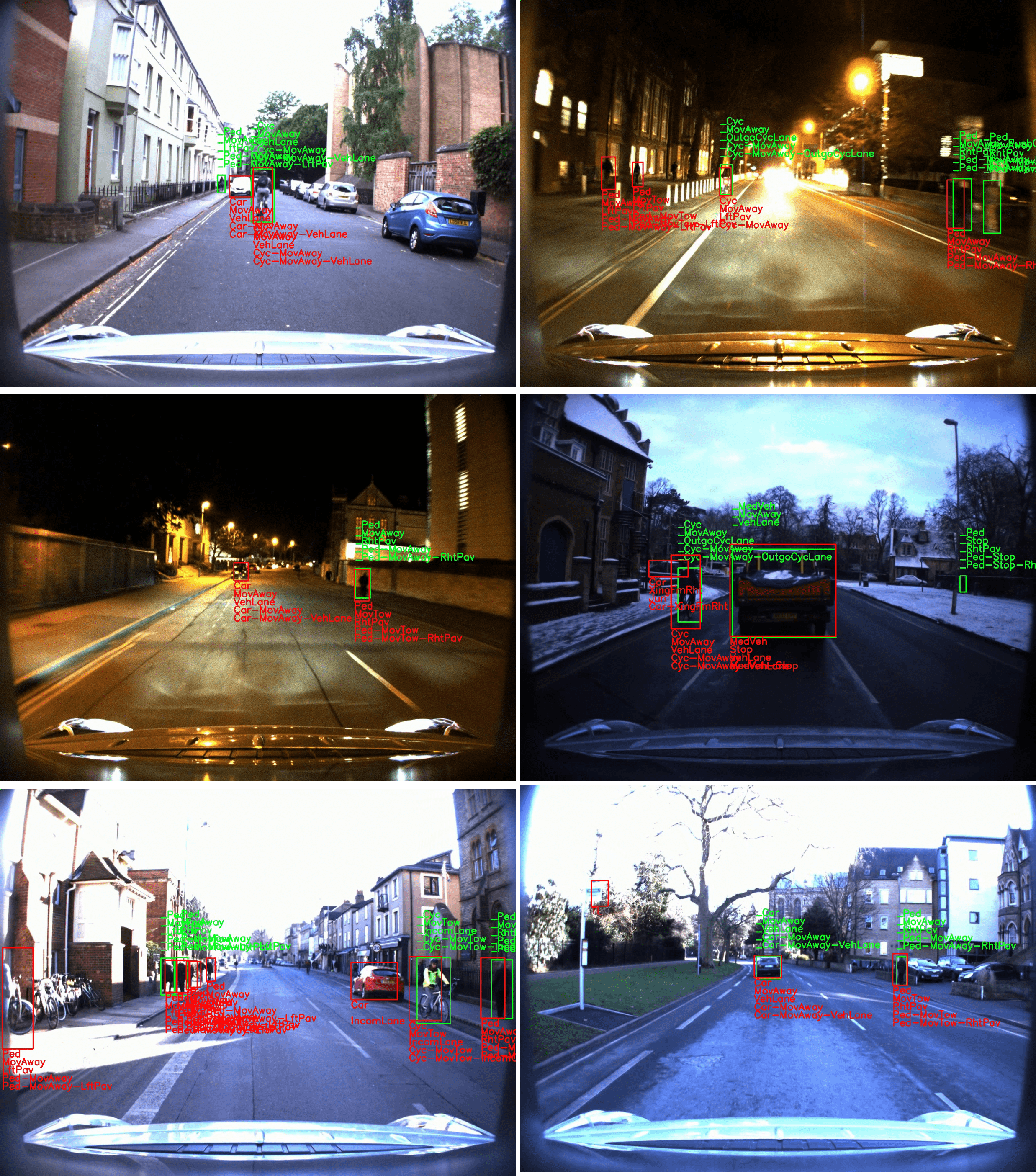}
\caption{Some of the failure modes of our model, shown detecting  inactive agents and/or assigning incorrect action and location labels. Ground truth bounding boxes and labels are shown in green, while the predictions of our model are shown in red.
}
\label{fig:failure}
}
\end{figure*}

\subsection{Qualitative Results}

Finally, we provide some qualitative results of our baseline model in terms of success and failure modes. Cases in which the baseline work accurately are illustrated in Figure~\ref{fig:Success}, where the model is shown to detect only those agents which are active (i.e., are performing some actions) and ignore all the inactive agents (namely, parked vehicles). Agent prediction is very stable across all the examples, whereas action and location prediction show some weakness in some case: for instance, the night-time example in the second row of the second column, where both the cars in front are moving away in the outgoing lane but our method fails to label their location correctly.

In contrast, the failure modes illustrated in Figure~\ref{fig:failure} are cases in which the model fails to assign to agents the correct label, and also detects agents which are not active (e.g. often parked cars, see the white vehicle in the top row, first column or the red vehicle in the third row, first column).

\begin{table}[t]
  %\vskip -3mm
  \centering
  \setlength{\tabcolsep}{4pt}
  \caption{AV \rev{temporal action detection} results (frame mAP$\%$) averaged across all three splits. 
  }\label{tab:classwise-av-actions} 
  {\footnotesize
  \scalebox{0.99}{
  \begin{tabular}{lccccc}
  \toprule
  & \multicolumn{1}{c}{No instances}  & \multicolumn{4}{c}{Frame-mAP@$0.5$ %/Video-mAP@$0.2$
  } \\ 
  \midrule
  Model & \multicolumn{1}{c}{} & \multicolumn{2}{c}{I3D} & \multicolumn{2}{c}{2D}
  \\ 
  \midrule 
  Eval subset & All & Val & Test  & Val & Test \\ 
  % \midrule
  % \multicolumn{5}{l}{Duplex results} \\ 
 \midrule
  Av-move                    & 81196/233    & 92.0     & 96.6  & 83.0     & 87.8     \\ 
  Av-stop                   & 31801/108    & 92.2     & 98.5  & 65.3     & 68.4     \\ 
  Av-turn-right                 & 3826/50      & 46.1     & 63.0  & 35.0     & 57.7     \\ 
  Av-turn-left                 & 3787/56      & 69.0     & 59.8  & 55.1     & 42.9     \\ 
  Av-overtake                  & 599/12       & 4.9      & 1.1   & 2.7      & 2.5      \\ 
  Av-move-left                 & 537/15       & 0.5      & 0.8   & 0.5      & 0.5      \\ 
  Av-move-right                 & 408/16       & 10.5     & 0.6   & 4.0      & 2.0      \\ 
  Total/Mean                & 122154/490   & 45.0     & 45.8  & 35.1     & 37.4     \\ 
 \bottomrule
  \end{tabular}
  }
  }

\end{table}

\subsection{Challenge Results} \label{road:sec:experiments-challenge}

Table~\ref{tab:challenge_results} compares the results of the top teams participating in our ROAD @ ICCV 2021 challenge with those of the Slowfast and YOLOv5 baselines, at a tube detection threshold of 0.2. The challenge server remains open at \url{https://eval.ai/web/challenges/challenge-page/1059/overview}, where one can consult the latest entries.

\begin{table}[t]
    \centering
    \caption{Results (in video-mAP) of the winning entries to the ICCV 2021 ROAD challenge compared with the Slowfast and YOLOv5 baselines, at a detection threshold of $0.2$.}
    \label{tab:challenge_results}
    \begin{tabular}{llccc}
        \toprule
         Task & Top team & Slowfast & YOLOv5 & Winners \\ \midrule
         Agent detection & Xidian & 29.0 & 43.3 & \textbf{52.4} \\
         Action detection & CMU-INF & 20.5 & - & \textbf{25.6} \\
         Event detection & IFLY & 22.4 & - & \textbf{24.7} \\
         \bottomrule
    \end{tabular}
\end{table}

\emph{Agent detection}. The agent detection challenge was won by a team formed by Chenghui Li, Yi Cheng, Shuhan Wang, Zhongjian Huang, Fang Liu of Xidian University, with an entry using YOLOv5 with post-processing. In their approach, agents are linked by evaluating their similarity between frames and grouping them into a tube. Discontinuous tubes are completed through frame filling, using motion information. Also, the authors note that YOLOv5 generates some incorrect bounding boxes, scattered in different frames, and take advantage of this by filtering out the shorter tubes. As shown in Table \ref{tab:challenge_results}, the postprocessing applied by the winning entry significantly outperforms our off-the-shelf implementation of YOLOv5 on agent detection.

\emph{Action detection}. The action detection challenge was won by Lijun Yu, Yijun Qian, Xiwen Chen, Wenhe Liu and Alexander G. Hauptmann of team CMU-INF, with an entry called ``ArgusRoad: Road Activity Detection with Connectionist Spatiotemporal Proposals", based on their Argus++ framework for real-time activity recognition in extended videos in the NIST ActEV (Activities in Extended Video ActEV) challenge\footnote{\url{https://actev.nist.gov/}.}. The had to adapt their system to be run on ROAD, e.g. to construct tube proposals rather than frame-level proposals. The approach is a rather complex cascade of object tracking, proposal generation, activity recognition and temporal localisation stages \cite{liu2020argus}. Results show a significant (5\%) improvement over the Slowfast baseline, which is close to state-of-the-art in action detection, but still at a relatively low level (25.6\%)

\emph{Event detection}. The event detection challenge was won by team IFLY (Yujie Hou and Fengyan Wang, from the University of Science and Technology of China and IFLYTEK). The entry consisted in a number of amendments to the 3D-RetinaNet baseline, namely: bounding box interpolation, tuning of the optimiser, ensemble feature extraction with RCN, GRU and LSTM units, together with some data augmentation. Results show an improvement of above 2\% over Slowfast, which suggests event better performance could be achieved by applying the ensemble technique to the latter.

\subsection{Experimental Results of ROAD-R}
\label{road_r:sec:sota}

As a first step, we ran 6 SOTA temporal feature learning architectures as part of a 3D-RetinaNet model~\cite{gurkirt2021} (with a 2D-ConvNet backbone made of Resnet50~\cite{he2016deep}) for event detection and evaluated 
to which extent constraints are violated. We considered:

\begin{enumerate}
\item {\sl 2D-ConvNet} (C2D)~\cite{wang2018non}:
     a Resnet50-based architecture with an additional temporal dimension for learning features from videos. The extension from 2D to 3D is done by adding a pooling layer over time to combine the spatial features.
\item {\sl Inflated 3D-ConvNet} (I3D)~\cite{carreira2017quo}:
      a sequential learning architecture extendable to any SOTA image classification model (2D-ConvNet based), 
    able to learn continuous spatio-temporal features from the sequence of frames.
\item {\sl Recurrent Convolutional Network} (RCN)~\cite{singh2019recurrent}:
     a 3D-ConvNet model that relies on recurrence for learning the spatio-temporal features at each network level. During the feature extraction phase, RCNs exploit both 2D convolutions across the spatial domain and
    1D convolutions across the temporal domain.  
\item {\sl Random Connectivity Long Short-Term Memory} (RCLSTM) \cite{hua2018traffic}:
     an updated version of LSTM in which the neurons are connected in a stochastic manner, rather than fully connected. In our case, the LSTM cell is used as a bottleneck in Resnet50 for learning the features sequentially.
\item {\sl Random Connectivity Gated Recurrent Unit} (RCGRU)  \cite{hua2018traffic}:
    an alternative version of RCLSTM where the GRU cell is used instead of the LSTM one. GRU makes the process more efficient with fewer parameters than the LSTM.
\item {\sl SlowFast} \cite{feichtenhofer2019slowfast}: a 3D-CNN architecture that contains both slow and fast pathways for extracting the sequential features. A Slow pathway computes the spatial semantics at low frame rate while a Fast pathway processes high frame rate for capturing the motion features. Both of the pathways are fused in a single architecture by lateral connections.
\end{enumerate}
We trained 3D-RetinaNet\footnote{https://github.com/gurkirt/3D-RetinaNets.} using the same hyperparameter settings for all the models: (i) batch size equal to 4, (ii)~sequence length equal to 8, and (iii) image input size equal to $512\times682$. All the models were initialized with the  Kinetics pre-trained weights. An SGD optimizer~\cite{lecun2012} with step learning rate was used. The initial learning rate was set to 0.0041 for all the models except SlowFast, for which it was set to 0.0021 due to the diverse nature of slow and fast pathways. All the models were trained for 30 epochs and the learning rate was made to drop by a factor of 10 after 18 and 25 epochs. The machine used for the experiments has 64 CPUs (2.2 GHz each) and 4 Titan RTX GPUs having 24 GB of RAM each. 

\begin{figure*}[t]

\begin{subfigure}[b]{0.32\textwidth}
         \centering
         \includegraphics[width=\textwidth,height=5.4cm, trim={50pt 30pt 50pt 10pt},clip]{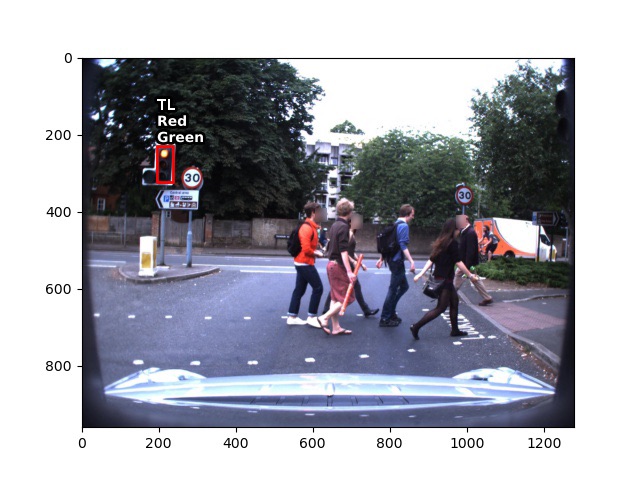}
         \caption{I3D}
         \label{I3D1}
\end{subfigure}
\begin{subfigure}[b]{0.32\textwidth}
         \centering
         \includegraphics[width=\textwidth, height=5.4cm, trim={20pt 20pt 50pt 10pt},clip]{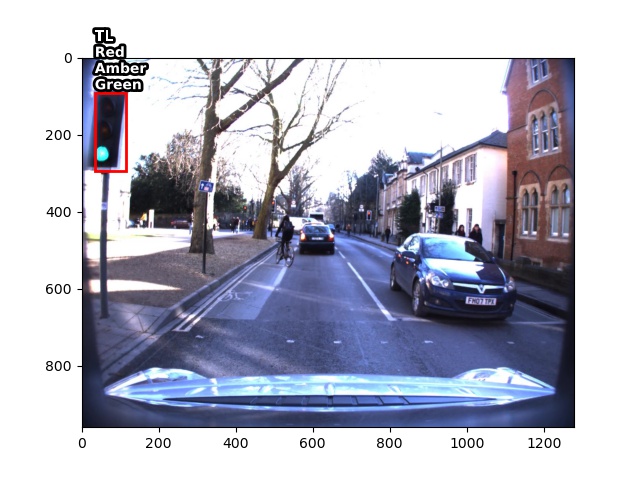}
         \caption{C2D}
         \label{C2D1}
\end{subfigure}
\begin{subfigure}[b]{0.32\textwidth}
         \centering
         \includegraphics[width=\textwidth, height=5.4cm, trim={20pt 20pt 50pt 10pt},clip]{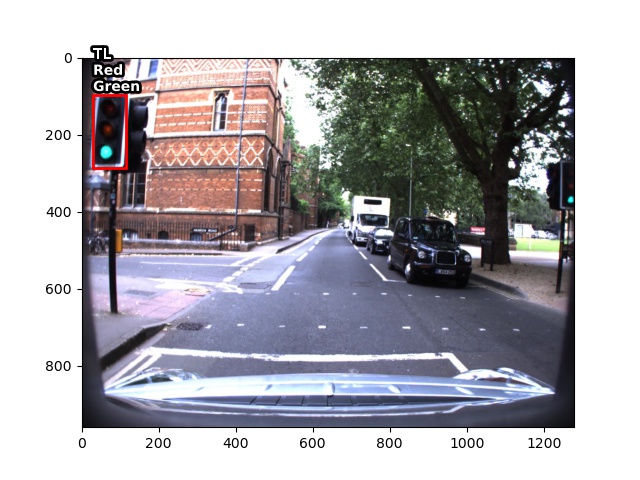}
         \caption{RCGRU}
         \label{RCGRU1}
\end{subfigure}

\begin{subfigure}[b]{0.32\textwidth}
         \centering
         \includegraphics[width=\textwidth, height=5.4cm, trim={40pt 20pt 40pt 10pt},clip]{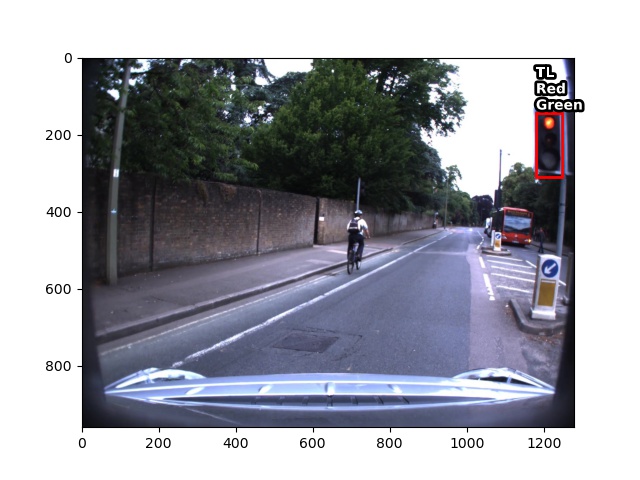}
         \caption{RCLSTM}
         \label{RCLSTM1}
\end{subfigure}
\begin{subfigure}[b]{0.32\textwidth}
         \centering
         \includegraphics[width=\textwidth, height=5.4cm, trim={20pt 20pt 50pt 10pt},clip]{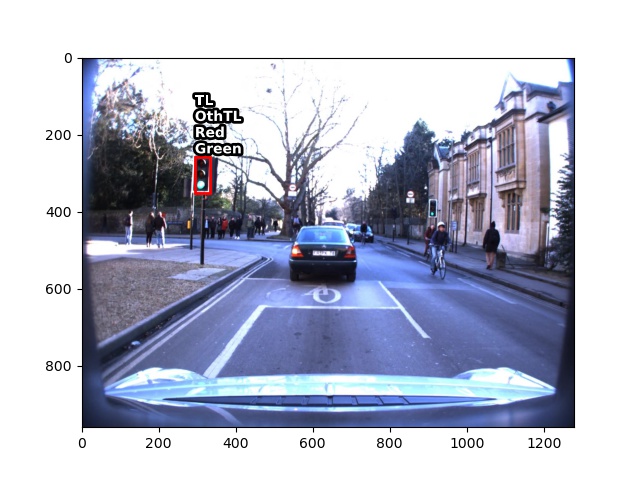}
         \caption{RCN}
         \label{RCN1}
\end{subfigure}
\begin{subfigure}[b]{0.32\textwidth}
         \centering
         \includegraphics[width=\textwidth,  height=5.4cm, trim={20pt 20pt 50pt 10pt},clip]{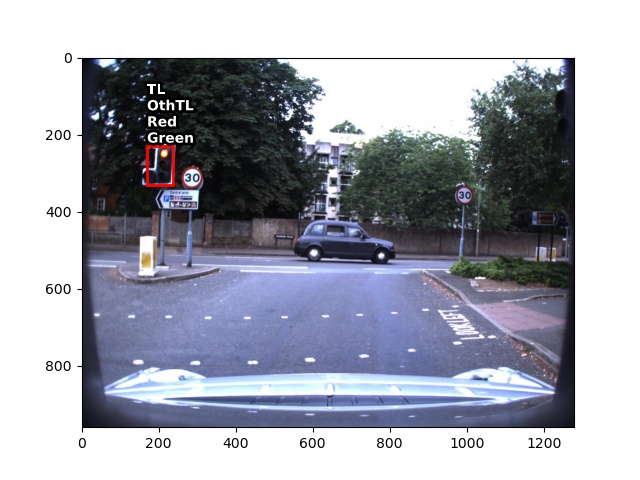}
         \caption{SlowFast}
         \label{Slowfast}
\end{subfigure}
\caption{Examples of violations of $\{\neg{\text{RedTL}},\neg{\text{GreenTL}}\}$.}
\label{fig:violRedTL}
\end{figure*}

To measure the models' performance, we used the {\sl frame mean average precision} (f-mAP), which is the standard metric used for action detection (see, e.g.,~\cite{kalogeiton2017a,li2018map}), with IoU threshold equal to 0.5 and 0.75, indicated as f-mAP@0.5 and f-mAP@0.75,  respectively. 
The results for the SOTA models at IoU threshold 0.5 and 0.75 are reported in Table \ref{tab:map5}, 
column ``SOTA''.

 To measure the extent to which each system violates the constraints, we used the following metrics: 
 \begin{itemize}
     \item the percentage of non-admissible predictions,
     \item the average number of violations committed per prediction, and
     \item  the percentage of constraints violated at least once,
 \end{itemize}
while varying the threshold $\theta$ from 0.1 to 0.9 with step 0.1. 
The results are in Fig.~\ref{fig:sota}, where (to improve readability) we do not plot the values corresponding to $\theta=0.0$ and $\theta=1.0$. For $\theta=0.0$ (resp., $\theta = 1.0$), all the predictions are positive (resp., negative), and thus the corresponding values are (in order) 100\%, 214, and 214/243 (resp., 100\%, 2, and 2/243).

\begin{figure*}
\centering
    \begin{subfigure}[t]{0.32\textwidth}
    \includegraphics[width=0.98\textwidth,trim={0 0 7cm 7cm}]{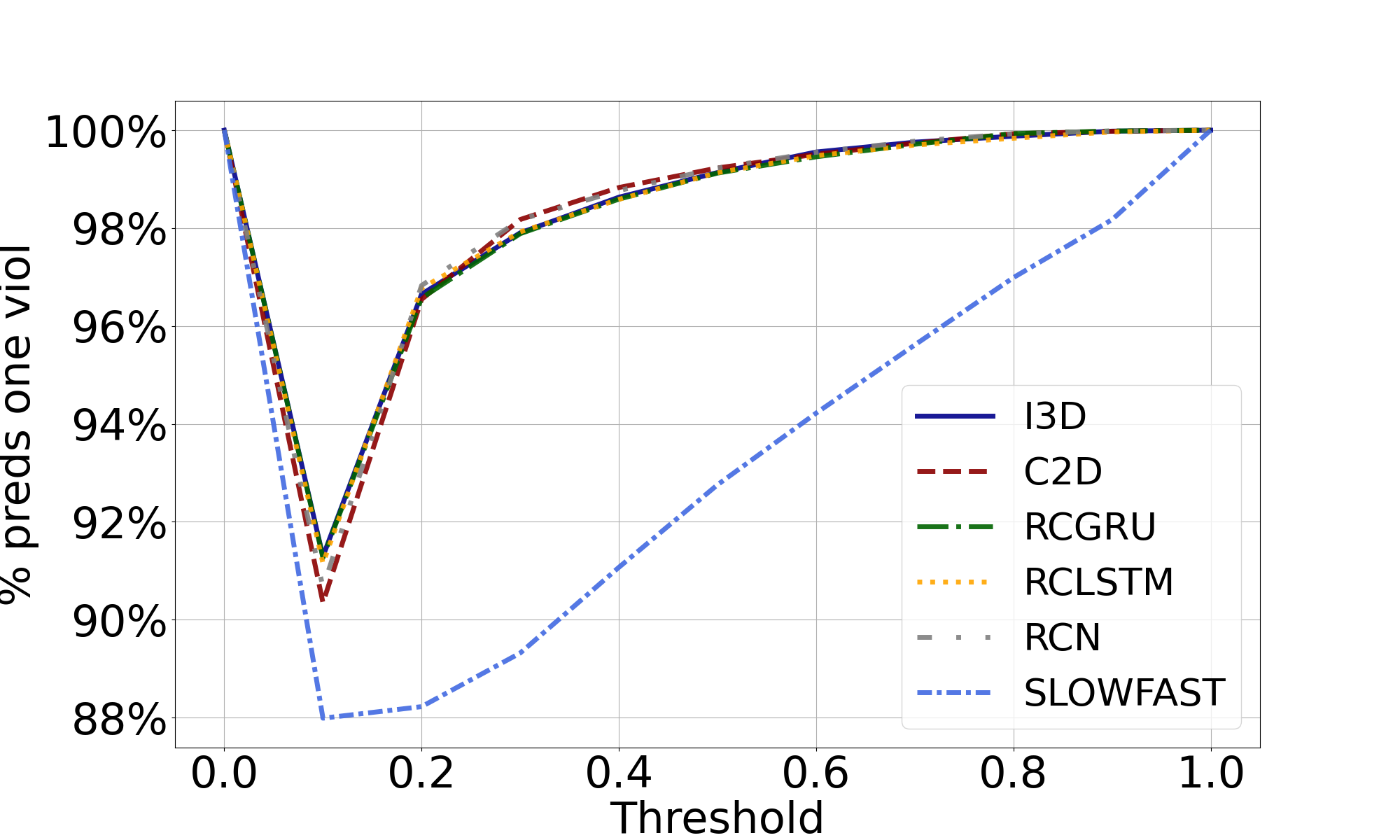}
    \caption{Percentage of predictions violating at least one constraint.} % for $\theta \in [0.1,0.9]$.}
    \label{fig:sota-viola}
    \end{subfigure}
    \hfill
    \begin{subfigure}[t]{0.32\textwidth}
    \includegraphics[width=0.98\textwidth,trim={0 0 7cm 7cm}]{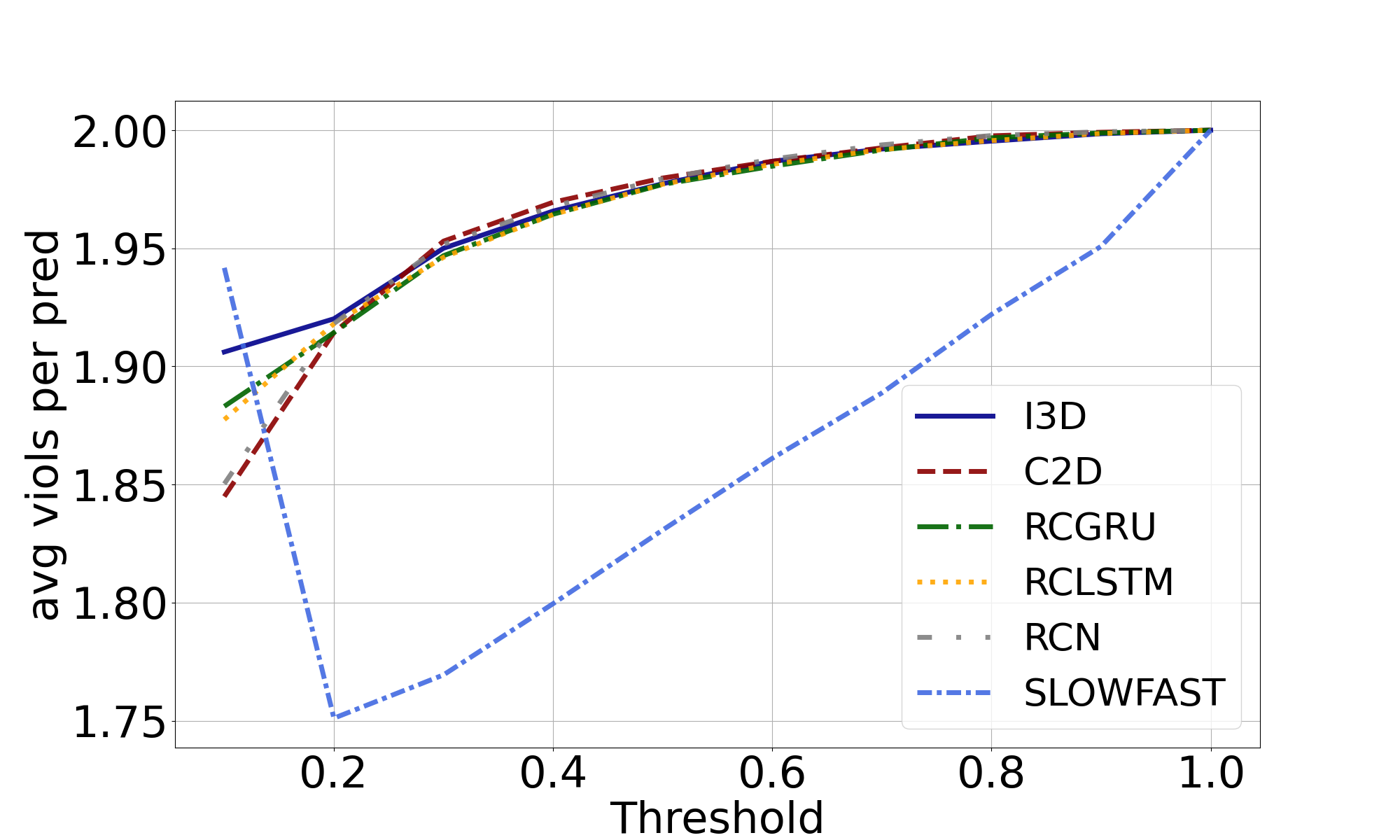}
    \caption{Average number of violations committed per prediction.} % for $\theta \in [0.1,0.9]$.}
    \label{fig:sota-violb}
    \end{subfigure}
    \hfill
    \begin{subfigure}[t]{0.32\textwidth}
    \includegraphics[width=0.98\textwidth,trim={0 0 7cm 7cm}]{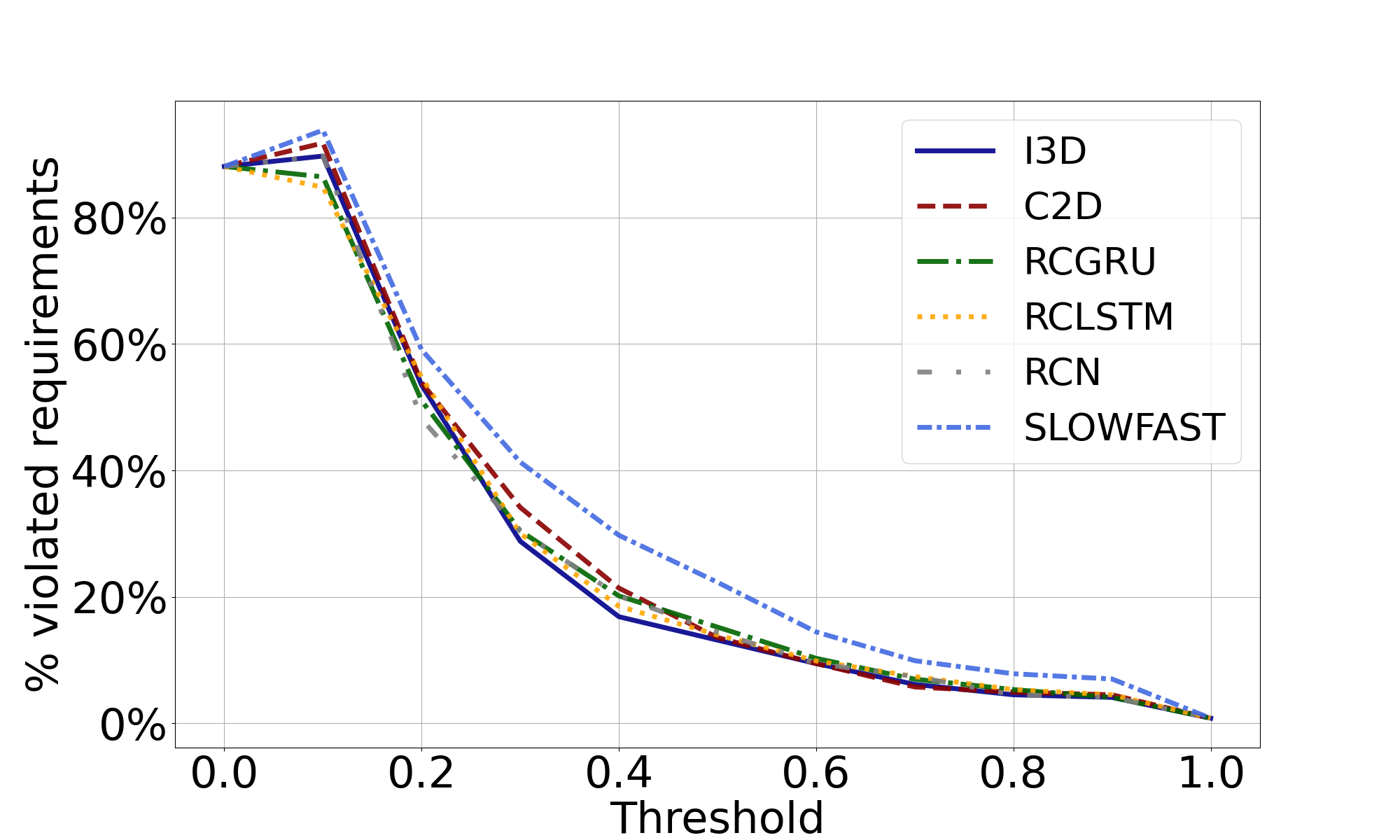}
    \caption{Percentage of constraints violated at least once.} % for $\theta \in [0.1,0.9]$.}
    \label{fig:sota-violc}
    \end{subfigure}
    \caption{ROAD-R and SOTA models. In the $x$-axis, there is the threshold $\theta \in [0.1,0.9]$, step 0.1.}\label{fig:sota}
\end{figure*}

% Consider the results in Table \ref{tab:map5}, 
% column ``\SOTA'', and in Fig.~\ref{fig:sota}. 
% First, note that the performances are not an indicator of the ability of the model to satisfy the constraints. Indeed, higher f-mAPs do not correspond to lower trends in the plots of Fig.~\ref{fig:sota-violb}. For example,
% RCGRU performs better than C2D for both IoU = 0.5 and IoU = 0.75, however, its curve is above C2D's in both Figs.~\ref{fig:sota-viola} and~\ref{fig:sota-violb}.
% Then, note that the percentage of non-admissible predictions is always very high for every model: at its minimum, for $\theta=0.1$, more than 90\% of the predictions are non-admissible, and this percentage reaches 99\% for $\theta = 0.9$ (see Fig.~\ref{fig:sota-viola}). In addition, most predictions violate roughly two constraints, as shown  by Fig.~\ref{fig:sota-violb}. 
% Considering that we are in an autonomous vehicle setting, such results are critical: one of the constraints that is  violated  
% by all the baseline models
% is $\{\neg{\text{RedTL}}, \neg{\text{GreenTL}}\}$, corresponding to predictions according to which there is a traffic light with both the red and the green lights on. Fig.~\ref{fig:violRedTL} shows an image for each of the SOTA models where such a prediction (for $\theta=0.5$) is made.  
% % Appendix~\ref{app:imgs} contains qualitative examples of all the SOTA models making  predictions violating other constraints. 

\begin{table*}[t]
\setlength{\tabcolsep}{6.4pt}
    \centering
    \caption{f-mAP@0.5 and f-mAP@0.75 for the current SOTA models.}
    \label{tab:map5}
    \footnotesize
    \scalebox{0.99}{
    \begin{tabular}{l|c c}
    \toprule
     & \multicolumn{2}{c}{{\textbf{IoU threshold}}} \\
         \textbf{Model}  & \textbf{0.5} & \textbf{0.75} \\
         \midrule
         C2D & 27.57 & 12.66 \\
         I3D & 30.12 & 16.29 \\
         RCGRU &  30.78 &  15.98 \\
         RCLSTM & 30.49 & 16.64 \\
         RCN & 29.64 & 15.82 \\
         SlowFast & 28.79 & 14.40 \\
    \bottomrule
\end{tabular}
    }

\end{table*}

\section{Further Extensions} \label{road:sec:extensions}

By design, ROAD is an open project which we expect to evolve and grow over time.

\emph{Extension to other datasets and environments}. 
In the near future, we will work towards completing the multi-label annotation process for a larger number of frames coming from videos spanning an even wider range of road conditions.
Further down the line, we plan to extend the benchmark to other cities, countries and sensor configurations, to slowly grow towards an even more robust, 'in the wild' setting.
In particular, we will initially target the Pedestrian Intention Dataset (PIE, \cite{rasouli2019pie}) and Waymo \cite{sun2020scalability}. 
The latter one comes with spatiotemporal tube annotation for pedestrian and vehicles, much facilitating the extension of ROAD-like event annotation there.

\emph{Event anticipation/intent prediction}.
ROAD is an oven-ready playground for action and event anticipation algorithms, a topic of growing interest in the vision community \cite{kong2017deep,kong2018adversarial}, as it already provides the kind of annotation that
allows researchers to test predictions of both future event labels and future event locations, both spatial and temporal.
Anticipating the future behaviour of other road agents is crucial to empower the AV to react timely and appropriately. The output of this Task should be in the form of one or more future tubes, with the scores of the associated class labels and the future bounding box locations in the image plane \cite{singh2018predicting}.
We will shortly propose a baseline method for this Task, but we encourage researchers in the area to start engaging with the dataset from now.

\emph{Autonomous decision making}.
In accordance with our overall philosophy,
we will design and share a baseline for AV decision making from intermediate semantic representations.
The output of this Task should be the decision made by the AV in response to a road situation \cite{hubmann2017decision}, represented as a collection of events as defined in this chapter. As the action performed by the AV at any given time is part of the annotation, the necessary meta-data is already there. Although we did provide a simple temporal segmentation baseline for this task seen as a classification problem, we intend in the near future to propose a baseline from a decision making point of view, making use of the intermediate semantic representations produced by the detectors.

\emph{Machine theory of mind} \cite{rabinowitz2018machine} refers to the attempt to provide machines with (limited) ability to guess the reasoning process of other intelligent agents they share the environment with. Building on our efforts in this area \cite{Cuzzolin2020tom}, we will work with teams of psychologists and neuroscientists to provide annotations in terms of mental states and reasoning processes for the road agents present in ROAD. Note that theory of mind models can also be validated in terms of how close the predictions of agent behaviour they are capable of generating are to their actual observed behaviour.
Assuming that the output of a theory of mind model is intention (which is observable and annotated) the same baseline as for event anticipation can be employed.

\emph{Continual event detection}.
ROAD's conceptual setting is intrinsically incremental, one in which the autonomous vehicle keeps learning from the data it observes, in particular by updating the models used to estimate the intermediate semantic representations. The videos forming the dataset are particularly suitable, as they last 8 minutes each, providing a long string of events and data to learn from.
To this end, we plan to set a protocol for the continual learning of event classifiers and detectors and propose ROAD as the first continual learning benchmark in this area \cite{parisi2019continual}.

\rev{
\emph{Alternatives to ROAD-R}.
Throughout our ROAD-R, we used logical constraints to satisfy the requirements and avoid the prediction of impossible scenarios. The main intuition behind using logical constraints instead of probabilistic priors is that probabilistic priors do not give any guarantee that the constraints will be satisfied. In ROAD-R, the constraints we employed predominantly pertained to impossible scenarios. This means that the majority of our constraints are rooted in expressing physical impossibilities. This distinction between impossible and improbable scenarios is pivotal when considering alternative probabilistic methods. In the future, we are aiming to consider alternative probabilistic methods that can effectively handle both impossible and improbable scenarios. This reflection can enhance the robustness and applicability of our research findings, ensuring that they are not limited solely to situations of physical impossibility but also account for highly improbable yet impactful events. By doing so, we aim to provide a more comprehensive and versatile framework for decision-makers and stakeholders who rely on our research to navigate uncertain real-world situations.}

\section{Summary of the Chapter} \label{road:sec:conclusion}

This chapter proposed a strategy for situation awareness in autonomous driving based on the notion of road events, and contributed a new ROad event Awareness Dataset for Autonomous Driving (ROAD) as a benchmark for this area of research. The dataset, built on top of videos captured as part of the Oxford RobotCar dataset~\cite{maddern20171}, has unique features in the field. Its rich annotation follows a multi-label philosophy in which road agents (including the AV), their locations and the action(s) they perform are all labelled, and road events can be obtained by simply composing labels of the three types. The dataset contains
22 videos with 122K annotated video frames, for a total of 560K detection bounding boxes associated with 1.7M individual labels.

Baseline tests were conducted on ROAD using a new 3D-RetinaNet architecture, as well as a Slowfast backbone and a YOLOv5 model (for agent detection). Both frame--mAP and video--mAP were evaluated. 
Our preliminary results highlight the challenging nature of ROAD, with the Slowfast baseline achieving a video-mAP on the three main tasks comprised between 20\% and 30\%, at low localisation precision (20\% overlap). YOLOv5, however, was able to achieve significantly better performance. These findings were reinforced by the results of the ROAD @ ICCV 2021 challenge, and support the need for an even broader analysis, while highlighting the significant challenges specific to situation awareness in road scenarios.

Our dataset is extensible to a number of challenging tasks associated with situation awareness in autonomous driving, such as event prediction, trajectory prediction, continual learning and machine theory of mind, and we pledge to further enrich it in the near future by extending ROAD-like annotation to major datasets such as PIE and Waymo.

We also proposed a new dataset for the task of learning with requirement, called ROAD-R. We showed that SOTA models most of the time violate the requirements, and how it is possible to exploit the requirements to create models that are compliant with (i.e., strictly satisfy) the requirements while improving their performance.

    \pagestyle{plain}
    \cleardoublepage
\phantomsection
\renewcommand{\pname}{Part II : Complex Activity Detection} \label{part:part2}
\addcontentsline{toc}{chapter}{\pname}\label{partII}

\pagebreak
\hspace{14pt}
\vfill
\begin{center}
\textbf{\pname}
\end{center}
\vfill
\hspace{0pt}
\pagebreak
    
    \pagestyle{fancy}
    \chapter{Spatiotemporal Deformable Scene Graphs for Complex Activity Detection}
\label{chapter:deformsgraph}

\section{Introduction} 
\label{deformsgraph:intro}

In the second part of the thesis, we are taking one step ahead of ROAD event detection presented in the previous chapter (Chapter \ref{chapter:road}) and address a novel problem of Complex Activity Detection. In this chapter, long-term complex activity recognition and localisation framework is presented and can be crucial for decision-making in autonomous systems such as smart cars and surgical robots. Here we address the problem via a novel deformable, spatiotemporal scene graph approach, consisting of three main building blocks: (i) action tube detection, (ii) the modelling of the deformable geometry of parts, and (iii) a graph convolutional network. Firstly, action tubes are detected in a series of snippets. Next, a new 3D deformable RoI pooling layer is designed for learning the flexible, deformable geometry of the constituent action tubes. Finally, a scene graph is constructed by considering all parts as nodes and connecting them based on different semantics such as order of appearance, sharing the same action label and feature similarity. We also contribute fresh temporal complex activity annotation for the recently released ROAD autonomous driving and SARAS-ESAD surgical action datasets and show the adaptability of our framework to different domains. Our method is shown to significantly outperform graph-based competitors on both augmented datasets.

\subsection{Motivation}
Complex activity recognition is attracting much attention in the computer vision research community due to its significance for a variety of real-world applications, such as autonomous driving \cite{caesar2020nuscenes,camara2020pedestrian}, surveillance \cite{liang2019peeking}, medical robotics \cite{zia2018surgical} or team sports analysis \cite{hu2020progressive}. In autonomous driving, for instance, it is extremely important that the vehicle understands dynamic road scenes, in order, e.g., to accurately predict the intention of pedestrians and forecast their trajectories to inform appropriate decisions. 
In surveillance, group activities rather than actions performed by individuals need to be monitored.
Robotic assistant surgeons need to understand what the main surgeon is doing throughout a complex surgical procedure composed of many short-term actions and events \cite{singh2021saras}, in order to suitably assist them.

\rev{
Prior to the advent of deep learning techniques, the study of \emph{atomic} and \emph{complex} activities was a vibrant research area that encompassed a range of methodologies \cite{laxton2007leveraging,ryoo2006recognition}. The fundamental works from the 1990s and the early 2000s focused on understanding and modeling these activities through various approaches such as grammars \cite{ivanov2000recognition}, Bayesian networks \cite{park2004hierarchical}, and graph-based recognition \cite{wolf2001smart}. Researchers explored the decomposition of complex actions into fundamental \emph{atomic actions}, which were basic and indivisible components of more intricate behaviors. These foundational works laid the groundwork for understanding how activities are composed and structured, providing insights into the underlying patterns and dependencies. By analyzing the relationships between atomic actions and their compositions, researchers were able to formulate models that captured the essence of human actions and interactions. This chapter revisits the concepts of \emph{atomic} and \emph{complex} activities within the framework of the present era, building upon the legacy of past research while also addressing the novel challenges and opportunities introduced by deep learning techniques.
}

Recent methods for action or activity recognition and localisation  can be broadly divided into two categories; single atomic action \cite{long2020learning,shi2020weakly,gong2020learning,xu2020g} and multiple atomic action recognition/localisation \cite{huang2020improving,wu2019long,zhao2019hacs,sudhakaran2019lsta,luo2019grouped,ji2020action}. The former methods only focus on identifying the start and end of an action performed in a short video portraying a single instance, leveraging datasets such as UCF-101 \cite{soomro2012ucf101} or Charades \cite{sigurdsson2016hollywood}. The latter set of approaches consider videos which contain a number of atomic actions 
or multiple repetitions of the same action. Methods in this category do address complex activity recognition, as their aim is to understand an overall, dynamic scene by detecting and identifying its constituent components. Datasets used for complex activity detection are Epic-kitchens \cite{damen2018scaling}, THUMOS14 \cite{idrees2017thumos} or ActivityNet v1.3 \cite{caba2015activitynet}.
Both classes of methods are geared towards merely recognising and localising short term action or activities that lasts for only a few frames or seconds.

\subsection{Objective}
Unlike all existing methods, in this chapter, we present a framework capable of recognising \emph{complex, long-term activities}, validated in {the fields of} autonomous driving and surgical robotics but of general applicability and extendable to other domains.
More precisely, by `complex activity' we mean \emph{an ensemble of `atomic' actions, each performed by an individual agent present in the environment, which extends over a period of time and which collectively has a meaning}. \rev{ These activities can be contiguous, occurring sequentially without any gap or background (in the case of surgical robotics). Alternatively, these activities may occur at different positions within the video, with background segments interspersed between them (in the case of autonomous driving).} For instance, in the autonomous driving context, one can define the complex activity \emph{Negotiating intersection} as composed of the following atomic actions: an autonomous vehicle (AV) moves along its lane; the traffic light regulating the vehicle's lane turns red, while the light turns green for a traversing road; a number of other vehicles pass through the intersection from both sides; the AV's light turns green again; the AV resumes moving and crosses the intersection.
\\
The proposed pipeline (Fig. \ref{fig:deformsgraph_framework}) is divided into three parts: (i) action tube detection, (ii) part-based feature extraction and learning via 3D deformable RoI pooling, and {(iii) a graph generation strategy to process a variable number of parts and their connections, aimed at learning the overall semantics of a dynamic scene representing a complex activity}. Action tube detection is a necessary pre-processing step, aimed at spatially and temporally locating %detecting the spatiotemporal location of all 
the atomic actions present \cite{de2014online,jetley20143d,singh2017online,saha2017amtnet,behl2017incremental,singh2018predicting,saha2016deep}. 
Note that the tube detector needs to ensure a fixed-size representation for each activity part (atomic action). 
%Towards detection of action tubes with same constraints, we used an existing 
Here, in particular, we adopt
%action tubes detector called 
AMTNet \cite{saha2020two}, as the latter describes action tubes of any duration using a fixed number of bounding box detections.

\begin{figure*}[h]
    \centering
    \includegraphics[width=0.99\textwidth]{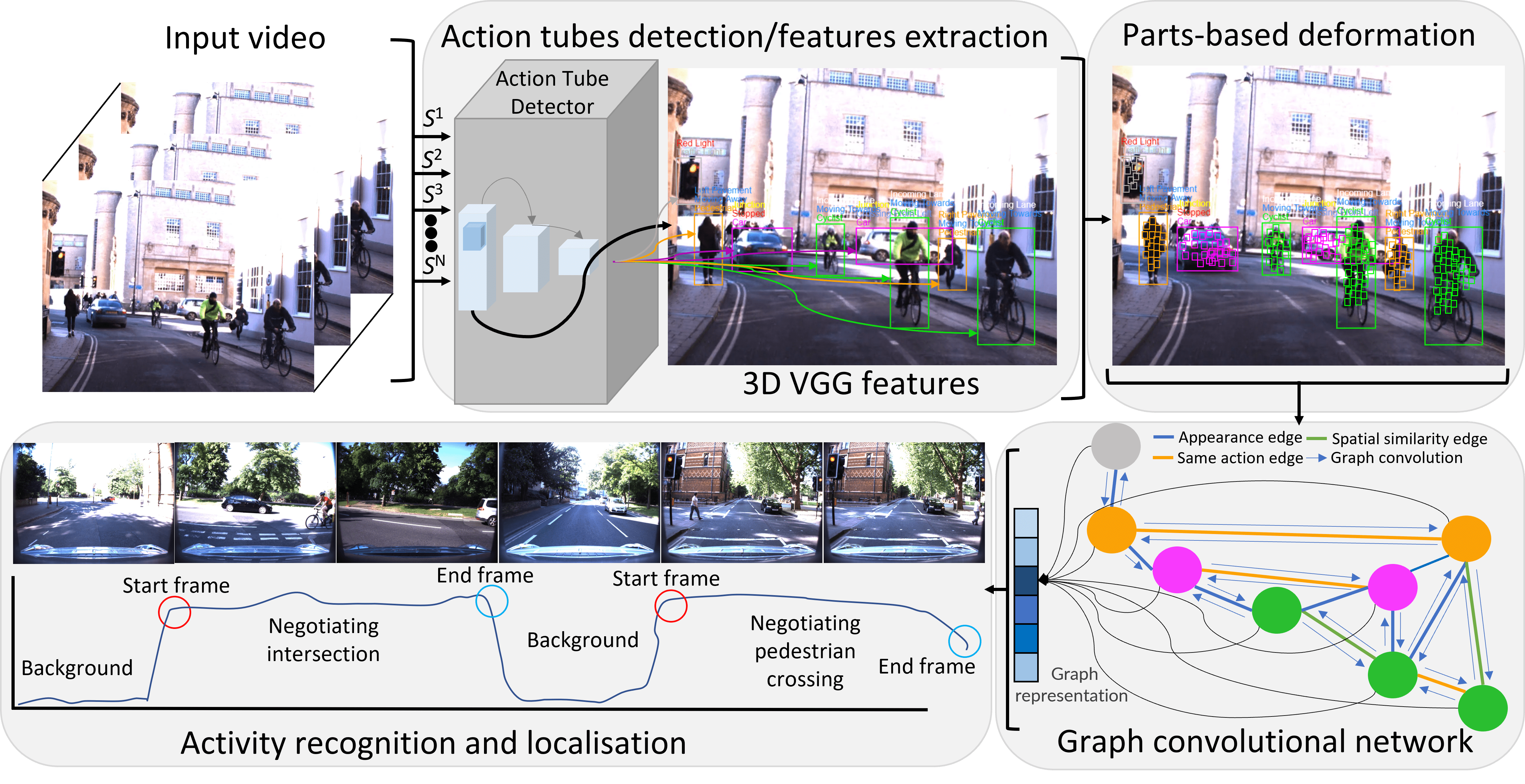}
    \caption{Overall pipeline of our {long-term} complex activity detection framework. (i) The input video is first divided into snippets. (ii) Snippets are passed to an action tube detector module one by one. (iii) A part-based deformation module receives 3D VGG features and action tube locations and returns  features for the salient (non-background) parts of each tube instance. (iv) {A GCN module represents activity parts (action tubes) as nodes with the features generated by ROI pooling and builds edges with different semantics to construct a spatiotemporal graph representation. Finally (v), the graph representation features produced by GCN inference on the consolidated graph are used for temporal activity detection.}} 
    \label{fig:deformsgraph_framework}
\end{figure*}

\subsection{Main Contributions}
Our contribution is twofold. Firstly, our novel 3D deformable RoI pooling layer, inspired by standard deformable and modulated RoI pooling \cite{dai2017deformable,zhu2019deformable}, %Unlike existing deformable models, our %part-based deformation module 
is not only designed to work with 3D data but is also capable of learning feature representations for tubes of variable spatiotemporal shape.
%spatial shape in a fixed temporal sequence. 
{Secondly, an original Graph Convolutional Network (GCN) module constructs a graph by 
%taking each of the part features individually and consider 
considering individual tubes as nodes and connecting them via edges encoding diverse semantics, namely: appearance order, sharing of the same action label, and spatial feature similarity. The spatiotemporal scene graph so constructed is then processed by a stack of graph convolutional layers resulting in graph representation features, which are used to train a classifier for recognising complex activities, followed by a localisation stage which uses a sliding window approach.}

The framework is evaluated using two {real-world datasets springing from} completely different domains: ROAD \cite{singh2021road} for situation awareness in autonomous driving and SARAS-ESAD \cite{bawa2020esad} for surgical action detection, both providing video-level annotation in the form of (atomic) action tubes. %The main reason behind using these datasets are the availability of action bounding boxes.
In this work we augment these datasets with suitable annotation on the start and end time of each instance of complex activity (road activities in ROAD vs surgical phases in SARAS-ESAD).
The main contributions of this work are therefore:

\begin{itemize}
\item 
A novel framework for long-term complex activity recognition and localisation. % in diverse environments.

\item 
An original deformable 3D RoI pooling approach for flexibly pooling features from the various components of the detected tubes, to create an overall representation for activity parts.

\item 
{A spatiotemporal scene graph generation and processing mechanism able to cope with a variable number of parts while learning the overall semantics of an activity class.}

\item 

Augmented annotation for two newly-released datasets 
%in autonomous driving and robotic surgery, 
aimed at making them suitable benchmarks for future work on complex activity detection.
%We annotate two datasets temporally from completely different domains i.e., autonomous driving and surgical robotics for evaluation of our framework in diverse scenarios.  
%\item  {\color{blue} here we need to summarise our experimental findings---Sal: Do we need an extra contribution for that?}
%\textcolor{red}{Reimplement and outperformed state-of-the-art methods over our augmented datasets.}
\end{itemize}
{Our results clearly indicate that the detection task (both at atomic and complex activity level) is extremely challenging on the real-world data which forms these newly annotated benchmarks, when compared to existing academic datasets. We hope this will stimulate further original thinking to address these challenges}.
Our method is shown to clearly outperform two recent state of the art graph-based competitors \cite{zeng2019graph,xu2020g} on both augmented datasets.

\paragraph{Related publications:}
The work presented in this chapter is published in BMVC 2021 ~\cite{skhan2021comp}. 
The dissertation author was the primary investigator in ~\cite{skhan2021comp}.
Our code is also available online\footnote{\url{https://github.com/salmank255/SDSG_Complex_Activity}}.

\paragraph{Outline:}\label{online:outline}
This chapter is organised as follows.
Firstly, we start by presenting the proposed methodology of our spatiotemporal deformable scene graphs framework in Section ~\ref{deformsgraph:sec:overview}. The proposed methodology is divided into building blocks including Action Tube Detection,  3d Part-Bsed Deformable Layer, and Graph Convolutional Network described in Section \ref{deformsgraph:ssec:atd}, Section \ref{deformsgraph:ssec:3dd}, and Section \ref{deformsgraph:ssec:gcn}, respectively. Next, the detailed experimental evaluation of our proposed method is reported in Section \ref{deformsgraph:sec:exp}. Finally, a summary with limitations of the proposed method is given in Section~\ref{deformsgraph:ssec:summary}.

\section{Proposed Method}\label{deformsgraph:sec:overview}

% \subsection{Motivation}
Whereas modern complex activity recognition and localisation methods focus on single action or collections of short-term atomic actions, the issue of detecting complex activities articulated into a number atomic actions and spanning longer intervals of time is still rather unexplored. 

Crucial to the identification of complex video activities is the modelling of the relations among the constituent actions. Deformable models that represented activities as graphs of atomic parts provide such a framework. The existing part-based methods are investigated for action detection and prediction tasks, but their potential for the recognition of \emph{long-term} activities has not been explored so far.

In this chapter, we propose to achieve this via a combination of deformable pooling of features and a spatiotemporal graph representation that employs multiple semantics. The workflow of our approach is illustrated in Figure ~\ref{fig:deformsgraph_framework}. The input video is divided into fixed-sized snippets which are fed to an action tube detector, followed by a novel 3D deformable RoI pooling stage which computes parts features, which are later processed by GCN for activity detection.

\subsection{Action Tube Detection}\label{deformsgraph:ssec:atd}

To provide a fixed-size representation for the instances of atomic actions composing a complex activity,
%To detect action tubes of fixed size, we utilized an existing action tubes detector termed as 
we adopt AMTnet \cite{saha2020two}. AMTnet is a two-stream online action tube detector that uses both RGB and optical flow information (although here we only use the RGB stream). 
%in which we only adopt the RBG steam for action tubes detection. 
The main rationale for using AMTnet is that it generates tubes in an incremental manner while preserving a fixed-size representation. % as required by our feature extraction stage. %with the fixed size of frames that we need in our next feature extraction module.

\textbf{Architectural Details}. AMTNet uses VGG-16 \cite{simonyan2014very} as baseline CNN feature extractor. The last two fully-connected layers of VGG-16 are replaced by two convolutional layers, and add four extra convolutional layers at the end. % FAB: check if this is correct-Sal:confirmed
%For finding the spatio-temporal features and correlation, the 
AMTNet takes sequence of RGB frames as an input with a fixed temporal interval $\triangle$ between consecutive frames, i.e., $\{ {f}_t, {f}_{t + \triangle} \}$. The input to AMTNet is in the format [$BS \times {Sq} \times {D} \times {H} \times {W}$], where ${BS}$ is the training batch size, ${Sq}$ is the sequence length (in this case a pair), $D$ is the dimensionality (equal to 3 as we are dealing with RGB frames), while ${H}$ and ${W}$ are the height and width of each frame ($300 \times 300$ in our case). As typical in action detection, AMTNet uses both a classification and a regression layer for recognition and detection, respectively, with the goal of predicting action `micro-tubes' defined by pairs of consecutive detections. %which are combined with interpolation.
The method predicts bounding boxes for a pair of frames separated by fixed gap $\triangle$, while the bounding boxes for intermediate frames are generated by interpolation. In this work, atomic action instances are represented as 3 micro-tubes with $\triangle$ = 3 for an overall tube length of $L = 12$ frames, aligned with our snippet length.
%The next most important aspect of AMTNet is incremental action tube generation from the micro-tubes predicted by the architecture as illustarted in Figure ~\ref{fig:AMTNet}. 
Complete action tubes are incrementally generated by AMTNet by temporally linking the micro-tubes predicted by the network. (see Figure ~\ref{fig:AMTNet}).

\begin{figure*}[h]
    \centering
    \includegraphics[width=0.99\textwidth]{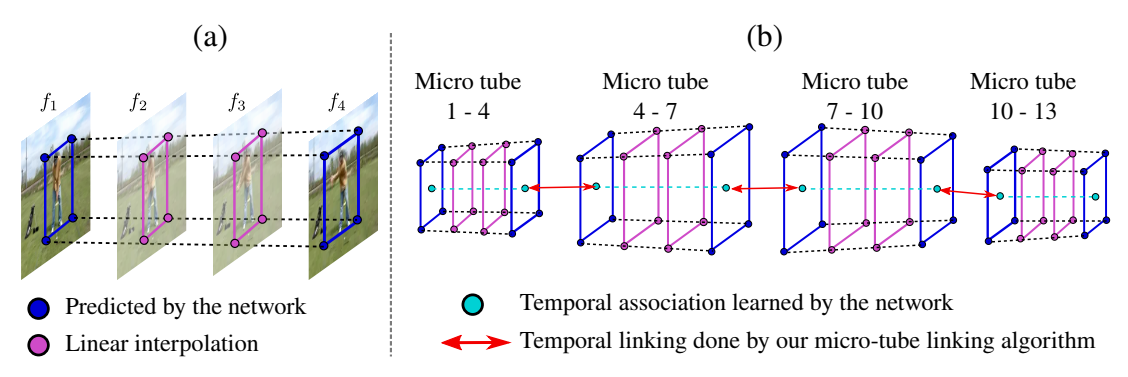}
    \caption{(a)
    AMTNet \cite{saha2020two} generates micro-action tubes  %represents the micro action tube generation where 
    by predicting bounding boxes for the start and the end frame of a snippet. Predictions for a fixed number of
    intermediate frames are constructed by bilinear interpolation. (b) Complete action tubes are created by temporally linking micro action tubes by dynamic programming. 
    % {\color{blue} [Fig 2 is nice but if short of space we can remove it]}
    }
    \label{fig:AMTNet}
\end{figure*}

\subsection{3D Part-Based Deformable Layer}\label{deformsgraph:ssec:3dd}

The feature extractor in our framework is a novel 3D deformable RoI pooling layer which encodes the spatiotemporal geometry of the action tubes which correspond to the activity parts. This is an extension of the existing standard deformable RoI pooling layer \cite{dai2017deformable} that has the ability to extract and learn features from an action tube rather than a 2D bounding box. 
{ The rationale behind using the 3D deformable RoI pooling layer is that it allow us to learn at training time how the %geometric transformation of each atomic action during the
geometric shape of atomic action tubes (as regions of the video considered as a spatiotemporal volume) varies across instances of the same class.
%at training time rather than using the feature of pretrained model.
Intuitively, the shape of the bounding box detections forming a tube (and therefore the shape of the tube itself) will vary with e.g. the viewpoint, as well as the particular style with which the action is performed by a certain agent (for instance, a cyclist can turn right making a narrower as opposed to a wider turn).
} 

The principle of our 3D deformable RoI pooling operation is shown in Figure ~\ref{fig:droi}. Like the classical deformable RoI pooling layer, our module also includes standard RoI pooling (used in all region proposal-based object detection methods), a fully connected layer, and {\emph{offsets} which encode the amount of geometric deformation. 
Firstly, standard RoI pooling is applied to the provided feature map $X$ and bounding boxes locations forming an action tube ($L$ $\times$ [$x$,$y$,$w$,$h$]), by subdividing the tube into a pooled feature map grid of fixed-size in both the spatial and the temporal dimensions: $L$ $\times$ $k$ $\times$ $k$. Here $L = 12$ is the fixed action tube length,
%fixed during action tube detection
while $k$ is a free parameter which determines the `bin size', i.e., the number of grid locations detections are divided into in each spatial dimension (see Figure \ref{fig:droi} again). % in RoI pooling. 
Next, for each bin in the grid, normalised offsets (representing the degree of deformation of the grid components of each action tube) are generated for these feature maps using a fully-connected layer, which are then transformed using an element-wise product with the original RoI's width and height. Offsets are also multiplied by a scalar value to modulate their magnitude (empirically set to 0.1), making them invariant to the different possible sizes of the RoI. In our framework, this layer takes the VGG features extracted by AMTNet and each detected action tube separately as an input and returns an overall feature map which encodes both the appearance and the shape (through the above offsets) of each atomic action.}

\begin{figure}
    \centering
    \includegraphics[width=0.95\textwidth]{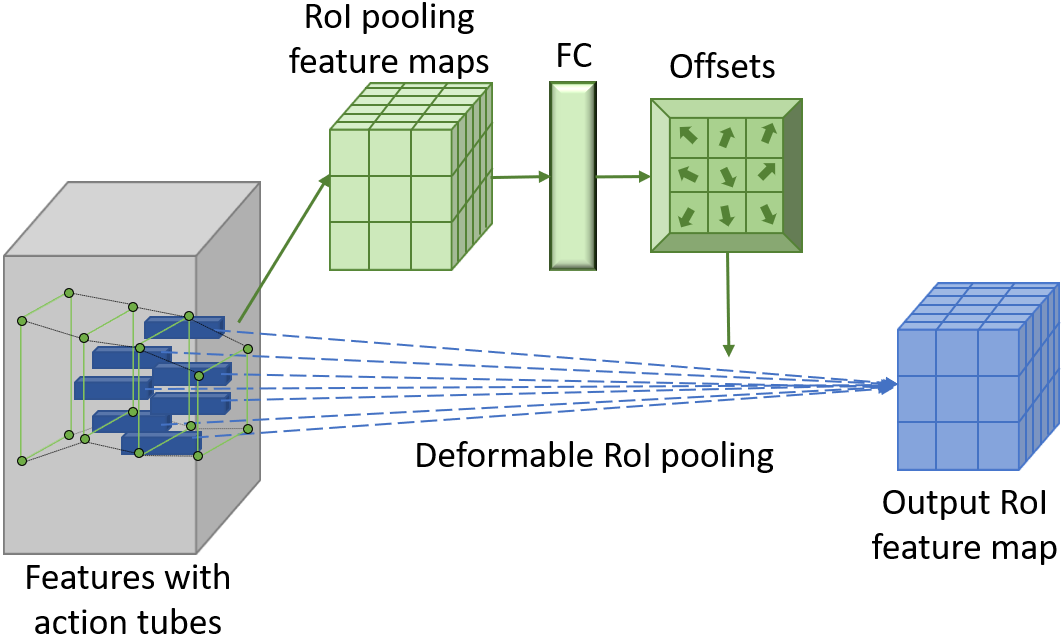}
    \caption{Our 3D deformable RoI pooling layer takes feature maps and action tube locations as input and arranges them into a fixed-size grid of components (here illustrated for size $3\times 3$). For each grid component an offset is generated and multiplied by the original tube feature to produce the final component features.}
    \label{fig:droi}
\end{figure} 

\subsection{Graph Convolutional Network}\label{deformsgraph:ssec:gcn}

As our purpose is to achieve a comprehensive understanding of the dynamic scene which comprises a complex activity,
%Of the features extracted by our 3D part-based deformable layer, need to be learn for understanding the context of overall scene representing the complex activity. Thus, in our approach we use GCN 
we propose to use a graph convolutional network to model and exploit the relations between the constituent action tubes. %contributing to the main activity. 
%The main rational behind using GCN is that small atomic actions and their coordination play a vital role in decision making and highly depend on each other. Therefore, these actions and their connections are represented in a graph structure. 
Unlike the tree structure of classical part-based models (which requires to fix the number of parts \cite{chen2018part}), %FAB: please add reference here 
(spatiotemporal) graphs allow us to flexibly describe a complex activity composed by a variable number of actions (nodes) of different type, and to encode the different semantic relationships between them.
The functioning of our GCN module is illustrated in Figure ~\ref{fig:gcn}.

\textbf{Graph Construction}. 
{When constructing the activity graphs, the input video is subdivided into consecutive, non-overlapping, fixed-length snippets.}
For each snippet, a separate graph is built with a variable number of nodes corresponding to the number of detected activity parts (action tubes) { within the snippet}. The initial representation of the nodes is provided by their RoI features. 
%while the edges as their connections. In our case, 
We consider three different types of connections: (i) the \emph{order of appearance} (from left to right) of each action tube (bearing in mind that in autonomous driving, for instance, road activities tend to follow a specific order, e.g., pedestrian crossing the road followed by vehicles engaging an intersection); (ii) the \emph{spatial similarity} \rev{ of the node features inspired from the edge convolution proposed in \cite{wang2019dynamic}, which dynamically constructs edges between the nodes according to their feature distances. The distance threshold in our case is fixed, but can be a learnable parameter by updating via an edge convolution  \cite{wang2019dynamic};} (iii) \emph{node type}, meant as the % Finally, we also connect the tubes of 
sharing the same action label, as this provides very relevant information for the determination of the activity class. %help in understating the activity from dominant actions. All these connections 
{As a result %of these connections 
three spatiotemporal scene graphs are constructed having the same nodes but with different edges. While the second and third graph are undirected, the appearance order graph is a directed one. However, when merging the three graphs an undirected version of the order graph is used. These graphs are then combined by taking a union of all edges to create a single homogeneous graph representing the overall scene. Namely, two nodes are connected in the merged graph iff they are connected in at least one of the three graphs.} %FAB: in the future we can much work on fusion and inference on this HG
% All edges are then refined by suppressing repetitions {\color{blue} [? meaning?]} and constructing a single homogeneous graph.} {\color{blue} [how do we merge the graphs for the different semantics? also we never say that our S/T scene graph is heterogenous. I feel more details are required here.]}

\textbf{Graph Convolution and Representation}. 
Given the final graph, global graph embedding is applied to extract the context of each snippet portraying a complex activity. In our GCN approach we apply a stack of three graph convolutional layers followed by a graph readout layer. 
The latter encapsulates the final graph representation by taking the mean of the hidden convolutional representation, resulting in a fixed-sized feature representation which is invariant to the number of nodes and edges.

\begin{figure}
    \centering
    \includegraphics[width=0.95\textwidth]{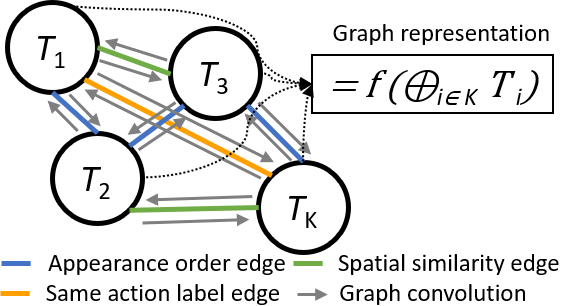}
    \caption{Our graph module takes as input the features generated for each tube by the ROI pooling layer $T_{1,2,3...K}$ and builds edges between them according to different semantics. ({ order of appearance, spatial similarity, label}). 
    The overall graph is processed by a GCN to deliver a fixed-size graph representation.}
    \label{fig:gcn}
\end{figure}

\begin{algorithm}[h]
\SetAlgoLined
\nl \KwIn{Input Video $V$ divided into $N$ number of Snippets  $S$ consists of fixed $M$ number of frames ${F}^{1,2,3,...,M}$}
\nl {\textbf{Initialization}: Load pretrained action tube detector ${ATD}$ on desire dataset, import our 3D deformable RoI pooling ${DRoI}$ and sparsity ${Sp}$ modules \; }
\nl \For{$i$ $\in$ 1,2,3,..., $N$}{
\nl  Extract features and detect action tubes ${B_A}^{M \times K \times 5}$, ${X_B}^{M \times 64 \times 300 \times 300}$  $\leftarrow$ ${ATD}$($S_i$) \;

\nl ${X_{DRoI}}^{K \times 64 \times M \times 7 \times 7}$ $\leftarrow$ ${DRoI({{B_A}^{M \times K \times 5}} ,{ {X_B}^{M \times 64 \times 300 \times 300} }  )}$ \;

\nl ${X_{Sp}}^{K \times 32 \times M \times 7 \times 7}$ $\leftarrow$ ${Sp}({X_{DRoI}}^{K \times 64 \times M \times 7 \times 7}$)  \;

\nl ${X_{Fc1}}^{K \times M \times 512}$ $\leftarrow$ ${Fc1}({X_{Sp}}^{K \times 32 \times M \times 7 \times 7})$  \;

\nl ${X_{Fc2}}^{2048}$ $\leftarrow$ ${Fc2}({X_{Fc1}}^{K \times M \times 512})$  \;

\nl ${CA}_i$ $\leftarrow$ ${Softmax}({X_{Fc2}}^{2048})$  \;

\nl  \eIf { ${CA}_{i-1}$ == ${CA}_{i}$}{

\nl   Same activity\;
    }{
\nl   \eIf{${CA}_{i-1}$ == ${CA}_{i+1}$}{
\nl   same activity\;
}{
\nl   end of activity\;

\nl   start new activity\;
}}}
\nl \KwOut{Start and end frames with activity labels}
 \caption{Complex Activity Detection}
 \label{algo:algo1}
\end{algorithm}

\subsection{Complete Framework}

The complete framework is the concatenation of the aforementioned three modules. Firstly, we divide the video ${V}$ into ${N}$ Snippets ${S}_{1,2,3,...,N}$, with each Snippet ${S}_{i}$ consisting of a fixed (${M}$) number of frames: $S_i = {F}^{1,2,3,...,M}$. Each snippet is passed to the action tube detection module ${AT}$ which returns ${K}$ action tubes each composed by $L = 12$ bounding boxes with labels ${B}_{A}$ and intermediate VGG features ${X}_{A}$ represented as ${B_A}^{M\times K\times5}$, ${X_A}^{M\times 64\times 300\times300 }$ $\in$ ${AT}$. % FAB: you mean X_A, right?
Action tube locations % FAB: be specific: how is the tube geometry represented? two vertices for each of the 12 bounding boxes?-sal: 2 vertices with 2 len and height and one for label total 5 values with each bounding box for 12 frames
and features are then passed to our 3D deformable RoI pooling layer ${DRoI}$ which returns a fixed-sized (i.e., $7 \times 7$) grid of components % FAB: again check the terminology
whose dimensionality is equal to the number (64) of convolutional layers: ${{X}_{DRoI}}^{K\times 64\times M\times 7\times7 }$. % $\in$ ${DRoI}$. 
{ These features are then fed to our GCN module ${G}$, where a graph with $K$ nodes and $E$ edges is processed to yield a fixed-sized feature representation ${X_{G}}^{2048}$ $\in$ ${G}$. 
Finally, the latter features are fed to a Softmax classifier to classify the snippet into their respective activity category.}

For localisation we use a sliding window approach {in which snippets are sequentially classified, using a dual verification mechanism. Namely, as our purpose is to recognise and detect \emph{long-term} activities, if there appears to be a random false positive or false negative between two snippets belonging to the same class we simply ignore it and consider it as having the same activity label. The detection algorithm is described in detail in Algorithm \ref{algo:algo1}.} 
% {\color{red} so all interruptions composed of 1 snippet are ignored? what about 2, or 3?} Sal: If more than one, we consider it as new activity as some of our activities are also very short (2 and 3 snippets). 

% FAB: a lot of this we say before, consider cutting some repetitions if lacking space
\textbf{Implementation}. Before training our overall architecture, we separately train AMTNet for action tube detection over both datasets. 
Note that we had to design from scratch suitable data loaders for the two datasets, as 
%In AMTNet, we contribute to the data loader part as the dataset 
the format of the annotation there is completely different from that of the original datasets AMTNet was validated upon.
%Next, we implement the 3D deformable RoI pooling layer based on the same principle of RoI pooling layer, however, we include a temporal dimension we learn and extract the parts in tube rather than a 2d object. 
As mentioned, our 3D RoI pooling layer includes a temporal dimension to learn and extract tube parts of a tube rather than of a 2D object.
In our experiments we also convert a more recent version of deformable RoI pooling called 'modulated' deformable RoI pooling to the 3D case. 
{In the GCN module we construct graphs online for each snippet during training using a PyTorch data loader \cite{NEURIPS2019_9015}. For the design of the GCN architecture we use the Deep Graph Library (DGL) \cite{wang2019dgl} with a PyTorch back-end, which supports the processing of graphs of various length in a single mini-batch.} Overall, architecture is implemented using the PyTorch \cite{NEURIPS2019_9015} deep learning library with OpenCV and Scikit-learn. For training we used a machine equipped with 4 Nvidia GTX 1080 GPUs with 12GB VRAM each.

\begin{table}
\begin{center}
\caption{List of ROAD complex activities with number of train and test snippets splits and brief description.}\label{tab:road_data_tab}
\footnotesize \scalebox{1} {
\begin{tabular}{p{0.016\textwidth}p{0.16\textwidth}p{0.06\textwidth}p{0.06\textwidth}p{0.50\textwidth}}
%\begin{tabular}{|l|c|c|c|c|}
\hline
No & Complex activity & Train snippets & Test snippets & Scenario/Description \\
\hline\hline
1 & Negotiating intersection &  523 &  414 & Autonomous vehicle (AV) stops at intersection/junction, waits for other vehicles to pass, then resumes.   \\
2 & Negotiating pedestrian crossing & 381  & 25  & AV Stops at pedestrian crossing signal, waits for pedestrian(s) and cycles to cross, then resumes.  \\
3 & Waiting in a queue & 570  & 447  & AV stops and waits in a queue for a signal (either pedestrian crossing or junction traffic light). %, as they both look similar from the AV’s perspective)   
\\
4 & Merging into vehicle lane & 68  & 3  & AV stops and waits for bus/car to merge into the vehicle lane from the opposite side of the road.  \\
5 & Sudden appearance &  3 & 3  & A pedestrian/cycle/vehicle suddenly appears in front of the vehicle, coming from
a side street.  \\
6 & Walking in the middle of road & 13  &  19 & A pedestrian walking in the middle of the road in front of the AV, causing the AV to slow down or stop.  \\

\hline
\end{tabular}
}
\end{center}
\end{table}

\section{Experiments} \label{deformsgraph:sec:exp}

\subsection{Datasets and Evaluation Metrics}

In this chapter, we used two datasets for evaluating our approach, %The main reason behind using these datasets is that both of them are 
both already annotated at video level for action tubes detection. % which is a prerequisite for our system to be able to perform complex activity recognition. 

\textbf{ROAD} \cite{singh2021road}: ROAD (the ROad event Awareness Dataset for autonomous driving) is annotated for road action and event detection. The complete detail of ROAD dataset is given in Section \ref{related_work:sec:datasets}.

In this work, we augmented the annotation of the ROAD dataset for complex road activity detection. We used a total of 19 videos with an average duration of 8 minutes each, 12 of which were selected for training and the remaining 7 for testing. We temporally annotated the ROAD videos by specifying the start and end frame for six different classes of complex road activities we inferred from video inspection. For example, a `Negotiating intersection' activity class can be defined which is made up of the following `atomic’ events: Autonomous Vehicle (AV)-move + Vehicle traffic light / Green + AV-stop + Vehicle(s) / Stopped / At junction+ AV-move. {Activity class statistics are listed and described in Table \ref{tab:road_data_tab}}.

\textbf{SARAS-ESAD} \cite{bawa2020esad}: ESAD (the Endoscopic Surgeon Action Detection Dataset) is a benchmark devised for surgeon action detection in real-world endoscopic surgery videos described in Section \ref{related_work:sec:datasets}.

In this work, we took a step forward and annotated ESAD 
%for recognition of complex activities. The activities in this dataset are refer to
in terms of complex activities corresponding to the different \emph{phases} of the surgical procedure portrayed by the videos (namely, radical prostatectomy). For example, Phase \# 3 corresponds to `Bladder neck transection', in which a scissor cuts the neck of the bladder until it is transected. {Phases and their statistics are again reported in Table \ref{tab:saras_data_tab}}. 
{The complex activities (surgical phases) in SARAS were defined by professional surgeons experts in radical prostatectomy. As standard in the surgical context, such phases are consecutive without the need for any background activity class.}
For more details please see \cite{bawa2020esad}.

\begin{table}
\begin{center}
\caption{SARAS video phases with number of train and test snippets and a brief description.}\label{tab:saras_data_tab} % FAB: where is this info actually coming from? Is it correct to cite Vivek's paper?-sal: no these are for temporal annotation provided by giocomo
\footnotesize \scalebox{1} {
\begin{tabular}{p{0.014\textwidth}p{0.2\textwidth}p{0.030\textwidth}p{0.030\textwidth}p{0.50\textwidth}}
%\begin{tabular}{|l|c|c|c|c|}
\hline
No & Complex activity/ Phase \# & Train snippets & Test snippets & Scenario/Description \\
\hline\hline
1 & Phase\#1 (Bladder mobilization) &  1808 & 999  & Tools only used for mobilization   \\
2 & Phase\#2 (AFP Dissection) &  6263 &  4455 & Anytime only Anterior Prostatic Fat (AFP) is removed  \\
3 & Phase\#3 (Bladder neck transection)& 8353  & 2419  & Whenever the scissor cut the neck, until all the neck is transected \\
4 & Phase\#4 (NVP dissection) &  361 & 46  & Located just under the seminal vescicles. \\
5 & Phase\#5 ( Exposure of seminal vescicles) & 2858  & 800  & White bags surrounding the bladder neck. Comprises fat removal.  \\
6 & Phase\#6 (Urethral division) & 2553  &  1685 & 	Start even when removing fat on the urethra. \\
7 & Phase\#7 (Prostate liberation)& 1092  & 2903  & 	After urethral division. It is not AFP. Stops only when completely free. \\
8 & Phase\#8 (Vesicourethral anastomosis)& 4430  & 4690  & 	Every time there is string and needle in the scene. Stops when urethra is linked to bladder\\
\hline
\end{tabular}
}
\end{center}
\end{table}

\textbf{Evaluation Metrics}: For the evaluation of action tube detection performance we used the standard frame/video mean Average Precision (\textit{mAP}) at different IoU thresholds $\delta$ (namely, 0.2, 0.3, 0.5, 0.75) on both datasets. Complex activity recognition was evaluated using classification accuracy, precision, recall and F-score. For complex activity localisation we used the standard protocol \textit{mAP} over the temporal dimension used by all relevant methods.

\subsection{Action Tube Detection}

%In this section, we show the evaluation of action tubes detector AMTNet as it is an important bit of our framework and the overall accuracy depends upon it. 
A detailed comparative analysis of AMTNet over different action detection datasets can be found in the original paper \cite{saha2020two}. Here %while in this section our focus is on 
we briefly report the performance of AMTNet on our two datasets of choice, as 
%We show these results because %The main reason behind showing these experiments are that both of them are exposed for the first time to 
%none of them was ever used to test 
AMTNet was never tested there. 
%The detailed results are reported in 
Table \ref{tab:tab1} reports both frame-\textit{mAP} and video-\textit{mAP} results, and compares AMTNet with the proposed baselines 
%proposed by the papers which released 
for the two datasets: the ROAD baseline (termed \emph{3D RetinaNet} \cite{singh2021road}), and % reports detailed results for action tube detection there at both frame-\textit{mAP} and video-\textit{mAP} level. The 
the ESAD baseline \cite{bawa2020esad}, a vanilla implementation of RetinaNet (only providing frame-level results). %for only frame-\textit{mAP}. 
%Thus, to provide a comparison between our method and both baselines we calculated  on both datasets. As it can be seen in Table ~\ref{tab:tab1}, 
AMTNet performed better than \cite{bawa2020esad} on SARAS-ESAD, while being inferior to \cite{singh2021road} on ROAD. Remember that the main rationale for using AMTNet is that it can provide a fixed-size representation for the tubes (as required by our framework), motivating us to compromise on accuracy.

\subsection{Complex Activity Recognition}

Next, we provide a detailed analysis of complex activity \emph{classification} using our approach on both the ROAD and SARAS-ESAD datasets. The performance for each class in both datasets is illustrated in Fig. ~\ref{fig:reg_res_road} and Fig. ~\ref{fig:reg_res_saras} using all metrics. %From the results, it is 
It is apparent that the ROAD dataset is characterised by significant fluctuations in class-wise performance, with higher recognition accuracy for activities that appear more often, e.g. `waiting in a queue', as opposed to infrequent ones (e.g. `sudden appearance'). %In other words, the recognition accuracy of activities that occur too often i.e., waiting in a queue is very higher than the other with very low occurrence i.e., sudden appearance.
%This fluctuation of activity occurrence makes the ROAD dataset more challenging during the training. 
%Unlike ROAD, 
In SARAS-ESAD each activity class does contain enough samples for good training, while the diversity in phases 
%also play a challenging role in recognition.
still poses a challenge. 
{ In fact, performance on this dataset is a function of the complexity of the phases and the visual similarity of the constituent atomic actions, not just the amount training data. For example, Phase 5 and 6 both include ‘fat removal’ actions, complicating their differentiation.}

\begin{table*}
\begin{center}
\caption{Action tube detection performance on both the ROAD and SARAS-ESAD datasets. Both Frame-\textit{mAP} and Video-\textit{mAP} at different IoU thresholds are reported for evaluation}\label{tab:tab1}
\footnotesize \scalebox{1} {
%\begin{tabular}{p{0.31\textwidth}p{0.31\textwidth}p{0.31\textwidth}}
\begin{tabular}{*{5}{p{0.050\textwidth}}|*{4}{p{0.050\textwidth}}|*{4}{p{0.050\textwidth}}}
\hline
 \multicolumn{5}{c}{}& \multicolumn{4}{c|}{ROAD}  & \multicolumn{4}{c}{SARAS-ESAD}\\
\hline
\multicolumn{5}{c|}{Methods / IoU threshold $\delta$}  & 0.2 & 0.3 & 0.5 & 0.75  &  0.2 &0.3 &0.5 & 0.75\\
\hline
\multicolumn{5}{c|}{Singh et al. \cite{singh2021road} (frame-\textit{mAP}) }  & - & - & 25.9 & -&  -&- &-&-   \\
\multicolumn{5}{c|}{Singh et al. \cite{singh2021road} (video-\textit{mAP}) }  & 17.5 & - & 4.6 & - &  -&- &-&-   \\
\multicolumn{5}{c|}{Bawa et al. \cite{bawa2020esad} (frame-\textit{mAP}) }  & -& - & - & - &  - &24.3 & 12.2 & -  \\
\multicolumn{5}{c|}{AMTNet (frame-\textit{mAP})}  & 22.3 & 18.1 & 15.4 & 11.0 &   30.4 &24.6 &18.7&7.9   \\
\multicolumn{5}{c|}{AMTNet (video-\textit{mAP})}  & 11.6 & 7.9 & 3.8 & - & 13.7 & 10.1 &8.8 &5.4  \\

\hline
\end{tabular}
}
\end{center}
\end{table*}

\begin{figure*}[h]
    \centering
    \includegraphics[width=0.95\textwidth]{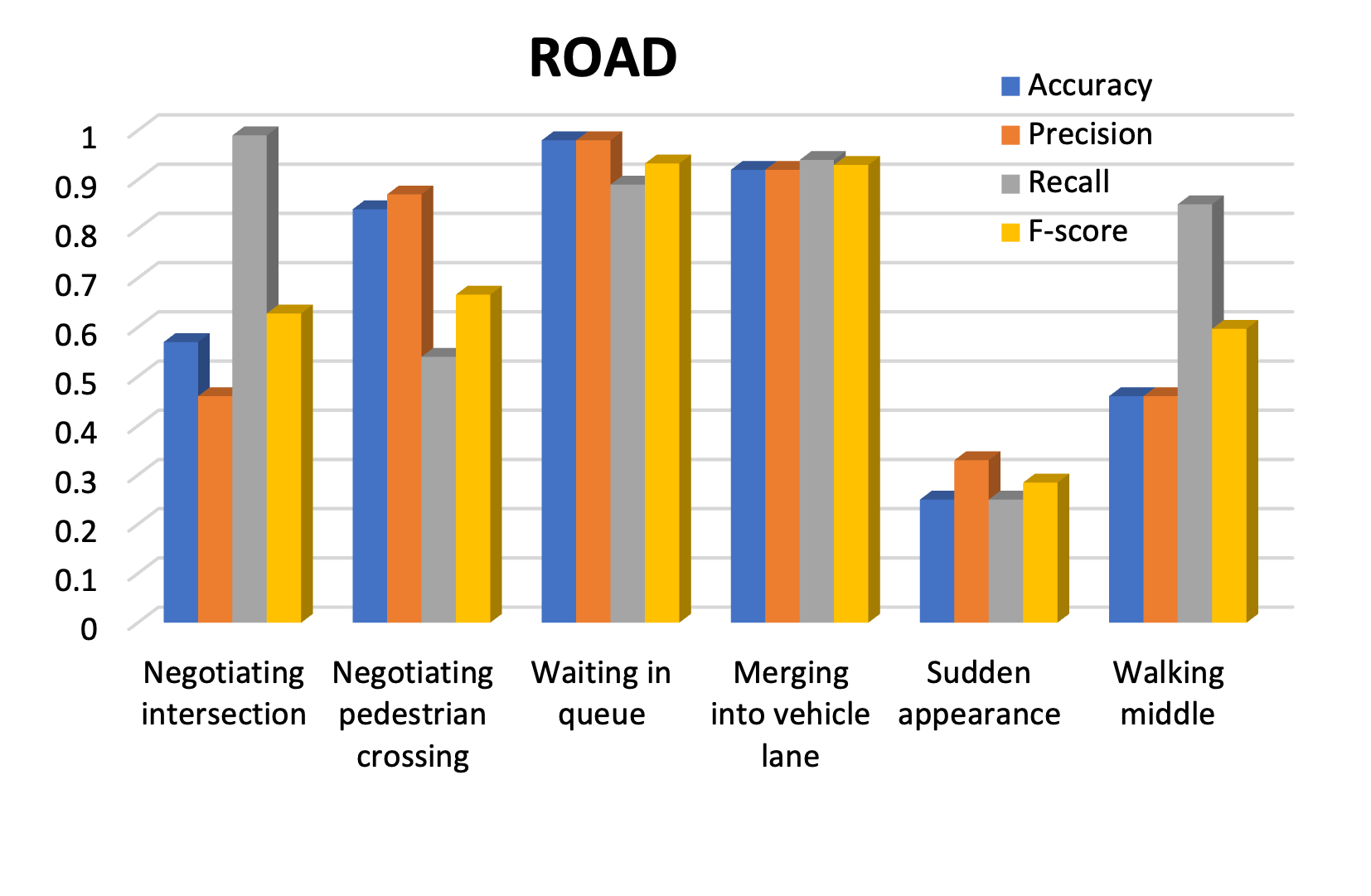}
    \caption{Complex activity classification performance over ROAD dataset.
    }
    \label{fig:reg_res_road}
\end{figure*}

\begin{figure*}[h]
    \centering
    \includegraphics[width=0.95\textwidth]{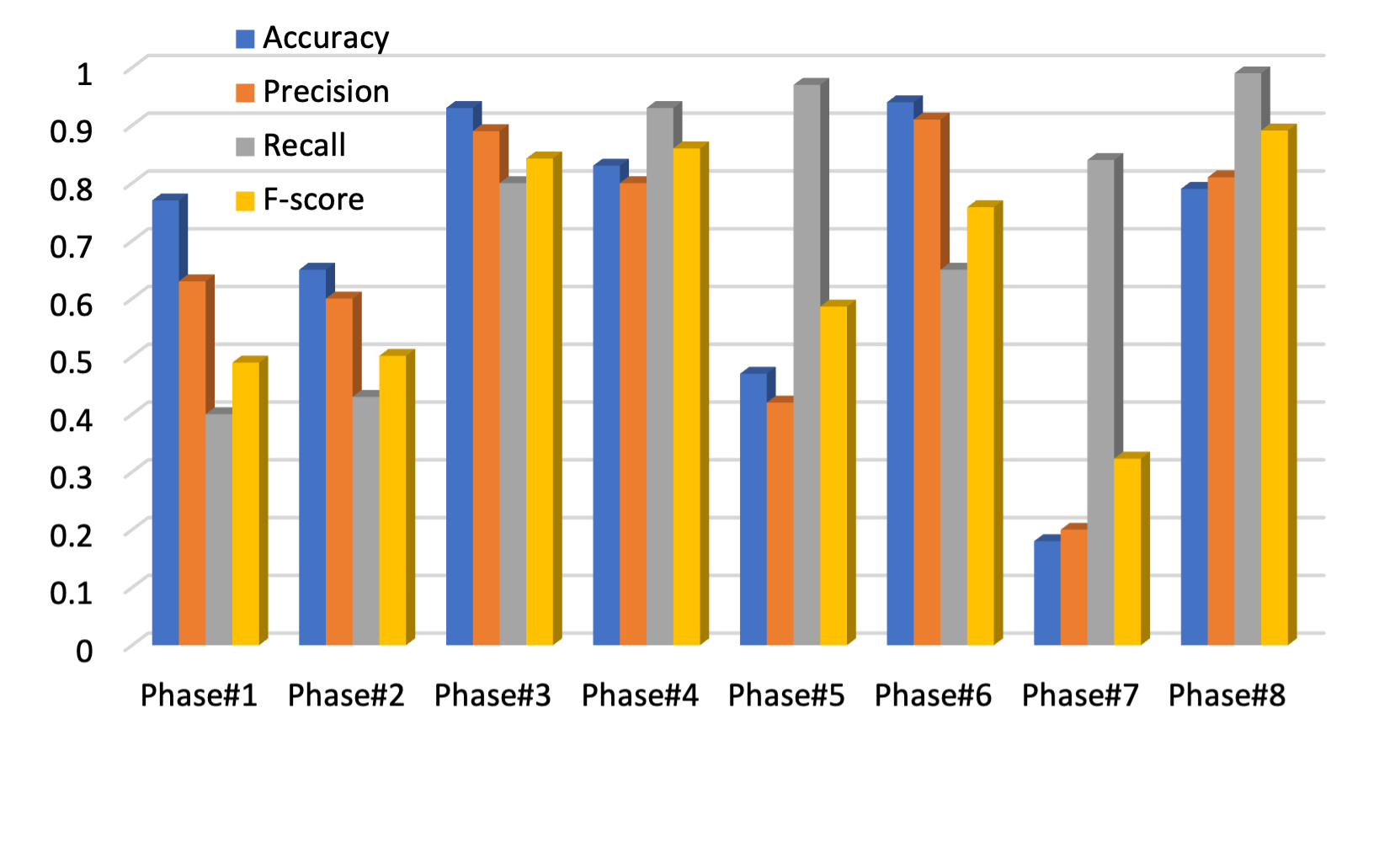}
    \caption{Complex activity classification performance on SARAS-ESAD dataset.
    }
    \label{fig:reg_res_saras}
\end{figure*}

\begin{table}
\begin{center}
\caption{{Comparative analysis of temporal activity localisation performance on ROAD and SARAS-ESAD, %with state-of-the-art 
reporting \textit{mAP} (\%) at five different IoU thresholds.}}
\label{tab:tab2}
\footnotesize \scalebox{1} {
%\begin{tabular}{p{0.31\textwidth}p{0.31\textwidth}p{0.31\textwidth}}
\begin{tabular}{*{5}{p{0.038\textwidth}}|*{5}{p{0.038\textwidth}}|*{5}{p{0.038\textwidth}}}
\hline
 \multicolumn{5}{c}{}& \multicolumn{5}{c|}{ROAD}  & \multicolumn{5}{c}{SARAS-ESAD}\\
\hline
\multicolumn{5}{c|}{Methods / IoU threshold $\delta$}  & 0.1 & 0.2 & 0.3 & 0.4& 0.5  &  0.1 &0.2 &0.3 & 0.4 & 0.5\\
\hline
\multicolumn{5}{c|}{P-GCN \cite{zeng2019graph} }  &60.0 & 56.7 & 53.9 & 50.5 & 46.4& 57.9 & 55.6 & 53.4 & 49.0 & 45.1   \\
\multicolumn{5}{c|}{G-TAD \cite{xu2020g} }  &62.1 & 59.8 & 55.6 & 52.2 & 48.7& 59.1 & 56.7 & 54.5 & 49.8 & 46.9 \\
\multicolumn{5}{c|}{\textbf{Ours} }  &\textbf{77.3} & \textbf{74.6} & \textbf{71.2} & \textbf{66.7} & \textbf{59.4}& \textbf{62.9} & \textbf{59.6} & \textbf{58.2} & \textbf{55.3} & \textbf{51.5}  \\

\hline
\end{tabular}
}
\end{center}
\end{table}

\subsection{Temporal Activity Detection - Comparison with State-of-the-Art}

To evaluate the performance of our complex activity detection approach we reimplemented two state-of-the-art activity localisation methods -- P-GCN \cite{zeng2019graph} and G-TAD \cite{xu2020g}, as 
%both datasets are exposed for the first time to 
neither ROAD nor ESAD were ever used for
complex activity detection. The major changes we made during re-implementation are: (i) data loading (as both P-GCN and G-TAD were designed to be trained and tested on pre-extracted features), and (ii) replacing the regression part with a sliding window approach for the SARAS-ESAD dataset, as the latter lacks a background class.

\begin{table}
\begin{center}
\caption{{ROAD activity localisation performance (\textit{mAP}, \%) for each complex road activity, at a standard IoU threshold of 0.5.} }
\label{tab:tab3}
\footnotesize \scalebox{1} {
%\begin{tabular}{p{0.31\textwidth}p{0.31\textwidth}p{0.31\textwidth}}
\begin{tabular}{*{1}{p{0.11\textwidth}}|*{6}{p{0.11\textwidth}}}
\hline
Method / Activities &  Negotiating intersection &Negotiating pedestrian crossing& Waiting in queue &Merging into vehicle lane& Sudden appearance & Walking middle of road\\
\hline
P-GCN \cite{zeng2019graph}  & 44.3 &	53.8	&74.4&	50.1&	21.7&	34.1\\
G-TAD \cite{xu2020g}  & 47.8&	57.3&	70.6&	55.2&	\textbf{24.3}&	37.1
  \\
\textbf{Ours} & \textbf{51.2}&	\textbf{72.3}&	\textbf{89.8}&	\textbf{84.1}&	17.8&	\textbf{41.3}  \\
\hline
\end{tabular}
}
\end{center}

\end{table}

{\textbf{ROAD}. For activity detection in ROAD we used an additional `background' class, which indicates either no action or presence of action(s) without any solid indication. %FAB: ? meaning? low scores?
%In first case, when there is 
Whenever no action tube was detected we would use the entire frame %dimensions as an 
as RoI for our parts deformation module to understand the overall scene. 
%The next scenario is considered as same as other activities to learn the model for actions without contribution to any activity. %FAB: I don't understand this sentence
%The comparison of ROAD 
Temporal activity detection performance on ROAD, measured via \textit{mAP} at five different IoU thresholds, is reported in Table ~\ref{tab:tab2} for both our approach and the two competitors. Class-wise results for each complex activity %While the results of each activity with 
at a standard IoU threshold of 0.5 are reported in Table ~\ref{tab:tab3}.   }

\begin{table}
\begin{center}
\caption{{SARAS-ESAD activity localisation performance (\textit{mAP}, \%) for each activity at a fixed IoU threshold of 0.5.}}
\label{tab:tab4}
\footnotesize \scalebox{1} {
%\begin{tabular}{p{0.31\textwidth}p{0.31\textwidth}p{0.31\textwidth}}
\begin{tabular}{*{10}{p{0.065\textwidth}}}
\hline
\multicolumn{2}{c|}{Method / Activities} & Phase\#1 & Phase\#2& Phase\#3& Phase\#4& Phase\#5 & Phase\#6 & Phase\#7 & Phase\#8\\
\hline
\multicolumn{2}{c|}{P-GCN \cite{zeng2019graph}} & 56.7&	43.2&	52.3&	59.1&	\textbf{33.8}&	59.4&	14.8&	41.2 \\
\multicolumn{2}{c|}{G-TAD \cite{xu2020g}} & 51.1&	46.6&	57.2	&63.8	&29.4	&62.2&	\textbf{19.3}&	45.7\\
\multicolumn{2}{c|}{\textbf{Ours}}& \textbf{57.5}&	\textbf{54.1}&	\textbf{69.3}&	\textbf{60.2}&	31.1&	\textbf{71.3}&	16.5&	\textbf{52.4} \\

\hline
\end{tabular}
}
\end{center}

\end{table}

{\textbf{SARAS-ESAD}. Temporal activity detection on this dataset much relates to %almost similar to the 
activity recognition, as surgical phases are contiguous. 
%For localisation, however, it is important to detect start and end time of each phase. 
%Like the ROAD dataset, we the average
Both the average \textit{mAP} of the methods at five IoU thresholds and the class-wise performance for each activity (phase) at a standard IoU threshold of 0.5 are reported in Table~\ref{tab:tab2} and \ref{tab:tab4}, respectively. From the results it is clear how our method outperforms the chosen state-of-the-art methods by a reasonable margin.}

\rev{
\subsection{Ablation Studies}\label{sec:ablation}
We ran a number of ablation studies on both datasets to understand the influence of the various components of our model.
}

 \rev{\textbf{Effect of GCN}. First of all, we demonstrated the significance of our GCN by conducting three different types of experiments; i) Snippet features from the entire scene without engaging in action tube detection or utilising GCN.. ii) Action features without GCN where we first detect the action tubes and then extract the deformable features from the tubes. It is worth noting that, in this particular instance, we constrained the number of action tubes to a fixed limit of 10. iii) Action features with GCN, our proposed framework, incorporating action features and integrating the GCN. The clear difference among these three types of setups can be observed in Fig. \ref{fig:ab_nodes}.
 }

\begin{figure}[h]
    \centering
    \includegraphics[width=0.95\textwidth]{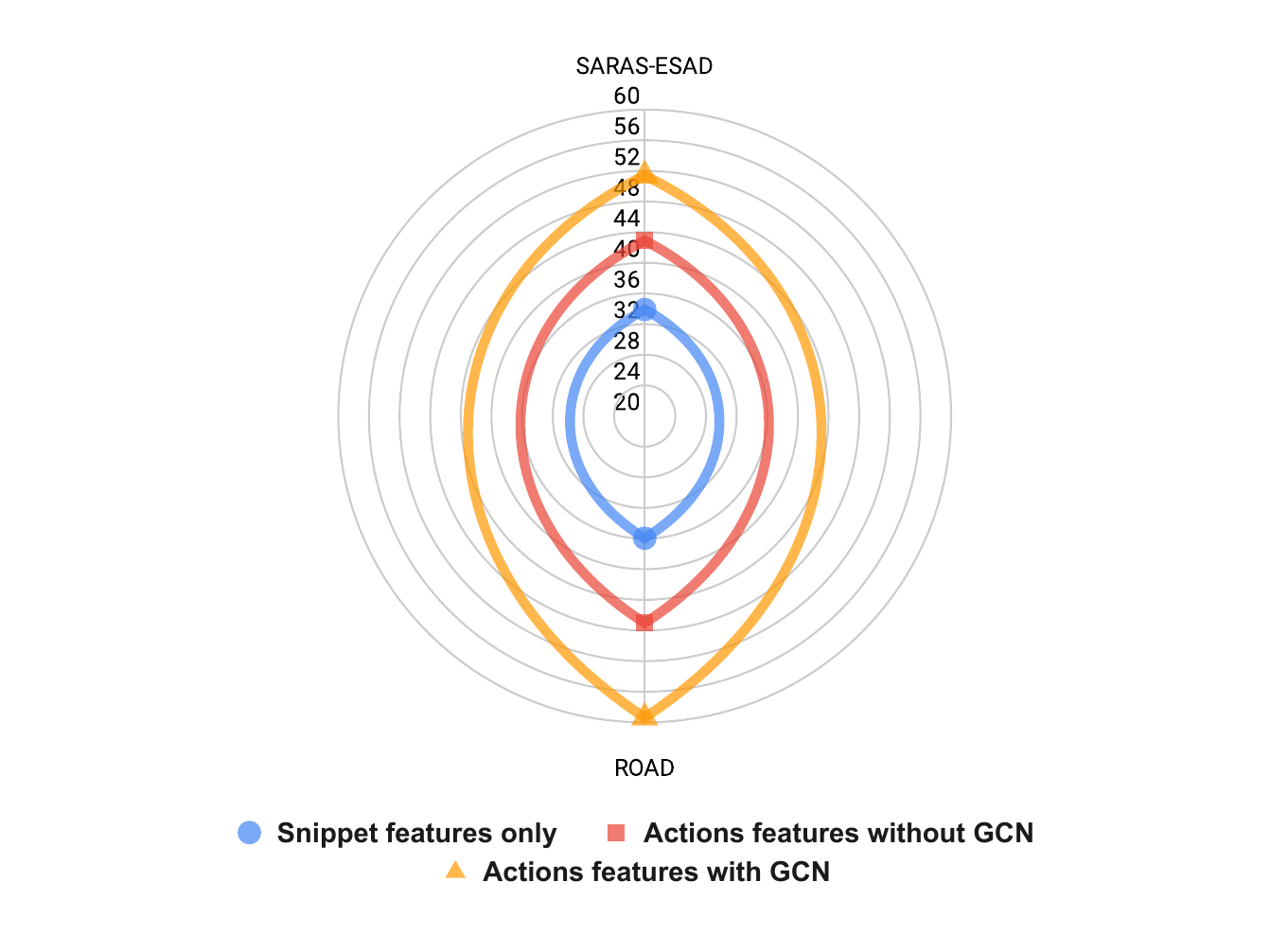}
    \caption{\rev{The performance of different components of our proposed method over ROAD and SARAS-ESAD dataset. The mAP at a threshold of 0.5 is reported for the test set of both datasets.}
    }
    \label{fig:ab_nodes}
\end{figure}

\rev{\textbf{Effect of Edges}. Next, we ablated the effect of various types of edges employed in our GCN for graph construction. This ablation analysis includes all three types of edges used for graph construction (explained in Section \ref{deformsgraph:ssec:gcn}) individually and combined. The comprehensive performance for each edge type over both ROAD and SARAS-ESAD datasets is presented in Fig. \ref{fig:ab_edges}.
}

\begin{figure}[h]
    \centering
    \includegraphics[width=0.95\textwidth]{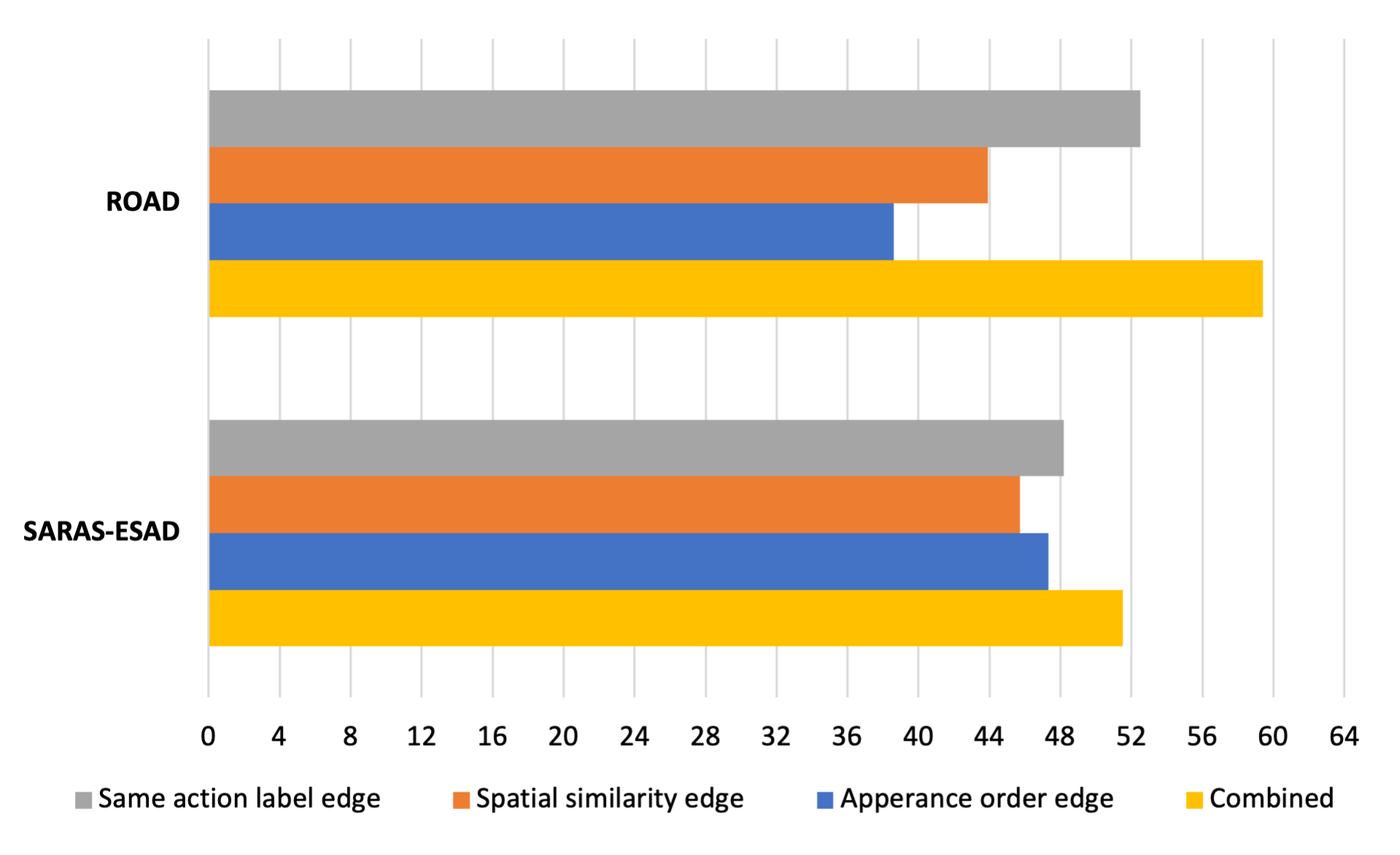}
    \caption{\rev{The performance comparison of various edge types in our GCN modules. The mAP at a threshold of 0.5 is reported for the test set of both datasets.}
    }
    \label{fig:ab_edges}
\end{figure}

\rev{\textbf{Effect of Sequence Length}. Lastly, we showed how snippet duration, which represents the time span of the local scene impacts our model. We conducted experiments with four different sizes (12, 18, 24, and 30), as given in Table \ref{tab:ab_seq_len}. Notably, when evaluating our model on ROAD, the most favorable results were achieved with a sequence length of 18. This choice is attributed to the datasets' characteristics, as the activities within them tend to be of average duration. Conversely, when evaluating on SARAS-ESAD, we observed superior performance with a sequence length of 24, aligning with the predominantly the much longer duration of activities in this dataset.
}

\begin{table}[h]
\begin{center}
\caption{\rev{The performance comparison of different sequence lengths over both ROAD and SARAS-ESAD dataset, reporting \textit{mAP} (\%) at two different IoU thresholds.}}
\label{tab:ab_seq_len}
\footnotesize \scalebox{1} {
%\begin{tabular}{p{0.31\textwidth}p{0.31\textwidth}p{0.31\textwidth}}
\begin{tabular}{*{2}{p{0.098\textwidth}}|*{2}{p{0.098\textwidth}}|*{2}{p{0.098\textwidth}}}
\hline
 \multicolumn{2}{c}{}& \multicolumn{2}{c|}{ROAD}  & \multicolumn{2}{c}{SARAS-ESAD}\\
\hline
\multicolumn{2}{c|}{Sequence length / IoU threshold $\delta$}  & 0.1 &0.5  &  0.1 & 0.5\\
\hline
\multicolumn{2}{c|}{12}  &77.3 & 59.4 & 62.9 & 51.5  \\
\multicolumn{2}{c|}{18}  &\textbf{79.7} & \textbf{60.8} & 64.6 & 52.8  \\
\multicolumn{2}{c|}{24}  &74.1 & 55.3 & \textbf{67.2} &\textbf{ 54.5}  \\
\multicolumn{2}{c|}{30}  &70.6 & 52.2 & 66.1 & 53.7  \\

\hline
\end{tabular}
}
\end{center}

\end{table}

\section{Summary of the chapter}\label{deformsgraph:ssec:summary}

\subsection{Limitations and Future Work}

The main limitation of this work is that it relies on action tube detection. From our results, the existing tube detectors are not reliable enough to perform well over challenging real-world datasets such as those we adopted here. 
Clearly, if the tube detector misses an important atomic action this will affect the overall activity detection performance. 
Nevertheless our results show that, even when using a suboptimal detector, our approach is capable of significantly outperforming state of the art methods on our new benchmarks.

Detection is challenging on SARAS and ROAD because of their real-world nature: surgical images are indistinct, road scenes come with incredible variations. These benchmarks show how even the best detectors suffer a huge drop in performance when moving from ‘academic’ benchmarks to real ones. We hope the realism of these two extremely challenging datasets will stimulate real progress and new original thinking in the field.

In the future our primary target will be the design of a more accurate action tube detector with the ability to perform better in challenging scenarios such as those portrayed in ROAD or SARAS-ESAD. 
We will also explore the end-to-end training of the entire model in all its three components.
Further down the line, 
%an interesting generalisation of deformable models is constituted by 
we will update our S/T scene graph approach to properly model the heterogenous nature of the graph \cite{zhang2019heterogeneous}, % FAB: you can cite this paper: https://dl.acm.org/doi/abs/10.1145/3292500.3330961
and extend it to a more complete representation of complex dynamic events
in which nodes (rather than correspond all to action tubes) may be associated with any relevant elements of a dynamic scene, such as objects, agents, actions, locations and their attributes (e.g. red, fast, drivable, etc). %with the aim of achieving a truly complete description of complex dynamic events.
%we are also intended to solve the problem of complex activities via heterogeneous, where each module such as actions, parts will be deal with a different graph CNN. 
%In addition, we are also facing the problem of lower accuracy for those classes with lower number of samples in ROAD dataset, which could be the possible target for future to augment the similar activities.

\subsection{Conclusion}

In this chapter, we presented a spatiotemporal complex activity detection framework which leverages both part deformation and a heterogenous graph representation. Our approach is based on three building blocks; action tube detection, part-based deformable 3D RoI pooling for feature extraction and a GCN module which processes the variable number of detected action tubes to model the overall semantics of a complex activity.
In an additional contribution, we temporally annotated two recently released benchmark datasets (ROAD and ESAD) in terms of {long-term} complex activities. Both datasets come with video-level action tube annotation, making them suitable benchmarks for future work in this area. We thoroughly evaluated our method, showing the effectiveness of our 3D part-based deformable model approach for the detection of complex activities.

    \chapter{A Hybrid Graph Network for Complex Activity Detection in Video}
\label{chapter:hgraph}

\section{Introduction} 
\label{hgraph:sec:intro}

Following the limitations of our previous work (relies on action tube detection) explained in Chapter \ref{chapter:deformsgraph}. In this Chapter, we propose a new \textbf{Comp}lex \textbf{A}ctivity \textbf{D}etection (\cad{}) paradigm that can be applicable to a variety of computer vision domains, including autonomous driving and sports analytics without additional action annotation. \cad{} can be seen as the advanced version of Temporal Action Localisation (TAL), which aims to recognise what actions take place in a video and identify their start and end time. While TAL mostly focuses on short-term actions, \cad{} extends the analysis to long-term activities, and does so by modelling a complex video activity's internal structure. In this chapter, we address the \cad{} problem using a hybrid graph neural network that combines  attention applied to a graph encoding the local (short term) dynamic scene with a temporal graph modelling the overall long-duration activity. Our contribution is threefold. i) Firstly, we propose a novel feature extraction technique which, for each video snippet, generates spatiotemporal `tubes' for the active elements (`agents') in the (local) scene by detecting individual objects, tracking them and then extracting 3D features from all the agent tubes as well as the overall scene. ii) Next, we construct a local scene graph where each node (representing either an agent tube or the scene) is connected to all other nodes. Attention is then applied to this graph to obtain an overall representation of the local dynamic scene. iii) Finally, all local scene graph representations are temporally connected to each other via a temporal graph, with the aim of estimating the complex activity class together with its start and end time. The proposed framework outperforms all previous state-of-the-art methods on three benchmark datasets: Thumos-14, ActivityNet-1.3, and the more recent ROAD. 

\subsection{Motivation}
{The ability to detect and recognise activities from untrimmed videos is a research problem which is attracting significant attention, }
%\emph{Temporal Action Localisation} (TAL) 
%or Complex Activity Detection (\cad{}) 
%in untrimmed videos is attracting significant attention,
due in part to the rapid growth of online video generation in domains including sports  \cite{hu2020progressive}, autonomous driving \cite{caesar2020nuscenes}, medical robotics \cite{lin2016video} and surveillance \cite{yuan2017temporal}. {The current state-of-the-art in action detection is \emph{Temporal Action Localisation} (TAL) - whose approaches not only} recognise the action label(s) present in a video, but also %to
identify the %temporal location - i.e. 
start and end time of each action instance occurring.
%Among its many applications, 
TAL can be instrumental in generating sports highlights \cite{li2021multisports,zeng2021graph}, understanding road scenes in autonomous driving \cite{skhan2021comp}, video summarisation %via key event extraction 
in surveillance videos \cite{xiao2020convolutional} and video captioning \cite{krishna2017dense,mun2019streamlined}.
A number of TAL methods \cite{lin2018bsn,zeng2019graph,lin2019bmn,long2019gaussian,liu2019multi,xu2020g,xia2022learning,liu2022empirical,hsieh2022contextual,bao2022opental} have recently been proposed, competing to achieve state-of-the-art performance on selected benchmark datasets.
Whereas various new datasets have been recently proposed, %but 
the two most common relevant benchmarks 
%used for the evaluation of TAL methods 
remain ActivityNet 1.3 \cite{caba2015activitynet}, and Thumos-14 \cite{idrees2017thumos}. %In the past four years only, 
State-of-the-art performance on Thumos-14 has improved in four years by some 19\% \cite{zhu2022learning}.
%demonstrating significant dynamism.
%AB:Not sure if 'dynamism' is the right word here. Perhaps "recent advances"?

\begin{figure*}[h]
    \centering
    \includegraphics[width=0.99\textwidth]{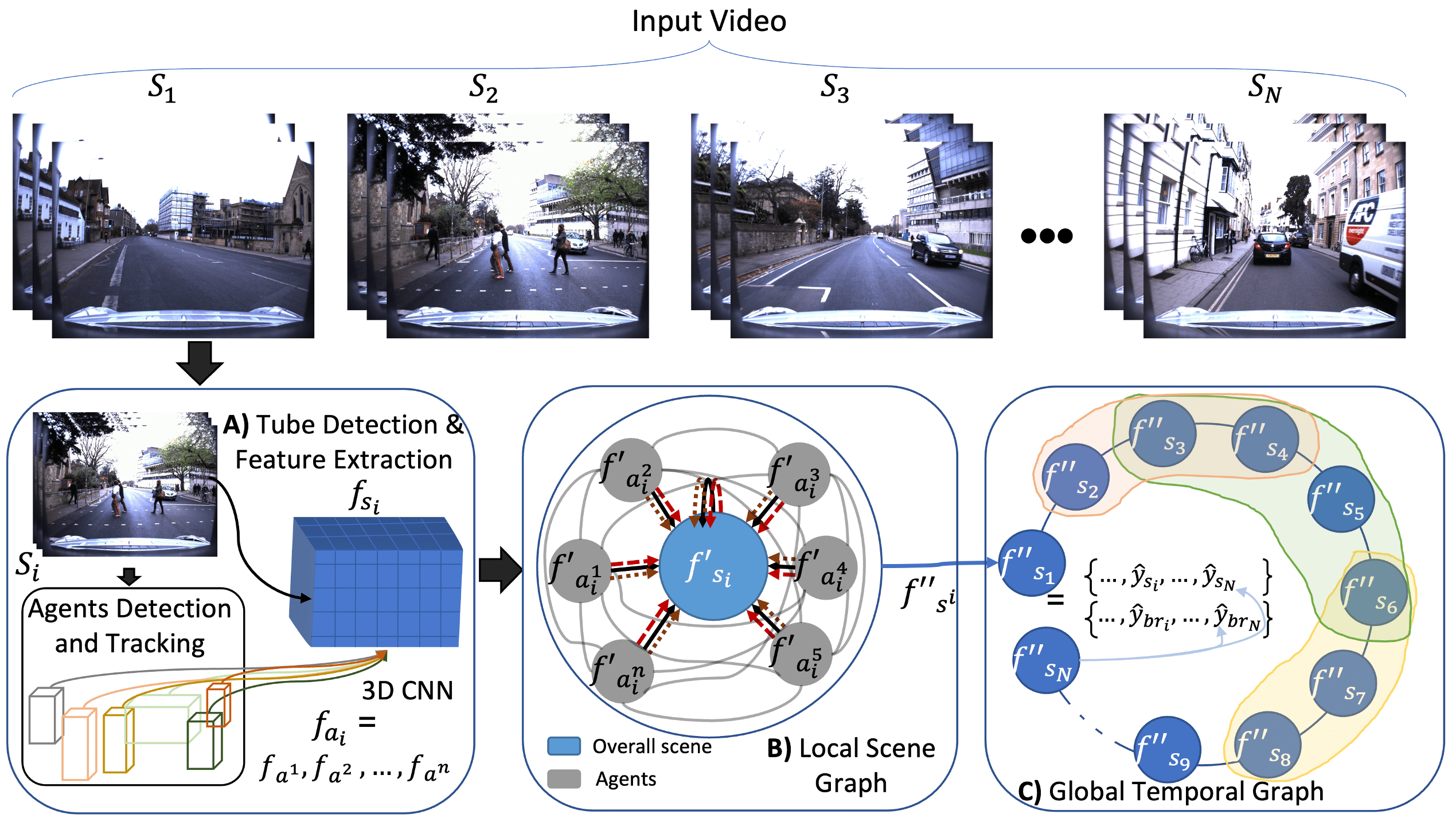}
    \caption{An overview of our \textbf{Comp}lex \textbf{A}ctivity \textbf{D}etection (\cad{}) framework. The input video is divided into fixed-size snippets $S_1,\ldots,S_N$ (top); each snippet is then processed in three major steps (bottom). A) Firstly, scene objects (agents) are detected and tracked throughout the snippet to form agent tubes. 3D features are then extracted from all the cropped agent tubes ($f_{a^i}$) as well as the local scene ($f_{s_i}$). B) Next, a fully-connected scene graph is constructed where agent nodes (in gray) are connected to each other and to the snippet node (in light blue). The local scene graph is processed using a graph attention
    network (GAT), resulting in intermediate scene features ($f''_{s_i}$). 
    %and the scene label ($y_{s^i}$). %FAB: I understand the label was not produced? Sal: yes
    C) {Finally, %the local scene feature of each snippet 
    all local scene features associated with individual snippets are temporally connected %to the next snippet 
    and processed using a global temporal graph to identify the boundaries of the activity %($y_{br^i}$) 
    using anchor proposals (shown in different colors). Each of the anchors represents a binary graph mask 
    %where the activity region is 1 while the background is 0.
    where snippets which belong to the activity receive label $y_{{br}_i} = 1$, 0 otherwise.
    }
    }
    % }
    % {\color{blue}FAB: i is used to index the local scene, so use a different notation for the activity; do you need an index at all? You could want to index the various temporal proposals for the sought activity, but the figure does not mention proposal generation at all. Also: what is $y_{br}$, a vector of two temporal instants? Finally, I understand the generated temporal proposals concern activity instances, which can overlap, while the local scenes remain about the snippets, which do not overlap. Is this correct? Lastly, use 1) 2) and 3) for the stages at the bottom of the figure to avoid confusion with the indexes used.}
    
    % the start ($y_{st^i}$) and end ($y_{ed^i}$) time of the activity,
    % as well as its label ($y_{s^i}$).}
    \label{fig:framework}
\end{figure*}

Almost all TAL approaches contemplate two major aspects: \emph{features/scene representation} and \emph{temporal localisation}.
In the scene representation stage, snippets (continuous sequences of frames) are processed to understand the local scene in the video.
Methods \cite{xia2022learning,liu2022empirical,hsieh2022contextual,bao2022opental} typically employ pre-extracted features obtained using a  sequential learning model (e.g., I3D \cite{carreira2017quo}), often pre-trained on the Kinetics \cite{kay2017kinetics} dataset. 
Such features are then processed, %in different ways, 
e.g., 
%by using 
via a temporal or semantic graph neural network, by applying appropriate feature encoding techniques 
%for learning purposes, 
or by generating temporal proposals, in an object detection style \cite{chao2018rethinking}.
In the second step of TAL, 
%is the {temporal localisation} of the action/activity. 
actions need to be temporally localised.
Existing methods 
%localise the action in various ways, including the 
vary in the way this is done, e.g. through temporal graphs~\cite{xu2020g,zeng2019graph}, boundary regression and proposals generation \cite{lin2018bsn,lin2019bmn,hsieh2022contextual} or encoder-decoder approaches \cite{zhu2022learning}. All such methods, however, are only applicable to short- or mid-duration actions that last for a few seconds ({e.g., a person jumping or pitching a baseball}).

\rev{
As recently pointed out in e.g. \cite{khan2021spatiotemporal,cheng2022tallformer}, in real-world applications a challenge is posed by \emph{complex activities}, longer-term events comprising a series of elementary actions, often performed by multiple agents.
For example, an Autonomous Vehicle (AV) negotiating a pedestrian crossing is engaged in a complex activity: First it drives along the road, then the traffic lights change to red, the vehicle stops and several pedestrians cross the road. Eventually, the lights turn green again and the AV drives off.}

\rev{Theoretically, TAL methods can be employed to temporally segment complex activities, in practice such approaches are only employed to detect short- or mid-duration actions lasting a few seconds at most ({e.g., a person jumping or pitching a baseball}). The activities contemplated by the most common datasets are of this nature.
%To distinguish complex activities from TAL, we present a comparison of datasets from both categories in 
Fig. \ref{fig:comp} compares two standard TAL benchmarks with the recently released ROAD dataset, explicitly designed for complex activity detection. %The comparison shows that 
Activities in the ROAD dataset last longer than those in ActivityNet or Thumos, with twice as many agents per snippet, making them more complex in nature.}

%{\color{red}[reasons for this limitation?]}
\rev{In this paper, we argue that standard TAL approaches are ill equipped to detect complex activities, as they fail to model both the global temporal structure of the activity and its fine-grained structure, in terms of the agents and elementary actions involved and their relations.}

%{\color{red}[formal definition of compAD and challenges]}
% {\color{red}[two versions of the task: with and without internal scene structure as output]}
\rev{We may thus define (strong) \emph{\textbf{Comp}lex \textbf{A}ctivity \textbf{D}etection} (\cad{}) as the task of recovering, given an input video, the temporal extent of the activities there present, \emph{as well as} their inner structure in terms of the agents or elementary actions involved. A weaker \cad{} is one in which the only expected output is the temporal segmentation,
with the internal structure of the activity estimated as a means for achieving segmentation - with no annotation available.}

%whereas the internal structure of the activities is estimated as a means for achieving segmentation but no annotation is available for it.

%{\color{red}[few exceptions exist]}
\rev{While a small number of studies have attempted to detect complex, long-duration activities \cite{khan2021spatiotemporal,cheng2022tallformer,shou2016temporal}, %but with limited success. %However, these approaches {mostly focus on activities performed by single actor}. {\color{red}[Sounds pretty vague and weak]}
%To the author's knowledge, 
to our knowledge \cite{skhan2021comp} is the only existing study which attempts to tackle \cad{} as defined above.
%complex activities, performed by multiple actors over an extended period of time.} %Further down the line, a long-term action localisation method has been proposed in 
%{\color{red}[however, they also have strong limitations]}
The work, however, relies upon the availability of %specifically-designed and 
heavily-annotated datasets which include granular labels for the individual actions which make up a complex activity, and the corresponding frame-level bounding boxes.
 %is dependent upon individual action annotations.}
This is a serious limitation, for it prevents  \cite{skhan2021comp} from being usable for pure temporal segmentation and compared with prior art there. 
Further, \cite{skhan2021comp} focuses only on the graph representation of snippets, neglecting the long-term modelling of complex activities.}

\begin{figure}[h]
    \centering
    \includegraphics[width=0.75\textwidth]{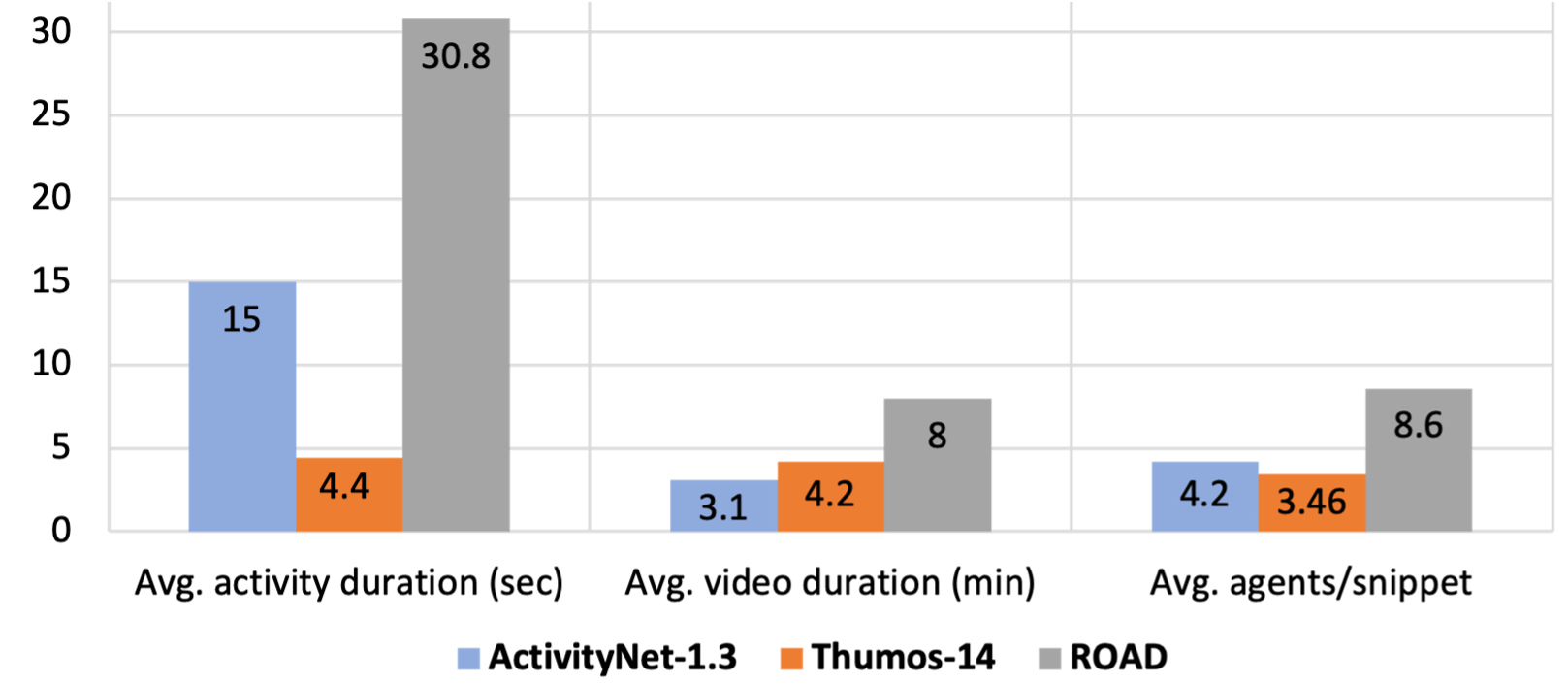}
    \caption{\rev{ Comparing TAL datasets (ActivityNet-1.3 and Thumos-14) with a CompAD dataset (ROAD) in terms of average activity and video durations and mean number of agents per snippet.}
    %\textcolor{cyan}{elhoseiny: what is the x-axis}
    %Sal: I updated the caption. I hope it answers your %question or do you want me to elaborate more
    }
    \label{fig:comp}
    \vspace{-2mm}
\end{figure}

\subsection{Objectives}
This work aims to push the boundaries of temporal action localisation towards full-blown \emph{\textbf{Comp}lex \textbf{A}ctivity \textbf{D}etection} (\cad{}), intended as the ability to detect long-term, complex activities, % \AB{from any generic action / activity dataset which includes temporal annotation.} 
leveraging datasets providing temporal segmentation annotation \emph{only}.
We do so by modelling and leveraging a complex video activity's internal structure, but
without resorting to any additional fine-grained annotation concerning individual actions. 
%The processing of a video is undertaken snippet by snippet.

Our \textbf{proposed framework} (Fig. \ref{fig:framework}) is composed of three stages: A) 
%the detection of agent tubes and extraction of features; 
feature extraction;
B) a scene graph attention network designed to learn the importance of each active object (`agent') within the local dynamic scene; and C) a temporal graph of attended 
scene graphs for the localisation of complex activities of arbitrary duration.
% Input videos are processed snippet by snippet.

%For feature extraction, in opposition to current TAL methods, we first detect the objects inside the snippet and track them using a tracker to build an \emph{agent tube} then use the pre-trained model to extract features from both the cropped tubes and whole snippet. Using such a building of agent tubes will allow the researcher to train the model on any dataset having a temporal annotations. To represent the local scene in a meaningful way these features are connect to each other using a scene graph for extracting a feature representation. 

Our {feature extraction} scheme (A) differs from what typically done in %existing methods 
prior art
- where spatiotemporal features are extracted from whole snippets only. In contrast, we first detect the relevant active objects in the scene and track them within each snippet to build for each of them an \emph{agent tube} (a series of related bounding boxes).
A pre-trained tracker is used to allow the method to be deployable to any datasets with only temporal annotation (no bounding box annotation for the scene agents is required), while enabling a finer-grained description of the internal structure of a complex activity.
%The series of bounding boxes from each 
Agent tubes are resized to a standard spatial size to create `mini videos' from which features can be extracted. 
A pre-trained 3D feature extraction model is then used to 
%extract features from 
encode both the `cropped' tubes and the whole snippet.

To suitably represent the local dynamic scene within each snippet, 
a \emph{local scene graph} (B) is constructed using three different topologies including a star graph, star graph with same agent label, and a fully-connected
% constructed in which all the agent nodes are connected to the snippet node (representing the local scene) and to each other 
(the effects of all three topologies are explained in Section \ref{sec:ablation} - Ablation study). The local scene graph is then processed using a \emph{scene graph attention (SGAT) network} for extracting the overall scene representation. The rationale behind using the SGAT is its ability to compute the importance of each node in the context of its neighbors, thus modelling the structure of complex scenes.

Finally (C), the learned local scene graphs are connected to each other by constructing a \emph{temporal graph}, with the aim of recognising the activity label and identifying its temporal boundaries (start and end time) {in a class-agnostic manner.}
Each node in the temporal graph learns two things; differentiating between {the background and the activity region and recognising the activity label}. Such a strategy enables us to model and detect both short and long-duration activities.
% {\color{blue}FAB: is this paragraph still accurate/up to date? This is the place where you should be talking about the new regression strategy/approach. This could be added as third bullet point to the list of contributions below.} salman - I think we don't need to focus more on the region proposal as it has been done by many methods, and we are generating these proposals after our temporal graph.

\subsection{Contributions} 
Our main contributions are, therefore:
\begin{itemize}

\item 
A  
%hybrid scene and temporal graph framework 
hybrid graph network approach for general complex activity detection, comprising a local scene graph as well as a global temporal graph, capable of localising both long and short-term activities.
%comprising:\elhoseiny{two level itemization can be too much, perhaps break this into two points?}
\iffalse
\begin{itemize}
\item
A scene graph activation network for learning the importance of each agent in the context of a (local) dynamic scene. 
\item 
A temporal graph of activated scene graphs for the detection of the start and end of an activity of arbitrary duration.
\end{itemize}
\fi
\item

A  feature extraction strategy for 
fine-grained temporal activity localisation
%temporal action localisation 
without the need for additional annotation, consisting in detecting and tracking the relevant scene agents to construct tubes, followed by feature extraction from both the whole scene and each agent tube individually, as opposed to existing methods which only rely on snippet-level features. %\elhoseiny{I think it better to mention code release once, including the feature once. }
\item

\rev{Comprehensive experiments showing how our approach leveraging weak \cad{} \emph{outperforms the most recent TAL state-of-the-art across the board} on both Thumos-14 and the recent ROad event Awareness Dataset (ROAD) \cite{singh2022road,skhan2021comp}, which portrays long-duration road activities involving multiple road agents over sometimes several minutes, while being extremely competitive on the standard ActivityNet 1.3 benchmark, showing a clear dominance on classical TAL approaches as the duration and the complexity of the activities increase.}
\end{itemize}

% We also run tests on the recent ROad event Awareness Dataset (ROAD) \cite{singh2022road,skhan2021comp}, which portrays long-duration road activities involving multiple road agents over sometimes several minutes, to show the effectiveness of our method in temporally localising long-duration activities.

\paragraph{Related publications:}
The work presented in this chapter is a preprint ArXiv. The dissertation author is the primary investigator of this research work. Our code is also available online\footnote{\url{https://anonymous.4open.science/r/HG_CompAD-7697/README.md}}.

\paragraph{Outline:}\label{online:hgraphoutline}
The remainder of the Chapter is organised as follows.
The proposed methodology of our framework is presented in Section ~\ref{hgraph:sec:overview} while the main building blocks including tube detection and features extraction in Section \ref{sec:feature-extraction}, scene graph attention in \ref{sec:scene-graph-attention}, temporal graph localisation in \ref{sec:temporal-graph}, and loss functions in \ref{sec:loss}, respectively. Further, the detailed experimental evaluation of the proposed method is reported in Section \ref{hgraph:sec:exp}. Finally, a summary with limitations of the proposed method is given in Section~\ref{hgraph:ssec:summary}.

% \begin{figure*}[h]
%     \centering
%     \includegraphics[width=0.99\textwidth]{figures/hgraph/Act_sing.png}
%     \caption{Visualisation of our agent detection and tracking stage using a bird’s-eye view of the spatiotemporal volume corresponding to a video segment of \emph{ActivityNet-1.3} dataset. In this figure, we present examples of activities performed by a single actor.}
%     \label{fig:act_sing}
% \end{figure*}

\begin{figure*}[h]
    \centering
    \includegraphics[width=0.85\textwidth]{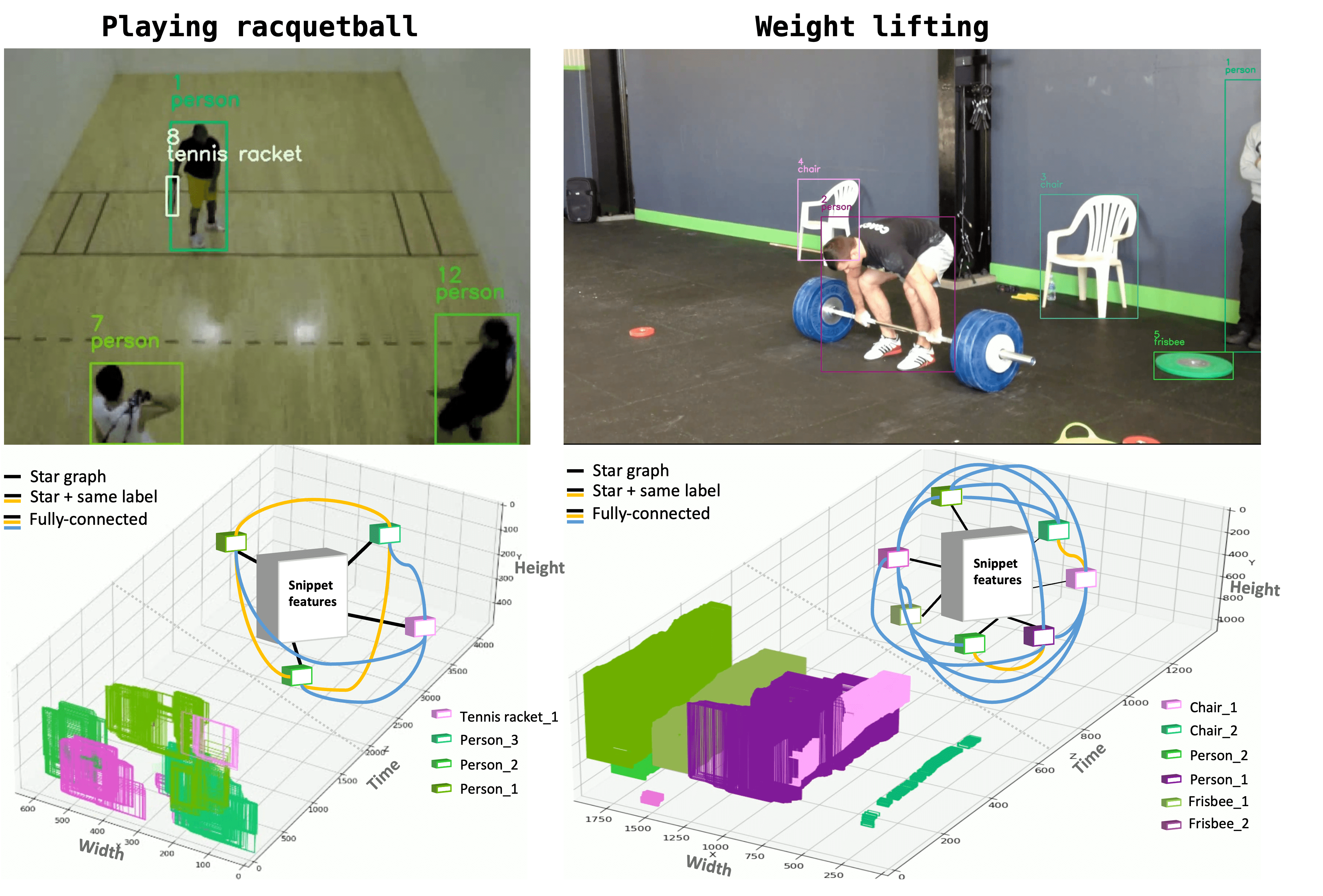}
    \caption{\rev{Visualisation of our agent detection and tracking stage using a bird’s-eye view of the spatiotemporal volume corresponding to a video segment of \emph{ActivityNet-1.3} dataset. The first half of the graph shows the plotting of agent tubes while the second half illustrates the scene graph of the detection agents. In this figure, we show examples of activities performed by multiple actors (agent tubes).} }
    \label{fig:act_multi}
\end{figure*}

\section{Proposed Method}
\label{hgraph:sec:overview}

The proposed framework, outlined in the Introduction, is illustrated in Fig. \ref{fig:framework}. An input untrimmed video $V$ is divided into $N$ snippets $S$ = $S_1$, ..., $S_i$, $S_{i+1}$, ..., $S_N$ (each snippet is a pre-defined constant length of consecutive frames). %FAB: wanna say constant length? - sal: yes, changed it
Each snippet is then passed to the tube detection and feature extraction module which returns a feature vector for both the snippet $f_{s_i}$ and the individual agent tubes $f_{a^1_i}$, $f_{a^2_i}$,...,$f_{a^n_i}$, where $n$ is the number of agents present in snippet $i$. 
% {\color{blue}FAB: please use the same notation for individual agent tubes in Figure 1}
These features are then forwarded to the local scene graph attention layer for learning the attention of each agent in the context of its neighbours. This returns an aggregate feature representation for the whole scene ($f''_{s_i}$). These aggregate local scene features for all the snippets, $f''_{s_1}$,$f''_{s_2}$,...,$f''_{s_N}$, are then connected
using a global temporal graph for the generation of the activity class label $\hat{y}_{s_i}$ and {activity boundary labels $\hat{y}_{br_i}$ using anchor proposals in a class-agnostic manner.} 
% of the start $y_{st_i}$ and end $y_{ed_i}$ labels.
% {\color{blue}FAB: the last sentence needs to be updated for the new regression approach}

\subsection{Tube Detection and Feature Extraction} \label{sec:feature-extraction}

As mentioned, one of the contributions of this chapter is a new strategy for feature extraction and representation which consists in analysing the finer structure of the local dynamic scene, rather extracting features from whole snippets only.  

\textbf{Objects Detection and Tracking}. We first detect scene objects in each frame of the snippet using an object detector pre-trained on the COCO dataset \cite{lin2014microsoft}, which comprises of 80 different types of objects. However, we select a relevant list of object types (termed \emph{agents}) which depends on the dataset (the lists of classes selected for each dataset is given in Section \ref{sec:implem} -- Implementation Details). The detected agents are then tracked using a pre-trained tracker throughout the snippet in order to construct agent {tubes}. 
%These agent tubes 
The latter are of variable length, depending on the agent itself and its role in each snippet. A pictorial illustration of agent tubes in an example video segment from all three datasets, showing both a sample frame, with overlaid the detection bounding boxes, and a bird’s-eye view of the agent tubes and a local scene graph for the snippet it belongs to.

\textbf{ActivityNet-1.3} is one of the largest temporal action localisation datasets which includes activities performed by both individual and multiple objects (agents). Therefore, we illustrate the example of multi-agent activities in Fig. \ref{fig:act_multi}.

\textbf{Thumos-14} covers different types of sports actions performed by single or mostly multiple agents. Two sample actions; baseball pitch and soccer penalty are visualised in Fig. \ref{fig:thumos}. In the figure, it can be seen that our model construct the seen graph of almost all agents despite the low-resolution videos.

\begin{figure*}[h]
    \centering
    \includegraphics[width=0.80\textwidth]{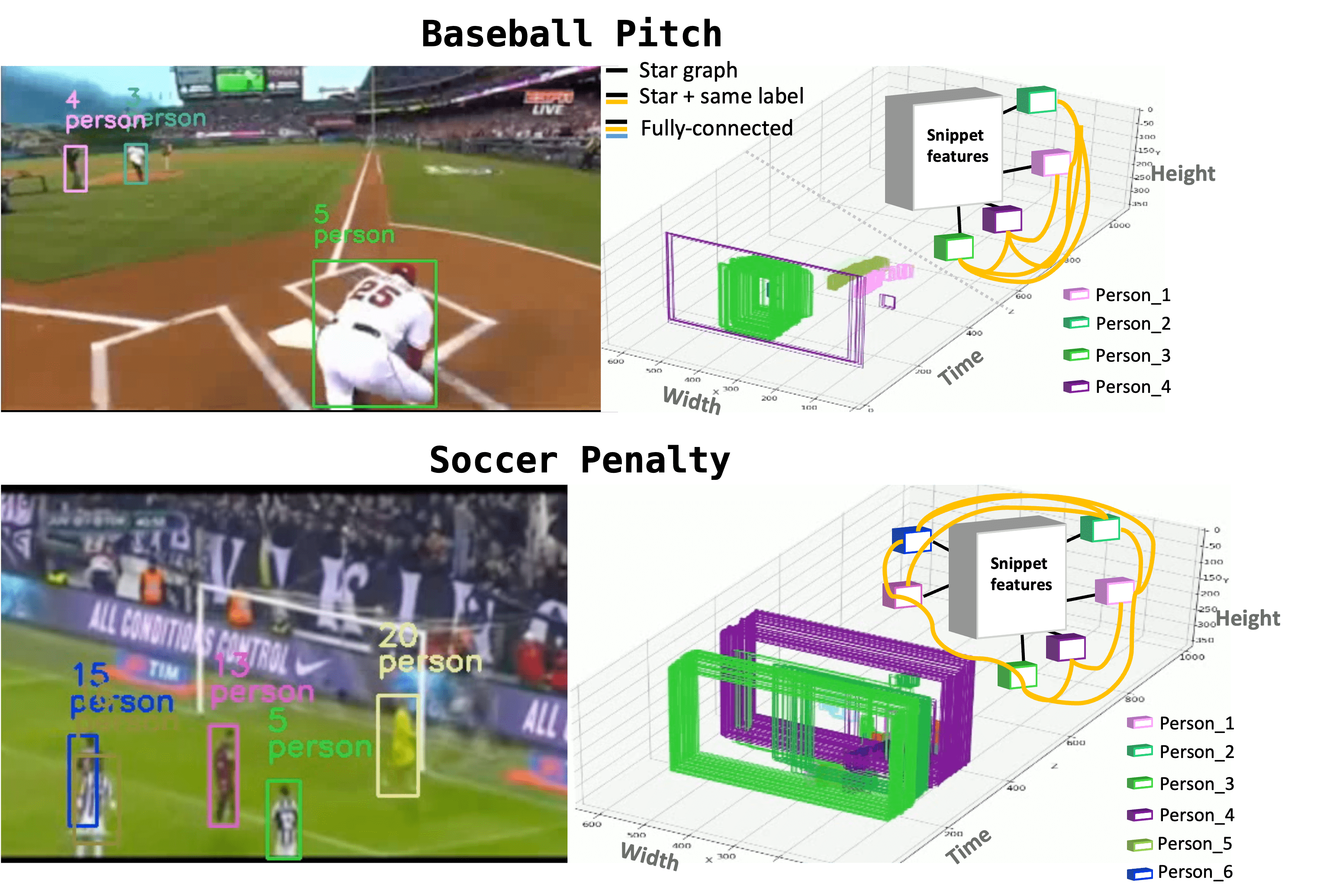}
    \caption{\rev{Visualisation of our agent detection and tracking stage using a bird’s-eye view of the spatiotemporal volume corresponding to a video segment of \emph{Thumos-14} dataset. The first half of the graph shows the plotting of agent tubes while the second half illustrates the scene graph of the detection agents. Both of the examples show the activities performed by multiple agents (recognising sports activities).}}
    \label{fig:thumos}
\end{figure*}

\textbf{ROAD}. The activities in the road dataset are mostly performed by a large number of agents, however, there are a few cases where the number of detected agents is less in number as shown in Fig. \ref{fig:road} (above). On the other hand, in Fig. \ref{fig:road} (below) we also show an example of a night video where the activity is performed by a large number of agents moving very promptly.

\begin{figure*}[h]
    \centering
    \includegraphics[width=0.95\textwidth]{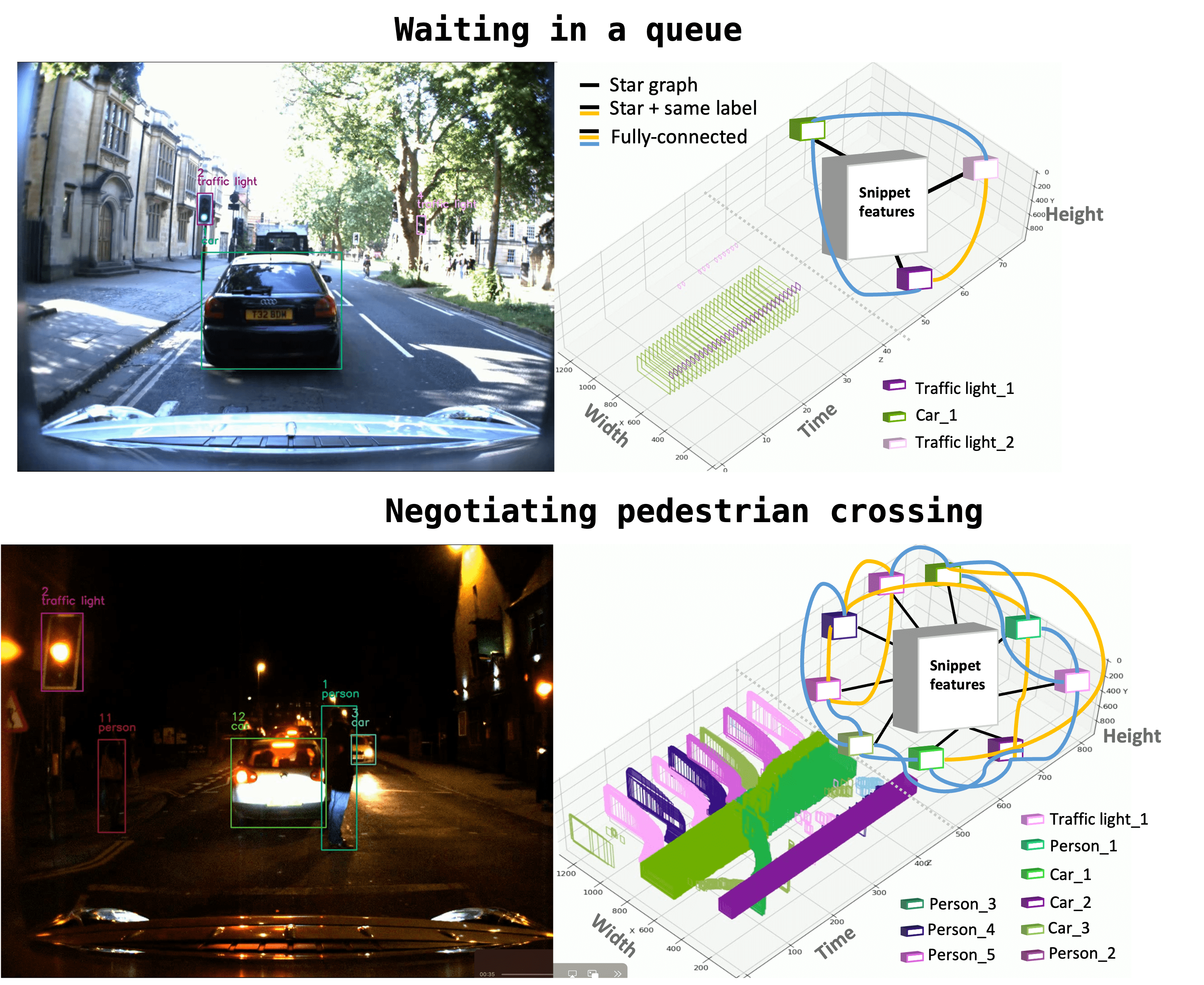}
    \caption{\rev{Visualisation of our agent detection and tracking stage
using a bird’s-eye view of the spatiotemporal volume corresponding to a video segment of the \emph{ROAD} dataset. The upper section
shows an example of activity performed by only three agents.
Below is the example of a night video where the activity is performed by multiple agents. The first half of the graph shows the plotting of agent tubes while the second half illustrates the scene graph of the detection agents.}}
    \label{fig:road}
\end{figure*}

\textbf{Feature Extraction}. Next, all the detected agent tubes are 
%cropped in both spatial and temporal domains 
brought to a standard size
and passed to the pre-trained 3D CNN model along with the whole snippet for spatiotemporal feature extraction. The adopted 3D CNN model allows variable length inputs, and outputs a fixed-sized feature vector for each of the agent tubes and the snippet. 

% \begin{figure}[h]
%     \centering
%     \includegraphics[width=0.95\textwidth]{figures/hgraph/tracking_scene_g.png}
%     \caption{{ Visualisation of our agent detection and tracking stage using a bird's-eye view of the spatiotemporal volume corresponding to a video segment of the ROAD dataset. The upper section shows a random frame from the segment (with bounding boxes). Below, the detected agent tubes are plotted in space and time together with different possible local scene graph representations. The agent tubes we extract from the scene are of variable sizes. The way scene object motion affects the spatial and temporal extent of the tubes can be appreciated. Additional scenarios with visualisation are illustrated in the \textbf{Supplementary material.}}}
%     \label{fig:tracking}

% \end{figure}

% \begin{figure}[h]
%     \centering
%     \includegraphics[width=0.50\textwidth]{cvpr2023_complex_activity/figs/tracking_scene_g_.png}
%     \caption{Visualisation of our agent detection and tracking stage using a bird's-eye view of the spatiotemporal volume corresponding to a video segment of the ROAD dataset. The left side shows a random frame from the segment (with bounding boxes). Right is the 3D plotting of the detected agent tubes in space and time. The agent tubes we extract from the scene are of variable sizes. The way scene object motion affects the spatial and temporal extent of the tubes can be appreciated.}
%     \label{fig:tracking}
% \end{figure}

\begin{figure*}[h]

    \centering
    \includegraphics[width=0.99\textwidth]{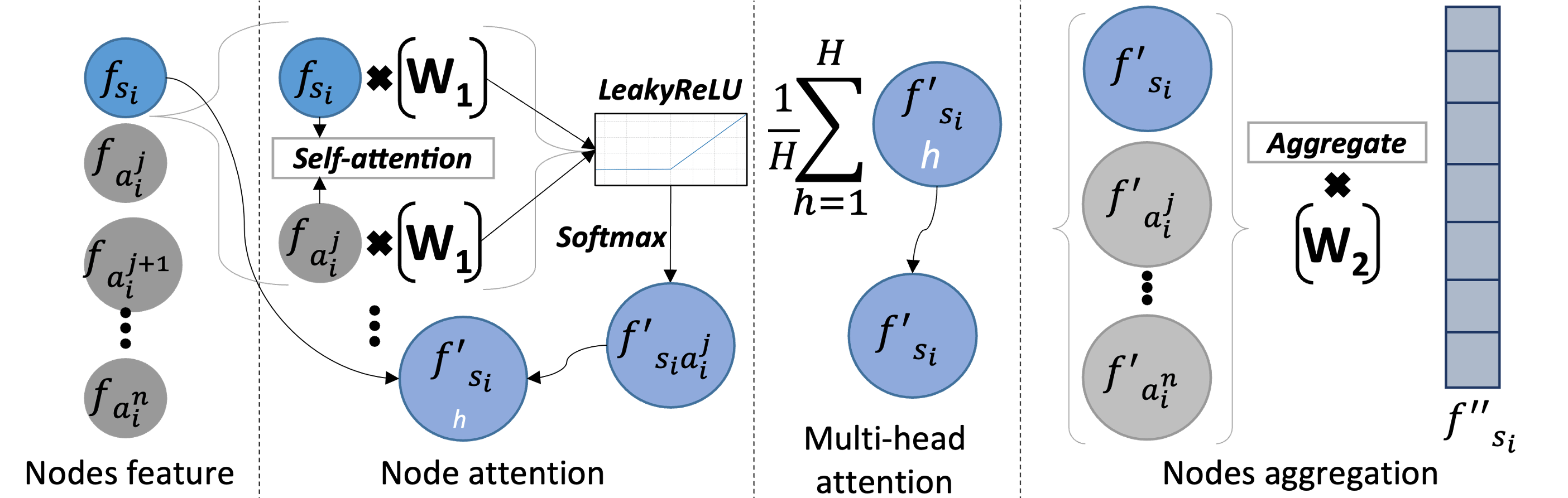}
    \caption{Our scene graph attention layer (Stage B of our approach, Fig. \ref{fig:framework}) takes the node features generated in the feature extraction Stage A as an input, and updates each node's features using node attention with respect to its all connected neighbours. Multi-head attention with $H$ heads is applied to each node to further robustify the representation -- averaging yields the final node features. To obtain a fixed-size overall representation for a specific local scene (snippet), the features of all its nodes are aggregated using a learnable weight matrix $W_2$. 
    % {\color{blue}FAB: the same notation $W$ is used in two different places here, please differentiate using $W_1$ and $W_2$ as in the text. Also the notation for the agent features has not been updated to reflect the text in Sec 3.1. In general make sure the notation is consistent throughout.}
    }
    \label{fig:sgat}

\end{figure*}

\subsection{Scene Graph Attention} \label{sec:scene-graph-attention}

In a second stage, a scene graph representation is used to describe the scene in terms of features extracted from both the overall snippet and the agent tubes.

\textbf{Graph Generation}. {The scene graph is constructed using three different topologies: a star graph connecting each agent node to the scene node, a star topology with also links between agents sharing a label, and a fully-connected one. 
The influence of topology is shown in Sec. \ref{sec:ablation}.}

\textbf{Graph Attention}. As mentioned, our scene graph attention layer is inspired by the GAT concept \cite{velivckovic2017graph}, originally designed for node classification. While we follow a similar attention mechanism, here we amend the attention layer in order to extract aggregate features from all the nodes, to be passed in turn to our localisation layer in the third stage. 

The workflow of our scene graph attention layer is shown in Fig. \ref{fig:sgat}. All input node features are linearly transformed using a {weight matrix} $W_1$, followed by \emph{self-attention} to find the importance of each node with respect to its connected neighbours. An activation function LeakyReLU is applied to the features for nonlinearity, resulting in final output features for all of the nodes. These are then normalised using a softmax function to make each node representation ($f'_{{s_i}{a}^j_i}$) comparable with that of all the nodes connected to it ($f'_{s_i}$).

The {self-attention} process is further improved by applying a \emph{multi-head attention} strategy inspired by 
%the success of 
transformers \cite{velivckovic2017graph}.
Attention is applied to the node features individually. For each node, the average over the $H$ heads is computed, resulting in an attended feature vector for each node.  

Finally, to get a fixed-size representation for the whole scene, the output features of all the nodes are aggregated using another learnable weight matrix $W_2$, which outputs the final feature representation $f''_{s_i}$ for the whole (local) scene. 

\subsection{Temporal Graph Localisation}
\label{sec:temporal-graph}

In Stage C of our framework, activity recognition and localisation are performed using a GCN. The final features from all the local scenes  ($f''_{s_1}$, ..., $f''_{s_i}$,$f''_{s_{i+1}}$,...,$f''_{s_N}$), as outputted by the scene graph attention layer, are temporally connected to build a global temporal graph (see Fig. \ref{fig:framework}, C).

Our GCN network for processing the global temporal graph is divided into two parts. The first part consists of three 1D convolutional layers designed to learn the temporal appearance of all the local scenes with boundaries. Each convolutional layer is followed by a sigmoid activation for non-linearity. The second part generates the anchor proposals of the temporally learned features via pre-defined anchors, where each of the anchors acts as a binary mask over the whole graph.
% {\color{blue}FAB: is there anything special about this GCN architecture with proposal generation? This is the only paragraph about architecture, and it has not changed since last time, which is puzzling.}

Overall, the GCN module provides two different outputs. i) \textit{Activity classification}: the list of predicted activity labels for all the snippets in the video ($\hat{y}_{{s}_1}$, ..., $\hat{y}_{{s}_i}$, ..., $\hat{y}_{{s}_N}$) is produced, where the dimensionality of the output vector $\hat{y}$ is equal to the number of classes. ii) \textit{Activity localisation}: activities are localised using binary masked class-agnostic anchor proposals. The Intersection over Union (IoU) measure between each anchor and the ground truth (true temporal extension of the activity) is computed, and the anchors with maximum IoU
% \AB{above what threshold?}
are selected to train the model for localising the boundaries of any activity, regardless of its activity label. 
The final output of our temporal graph is a one-hot binary vector  ($\hat{y}_{{br}_1}$, ..., $\hat{y}_{{br}_i}$, ..., $\hat{y}_{{br}_N}$) for each series of snippets (video), where 
$\hat{y}_{{br}_i} = 1$ iff snippet $S_i$ belongs to the activity, %while is equal to 
$ = 0$ when the snippet does not belong to it.
%each of the element in vector represents 0 as a background and 1 as activity region from start till the end.
% {\color{blue}FAB: the last sentence is totally unclear/badly written - also, in Fig 1 you say that $y_{br}$ is a vector?}
% ($y_{{ed}^1}$, ..., $y_{{ed}^{i}}$,...,$y_{{ed}^N}$)
% marking whether the snippet coincides with the end of an activity.

\subsection{Loss Functions}
\label{sec:loss}

\rev{Our problem is multi-objective, as we aim at not only recognising the label of the activity taking place but also finding its boundary (start and end time). Therefore, we combined two losses one for learning the class label (classification) and the other for finding the start and end time (regression). The details of both losses are as follows:}

Given the ground truth 
$y_s = \{ y_{s_1}, \ldots, y_{s_i}, \ldots, y_{s_N} \}$
\\and predicted
$\hat{y}_s = \{ \hat{y}_{s_1}, \ldots, \hat{y}_{s_i}, \ldots, \hat{y}_{s_N} \}$ activity labels,
and the ground truth
\\ $y_{br} = \{ y_{{br}_1}, \ldots, y_{{br}_i}, \ldots, y_{{br}_N} \}$
and predicted
$\hat{y}_{br} = \{ \hat{y}_{{br}_1}, \ldots, \hat{y}_{{br}_i}, \ldots, \hat{y}_{{br}_N} \}$
temporal extent labels, our overall loss function is formulated as:
\begin{equation} %\label{eq:combine}
    \mathcal{L} = \lambda \cdot \mathcal{L}_{Act} + \mathcal{L}_{Br}.
\end{equation}
The second component denotes the binary cross entropy (BCE) loss\\
$\mathcal{L}_{Br} = \{l_1,..., l_i,...,l_N\}^T$, where
\[
\begin{array}{l}
    l_i = - w_i \Big [ {y_{{br}_i} \cdot \log \hat{y}_{{br}_i}} + (1 - {y_{{br}_i}) \cdot \log(1 - \hat{y}_{{br}_i}}) \Big ],
\end{array}
\]
%a binary classification loss 
driving the recognition of the activity's temporal extent in class-agnostic manner.

$\mathcal{L}_{Act}$ denotes, instead, the \emph{BCEWithLogitsLoss} \cite{BCEWithLogitsLoss},
defined as \\
$\mathcal{L}_{Act} = \{ l_{1,c}, \ldots, l_{i,c}, \ldots, l_{N,c}\}^T$, where
\begin{equation} %\label{eq:scene_loss}
\begin{aligned}
    l_{i,c} = - w_{i,c} \Big [ p_c \cdot y_{s_i,c} \cdot \log \sigma(\hat{y}_{s_i,c}) \\ + (1 - y_{s_i,c} \cdot \log (1 - \sigma(\hat{y}_{s_i,c})) \Big ].
\end{aligned}
\end{equation}
Here $i$ is the index of the sample in the batch, $c$ is the class number, $p_c$ is the weight of a positive sample for class $c$, and $\sigma$ is the sigmoid function.
This loss component is used to predict the activity labels $\{y_{s_i}\}$, and combines  classical binary cross entropy %Loss (BCELoss) 
with a sigmoid layer \cite{BCEWithLogitsLoss}. This combination is proven to be numerically stronger (as it leverages the log-sum-exp trick) than using a plain sigmoid separately followed by a BCE loss.

Finally, $\lambda$ is a weight term, which we set to the number of selected anchor proposals (128 in our case).

\section{Experiments} \label{hgraph:sec:exp}

\subsection{Datasets} \label{sec:datasets}

To evaluate the performance of our proposed method, we used three benchmark datasets including ROAD, Thumoas-14, and ActivityNet-1.3. Firstly, ROAD is used to compare our method with state-of-the-art CompAD methods. Next, Thumos-14 and ActivityNet-1.3 are used to show the effectiveness of our method over temporal action localisation methods. The complete detail of all these three datasets can be found in Section \ref{related_work:sec:datasets}.

\subsection{Implementation Details} \label{sec:implem}

\textbf{Evaluation metrics}. In our experiments, mean Average Precision (mAP) was used as an evaluation metric (explained in Section \ref{related_work:sec:evalmat}), using different IoU thresholds for the different datasets. According to the official protocols for the various benchmarks, the following lists of temporal IoU thresholds were selected: $\{ 0.1,0.2,0.3,0.4,0.5 \}$ for \\ROAD, $\{ 0.3,0.4,0.5,0.6,0.7 \}$ for Thumos-14 and $\{ 0.5, 0.7, 0.95 \}$ for ActivityNet-1.3.

\textbf{Feature extraction}. Firstly, the agent tubes are constructed by detecting scene objects using a YOLOv5 detector \cite{glenn_jocher_2022_6222936} pre-trained on the COCO dataset. Detections are then tracked throughout a snippet using DeepSort \cite{yolov5deepsort2020}. Then, features are extracted using an I3D network pre-trained on the Kinetics dataset \cite{kay2017kinetics}, from both the entire snippet and each cropped agent tube individually. The object categories were reduced to six for the ROAD dataset to only cover the agents actually present in the road scenes portrayed there. As the other two datasets (ActivityNet-1.3 and Thumos-14) are general purpose, in their case we retained all the 80 classes present in the COCO dataset.

\textbf{Scene Graph Attention}. The local scene graph was generated by producing a list of tuples $[(0, 1), (0, 2), ...]$, where the first index relates to the source node
and the second number indexes the target node. The reason for preferring this structure over an adjacency matrix was to limit memory usage. For node attention learning, {we initialised our architecture using the weights of the GAT model \cite{velivckovic2017graph} pre-trained on the PPI dataset \cite{zitnik2017predicting} and} applied 4 attention layers with \{4, 4, \emph{num of classes}, and \emph{num of classes}\} heads, respectively. The number of classes \emph{num of classes} is equal to 201, 21, and 7 in ActivityNet-1.3, Thumos-14 and ROAD, respectively.

The 
\textbf{Temporal Graph} is a stack of three 1D convolution layers on the final representation of the temporally connected local scenes. The size of the input to the first convolutional layer is the number $N$ of local scenes (snippets), 
%in the  fixed duration input video , 
multiplied by the number of heads in the last attention layer.

The length of our temporal graph is fixed to $N$. Videos with number of snippets less than or equal to $N$ are passed directly to the temporal graph; %while the videos with larger lengths 
longer videos
are split into multiple chunks containing $N$ snippets each.
% {\color{blue}Please explain how you deal with videos of different lengths, both at training and at testing time.}
The output is a one-hot vector of activity labels of size $N$ and %a fixed-sized array of binary mask anchor proposals.
a collection of 128 proposals (binary graph masks) also of length $N$.

\textbf{Loss Functions}. Our problem is multi-objective, as we aim at not only recognising the label of the activity taking place but also finding its boundary (start and end time). Our overall loss function is thus the weighted sum of \emph{BCEWithLogitsLoss} \cite{BCEWithLogitsLoss} (for activity classification) and standard binary cross entropy (for temporal localisation). Full details 
%about the loss functions 
can be found in the Section \ref{sec:loss}.

\subsection{Comparison with State-of-the-Art}

\textbf{ROAD}. \rev{To validate our \cad{} approach, we first compared ourselves with the prior art available for the ROAD dataset \emph{and} re-implemented the current state-of-the-art TAL methods; TallFormer \cite{cheng2022tallformer} and ActionFormer \cite{zhang2022actionformer} there 
%for ROAD dataset 
(see Table \ref{table:road}). The standard IoU thresholds for mAP computation on ROAD range from 0.1 to 0.5. The detailed class-wise average precision of our methods is visualised in Fig. \ref{fig:road-class}.}

\setlength{\tabcolsep}{4pt}
\begin{table}[h!]
\begin{center}
\caption{\rev{Comparing our approach with the state-of-the-art methods for \cad{} available on the ROAD dataset. The mAP at the various standard thresholds is reported. Best results are in \textbf{bold} and second best \underline{underlined}.} }
\label{table:road}
{ \begin{tabular}{lllllll}
\hline\noalign{\smallskip}
 &  \multicolumn{6}{c}{ROAD}\\
Methods $\qquad\qquad$ & 0.1 & 0.2 & 0.3 & 0.4 & 0.5 & Avg\\
\noalign{\smallskip}
\hline
\noalign{\smallskip}
P-GCN \cite{zeng2019graph}  &60.0 & 56.7 & 53.9 & 50.5 & 46.4& 53.5\\
G-TAD \cite{xu2020g}   &62.1 & 59.8 & 55.6 & 52.2 & 48.7& 55.6\\
STDSG \cite{skhan2021comp} &  77.3 & 74.6 & 71.2 & \underline{66.7} & \underline{59.4} & \underline{69.8} \\
TallFormer \cite{cheng2022tallformer} & \underline{78.4}	& \underline{74.9}	& 70.3	& 63.8	&57.1&	68.9 \\
ActionFormer \cite{zhang2022actionformer} & 76.5	& 73.7	& \underline{72.6} &	64.4	&58.2&	69.0 \\
\textbf{Ours} & \textbf{82.1} & \textbf{77.4} & \textbf{73.3} & \textbf{69.5} & \textbf{62.9} & \textbf{73.0}\\
\hline
\end{tabular}}
\end{center}
\end{table}

\begin{figure}[h!]
    \centering
    \includegraphics[width=0.9\textwidth]{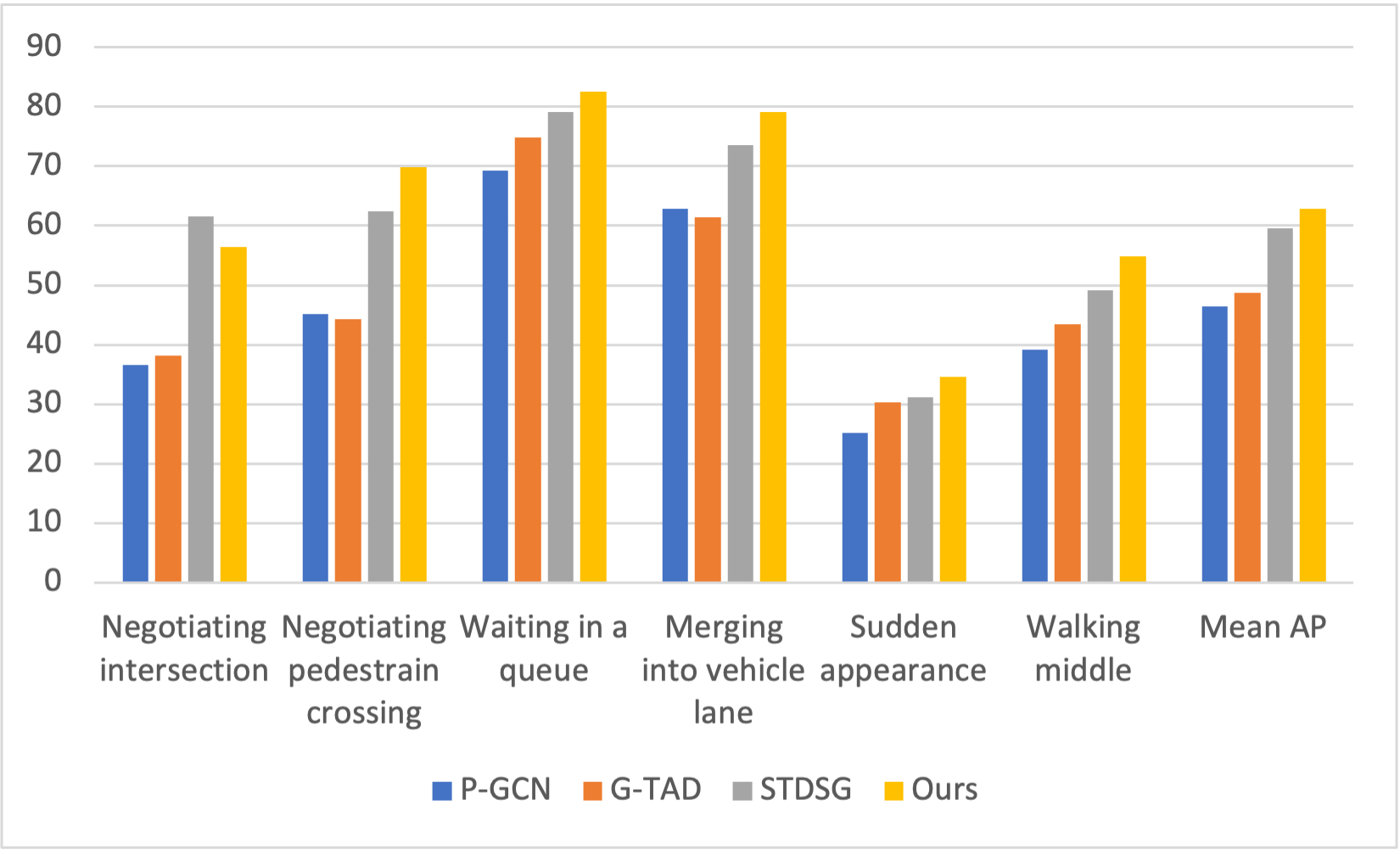}
    \caption{\rev{Class-wise average precision (\%) with IoU threshold of 0.5 for ROAD dataset. \label{fig:road-class}}}
\end{figure}

\textbf{Thumos-14}. \rev{ Our hybrid graph approach is compared with the state of the art on Thumos-14 in Table \ref{table:thum}. 
%For the comparison on Thumos-14, 
We followed the standard evaluation metric there: mAP with IoU thresholds ranging from 0.3 to 0.7 and their average. 
%The methods we used for comparison were published between 2018 and 2022 in various top computer vision 
Competitor methods were published in top vision
conferences in 2018-2022. 
%It can be noted from the results that the recent state-of-the-art method 
The recent ActionFormer \cite{zhang2022actionformer} approach achieves the highest average mAP 66.8.  The detailed class-wise average precision of our method is visualised in Fig. \ref{fig:road-class}.}

\setlength{\tabcolsep}{4pt}
\begin{table}[h!]
\begin{center}
\caption{\rev{Activity detection performance comparison with state-of-the-art methods on Thumos-14. mAP values at different IoU thresholds are reported for the test Thumos-14. Best results are in \textbf{bold} and second best \underline{underlined}.} }
\label{table:thum}
\begin{tabular}{llllllll}
\hline\noalign{\smallskip}
 & & \multicolumn{6}{c}{Thumos-14} \\
Methods & Conference & 0.3 & 0.4 & 0.5 & 0.6 & 0.7 & Average \\
\noalign{\smallskip}
\hline
\noalign{\smallskip}
BSN \cite{lin2018bsn} &ECCV'18  &53.5 & 45.0 &36.9 &28.4 &20.0 &36.8\\

P-GCN \cite{zeng2019graph} & ICCV'19 & 63.6& 57.8& 49.1& —& —& —\\
BMN \cite{lin2019bmn} &ICCV'19 & 56.0& 47.4& 38.8& 29.7& 20.5& 38.5\\
GTAN \cite{long2019gaussian} & CVPR'19  & 57.8& 47.2& 38.8& — &— &— \\
MGG \cite{liu2019multi} & CVPR'19 & 53.9& 46.8& 37.4& 29.5& 21.3& 37.8\\
G-TAD \cite{xu2020g} & CVPR'20 & 54.5& 47.6& 40.2& 30.8& 23.4& 39.3\\
BU-MR \cite{zhao2020bottom} & ECCV'20 & 53.9 &50.7& 45.4& 38.0& 28.5& 43.3\\
BC-GNN \cite{bai2020boundary} & ECCV'20 & 57.1& 49.1& 40.4& 31.2& 23.1& 40.2\\
BSN++ \cite{su2021bsn++} & AAAI'21 & 59.9& 49.5& 41.3& 31.9& 22.8& 41.1\\
TCANet \cite{qing2021temporal} & CVPR'21  & 60.6& 53.2& 44.6& 36.8& 26.7& 44.4\\
AFSD \cite{lin2021learning} & CVPR'21& 67.3 &62.4& 55.5& 43.7& 31.1& 52.0\\
MUSES \cite{liu2021multi} & CVPR'21 & 68.9& 64.0& 56.9& 46.3& 31.0& 53.4\\
VSGN \cite{zhao2021video} & ICCV'21 & 66.7& 60.4& 52.4& 41.0& 30.4& 50.2\\
ContextLoc \cite{zhu2021enriching} & ICCV'21 & 68.3& 63.8& 54.3& 41.8& 26.2& 50.9\\
RTD-Net \cite{tan2021relaxed} & ICCV'21  & 68.3& 62.3& 51.9& 38.8& 23.7& 49.0\\
CPN \cite{hsieh2022contextual} & WACV'22 & 68.2& 62.1& 54.1& 41.5& 28.0&50.7\\
E2E-TAD \cite{liu2022empirical} & CVPR'22 & 69.4& 64.3& 56.0& 46.4& 34.9& 54.2\\
RefactorNet \cite{xia2022learning} & CVPR'22 & 70.7& 65.4& 58.6 & 47.0& 32.1& 54.8\\
RCL \cite{wang2022rcl} & CVPR'22 & 70.1 &62.3& 52.9& 42.7& 30.7& 51.7\\
LDCLR \cite{zhu2022learning} & AAAI'22 & 72.1 & \underline{65.9}& 57.0& 44.2& 28.5& 53.5 \\
TallFormer \cite{cheng2022tallformer} & ECCV'22 & 76.1 &— &\underline{63.2} &— & 34.5& 59.2 \\
STALE \cite{nag2022zero} & ECCV'22 & 68.9 &64.1 &57.1 &46.7 &31.2 &52.9  \\
ActionFormer \cite{zhang2022actionformer} & ECCV'22 & \textbf{82.1} & \textbf{77.8} & \textbf{71.0} & \textbf{59.4} & \textbf{43.9} & \textbf{66.8} \\
\textbf{Ours} & — & \underline{78.2} & \underline{69.5} & 62.7 & \underline{50.1} & \underline{36.9} & \underline{59.8} \\

\hline
\end{tabular}
\end{center}

\end{table}

\begin{figure}[h!]
    \centering
    \includegraphics[width=0.9\textwidth]{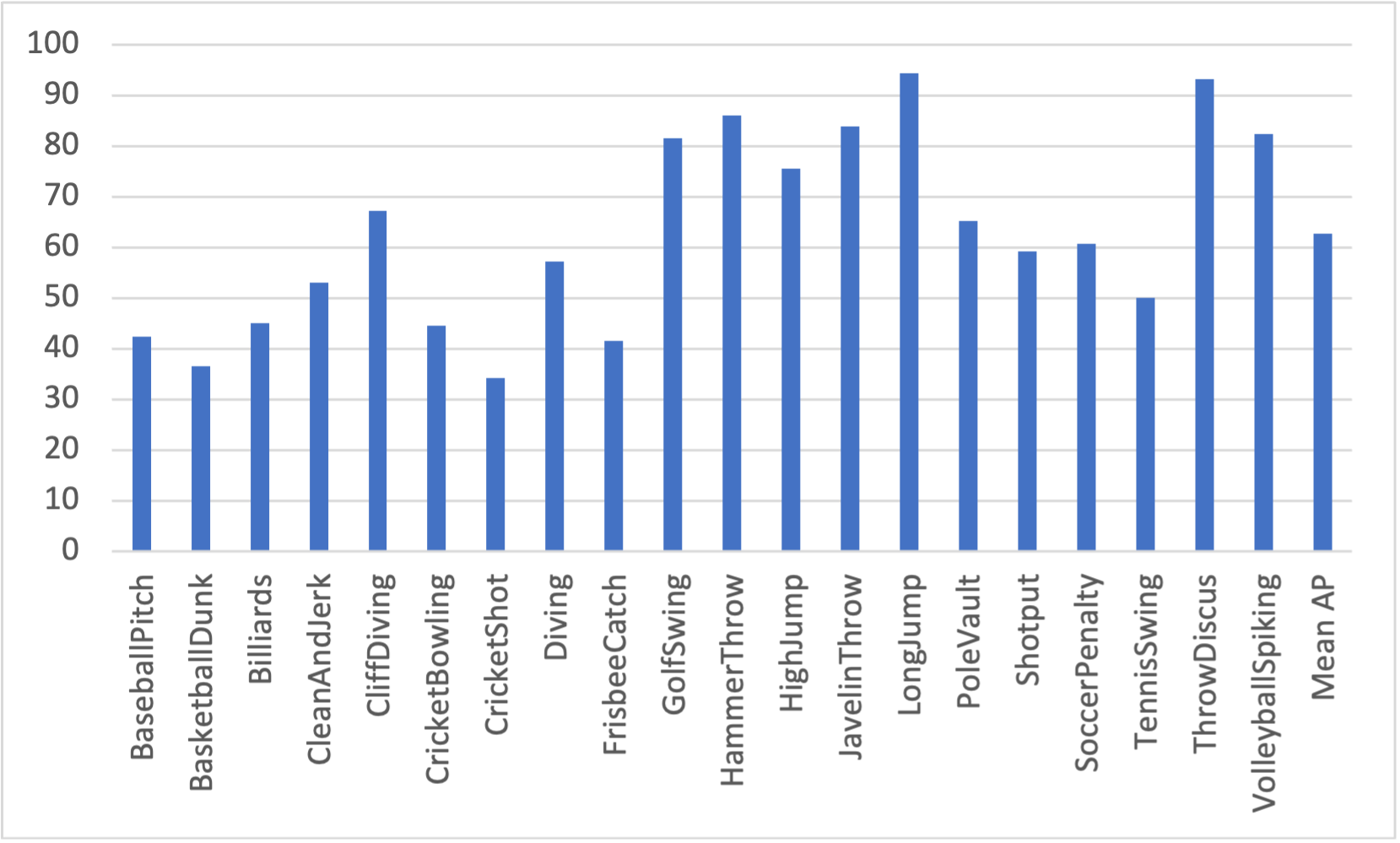}
    \caption{\rev{Class-wise average precision (\%) with IoU threshold of 0.5 for Thumos-14 dataset. \label{fig:thumos-class}}}
\end{figure}

\textbf{ActivityNet-1.3}. Table \ref{table:act} also compares our proposal with the state-of-the-art methods on ActivityNet-1.3. The mAP IoU thresholds used for comparison are, according to protocol, 0.5, 0.75 and 0.95. 
Our approach performs the best on average and for IoU = 0.5 and 0.95, while coming a very close second for IoU = 0.75 with RefactorNet \cite{xia2022learning},
proving our hypothesis that modelling the fine-grain structure of activity is conducive to better performance even for short-term actions.

\setlength{\tabcolsep}{4pt}
\begin{table}[h!]
\begin{center}
\caption{\rev{Activity detection performance comparison with state-of-the-art methods on ActivityNet-1.3. mAP values at different IoU thresholds are reported for validation sets of ActivityNet-1.3. Best results are in \textbf{bold} and second best \underline{underlined}.} }
\label{table:act}
\begin{tabular}{llllll}
\hline\noalign{\smallskip}
 & &  \multicolumn{4}{c}{ActivityNet-1.3}\\
Methods & Conference & 0.5 & 0.75 & 0.95 & Average\\
\noalign{\smallskip}
\hline
\noalign{\smallskip}
BSN \cite{lin2018bsn} &ECCV'18  & 46.45& 29.96& 8.02& 30.03\\

P-GCN \cite{zeng2019graph} & ICCV'19 &48.26& 33.16& 3.27& 31.11\\
% BMN \cite{lin2019bmn} & ICCV'19 & 56.0& 47.4& 38.8& 29.7& 20.5& 38.5& 50.07& 34.78& 8.29& 33.85\\
GTAN \cite{long2019gaussian} & CVPR'19  & 52.61 &34.14& 8.91& 34.31\\
% MGG \cite{liu2019multi} & CVPR'19 & 53.9& 46.8& 37.4& 29.5& 21.3& 37.8& —& —& —& —\\
G-TAD \cite{xu2020g} & CVPR'20 & 50.36& 34.60& 9.02& 34.09\\
BU-MR \cite{zhao2020bottom} & ECCV'20 & 43.47& 33.91& 9.21& 30.12\\
BC-GNN \cite{bai2020boundary} & ECCV'20 &  50.6& 34.8& 9.4& 34.3\\

BSN++ \cite{su2021bsn++} & AAAI'21 &  51.27& 35.70& 8.33& 34.88\\
TCANet \cite{qing2021temporal} & CVPR'21  &  51.91& 34.92& 7.46& 34.43\\
AFSD \cite{lin2021learning} & CVPR'21&  52.4 &35.3& 6.5 &34.4\\
MUSES \cite{liu2021multi} & CVPR'21 &  50.02& 34.97& 6.57& 33.99\\
VSGN \cite{zhao2021video} & ICCV'21 &  52.4& 36.0& 8.4& 35.1\\
ContextLoc \cite{zhu2021enriching} & ICCV'21 & 56.01& 35.19& 3.55 & 34.23\\
RTD-Net \cite{tan2021relaxed} & ICCV'21  &  47.2& 30.7& 8.6& 30.8\\
% CPN \cite{hsieh2022contextual} & WACV'22 & 68.2& 62.1& 54.1& 41.5& 28.0&50.7& — &—& —& —\\
E2E-TAD \cite{liu2022empirical} & CVPR'22 & 50.47& 35.99& \underline{10.83} & 35.10\\
RefactorNet \cite{xia2022learning} & CVPR'22 &  56.6& \textbf{40.7}& 7.4& \underline{38.6}\\
RCL \cite{wang2022rcl} & CVPR'22 &  55.15& 39.02& 8.27& 37.65\\
LDCLR \cite{zhu2022learning} & AAAI'22 &  \underline{58.14} & 36.30& 6.16& 35.24\\
TallFormer \cite{cheng2022tallformer} & ECCV'22 &  54.1 &36.2&7.9 &35.6 \\
STALE \cite{nag2022zero} & ECCV'22 & 56.5 &36.7&9.5 &36.4 \\
ActionFormer \cite{zhang2022actionformer} & ECCV'22 & 53.5 & 36.2 & 8.2 & 35.6 \\
\textbf{Ours} & — & \textbf{60.6} & \underline{40.3} & \textbf{11.1} & \textbf{39.3}\\

\hline
\end{tabular}
\end{center}

\end{table}

\rev{To sum up, our proposal clearly outperforms all previous approaches over the ROAD dataset, showing the potential of this approach for modelling and detecting long, complex activities. Further, our approach outperforms all competitors over ActivityNet and achieves a reasonable performance (second best) on Thumos-14, which is characterised by shorter, simpler activities. This proves %our hypothesis 
that modelling the fine-grain structure of activities is conducive to better performance, even for simpler events.
Fig. \ref{fig:sota} illustrates how our method increasingly outperforms the prior art as the duration and complexity of the activities increases.}

\begin{figure}[h!]
    \centering
    \includegraphics[width=0.9\textwidth]{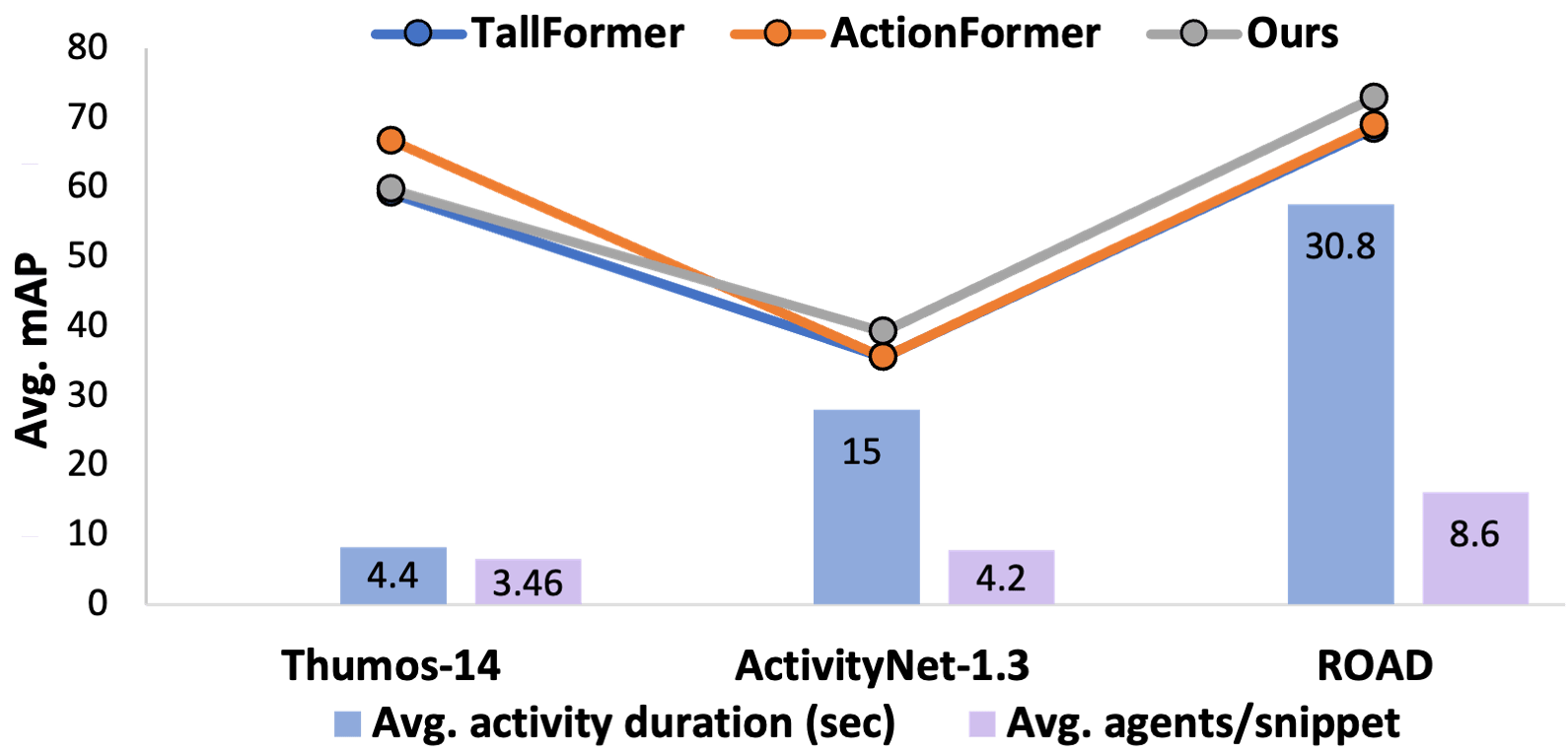}
    \caption{\rev{Performance comparison with the state-of-the-art over all the three datasets considered, as a function of %while considering the 
    average activity duration and average no. of agents per snippet. \label{fig:sota}}}
    \vspace{-4mm}
\end{figure}

% \setlength{\tabcolsep}{4pt}
% \begin{table*}
% \begin{center}
% \caption{Effect of different local scene graph topologies and feature aggregation strategies on the performance of our proposal on the three datasets considered.
% }
% \label{table:edge_agg}
% \begin{tabular}{lll|ll|ll}
% \hline\noalign{\smallskip}
% $\qquad\qquad$ &  \multicolumn{2}{c}{Thumos-14} & \multicolumn{2}{c}{ActivityNet-1.3} & \multicolumn{2}{c}{ROAD}\\
% Methods $\qquad\qquad$ & Aggregated & Scene & Aggregated & Scene & Aggregated & Scene\\
% \noalign{\smallskip}
% \hline
% \noalign{\smallskip}
% Fully connected & \textbf{59.8} & 49.2 & 37.4 & 31.4 & \textbf{73.0} & 62.7\\
% Star structure & 51.2 & 44.8 & \textbf{39.3} & 36.6 & 62.3 & 57.9\\
% Star + same label & 52.2 & 41.9 & 35.3 &32.7 &64.9   & 59.2\\
% \hline
% \end{tabular}
% \end{center}
% \end{table*}
% \setlength{\tabcolsep}{1.4pt}

\subsection{Ablation Studies}\label{sec:ablation}

We ran a number of ablation studies to understand the influence of the various components.

\textbf{Effect of Agent Nodes}. Firstly, we showed the advantage of using a scene graph modelling and connecting single agents compared to only using whole scene features. Namely, we removed the local scene graph from our pipeline (Fig. \ref{fig:framework}, stage B) and passed the 3D features of the whole scene as a node to the global temporal graph. The significant performance drop can be clearly observed in Fig. \ref{fig:agent_nodes} over all three datasets.

\begin{figure}[h]
    \centering
    \includegraphics[width=0.9\textwidth]{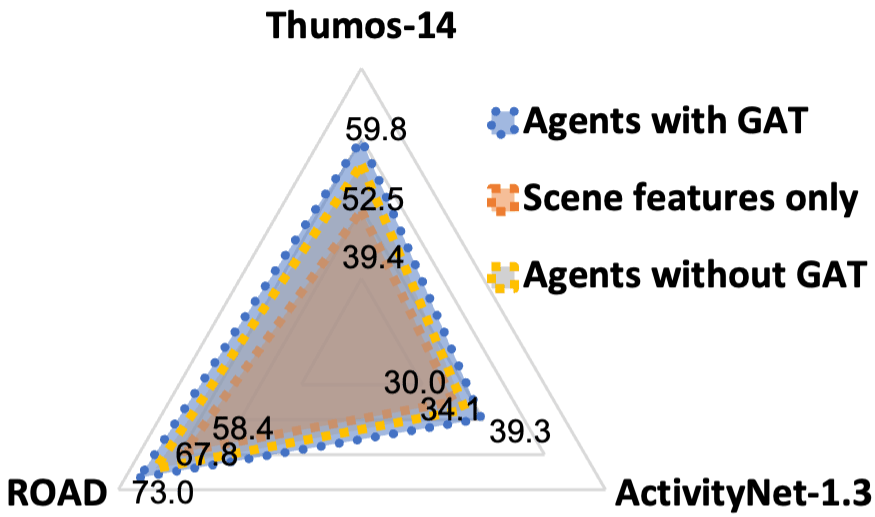}
    \caption{\rev{Average mAP of variants of our method with: attended scene graph ('Agents with GAT'), scene graph ('Agents without GAT') and 'Scene features only', over the all three datasets.}}
\label{fig:agent_nodes}
\end{figure}

\textbf{Effect of Edges and Aggregation}. 
%In our experiments, 
Next, we constructed our local scene graph using three different types of edge connections. Firstly, a fully-connected scene graph where each node is connected to every other node. Secondly, a star structure where all of the agent nodes are connected to the scene node only. Lastly, a star structure with additional connections between agent nodes sharing the same label. 

We also validated two different techniques for extracting the final representation from the local scene graph, which we named `Aggregated' and `Scene'. In the former, the feature representation is extracted by aggregating %FAB: how?
those of all the attended nodes. In the latter, only the feature vector related to the scene node (after attention) is retained. The effect of the possible combinations of graph topologies and aggregation strategies is shown in Table \ref{table:edge_agg} -- a fully-connected scene graph with Aggregated features performs the best in two cases over three, while the star topology performed best for ActivityNet. 
% {\color{blue}FAB: any idea why?}

\textbf{Effect of Sequence Length}. To explore the effect on our model of snippet duration (the temporal extent of the local dynamic scene), 
%sequence length at the level of local scene graph (snippet size), 
we performed experiments with four different sizes (12, 18, 24, and 30), reported in Table \ref{table:seq_lens}. On Thumos-14 and ROAD the top scores were obtained by selecting a sequence length of 24, due to the nature of the activities present in these datasets, which last longer. On the other hand, on ActivityNet-1.3 we achieved the best performance using a sequence length of 18, as most activities there are shorter in duration.

% \includegraphics[width=0.95\textwidth]{figures/hgraph/nodee.png}
%     \caption{Performance of our scene graph with scene features only, compared to that of the full hybrid graph with agents, over all three datasets in terms of average mAP over all thresholds.}
%     \label{fig:agent_nodes}
%     \caption{Effect of different local scene graph topologies and feature aggregation strategies on the performance of our proposal on the three datasets considered.
% }

\begin{table}
\caption{\rev{Effect of local scene graph topologies and feature aggregation strategies on the performance of our proposal. The average mAP for the relevant IoU thresholds is reported for each dataset. %on the three datasets considered.
\label{table:edge_agg}
}}
\begin{tabular}{lll|l|l}
\hline\noalign{\smallskip}
\multicolumn{2}{c}{Methods $\qquad\qquad$} & Thumos-14 & ActivityNet-1.3 & ROAD\\
\noalign{\smallskip}
\hline
\noalign{\smallskip}
\multicolumn{2}{c}{Fully connected (Aggregated)} & \textbf{59.8} & 37.4 & \textbf{73.0} \\
\multicolumn{2}{c}{Fully connected (Scene)} & 49.2 & 31.4 & 62.7\\

\multicolumn{2}{c}{Star structure (Aggregated)} & 51.2 & \textbf{39.3} & 62.3 \\
\multicolumn{2}{c}{Star structure (Scene)} &  44.8 &  36.6 & 57.9\\

\multicolumn{2}{c}{Star + same label (Aggregated)}  & 52.2 & 35.3 &64.9\\
\multicolumn{2}{c}{Star + same label (Scene)} & 41.9 & 35.3 & 59.2\\

\hline
\end{tabular}
\end{table}

\setlength{\tabcolsep}{4pt}
\begin{table}
\centering
\begin{center}
\caption{Performance of our method as a function of different snippet lengths over the three datasets. The average mAP for the relevant IoU thresholds is reported for each dataset.}
\label{table:seq_lens}
\begin{tabular}{ll|l|l}
\hline\noalign{\smallskip}
Snippet length & Thumos-14 & ActivityNet-1.3 & ROAD\\
\noalign{\smallskip}
\hline
\noalign{\smallskip}
12 & 52.7 & 31.3 & 56.9 \\
18 & \textbf{59.8} & 37.6 & 62.6 \\
24 & 54.3 & \textbf{39.3} & \textbf{73.0} \\
30 & 49.7 & 34.5 & 70.0 \\
\hline
\end{tabular}
\end{center}

\end{table}

\setlength{\tabcolsep}{4pt}
\begin{table}
\centering
\begin{center}

\caption{Performance of our method as a function of different temporal graph lengths over the three datasets. The average mAP for the relevant IoU thresholds is reported for each dataset.}
\vspace{-2mm}
\label{table:temp_len}
\begin{tabular}{ll|l|l}
\hline\noalign{\smallskip}
Temporal length & Thumos-14 & ActivityNet-1.3 & ROAD\\
\noalign{\smallskip}
\hline
\noalign{\smallskip}
128 & \textbf{59.8} & 29.6 & 48.3 \\
256 & 52.9 & 32.7 & 58.5 \\
512 & 50.3 & \textbf{39.3} & 69.2  \\
1,024 & 46.8 & 34.1 & \textbf{73.0} \\
\hline
\end{tabular}
\end{center}

\end{table}
\setlength{\tabcolsep}{1.4pt}

\begin{figure}[h]
    \centering
    \includegraphics[width=0.85\textwidth]{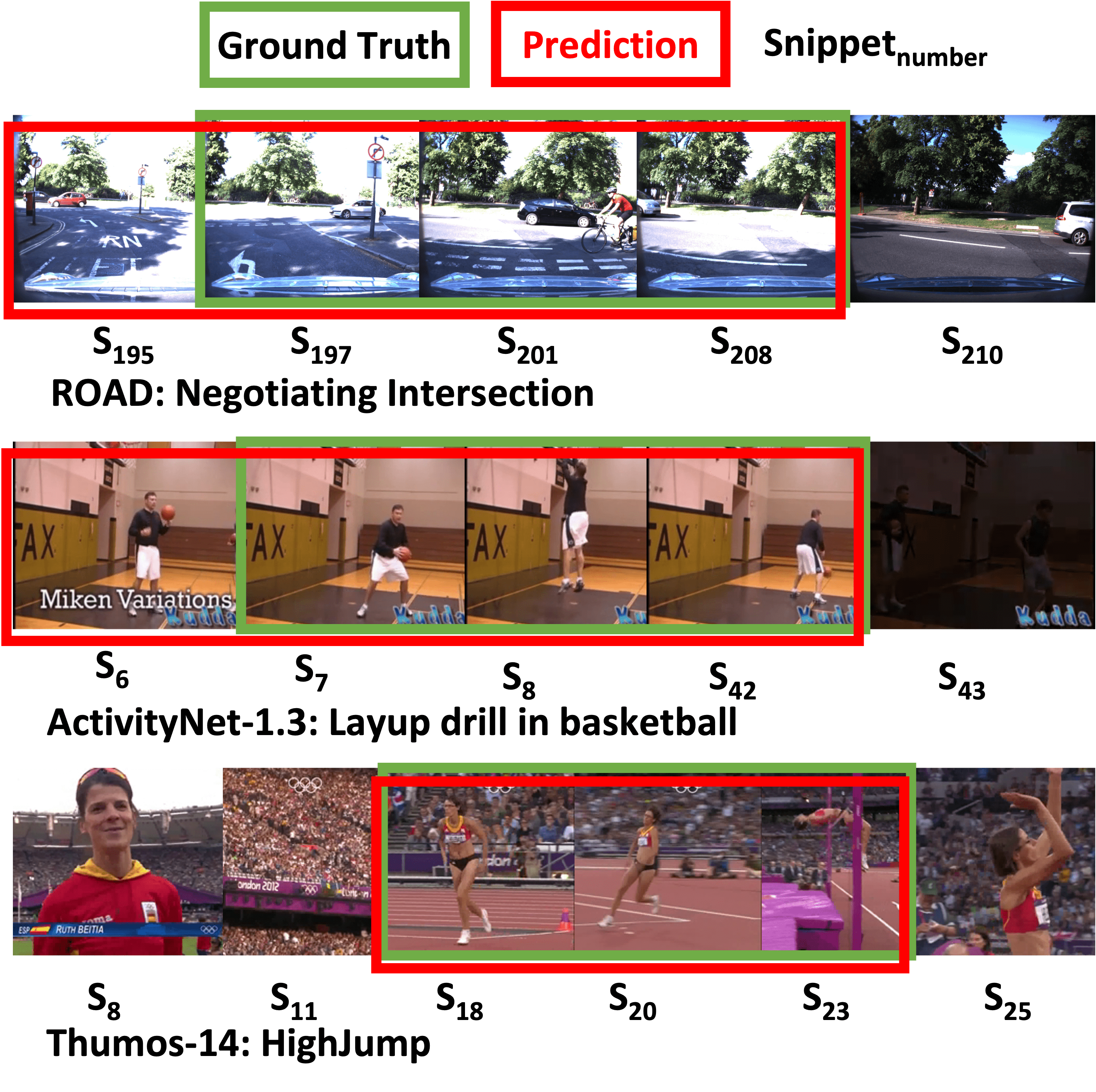}
    \caption{The qualitative results of our proposed method for all three datasets. The green rectangles covering the snippets (local scenes) are the ground truth while the yellow boxes show the prediction of our model.}
    \label{fig:qaul_res}

\end{figure}

\begin{figure}[h]
    \centering
    \includegraphics[width=0.80\textwidth]{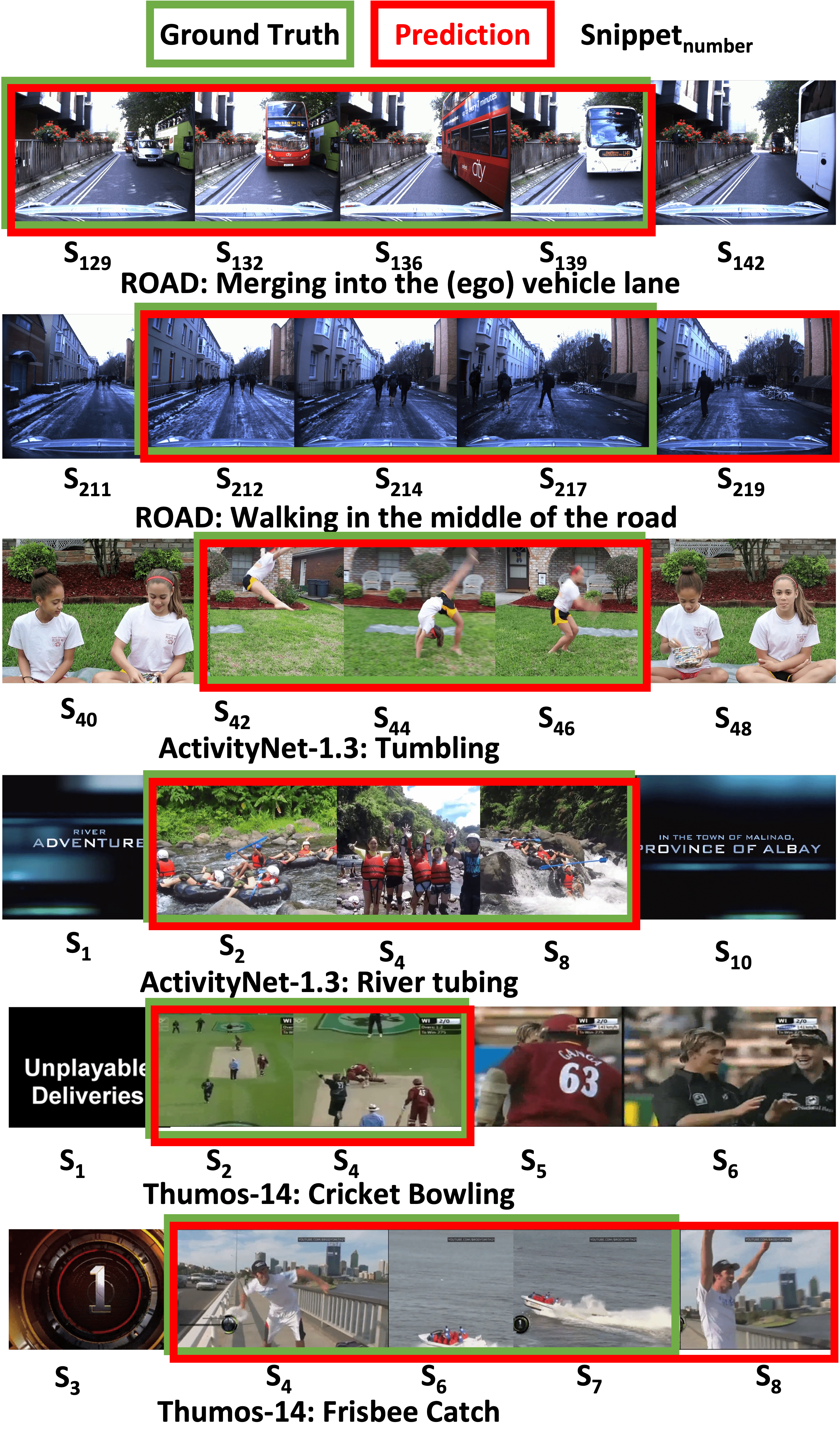}
    \caption{The qualitative results of our proposed method for all three datasets. The green rectangles covering the snippets (local scenes) are the ground truth while the yellow boxes show the prediction of our model.}
    \label{fig:qaul}
\end{figure}

\textbf{Effect of Temporal Graph Length}. In addition to the temporal span of the local scene graph, we also ablated the effect of the length of the temporal graph. By this, we mean the number of local scene nodes used in the temporal graph for recognition and localisation. Four different temporal lengths $(128, 256, 512, 1024)$ are compared in Table \ref{table:temp_len}. The best performance is achieved on ActivityNet-1.3 using a shorter temporal length, due to the average duration of the videos there and its portraying shorter-term activities. In contrast, on Thumos-14 and ROAD our approach performed the best using longer temporal graphs, as activities there are typically of longer duration.

\textbf{Qualitative Results}. To help the reader visualise the output of our proposed method, we show some qualitative detection results on all the three datasets in Fig. \ref{fig:qaul_res} and Fig. \ref{fig:qaul}. The figure shows one sample per dataset, and portrays a series of local scenes (snippets), skipping some for visualisation purposes, with superimposed the ground truth (in green) and the prediction of our model (in red). For example, for ActivityNet-1.3 an instance of the `Layup drill in basketball' class is shown in which the activity starts with snippet 7 and ends with snippet 42. Our model predicts the activity to start from snippet 6 and end with snippet 43.

\section{Summary of the chapter}\label{hgraph:ssec:summary}

\subsection{Future Work}

In the future we intend to progress further from incremental inference to incremental training, by learning to constructing activity graphs in an incremental manner, thus opening the way to applications such as future activity anticipation \cite{liu2022hybrid} and pedestrian intent prediction \cite{cadena2022pedestrian}, while overcoming the need for a fixed-size snippet representation. A further exciting line of research is the modelling of the uncertainty associated with complex scenes, in either the Bayesian \cite{kendall2017uncertainties} 
or the full epistemic setting \cite{manchingal2022epistemic,osband2021epistemic}.

\subsection{Conclusion}

Whereas most existing work focuses on short-term actions, this chapter explicitly addressed the problem of detecting longer-term, complex activities using a novel hybrid graph neural network-based framework which combines both scene graph attention and a temporal graph to model activities of arbitrary duration. Our proposed framework is divided into three main building blocks: agent tube detection and feature extraction; a local scene graph construction with attention; %, for updating the node features; 
and a temporal graph for recognising the class label and localising each activity instance.
We tested our method on three benchmark datasets, showing the effectiveness of our method in detecting both short-term and long-term activities, thanks to its ability to model their finer-grained structure without the need for extra annotation.

    \pagestyle{plain}
    \cleardoublepage
\phantomsection
\renewcommand{\pname}{Part III : Continual Learning} \label{part:part3}
\addcontentsline{toc}{chapter}{\pname}\label{partIII}

\pagebreak
\hspace{14pt}
\vfill
\begin{center}
\textbf{\pname}
\end{center}
\vfill
\hspace{0pt}
\pagebreak
    \pagestyle{fancy}
    \chapter{Continual Semi-Supervised Learning}
\label{chapter:continual}
\renewcommand{\imagepath}{figures/rcn} 
\section{Introduction} 
\label{continual:intro}

The last part of the thesis is about continual learning, while interesting works have been recently directed at continual learning from streaming data in a fully supervised setting, especially focusing on avoiding catastrophic forgetting, continual learning in a semi-supervised setting remains a wide-open research question. Thus in this chapter, we present a new continual semi-supervised learning (CSSL) paradigm, proposed to the attention of the machine learning community via the IJCAI 2021 International Workshop on Continual Semi-Supervised Learning (CSSL@IJCAI)\footnote{\url{https://sites.google.com/view/sscl-workshop-ijcai-2021/}}, with the aim of raising the field's awareness about this problem and mobilising its effort in this direction. After a formal definition of continual semi-supervised learning and the appropriate training and testing protocols, the chapter introduces two new benchmarks specifically designed to assess CSSL on two important computer vision tasks: activity recognition and crowd counting. We describe the Continual Activity Recognition (CAR) and Continual Crowd Counting (CCC) challenges built upon those benchmarks, the baseline models proposed for the challenges, and describe a simple CSSL baseline which consists in applying batch self-training in temporal sessions, for a limited number of rounds. The results show that learning from unlabelled data streams is extremely challenging, and stimulate the search for methods that can encode the dynamics of the data stream. 

In addition to the challenge baseline, we also propose to formulate the continual semi-supervised learning problem as a latent-variable one, in which the labels of a series of unlabelled, streaming datapoints play the role of the hidden states, while the instances play that of the observations. In any realistic setting the process generating instances and labels, modelled by a probability distribution, will vary in time. The aim is for the model to be updated in such a way as to adapt to a (slowly) shifting target domain, without artificial subdivisions into tasks. To accommodate for this, latent variable temporal graphical models whose parameters also vary in time need to be considered. In particular, we thoroughly analyse the proposed framework in a classification setting, for the case in which the data stream is modelled by a time-varying hidden Markov model.

\subsection{Motivation}

The continual learning problem has been recently the object of much attention in the machine learning community, with a focus on preventing the model updated in the light of new data from `catastrophic forgetting' its initial abilities. A typical example is that of an object detector which needs to be extended to include classes not originally in its list (e.g., `donkey') while retaining its ability to correctly detect, say, a `horse'.
The unspoken assumption there is that we are quite satisfied with the model we have, and merely wish to extend its capabilities to new settings and classes.

This way of posing the continual learning problem, however, is in rather stark contrast with widespread real-world situations in which an initial model is trained using limited data, only for it to then be deployed without any additional supervision.
Think of a person detector used for surveillance purposes on a busy street. Even after having been trained extensively on public datasets, experience shows that its performance in its target setting will suboptimal. In this scenario, the goal is for the model to be incrementally updated using the new (unlabelled) data, in order to adapt to its target domain.

In such settings, the goal is quite the opposite of that of the classical scenario described above: the initial model is usually poor and not a good match for the target domain. To complicate things further, the target domain itself may change over time, both periodically (e.g., night/day cycles) and in asynchronous, discrete steps (e.g., when a new bank opens within the camera's field of view or a new building is erected with a new entrance).

\subsection{Overview of the CSSL Challenge} \label{subsec:ov_cssl_ch}

In the first part of this chapter, we formalise this problem as one of \emph{continual semi-supervised learning}, in which an initial training batch of labelled data points is available and can be used to train an initial model, but then the model is incrementally updated exploiting the information provided by a time series of unlabelled data points, each of which is generated by a data generating process (modelled by a probability distribution) which \emph{varies with time}. \rev{Our formulation of CSSL is similar to \emph{online domain adaptation} (ODA). Both CSSL and ODA involve adaptation to changing data distributions, they have distinct focuses and characteristics. The main difference between ODA and our CSSL is that ODA primarily deals with adapting a model to changes in the data distribution across different domains (tasks). This is particularly relevant in applications where the target domain is dynamic and constantly evolving \cite{liu2020open,panagiotakopoulos2022online,vs2023towards}. CSSL, on the other hand, focuses on the scenario where a model incrementally learns from both labeled and unlabeled data over time. The key challenge is to continually adapt to new data without catastrophic forgetting of previously learned knowledge. While this may involve adapting to some degree of changing data distributions, the primary emphasis is on incremental learning and avoiding model degradation over time. In CSSL,} we do not assume any artificial subdivision into `tasks', but allow the data-generating distribution to be an arbitrary function of time.

While batch semi-supervised learning (SSL) has seen renewed interest in recent times, thanks to relevant work in the rising field of unsupervised domain adaptation \cite{Tzeng2017adversarial} and various approaches based on classical SSL self-training \cite{rosenberg2005semi,chen2019progressive}, where the dominant trend is based on adversarial approaches to reduce the discrepancy between source and target domain features \cite{Tzeng2017adversarial}. A panoply of deep self-training approaches have been proposed, inspired by classical SSL self-training \cite{rosenberg2005semi}, which alternate between generating pseudo-labels for the unlabelled data-points and re-training a deep classifier \cite{chen2019progressive}. In opposition, continual SSL is still a rather unexplored field. The reason is that, while in the supervised case it is clear what information the streaming data points carry, in the semi-supervised case it is far from obvious what relevant information carried by the streaming instance should drive model update.

\subsection{Overview of Hidden Markov Model as Latent Variable}

\rev{In response to the CSSL challenge}, we propose to model the stream of instances and ground truth values as a (temporal) \emph{latent variable model}, in which the instances play the role of the observations, and the targets that of the hidden variables.
To accommodate for data distributions that shift in time, this latent variable model needs to be time-variant, i.e., its parameters must be a function of time.
In this chapter we investigate, in particular, the case in which the latent variable model belongs to the class of \emph{time-varying hidden Markov models} (T-HMMs) \cite{Otranto2008thmm,wang2009event}, although deep generative models can also be envisaged. We focus on \emph{classification} tasks, applied to the \emph{continual activity recognition} problem setting defined in \ref{subsec:ov_cssl_ch}.

Our approach is two-step. For each time instant, $t$, in the first stage, the parameters of the latent model of the data stream are updated incrementally, together with the sequence of state (label) estimates up to the present time instant.  In the case of T-HMMs, the parameters are: i) the time-varying matrix of transition probabilities between labels and ii) the time-varying parameters of the emission probability densities (typically, Gaussian) for generating observations given the current state (label).
Such a T-HMM latent variable model can be suitably initialised using the information provided by the initial, supervised training set.
Based on the obtained sequence of most likely (pseudo-)label estimates\footnote{In fact, an HMM outputs a series of probability vectors $P(y_t|x_1,\cdots,x_t)$, where $y_t$ is the current label and $x_\tau$ denotes the observation at time $\tau$. See Conclusions.}, 
a suitable continual supervised learning approach (in our case, memory replay with random sampling \cite{Aljundi2019online}) can be applied to incrementally update the model. Figure \ref{fig:overview} shows the overall pipeline of the proposed framework.

\begin{figure}[t!]
\centering
\includegraphics[width=0.85\textwidth]{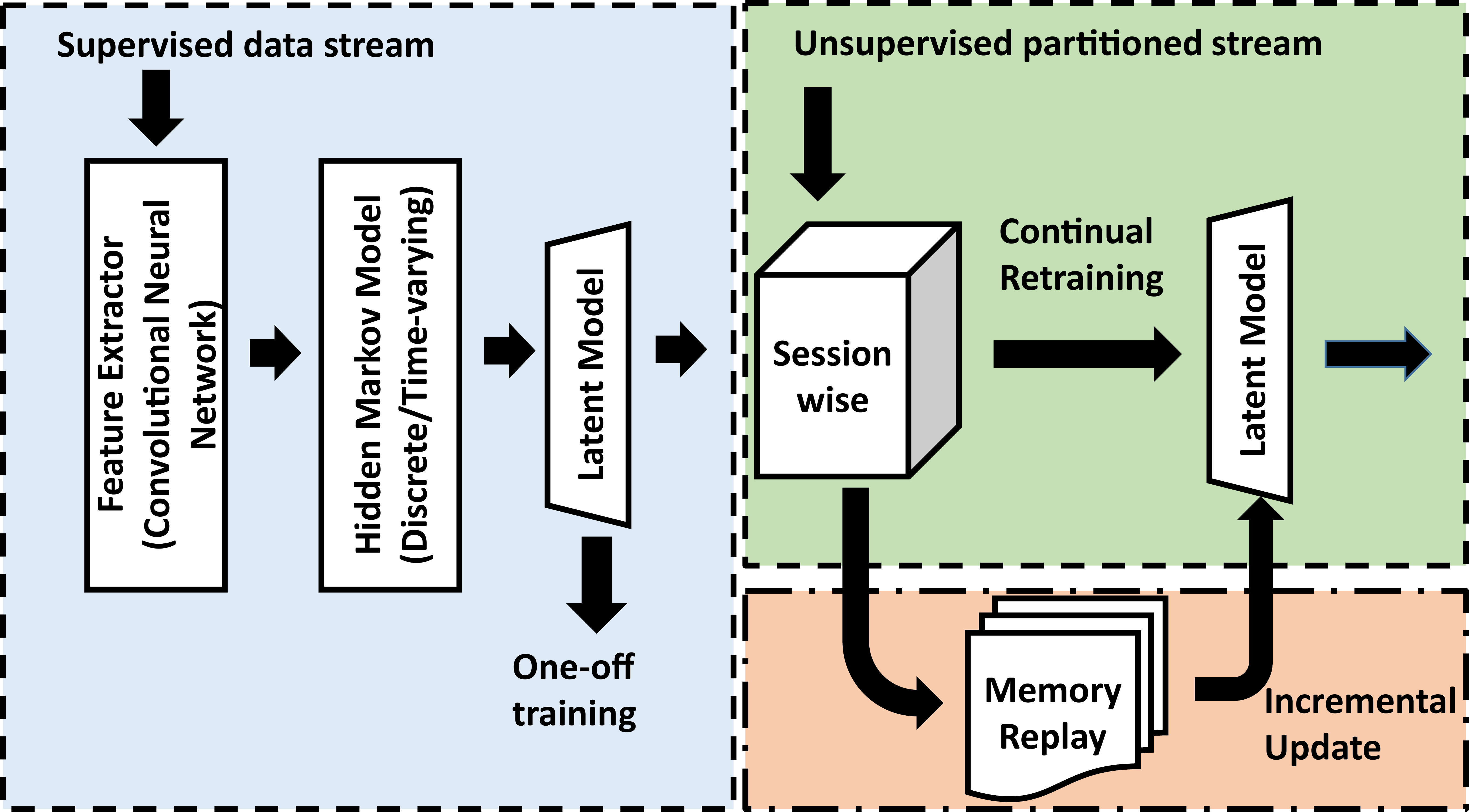}
\caption{Overview of the proposed latent variable pipeline.}
\label{fig:overview}
\end{figure}

\subsection{Main Contributions}

Summarising, the main contributions of this chapter are:
\begin{enumerate}
\item
A formal definition of the continual semi-supervised learning problem, and the associated training and testing protocols.
\item
The first two benchmark datasets for the validation of semi-supervised continual learning approaches, one for classification (continual activity recognition, CAR) and one for regression (continual crowd counting, CCC), which we propose as the foundations of the first challenges in this domain.
\item
Results produced by a simple strategy in which self-training is applied in sessions to competitive baseline models on both benchmarks, as the foundations for the above challenges.
\item
A principled two-stage approach to the problem based on a latent model formulation reducing the task to continual supervised learning.
\item
An efficient solution based on a time-varying hidden Markov models, able to model streaming data issued from a time-shifting target data distribution. 
\item
State-of-the-art results produced by our approach  on the recently released Continual Activity Recognition (CAR) dataset, as well as the 50Salads dataset adapted to continual learning thanks to a new, original protocol.

\end{enumerate}

\paragraph{Related publications:}
The work of this chapter appeared in International Workshop on Continual Semi-Supervised Learning \cite{shahbaz2022international}, Continual Semi-Supervised Learning Workshop and Challenge at IJCAI 2021\footnote{\url{https://sites.google.com/view/sscl-workshop-ijcai-2021/}}, and under preparation 
IEEE transactions on pattern analysis and machine intelligence (IEEE TPAMI). The dissertation's author was co-organiser of the workshop and challenge and co-author of the IEEE TPAMI paper. 
The main contributions of the dissertation's author to this project are: leading the effort of continual activity recognition (continual semi-supervised classification), HMM latent baseline implementation, and CAR baseline for challenge. All the resources of this project are publically available and are accessed from \emph{Code}\footnote{\url{https://github.com/salmank255/cssl}}, \emph{CAR challenge baseline}\footnote{\url{https://github.com/salmank255/IJCAI-2021-Continual-Activity-Recognition-Challenge}}, \emph{CAR evalAI}\footnote{\url{https://eval.ai/web/challenges/challenge-page/984/overview}},\emph{CCC challenge baseline}\footnote{\url{https://github.com/Ajmal70/IJCAI_2021_Continual_Crowd_Counting_Challenge}}, and \emph{CCC evalAI}\footnote{\url{https://eval.ai/web/challenges/challenge-page/986/overview}}.

\paragraph{Outline:}
The structure of the chapter is as follows. Firstly, we explained Continual Semi-Supervised Learning Workshop and Challenge in Section \ref{continual:overview_challenge} including learning (Section \ref{sec:problem-learning}) and testing (Section \ref{sec:problem-testing}) protocols. The CSSL workshop and challenge also include benchmarks (Section \ref{sec:benchmarks}), challenges (Section \ref{sec:challenges}), and baselines (Sections \ref{sec:baseliness}. Secondly, we then explained HMM latent variable model in Section \ref{continual:overview_hmm}. The latent variable model is then divided into principles (Section \ref{sec:problem-principles}), learning process (Section \ref{sec:problem-approach}), distribution of data stream (Section \ref{sec:jdds}, discrete-state T-HMM for continuous classification (Section \ref{sec:latent-thmm}), and memory replay (Section \ref{sec:memory_replay}). Next, the experimental results of CSSL workshop and challenge and latent variable are given in Section \ref{continual:sec:exp} and Section \ref{continual:sec:exphmm}, respectively. Finally, in Section \ref{continual:summary} we provide the summary of the chapter.

\section{Continual Semi-Supervised Learning Workshop and 
Challenge}\label{continual:overview_challenge}

We begin by formalising the continual semi-supervised learning problem, and by defining the appropriate training and testing protocols for this new learning setting.

\subsection{Learning} \label{sec:problem-learning}

The \emph{continual semi-supervised learning} problem can be formulated as follows.

Given an initial batch of supervised training examples, 
\[
\mathcal{T}_0 = \{ (x^j,y^j), j = 1, \ldots, N \}, 
\]
and a stream of unsupervised examples, 
\[
\mathcal{T}_t = \{ x_1, \ldots, x_t, \ldots \}, 
\]
we wish to learn a model $f(x|\theta)$ mapping instances $x \in \mathcal{X}$ to 
target values $y \in \mathcal{Y}$, 
depending on a vector $\theta \in \Theta$ of parameters. 

When $\mathcal{Y}$ is discrete (i.e., a list of labels) the problem is one of continual semi-supervised \emph{classification}; when $\mathcal{Y}$ is continuous we talk about continual semi-supervised \emph{regression}.

In both cases we wish the model to be updated incrementally as the string of unsupervised samples come in: namely, based on the current instance $x_t$ at time $t$, model $\theta_{t-1}$ is mapped to a new model $\theta_{t}$.

\subsection{Testing} \label{sec:problem-testing}

How should such a series of models be evaluated?
In continual supervised learning (see e.g. \cite{Aljundi2019online}), a series of ground-truth target values is available and can be exploited to update the model. 
To avoid overfitting, the series of models outputted by a continual supervised learner, $f(x|\theta_1), \ldots, f(x|\theta_t), \ldots$ cannot be tested on the same data it was trained upon. 
Thus, basically all continual (supervised) learning papers set aside from the start a set of instances upon which the incremental models are tested (e.g., in the Core50 dataset the last three sessions are designated as test fold \cite{lomonaco2017core50}).
\\
In principle, however, as each model $f(x|\theta_t)$ is estimated at time $t$ it should also be tested on data available at time $t$, possibly coming from a parallel (test) data stream. 
An idealised such scenario is one in which a person detector is continually trained on data coming from one surveillance camera, say in a shopping mall, and the resulting model is deployed (tested) in real time on all other cameras in the same mall. 

In our continual semi-supervised setting, model update cannot make use of any ground truth target values. 
If the latter are somehow available, but not shown to the learner for training purposes, the performance of the learner can in fact be evaluated on the stream of true target values by testing at each time $t$ model $f(x|\theta_t)$ on the data pair $(x_t,y_t)$.
A reasonable choice for a performance measure is the average loss of the series of models on the \emph{contemporary} data pair:
\begin{equation} \label{eq:performance}
\sum_{t = 1, \ldots, T} l( f(x_t|\theta_t), y_t ).
\end{equation}

In our experiments, a standard 0/1 loss, which translates into classical accuracy, is used for classification tasks.

\subsection{Benchmarks} \label{sec:benchmarks}
To empirically validate CSSL approaches in a computer vision setting we created two benchmark datasets, one designed to test continual classification (CAR), and one for continual regression (CCC).

\subsubsection{Continual activity recognition (CAR) dataset}

\subsubsection{The MEVA dataset}

To allow the validation of continual semi-supervised learning approaches in a realistic classification task we created a new \emph{continual activity recognition} (CAR) dataset derived from the very recently released MEVA (Multiview Extended Video with Activities) dataset \cite{corona2021meva}. 

MEVA is part of the EctEV (Activities in Extended Video) challenge\footnote{\url{https://actev.nist.gov/}}. As of December 2019, 328 hours of ground-camera data and 4.2 hours of Unmanned Arial Vehicle video had been released, broken down into 4304 video clips, each 5 minutes long. These videos were captured at 19 sites (e.g. School, Bus station, Hospital) of the Muscatatuck Urban Training Center (MUTC), using a team of over 100 actors performing in various scenarios.  There are annotations for 22.1 hours (266 five-minute-long video clips) of data. The original annotations are available on GitLab\footnote{\url{https://gitlab.kitware.com/meva/meva-data-repo/tree/master/annotation/DIVA-phase-2/MEVA/}}.
Each video frame is annotated in terms of 37 different activity classes relevant to video surveillance (e.g. \textit{person\textunderscore opens\textunderscore facility\textunderscore door}, \textit{person\textunderscore reads\textunderscore document}, \textit{vehicle\textunderscore picks\textunderscore up\textunderscore person}). Each activity is annotated in terms of class labels and bounding boxes around the activity of interest. Whenever activities relate to objects or other persons (e.g., in \textit{person\textunderscore loads\textunderscore vehicle} the person usually puts an object into the vehicle’s trunk; in \textit{person\textunderscore talks\textunderscore to\textunderscore person} a number of people listen to the person speaking), these object(s) or people are also identified by a bounding box, to allow human-object interaction analyses.

\subsubsection{The CAR dataset}

The original MEVA dataset comes with a number of issues, from our standpoint: (i) multiple activity classes can take place simultaneously, whereas in our formulation (at least for now) only one label can be emitted at any given time instant; (ii) the quality of the original annotation is uneven, with entire instances of activities missing. As our tests are about classification, we can neglect the bounding box information.

For these reasons we set about creating our own continual activity recognition (CAR) dataset by selecting 45 video clips from MEVA and generating from scratch a modified set of annotations spanning a reduced set of 8 activity classes (e.g. \emph{person\_enters\_scene\_through\_structure}, \emph{person\_exits\_vehicle}, \emph{vehicle\_starts}). Those 8 classes have been suitably selected from the original 37 %original classes in MEVA 
to ensure that activity instances from different classes do not temporally overlap, so that we can assign a single label to each frame. Each instance of activity is annotated with the related start and end frame. %of the activity of interest. 
Frames that do not contain any relevant activity label are assigned to a `background' class.
The goal is to classify the activity label of each input video frame.

For some MEVA sites, contiguous annotated videos exist with no gap between the end of the first video and the start of the second video. For some sites, ‘almost’ contiguous videos separated by short (5 or 15 min) gaps are available. Finally, videos from a same site separated by hours or days exist.
Accordingly, our CAR benchmark is composed of 15 sequences, broken down into three groups: 
\begin{enumerate}
\item
Five 15-minute-long sequences from sites G326, G331, G341, G420, and G638 formed by three original videos which are contiguous.
\item
Five 15-minute-long sequences from sites G329, G341, G420, G421, G638 formed by three videos separated by a short gap (5-20 minutes).
\item
Five 15-minute-long sequences from sites G420, G421, G424, G506, and G638 formed by three original videos separated by a long gap (hours or days).
\end{enumerate}
Each of these three evaluation settings is intended to simulate a different mix of continuous and discrete domain dynamics.

The CAR dataset including annotation and scripts is available on GitHub\footnote{\url{https://github.com/salmank255/IJCAI-2021-Continual-Activity-Recognition-Challenge}}.

\subsubsection{The continual crowd counting (CCC) dataset}

\emph{Crowd counting} is the problem of, given a video frame, counting the number of people present in the frame. While intrinsically a classification problem, crowd counting can be posed as a regression problem by manually providing for each training frame a density map \cite{boominathan2016crowdnet}. 
To date, crowd counting is mostly considered an image-based task, performed on single video frames. Few attempts have been made to extend the problem to the video domain \cite{hossain2020video}, \cite{fenget}, \cite{liuet}.

To the best of our knowledge, continual crowd counting has never been posed as a problem, not even in the fully supervised context -- thus, there are no standard benchmarks one can adopt in this domain.
For this reason we set about assembling the first benchmark dataset for \emph{continual crowd counting} (CCC).
Our CCC dataset is composed by 3 sequences, taken from existing crowd counting datasets:
\begin{enumerate}
\item
A single 2,000 frame sequence originally from the Mall dataset\footnote{\url{https://www.kaggle.com/c/counting-people-in-a-mall}} \cite{chen2012feature}.
\item
A single 2,000-frame sequence originally from the UCSD dataset\footnote{\url{http://www.svcl.ucsd.edu/projects/peoplecnt/}} \cite{chan2008privacy}.
\item
A 750-frame sequence from the Fudan-ShanghaiTech (FDST) dataset\footnote{\url{https://drive.google.com/drive/folders/19c2X529VTNjl3YL1EYweBg60G70G2D-w}}, 
composed by 5 clips, 150 frames long, portraying a same scene \cite{fang2019locality}.
\end{enumerate}
The ground truth for the CCC sequences (in the form of a density map for each frame) was generated by us for all three datasets following a standard annotation protocol\footnote{\url{https://github.com/svishwa/crowdcount-mcnn}}.

The CCC dataset complete with scripts for download and generating the baseline results is available on GitHub\footnote{\url{https://github.com/Ajmal70/IJCAI_2021_Continual_Crowd_Counting_Challenge}}.

\subsection{Challenges} \label{sec:challenges}
\subsubsection{CAR challenge}

From our problem definition (Sec. \ref{continual:overview_challenge}), once a model is fine-tuned on the supervised portion of a data stream it is then both incrementally updated using the unlabelled portion of the same data stream and tested there, using the provided ground truth.
Incremental training and testing are supposed to happen independently for each sequence, as the aim is to simulate real-world scenarios in which a smart device with continual learning capability can only learn from its own data stream. % after deployment.
However, as CAR sequences do not contain instances of all 9 activities, the initial supervised training is run there on the union of the supervised folds for each sequence.

\emph{Split}. 
Each data stream (sequence) in our benchmark(s) is divided into a supervised fold ($S$), a validation fold ($V$) and a test fold ($T$), the last two unsupervised.
Specifically, in CAR given a sequence the video frames from the first five minutes (5 x 60 x 25 = 7,500 samples) are selected to form the initial supervised training set $S$.
%$\mathcal{T}_0$. 
The second clip (another 5 minutes) is provided as validation fold to tune the %latent variable model of the dynamics of the data. 
CSSL strategy.
The remaining clip (5 minutes) is used as test fold for testing the performance of the (incrementally updated) classifier. 

\emph{Evaluation}. 
Performance is evaluated as the average performance of the incrementally updated classifier over the test fold for all the 15 sequences.
More specifically, we evaluate the average 
%For continual classification, this amounts to the 
accuracy (percentage of correctly classified frames) over the test folds of the 15 sequences.
Remember that, however, in our setting each test frame
%unlabelled stream $\mathcal{T}_t$, where, however, each frame in $\mathcal{T}_t$ 
is classified by the \emph{current} model available at time $t$ \emph{for that specific sequence}.
This distinguishes our evaluation setting from classical ones in which all samples at test time are processed by the same model, whose accuracy is then assessed.

\subsubsection{CCC challenge}
Unlike the CAR challenge, in the crowd counting case (as this is a regression problem) each sequence is treated completely independently.

\emph{Split}.
For the CCC challenge we distinguish two cases. For the 2,000-frame sequences from either the UCSD or the Mall dataset, $S$ is formed by the first 400 images, $V$ by the following 800 images, and $T$ by the remaining 800 images. For the 750-frame sequence from the FDST dataset, $S$ is the set of the first 150 images, $V$ the set of the following 300 images, and $T$ the set of remaining 300 images.

\emph{Evaluation}.
For continual crowd counting, MAE (Mean Absolute Error) is adopted (as standard in the field) to measure performance. MAE is calculated using predicted and ground truth density maps in a regression setting. 

\begin{figure*}[h]
    \centering
    \includegraphics[width=1\textwidth]{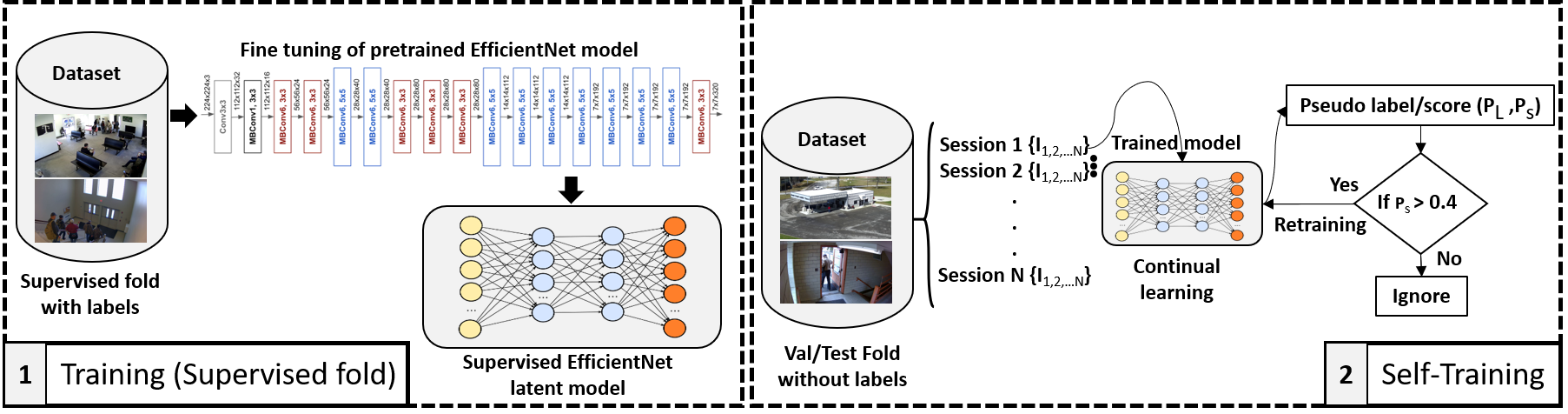}
    \caption{{Overall pipeline of our CAR baseline. 1) Firstly, a pre-trained EfficientNet model is fine-tuned over the supervised fold of the dataset (jointly over the 15 sequences). 
    %to get a latent model for the continuous learning of unlabeled data. 
    2) The unlabeled validation and test folds are divided into subfolds. In each session 
    %and pass the images of each session to trained latent model in a mini-batch wise
    video frames from each sub-fold are used in a self-training cycle in which a pseudo label $P_L$ with prediction probability score $P_s$ is generated for each frame, but only the pseudolabels with %. Each of the label with a certain 
    $P_s$ above a fixed threshold (namely, 0.4) are considered as ground truth for retraining the model. Frames with %The images with 
    lower $P_s$ are ignored. % during the self-training process.
    The model updated by self-training in session $n$ is used as initial model in session $n+1$.
    }}
    \label{fig:CAR}
\end{figure*}

\subsection{Tasks}

For our IJCAI challenge we therefore set four different validation experiments (\emph{tasks}).
\begin{enumerate}
\item
In the first task (\emph{CAR-absolute}) the goal is to achieve the best average performance across all the (test folds of the) 15 sequences in the CAR dataset. The choice of the baseline action recognition model is left to the participants.
\item
In the second task (\emph{CAR-incremental}) the goal is to achieve the best performance differential between the model as updated through the chosen CSSL strategy and the original model fine-tuned over the supervised fold. 
%over time, measured from start to end of each unlabelled data stream, on average over the 15 sequences. 
%{\color{red} The final evaluation is 
We thus evaluate
the difference between the average performance of the incrementally updated model on the test fold and the average performance of initial model, %from the supervised fold (S) 
also on the test fold. The baseline recognition model is set by us (see Baselines).
\item
Task \emph{CCC-absolute} seeks the best average performance over the 
%unlabelled section 
test fold of the 3 sequences of the CCC dataset. The choice of the baseline crowd counting model is left to the participants to encourage them to push its performance to the limit.
\item
Finally, task \emph{CCC-incremental} seeks the best 
performance differential between the initial and the updated model over the test fold, averaged across the three sequences.
%performance improvement over time, measured from start to end of each unlabelled data stream, on average over the 3 sequences of CCC. {\color{red} The final evaluation is the sum of the average performance of the incrementally updated model on test fold and average performance of initial model from the supervised fold (S) on the test fold.} 
The baseline crowd counting model is set.
\end{enumerate}

\subsection{Baselines}\label{sec:baseliness}

%To put the level of performance of our method in context we
To provide a baseline for the above tasks, and assess the degree of difficulty of the challenges,
%For the above challenges 
we decided to adopt a simple strategy which consists of classical (batch) semi-supervised self-training \emph{performed in a series of sessions}, for both the activity and the crowd counting experiments.
Baselines are provided regarding the initial action recognition model to be used in the CAR tests, the base crowd counter to be used in the CCC tests, and the semi-supervised incremental learning process itself.

\subsubsection{Baseline activity recognition model}

The baseline model is the recent EfficientNet network %FAB: put reference
(model EfficientNet-B5) \cite{tan2019efficientnet}, pre-trained on the large-scale ImageNet dataset over 1000 classes. For our tests we initialised the model using those weights and changed the number of classes to 9 activities (see Section \ref{sec:benchmarks}). Detailed information about its implementation, along with pre-trained models, can be found on Github\footnote{\url{https://github.com/lukemelas/EfficientNet-PyTorch}}. The model is easily downloadable using the Python command “pip” (\textit{pip install efficientnet-pytorch}).

\subsubsection{Baseline crowd counter}

For the baseline crowd counting model, we selected the Multi-Column Convolutional Neural Network (MCNN) \cite{zhang2016single}, whose (unofficial) implementation is publicly available\footnote{\url{https://github.com/svishwa/crowdcount-mcnn}} and uses PyTorch. {MCNN was considered state-of-the-art when released in 2016, but is still commonly used as a standard baseline in more recent works, due to its competitive performance on public datasets (e.g., \cite{Jinag2019,Xiong2019}. MCNN made significant contributions to crowd counting by proposing a network architecture better-equipped to deal with differing head sizes due to image resolution, perspective effects, distances between the camera and the people within the scene, etc.  In a nutshell, this robustness to scale was achieved via a network composed of three parallel CNN columns, each of which used filters with different receptive field sizes, allowing each column to specialize for a particular scale of human head.}

Pre-trained MCNN models are available for both the ShanghaiTech A and the ShanghaiTech B datasets. For our tests, as well as the Challenge, we chose to adopt the ShanghaiTechB pre-trained model.

\subsubsection{Baseline incremental learning strategy}

As mentioned, our baseline for incremental learning from unlabelled data stream is instead based on a \emph{vanilla self-training} approach \cite{Trigueroet}.
For each sequence, the unlabelled data stream (without distinction between validation and test folds) is partitioned into a number of sub-folds. Each sub-fold spans 1 minute in the CAR challenges, so that each unlabelled sequence is split into 10 sub-folds. Sub-folds span 100 frames in the CCC challenges, so that the UCSD and MALL sequences comprise 16 sub-folds whereas the FDST sequence contains only 6 sub-folds.

Starting with the model initially fine-tuned on the supervised portion of the data stream, self-training is successively applied in a batch fashion in sessions, one session for each sub-fold, for a single epoch (as we noticed that using multiple epochs would lead to results degradation). {Self-training requires to select a confidence threshold above which predictions are selected as pseudolabels.
In our tests we 
%selected as pseudolabels predictions with probability greater than 
set a confidence threshold of 0.4.} The predictions generated by the model obtained after one round of self-training
upon a sub-fold are stored as baseline predictions for the current sub-fold. The model updated after each self-training session is used as initial model for the following session.

\section{HMM Latent Variable Model for Unsupervised Data-Streams}\label{continual:overview_hmm}

In the second part of the chapter, the problem is defined as the same as in Section \ref{continual:overview_challenge} including \emph{Learning} and \emph{Testing}. However, in this part, we only consider the continual semi-supervised classification.

\subsection{Principles of HMM Latent Variable Model} \label{sec:problem-principles}

In this chapter, we wish to devise an approach to the continual semi-supervised learning problem adhering to the following principles.
\begin{enumerate}
\item
Continual semi-supervised learning, as defined in Section \ref{continual:overview_challenge}, should reduce to continual supervised learning 
once one or more hypotheses are made about the target values of the unsupervised stream.
\item
At each time step $t$, a constant number, $M$, of such hypotheses 
should be examined.
\item
The admissible target assignments for the unlabelled data stream up to time $t$ should be sampled according to their likelihood to be correct, assuming this information is available. 
\item
For each target assignment hypothesis under consideration, the model should be updated incrementally.
\item
To realistically model real-world applications, a budget should be enforced on the total number of (past) example instances one has access to at any given time step, $t$.  
\item
Assuming ground truth target values $y_1, \ldots, y_t, \ldots$ are available, the validation of the learner should take place simultaneously with the training process, by measuring at each time instant, $t$, the loss of the current model, $f(x|\theta_t)$, on the current data pair, $(x_t,y_t)$, and taking the average over the entire data stream.
\end{enumerate}
Principle 3) assumes that a likelihood model for the data stream of targets
is maintained at each time step, $t$, so that we can sample from it. 
In the classification case (when $\mathcal{Y}$ is finite),
a likelihood model for the admissible label assignments over the (current) sequence of instances, $x_1, \ldots, x_t$, is a finite probability distribution,\footnote{In this chapter we use $P$ (uppercase) to denote finite probability distributions, and $p$ (lowercase) to denote probability densities.} $P_Y ( y_1, \ldots, y_t | \Theta_Y^t )$ over $\mathcal{Y}^t \doteq \mathcal{Y}\times \cdots \times \mathcal{Y}$ ($t$ times), conditional on a vector of parameters 
$\Theta_Y^t = \{ \theta_Y^1, \ldots, \theta_Y^t \}$. 
A similar likelihood model can be defined for continual regression problems.

\subsection{A Two-Stage Learning Process} \label{sec:problem-approach}

In this chapter, we propose a general approach to continual semi-supervised learning (Section \ref{continual:overview_challenge}) which adheres to the principles stated in Section \ref{sec:problem-principles}. The approach is incremental, i.e., the `best' model, $f(x|\theta)$, mapping instances, $x \in \mathcal{X}$, to target values, $y \in \mathcal{Y}$, is updated at each time step, $t$, based on the new evidence, $x_t$.
Namely, at each time step $t$:
\begin{enumerate}
\item
The most likely sequence(s) of target values $y_1, \ldots, y_t$ are estimated based on the new piece of evidence (unlabelled instance) $x_t$.
\item
Based on the current estimate of the most likely labellings, a continual supervised learning task is posed and a memory replay-based approach is used to update the model. 
\end{enumerate}

More specifically, in this work we propose to estimate the most likely target assignments for the current data-stream (Step 1) via a \emph{latent variable model} of the data stream of pairs $(x_t, y_t)$. This is discussed in Section \ref{continual:overview_hmm}.
Although in principle any such model can be employed, including deep generative models, we elaborate in particular on the case in which \emph{time-varying hidden Markov models} are used to model the data and the way the data-generating distribution changes over time for a continual classification task. 

While the pseudo-labels estimated by the latent variable model are used to incrementally update the model at training time, at testing time the performance of the series of models (and, as a consequence, of the overall learner) can be measured versus the series of ground-truth labels, assuming those are available, as dictated by Principle 6.

\subsection{Joint Distribution of the Data Stream} \label{sec:jdds}

We assume that the unsupervised data stream, up to the current time instant $t$, can be modelled by the joint distribution 
\begin{equation} \label{eq:joint-data-distribution}
p_D^{1:t} ( (x_1,y_1), \ldots, (x_t, y_t) | \theta_D^1, \ldots, \theta_D^t),
\end{equation}
%which is equivalent to 
under the additional principle that:
\begin{enumerate}
\setcounter{enumi}{6}
\item
The data-generating probability distribution $p_{D}^t = p ( (x_t, y_t) | \theta_D^t )$, from which the pair $(x_t, y_t)$ is assumed to be sampled, can be a function of time.
\end{enumerate}
Any likelihood model for the target assignments (Principle 3) should thus be derived from the joint, time-varying probability distribution (\ref{eq:joint-data-distribution}), possibly under sensible simplifying assumptions.

Principle 7) allows the target distribution to vary over time, reflecting material, realistic real-world conditions.
This principle makes our approach more general than those which assume the existence, next to the source domain, of a single target domain, as in classical domain adaptation \cite{pan2010domain}. It is also more general than `task-based' approaches to continual learning \cite{thrun2012learning}, in which the data are assumed to be generated in a series of discrete tasks, each associated with its own data generating process, whose boundaries typically are given.

\subsection{Discrete-State T-HMM for Continuous Classification} \label{sec:latent-thmm}

%We can take a step further by acknowledging that 
The data stream of pairs is composed by two parallel series $x_1, \ldots, x_t, \ldots$ and $y_1, \ldots, y_t, \ldots$, of which only the sequence of  instances is observed, while the (unknown) 
%labels constitute values from 
target values form a series of hidden variables.
This amounts to representing the unsupervised data stream using a \emph{latent variable model} in which the targets $y_1, \ldots, y_t, \ldots$ play the role of the hidden variables. %, in turn determining the distribution of the observed instances (see Table \ref{tab:hmm}).

In this chapter, we consider \emph{hidden Markov models} (HMMs) \cite{elliott2008hidden} as one specific instance of a latent variable model that is applicable to the posed problem setup. 
In the continual classification case, states are discrete and correspond to the labels. %; in the regression case, states are real-valued vectors and correspond to the target vectors (Section \ref{sec:continual-regression}).
%where the labels play the role of the hidden variables (states) and the instances that of the observations.
%In both cases, 
To cater to Principle 7), we seek to learn a \emph{time-varying} HMM from the data stream which can model a data distribution that shifts with time.

\iffalse
% this can probably be removed from the final version
\begin{table}[ht!]
\caption{Time-variant HMM formulation of the unlabelled data stream of the continual semi-supervised learning problem \label{tab:hmm}}
\centering
\begin{tabular}{ccccccccc}
$y_1$ & $\rightarrow$ & $y_2$ & $\rightarrow$ & $\cdots$ & $\rightarrow$ & $y_t$ & $\rightarrow$ & $\cdots$
\\
$\downarrow$ & & $\downarrow$ & & & & $\downarrow$ & &
\\
$x_1$ &  & $x_2$ &  & $\cdots$ &  & $x_t$ &  & $\cdots$
\end{tabular} % need to add to the arrows the time-variant parameters
\end{table}
\fi

\subsubsection{Model formulation} \label{sec:thmm-formulation-discrete}

%For continual \emph{classification} problems, the HMM is a finite-state, continuous-observation one (as labels live in $\mathcal{Y}$, whereas the instances $x$ belong to $\mathbb{R}^k$). For such models, the joint distribution (\ref{eq:joint-data-distribution}) assumes the form:

In a discrete-state, continuous observation time-varying hidden Markov model \cite{Otranto2008thmm,wang2009event,bazzi2017time} the joint distribution (\ref{eq:joint-data-distribution}) assumes the form: 
\begin{equation} \label{eq:joint-hmm}
\begin{array}{l}
\quad p_D^{1:t} ( (x_1, y_1), \ldots, (x_t, y_t) | \theta_D^1, \ldots, \theta_D^t) =
\\
\displaystyle
P (y_1 | \theta_D^1) p(x_1 | y_1, \theta_D^1)
\cdot
\prod_{\tau = 2}^t 
P( y_\tau | y_{\tau - 1}, \theta_D^{\tau - 1}) p (x_\tau | y_\tau, \theta_D^\tau).
\end{array}
\end{equation}
%Equation (\ref{eq:joint-hmm}) makes clear that, under Principle 6, in a 
The model parameters $\theta_D^\tau = \{ \theta^0, \theta_T^\tau, \theta_\Gamma^\tau \}$ are \emph{time-varying}, to model the fact that, according to Principle 7), the (target) data distribution varies in general with time.
%\footnote{In the relevant literature there are often called \emph{covariates}, to distinguish them from the constant parameters of a traditional HMM}.
These parameters describe\footnote{The notation used for HMM parameters follows that in \cite{elliott2008hidden}.}: (i) the (discrete) probability of the initial state, $P(y_1 | \theta^0)$, (ii) the discrete, time-dependent transition probability for the hidden variables (in this case, the labels), $A^\tau \doteq P^\tau ( y_{\tau+1} | y_\tau ) = P ( y_{\tau + 1} | y_\tau, \theta_T^\tau )$, where $\tau \in \{1, \ldots, t \}$ denotes any time instant in the current data-stream, and (iii) the conditional density of the (continuous) observations (instances), given the state (label), $\Gamma^\tau ( x_\tau | y_\tau) = p ( x_\tau | y_\tau, \theta_\Gamma^\tau )$.

Under this model, the structure of the time-dependent data probability distribution is
\begin{equation} \label{eq:data-distribution-1}
p_D^t ( ( x_t, y_t ) | \theta_D^t ) = p ( x_t | y_t, \theta_\Gamma^t ) \cdot P ( y_t ),
\end{equation}
where
\begin{equation} \label{eq:data-distribution-2}
P ( y_t ) = P (y_t | y_{t - 1}, \theta_T^{t - 1}) \cdot \cdots \cdot P (y_2 | y_1, \theta_T^1) \cdot P (y_1 | \theta^0).
\end{equation}
In other words, the probability distribution generating the data at time $t$ is parameterised by: the parameters, $\theta_\Gamma^t$, that determine the distribution of the instances given the label at time $t$, those that determine the series of transitions $1 \rightarrow 2 \rightarrow  \cdots \rightarrow t$ of the Markov chain of labels, and the probability of the initial label.
This formulation constrains the way the target domain distribution $p ((x,y) | \theta_D)$ can vary in time, while still allowing for a rather general process model.

%In traditional, constant-parameter discrete-state hidden Markov models, the state-output (in our case, label-instance) `emission' distribution $\Gamma$ is often assumed to be a multivariate Gaussian $\mathcal{N}(\mu,\Sigma)$, with mean $\mu$ and covariance $\Sigma$. In the case of interest to us, both $\mu^t$ and $\Sigma^t$ are functions of time.

\subsubsection{Learning the Parameters of a Discrete-State T-HMM} \label{sec:thmm-learning}
\

\emph{Modelling time dependency with covariates}.
While the above formulation is completely general, the mainstream approach to T-HMM is to model the dependency of the parameters on time (and the transition probabilities, in particular) using a multinomial logistic model \cite{agresti2003categorical} of the kind
\begin{equation} \label{eq:multinomial-logistic}
A^\tau (i,j) =
P(y_\tau = i | y_{\tau - 1} = j) 
\doteq 
\frac{ \exp ( g (z_\tau, i, j) ) }{1 + \sum_{h = 1}^{N-1} \exp ( g (z_\tau, i, h) )},
\end{equation}
where $N$ is the number of states, $i,j \in \{ 1,\ldots,N \}$, and $g (z_\tau, i, j)$ is a function of a \emph{covariate} process $\{ z_\tau, \tau = 1, \ldots \}$, which is also \emph{observable}.
As each entry of the transition matrix is a function of the covariate variables, 
their values drive the dynamics of the transition probabilities. 
\\
The transition matrix can then be expressed as
\[
A^\tau = %P^\tau ( y_{\tau+1} | y_\tau ) = 
P ( y_{\tau + 1} | y_\tau, \theta_T^\tau ) = P ( y_{\tau + 1} | y_\tau, \theta_L, z_\tau ),
\]
where $\theta_L$ is the vector of parameters of the logistic model (\ref{eq:multinomial-logistic}) (in particular, of the function $g$), and $z_\tau$ is the covariate vector at time $\tau$.
Covariates on the initial probability can also be accounted for, in the form of a single vector, $z_0$.

\emph{Covariates as functions of time}. The shape of the function, $g$, is a design choice. A common option is $g (z_\tau, i, j) = \phi_{ij} + z_\tau \vartheta_{ij}$, where $\phi_{ij}$ and $\vartheta_{ij}$ are unknown coefficients, which are unique to the transition between state $i$ and state $j$. The resulting transition matrix has $|\theta_L| = N \cdot (N-1)$ parameters, twice those of a traditional, time-invariant transition matrix.

In many practical applications, time and date stamps can act as effective covariates. For instance, in surveillance settings, the time stamp can induce a suitable periodic behaviour in a system, e.g. to detect people entering a building. 
The date stamp can induce annual cycles, or facilitate the gradual shift of the parameter values over time.
As shown in \cite{li2017incorporating}, a range of transition models can be designed by plugging in a different function of time (intended as the main covariate quantity) as the argument $g (z_\tau, i, j)$ of the exponential in (\ref{eq:multinomial-logistic}).

\emph{Parameter estimation via Expectation Maximisation}. Just as in traditional HMMs, the Expectation-Maximisation (EM) algorithm \cite{dempster1977maximum} can be used to recursively estimate the parameters of a T-HMM in the logistic dependency from covariates formulation, as well as to estimate the most likely series of states \cite{visser2010depmixs4}, 
given one or more training sequences of observations: $\{ ( x^m_1, \ldots, x^m_{T_m} ), m = 1, \ldots, M \}$.

In the T-HMM case, the parameter estimation algorithm takes as input, in addition to each training sequence of observations $x_1, \ldots, x_T$, a parallel sequence of covariate vectors, $z_1,\ldots,z_T$, and outputs both the parameters $\theta_L$ of the logistic model (\ref{eq:multinomial-logistic}) and those of the emission probabilities.
%A software package implementing parameter estimation for time-varying hidden Markov models in the logistic model approach (Section \ref{sec:related-thmm}) is available in R at \url{https://cran.r-project.org/web/packages/depmixS4/} \cite{visser2010depmixs4}.
% FAB: make sure this is actually done, otherwise remove
EM is well-known to be prone to converge to local, as opposed to global, minima of the log-likelihood of the data. In our experiments, EM is run multiple times on the same sequence in order to extract `consensus' parameters. % how exactly?
%We will also examine the stability of the method as a function of the complexity of the model encoding the dependency on time (and therefore the number of parameters to estimate).
% we need to design the actual experiments:  what time dependencies, on what data

\emph{Initialisation via the supervised fold}. While the initial probabilities of the hidden variables (labels) can be estimated via EM with covariates in the same framework explained above for the transition matrix, for our problem it makes more sense to estimate them using the information provided by the initial, labelled training set $\mathcal{T}_0$, sampled from a `source' data distribution $p( (x,y) | \theta_D^0 )$. 
In fact, from $\mathcal{T}_0$ we can extract the empirical marginal of the labels, $\hat{P} (y | \theta_D^0 )$, by counting and normalising the occurrences of the labels in the initial training data, and set: 
\begin{equation} \label{eq:initialisation}
P(y_1 | \theta^0) = \hat{P} (y | \theta_D^0 ).
\end{equation}

\emph{Covariates in the emission probabilities}.
To the best of our knowledge, the extension of the dependence on covariates to the emission probabilities $p ( x_\tau | y_\tau, \theta_\Gamma^\tau )$ has not been discussed in the literature. Nevertheless, this appears to be conceptually straighforward, for instance by plugging a logistic dependency of the mean vector $\mu^\tau$ and covariance matrix $\Sigma^\tau$ in the classical multivariate Gaussian approach from a sequence of covariates (e.g., time or functions thereof). We will address this extension in future work.

\rev{\emph{Discrete state HMM vs. continuous state space }
The choice of using labels as the state space in HMMs is often a strong and meaningful solution for specific tasks. For example, in our case, the primary objective is to train the model on the sequences that are already available and are inherently discrete in nature (train set). We are not expecting a change in the number of states as each state represents a specific class and these classes will remain the same during the training and prediction. In contrast, using a continuous state space related to physics or perception may introduce complexity that is unnecessary for tasks focused on discrete events. This complexity can hinder model interpretability and complicate the mapping between states and meaningful labels. The model with continue state space would be useful in other continual learning settings such as class-incremental learning, where the objective is to train the model of new classes.
}

% \emph{Alternative approaches to time dependency}.
% As mentioned previously, alternative methods which do not require observable covariates exist, such as \cite{Otranto2008thmm} or \cite{chib2004non-markovian}.
% In the experiments presented in this chapter, the more `classical' approach based on a logistic dependency of the parameters from observable covariates is adopted.

\subsubsection{Protocols for training the latent model} \label{sec:latent-model-protocols}

%In classical HMM formulations, model parameters are learned in a batch way from a number (usually more than one) of training sequences of observations, $X^1_{T_1}, \ldots, X^m_{T_m}$.

In the incremental learning setting described in Section \ref{continual:overview_hmm}, the desired model (in this case, the classifier) is incrementally updated at each time instant $t$ using the unlabelled observation $x_t$, and tested on the ground-truth output $y_t$.
In our latent-model approach to the problem, this incremental updating exploits the series of pseudo-outputs (labels) generated by the latent model.

Classical (T-)HMM training, however, assumes that model parameters are learned in a batch way from one or more training sequences of observations.
The question thus arises of what training sequences the latent model should be learned from, whether or not in fact the latent model should be learned from the single unsupervised data stream at disposal, and at what time instants in the data-stream T-HMM learning should take place. %$1, \ldots, t, \ldots$.

We can identify a number of options. %, which we will explore for our experimental validation.
\begin{enumerate}
\item
\textbf{Separate (batch) training}.
The option which is closest to the traditional setting is to identify the latent variable model once and for all from one or more training sequences of instances gathered from the site of interest, and set aside for this purpose. 
This obviously assumes that such sequences are available, which might not be the case. %, and requires us to set aside one or more of those for the training of the latent model.
Alternatively, if only one sequence is available for both training and testing, one can learn the latent model from the labelled section $\mathcal{T}_0$ of the data stream and never update it, but simply use it to generate pseudolabels for the unlabelled section $\mathcal{T}_t$.
%Given an unlabelled, test data-stream, at each time instance $t$ we simply use the learnt latent model to estimate the probability of the current label $P (y_t = k)$, $k = 1, \ldots, | \mathcal{Y} |$ (the dependency on the model parameters is neglected for the sake of readability). Once again, sampling $M$ most likely labels from $P(y_t)$ drives $M$ optimisation problems whose solution generates $M$ possibly updates for the classifier under continual learning (see Section \ref{sec:optimisation} again).

\item
\textbf{Continuous (batch) training}. The second option is to learn the latent model from the test unsupervised data stream itself (as the only one available), but to learn a new version of the T-HMM as each new instance becomes available.
Given a new instance $x_t$:
\begin{enumerate}
\item
We use the current sequence $x_1, \ldots, x_t$ to re-estimate the parameters $\{ \theta_L, \theta_\Gamma \}$ of the T-HMM.
\item
We estimate the probability $P ( y_1, \ldots, y_t)$ of all possible labellings for the current sequence under the newly learnt model.
\end{enumerate}
%The $M$ mostly likely sequence labellings at time $t$ can then be used to solve a chain of suitable constrained optimisation problems (Section \ref{sec:optimisation}), from $\tau = 1$ to $\tau = t$. As explained, sampling a constant number $M$ of labellings can help deal with multi-modality issues.
This protocol fits well our philosophy of continually learning and revising our view of the world as new observations are gathered. However, it is very computationally expensive and possibly unfeasible for deployment on edge devices such as surveillance cameras. %Not only are the T-HMM parameters reestimated at each time instant, but for each $t$, a new cascade of $t$ optimisation problems driven by the new most likely labelling needs to be solved from scratch %(see Section \ref{sec:optimisation})
%to reflect the fact that under the new latent model the most likely sequence of labels may have changed. 

\item
\textbf{Continual (batch) training}.
An intermediate, less expensive option is to update the latent model at regular intervals, rather than for each new unlabelled sample (hence, \emph{continually}, rather than \emph{continuously}). 
In this case:
\begin{enumerate}
\item
The latent model is first initialised using the string of instances (which are temporally ordered) in the supervised training set $\mathcal{T}_0$ (without making use of the label information).
\item
The latent model is then re-trained in \emph{sessions}, each corresponding to a section of the unlabelled data stream.
\end{enumerate}
This approach requires us to suitably partition the data stream $\mathcal{T}_t$ into sessions.
The limit case is that in which a single learning session is used, after which the latent model is not updated further.
We will elaborate on this option in our Experiments.

\end{enumerate}

\subsubsection{Sampling labellings as a state estimation problem} \label{sec:thmm-estimation}

Once the parameters  $\theta_L$, $\theta_\Gamma$ of the latent T-HMM model have been learned from the training set of the unsupervised data streams, given a new unsupervised data stream $x_1, \ldots, x_T$ computing its most likely labelling  
%In our time-variant HMM formulation, the problem of sampling at each time instant $t$ the most likely labelling(s) 
%$\hat{y}_1, \ldots, \hat{y}_\tau, \ldots, \hat{y}_t$ up to $t$ 
$\hat{y}_1, \ldots, \hat{y}_T$
reduces to estimating the most likely sequence of hidden states.
%As in classical HMMs, once the parameters of the model $\theta_L$, $\theta_\Gamma$ have been learned, given a new paired sequence of instances and covariates, 
In fact, a sequence of discrete probability distributions $P(y_t | x_1,\ldots,x_t; z_1,\ldots,z_t; \theta^0, \theta_L, \theta_\Gamma)$ over the hidden variables can be obtained by simply `running' the model in a forward pass.

\subsection{Random Sampling based Memory Replay} \label{sec:memory_replay}

After obtaining the estimated pseudo-labels from session-wise continual retraining, the continual semi-supervised learning problem is simplified to continual supervised learning. A random sampling \cite{Aljundi2019online} based memory replay method is employed to incrementally update the model. An initial memory buffer of size $M$ is comprised of 1000 samples randomly selected from the supervised training set, $\mathcal{T}_0$, and the first session, $\mathcal{T}_1$, of the suitably partitioned data stream.

The model is then incrementally updated at the end of each session via a randomly selected memory buffer. At the end of every session, the memory buffer is filled with new samples. Then, the memory buffer is updated to contain 1000 samples that are obtained through random selection. This incremental updating process continues until the last session of the partitioned unlabelled stream is reached. Details on how the data stream is partitioned can be found in Section \ref{continaul:sec:datasets}. The effectiveness of random sampling-based memory replay along with one-off training and continual retraining is detailed in  Section \ref{sec:analysis}.

\rev{
\subsection{Alternative to HMM as Latent Variable Model }
In our method, we used HMMs as latent variable modeling for pseudo-label generation, however, there are alternative models that you can consider. These alternatives may offer different strengths and may be more suitable for specific tasks or data characteristics. Some of the alternatives are as follows:
\begin{itemize}
    \item \textbf{Variational Autoencoders (VAEs)}:  In CSSL, you can train a VAE on the labeled data and then generate pseudo-labels for unlabeled data by mapping them to the learned latent space.
    \item \textbf{Gaussian Mixture Models (GMMs)}: GMMs are probabilistic models that assume data points are generated from a mixture of Gaussian distributions. They can be used for clustering and semi-supervised learning. In this context, GMMs can help cluster data points and assign pseudo-labels based on cluster assignments.
    \item \textbf{Recurrent Neural Networks (RNNs) and Long Short-Term Memory networks (LSTMs)}: RNNs and LSTMs are neural network architectures capable of modeling sequential data. They can be used to capture temporal dependencies in data and generate pseudo-labels based on the learned representations.
\end{itemize}
}

\section{Experimental Analysis of Challenge Baselines} \label{continual:sec:exp}

We ran experiments on both continual classification and continual regression using two new benchmarks of our own creation, which formed the basis of an IJCAI 2021 workshop and challenge\footnote{\url{https://sites.google.com/view/sscl-workshop-ijcai-2021/}}.
The following description of the experimental setups mimics very closely the guidelines provided for the IJCAI challenge.

\setlength{\tabcolsep}{4pt}
\begin{table}
\begin{center}
\caption{Performance of %CAR absolute and incremental methods over 
the initial supervised model versus that of the incrementally updated models (separately on $V$ and $T$ or in combination), using
three standard evaluation metrics. For each metric we report both class average (C) and weighted average (W).}
\label{tab:cont_ch}
\footnotesize \scalebox{1} {
\begin{tabular}{|*{3}{p{0.036\textwidth}}|*{6}{p{0.036\textwidth}}|*{6}{p{0.036\textwidth}}|}
\hline
 \multicolumn{3}{|c}{}& \multicolumn{6}{c|}{\textbf{Validation Fold}}  & \multicolumn{6}{c|}{\textbf{Test Fold}}\\
\hline
\multicolumn{3}{|c|}{Methods / Evaluation metrics}  & \multicolumn{2}{c|}{Precision} & \multicolumn{2}{|c|}{Recall} & \multicolumn{2}{c|}{F1-Score} & \multicolumn{2}{c|}{Precision} & \multicolumn{2}{|c|}{Recall} & \multicolumn{2}{c|}{F1-Score} \\
\multicolumn{3}{|c|}{}  & C & \multicolumn{1}{c|}{W} & C & \multicolumn{1}{c|}{W} & C & \multicolumn{1}{c|}{W} & C & \multicolumn{1}{c|}{W} & C & \multicolumn{1}{c|}{W} & C & W \\
\hline
\multicolumn{3}{|l|}{$sup-ft-union$}  & \textbf{0.20} & \textbf{0.74} & \textbf{0.25} & 0.68 & \textbf{0.16} & 0.70 & \textbf{0.16} & \textbf{0.82} & \textbf{0.15} & 0.77 & \textbf{0.14} & 0.79  \\
\multicolumn{3}{|l|}{$upd-V$ / $upd-T$}  & 0.18 & 0.73 & 0.19 & \textbf{0.73} & 0.14 & \textbf{0.72} & 0.14 & 0.81 & \textbf{0.15} & \textbf{0.81} & \textbf{0.14} & \textbf{0.80}  \\
\multicolumn{3}{|l|}{$upd-V+T$}  & 0.17 & 0.73 & 0.19 & \textbf{0.73} & 0.14 & \textbf{0.72} & 0.14 & 0.80 & 0.14 & 0.79 & 0.13 & 0.79  \\

\hline
\end{tabular}
}
\end{center}

\end{table}

% table for validation fold
\setlength{\tabcolsep}{2pt}
\begin{table}[h!]
\centering
\caption{Evaluation on the CCC validation ($V$) and test ($T$) folds for different experimental setups. Here $sup-ft$, $upd-V$, and $upd-V+T$ refer to the supervised model fine-tuned on $S$, the model incrementally updated via self-training in sessions on the validation fold $V$, and the model incrementally updated on the combined validation $V$ and test $T$ folds, respectively. The average test MAE is reported for each model.}
\label{table:cont_ch_ccc}
\footnotesize
\begin{tabular}{ |p{2.5cm}|p{1.0cm}|p{1.8cm}|p{1.8cm}|p{1.8cm}|p{1.8cm}|p{1.8cm}|  }
 \hline
 \multicolumn{4}{|c|}{\textbf{Validation Fold $V$}} & 
 \multicolumn{3}{|c|}{\textbf{Test Fold $T$}}  \\ 
 \hline
 
 & $FDST$ & $UCSD$ & $MALL$  & $FDST$ & $UCSD$ & $MALL$  \\
\hline
$sup-ft$  & \textbf{5.17} & \textbf{6.36}   &  \textbf{6.65}    & 9.59     & 8.34  & 13.56     \\ [0.5ex]

$upd-V$ &   6.05 &  7.58   & 7.40 & \textbf{8.00} & \textbf{7.45} & 16.38   
\\ [0.5ex] 

$upd-V+T$   &   %3.15 
8.32 &   
%1.92 
8.28
& 
%2.71  
9.36
& 
%-0.43 
9.16
&  
%0.59 
8.93
& 
%-4.1 
\textbf{9.46}
\\ [0.5ex] 

 \hline
\end{tabular}

\end{table}

\subsection{Results on Activity Recognition}

Three types of experiments were performed to measure: (i) the performance of the baseline model after fine-tuning on the union of the supervised folds for the 15 CAR sequences ($sup-ft-union$), without any continual learning; (ii) the performance obtained by incrementally updating the model (in sessions) using the unlabelled validation ($upd-V$) or test ($upd-T$) data streams, considered separately; (iii) the performance of self-training over validation and training folds considered as a single contiguous data stream, again in a session-wise manner ($upd-V+T$).

The results are shown in Table \ref{tab:cont_ch}.
The standard classification evaluation metrics precision, recall, and F1-score were used for evaluation. For each evaluation metric we computed both the \emph{class average} (obtained by computing the score for each class and taking the average 
%individually and calculate average 
without considering the number of training samples in each class, i.e., $F1_{class1} + F1_{class2} \cdots + F1_{class9} $ ) and the \emph{weighted average} (which uses the number of samples $W_i$ in each class $i$, i.e., $F1_{class1}*W_1 + F1_{class2}*W_2 \cdots + F1_{class9}*W_9 $ ). 

It can be noted that on the validation fold continual updating does improve performance to some extent, especially under Recall and F1 score. Improvements in Recall are visible on the test fold as well. All in all the baseline appears able to extract information from the unlabelled data stream.

\subsection{Results on Crowd Counting}

Table \ref{table:cont_ch_ccc} shows a quantitative analysis of the performance of the fine-tuned supervised model ($sup$) and two incrementally updated models ($upd$) on the validation and the test split, respectively, using the mean absolute error (MAE) metric. 

The experiments were performed under three different experimental settings: (i) the initial model fine tuned on the supervised folds ($sup-ft$), (ii) a model incrementally updated on the validation fold ($upd-V$), and (iii) a model incrementally updated on both the validation and test folds considered as a single data stream ($upd-V+T$). In Table \ref{table:cont_ch_ccc} all three models are tested on both the $V$ and $T$ portions of the data streams.
The initial model was trained for 100 epochs on the supervised fold ($S$), whereas the incremental models were self-trained for 5 epochs in each session (we also tried 2 epochs with no significant change observed). 

{When compared with classical papers in the crowd counting literature, the MAE values in Table \ref{table:cont_ch_ccc} are noticeable higher.  Concretely, a recent work that incorporates optical flow to improve crowd counting achieved MAEs of 1.76, 0.97, and 1.78 on the FDST, UCSD, and Mall dataset, respectively \cite{hossain2020video}.  Also, the original MCNN implementation yielded MAEs of 3.77 and 1.07 on FDST and UCSD (performance was not reported on Mall) \cite{zhang2016single,fang2019locality}. 
The higher MAEs reported in our work are expected,  due to the significantly different and more challenging training protocol (i.e., batch training in standard crowd counting work vs.\ continual learning here).  For example, standard papers in crowd counting following the typical evaluation protocol employ 800 training images for the Mall and UCSD datasets; whereas, in our problem setting 
only 400 images are used for supervised training. }

Previous batch results should be seen as upper bounds to our incremental task, whereas ours should be considered as new baselines for a new problem setting.
While model updating does not seem to be effective (on average) on the validation streams, the effect is quite significant on the test fold (right side of the Table), with important performance gains.

\section{Experimental Analysis of HMM latent Variable} \label{continual:sec:exphmm}

\subsection{Datasets and Setups} \label{continaul:sec:datasets}
\subsubsection{Continual activity recognition (CAR) dataset.}

The CAR dataset (explained in Section \ref{sec:benchmarks}) is used also for evaluating the latent variable and comparison with the previous challenge baseline.  According to the protocol of the CAR dataset, each video sequence is divided into a supervised fold (S), corresponding to the first 5 minutes, a validation fold (V), corresponding to the second 5-minute video, and a test fold (T), corresponding to the last 5-minute video. The last two folds are unsupervised.

In our experiments, for each sequence, the unlabelled data stream (i.e. the last 10 minutes, without distinction between validation and test folds) is partitioned into a number of sub-folds. Each sub-fold spans 1 minute in the CAR dataset, so that each unlabelled sequence is split into 10 sub-folds. 

\subsubsection{50 Salads activity detection dataset.}
We also validate our method using the 50 Salads dataset, which consists of 50 videos by 25 cooks (actors) preparing salads, where each cook recorded two videos of themselves preparing salads by following different sequence of steps. The dataset contains two levels of annotations: ``high-level" labels only span three classes, whereas ``low-level" annotation involves a total of 51 classes. We added a background class to both settings, yielding a total number of classes of 4 and 52, respectively. To enforce a similar protocol to that of the CAR dataset, we introduced two splits to the 50 Salad dataset for training and unsupervised continual updates. The splits are defined as follows:
\begin{enumerate}
\item
\textbf{Split by time} selects the first video for each cook as the training set and the second as the validation set. In this set-up, we do not have test set, so we  only update our model continually over the validation set.
\item
\textbf{Split by cook} divides the set of 25 cooks into 9, 8, and 8 for train, validation, and test sets, respectively. For evaluation, we perform 10-fold random split cross-validation and report the averaged results.
\end{enumerate}
Session sub-folds span 1.28 minutes on average for the 50 Salads dataset, as the duration of each video is slightly longer than in the CAR dataset.

 \begin{table}[t!]
\centering
\caption{Time-varying HMM (TV HMM) vs. traditional HMM (HMM) based on one-off training, where P, R, and F1 are precision, recall, and F1-score, respectively.}\label{tab:one-off}
% \resizebox{0.5\textwidth}{!}{%
\begin{tabular}{c|ccc|ccc}
\hline
\multirow{2}{*}{}              & \multicolumn{3}{c|}{HMM}                                    & \multicolumn{3}{c}{TV HMM}                                     \\ \cline{2-7} 
                               & \multicolumn{1}{c|}{P} & \multicolumn{1}{c|}{R} & F1 & \multicolumn{1}{c|}{P} & \multicolumn{1}{c|}{R} & F1 \\ \hline
CAR                          & \multicolumn{1}{c|}{0.86}          & \multicolumn{1}{c|}{0.18}       &  0.24        & \multicolumn{1}{c|}{0.83}          & \multicolumn{1}{c|}{0.37}       & 0.49         \\ \hline
\multirow{2}{*}{50Salad\_low ${\dfrac{time split}{cook split}}$}  & \multicolumn{1}{c|}{0.22}          & \multicolumn{1}{c|}{0.05}       & 0.08          & \multicolumn{1}{c|}{0.226}          & \multicolumn{1}{c|}{0.052}       & 0.077         \\ \cline{2-7} 
                               & \multicolumn{1}{c|}{0.076}          & \multicolumn{1}{c|}{0.023}       &  0.018        & \multicolumn{1}{c|}{0.076}          & \multicolumn{1}{c|}{0.023}       & 0.018         \\ \hline
\multirow{2}{*}{50Salad\_high ${\dfrac{time split}{cook split}}$} & \multicolumn{1}{c|}{0.39}          & \multicolumn{1}{c|}{0.23}       & 0.27          & \multicolumn{1}{c|}{0.4}          & \multicolumn{1}{c|}{0.24}       & 0.26         \\ \cline{2-7} 
                               & \multicolumn{1}{c|}{0.407}          & \multicolumn{1}{c|}{0.236}       &0.256          & \multicolumn{1}{c|}{0.405}          & \multicolumn{1}{c|}{0.238}       & 0.26          \\ \hline
\end{tabular}
% }

\end{table}

% \usepackage{multirow}
% \usepackage{graphicx}
\begin{table}[t!]
\centering
\caption{Time-varying HMM (TV HMM) vs. traditional HMM (HMM) based on continual retraining, where P, R, and F1 are precision, recall, and F1-score, respectively.}\label{tab:retraining}
% \resizebox{0.5\textwidth}{!}{%
\begin{tabular}{c|ccc|ccc}
\hline
\multirow{2}{*}{}              & \multicolumn{3}{c|}{HMM}                                    & \multicolumn{3}{c}{TV HMM}                                     \\ \cline{2-7} 
                               & \multicolumn{1}{c|}{P} & \multicolumn{1}{c|}{R} & F1 & \multicolumn{1}{c|}{P} & \multicolumn{1}{c|}{R} & F1 \\ \hline
                               
CAR & \multicolumn{1}{c|}{0.81} & \multicolumn{1}{c|}{0.28}  & 0.39          & \multicolumn{1}{c|}{0.81}          & \multicolumn{1}{c|}{0.40}       & 0.52         \\ \hline

\multirow{2}{*}{50Salad\_low {$\dfrac{time split}{cook split}$}}  & \multicolumn{1}{c|}{0.07} & \multicolumn{1}{c|}{0.02}       & 0.03 & \multicolumn{1}{c|}{0.07}          & \multicolumn{1}{c|}{0.028}       & 0.035         \\ \cline{2-7} 
                               & \multicolumn{1}{c|}{0.058}          &  \multicolumn{1}{c|}{0.020}       & 0.025         & \multicolumn{1}{c|}{0.059}          & \multicolumn{1}{c|}{0.018}       &  0.023        \\ \hline
                               
\multirow{2}{*}{50Salad\_high {$\dfrac{time split}{cook split}$}} & \multicolumn{1}{c|}{0.14}          & \multicolumn{1}{c|}{0.26}       &  0.17        & \multicolumn{1}{c|}{0.14}          & \multicolumn{1}{c|}{0.36}       & 0.20          \\ \cline{2-7} 
                               & \multicolumn{1}{c|}{0.402}          & \multicolumn{1}{c|}{0.257}       & 0.288         & \multicolumn{1}{c|}{0.407}          & \multicolumn{1}{c|}{0.237}       &   0.261       \\ \hline
\end{tabular}%
% }

\end{table}

\begin{table}[t!]
\centering
\caption{{Time-varying HMM (TV HMM) vs. traditional HMM (HMM) based on incremental update, where P, R, and F1 are precision, recall, and F1-score, respectively.}}\label{tab:memory}
% \resizebox{0.5\textwidth}{!}{%
\begin{tabular}{c|ccc|ccc}
\hline
\multirow{2}{*}{}              & \multicolumn{3}{c|}{ Traditional HMM}                                    & \multicolumn{3}{c}{Time-varying HMM}                                     \\ \cline{2-7} 
                               & \multicolumn{1}{c|}{P} & \multicolumn{1}{c|}{R} & F1 & \multicolumn{1}{c|}{P} & \multicolumn{1}{c|}{R} & F1 \\ \hline
                               
CAR                            & \multicolumn{1}{c|}{0.79}          & \multicolumn{1}{c|}{0.34}       &    0.46      & \multicolumn{1}{c|}{0.80}          & \multicolumn{1}{c|}{0.34}       &  0.46        \\ \hline
\multirow{2}{*}{50Salad\_low {$\dfrac{time split}{cook split}$}}  & \multicolumn{1}{c|}{0.25}          & \multicolumn{1}{c|}{0.15}       & 0.19          & \multicolumn{1}{c|}{0.309}          & \multicolumn{1}{c|}{0.179}       & 0.214         \\ \cline{2-7} 
                               & \multicolumn{1}{c|}{0.155}          & \multicolumn{1}{c|}{0.031}       &  0.044        & \multicolumn{1}{c|}{0.144}          & \multicolumn{1}{c|}{0.029}       & 0.04         \\ \hline
\multirow{2}{*}{50Salad\_high {$\dfrac{time split}{cook split}$}} & \multicolumn{1}{c|}{0.34}          & \multicolumn{1}{c|}{0.21}& 0.24         & \multicolumn{1}{c|}{0.42}          & \multicolumn{1}{c|}{0.25}       & 0.27         \\ \cline{2-7} 
                               & \multicolumn{1}{c|}{0.440}          & \multicolumn{1}{c|}{0.274}       & 0.306          & \multicolumn{1}{c|}{0.409}          & \multicolumn{1}{c|}{0.259} & 0.278          \\ \hline
\end{tabular}%
% }

\end{table}

\begin{figure*}[t!]
\centering
% \vspace{-7mm}
\includegraphics[width=\linewidth]{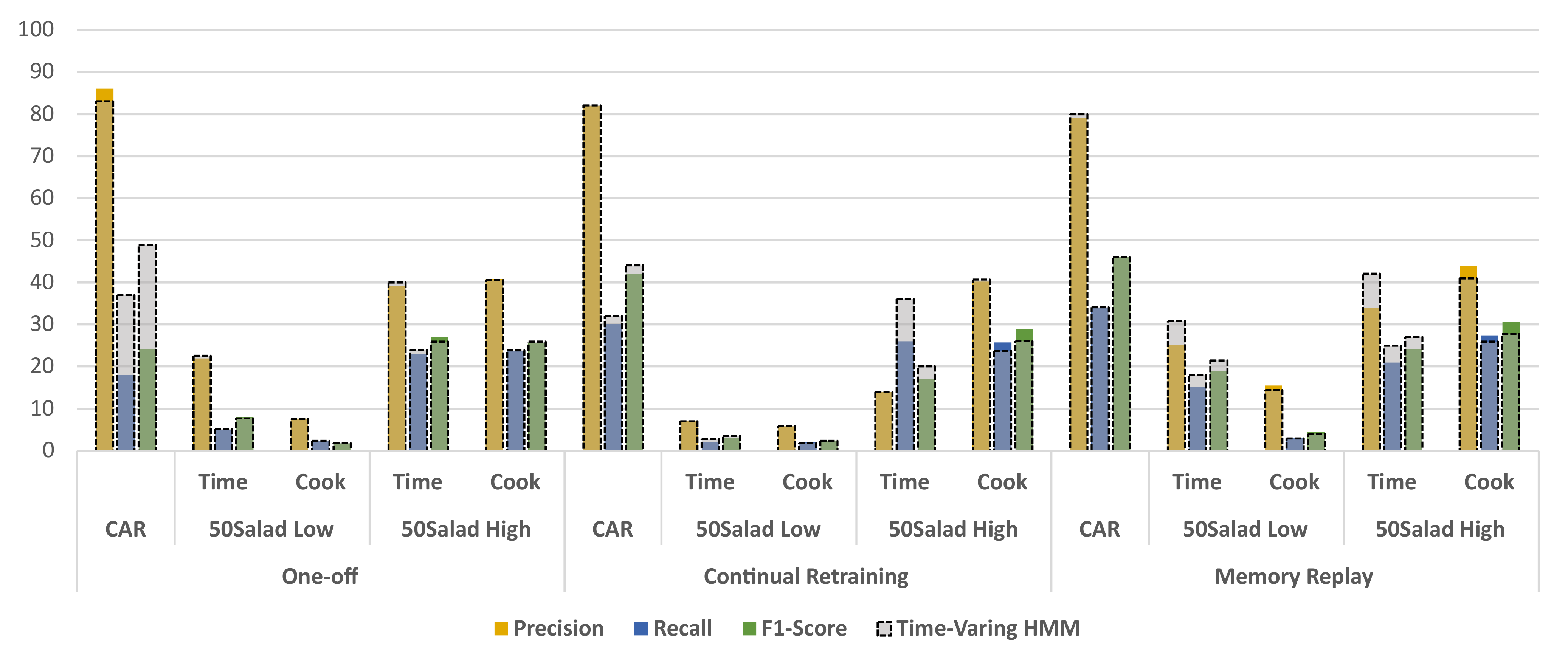}
\caption{Time-varying HMM (dashed bars) vs. traditional HMM (colored bars).}
\label{fig:results}
\end{figure*}

\subsection{Quantitative Analysis} \label{sec:analysis}
The performance of Time-varying HMM (T-HMM) and traditional HMM (HMM) %were tracked
was assessed % FAB: what are D-HMM? we do not introduce this acronym
using three different experimental setups, namely: single, one-off training; continual retraining; and random sampling-based memory replay. 
In the first two settings, only the accuracy of the pseudo-labels generated by the latent model was assessed, whereas the third setting evaluated the whole framework in which the image classifier is incrementally updating using the pseudo-labels generated in sessions by the latent model, also incrementally updated. The quantitative comparison was performed using standard precision (P), recall (R) and F1-score (F1) performance metrics. The quantitative analysis from the Table \ref{tab:one-off},\ref{tab:retraining} and \ref{tab:memory} are further summarized in the Figure \ref{fig:results}. 

\textbf{T-HMM vs. HMM}
It can be seen from Table \ref{tab:one-off}, \ref{tab:retraining} and \ref{tab:memory} that the T-HMM significantly performs better in terms of the F1 and R. T-HMM is better overall in all the three experimental setups. There were big gains in terms of R and F1 in the case of CAR dataset. The overall performance of T-HMM was quite low in case of 50Salad low as it comprises of relatively higher number of classes (51) as described in \ref{continaul:sec:datasets}.

\textbf{Continual retraining vs. One-off training}
It can be seen that the continual retraining improves the performance of not only T-HMM but also in the case of the HMM (Table \ref{tab:retraining}). The HMM performance improves significantly in terms of F1 (+15$\%$). Similarly, R also gained (+10$\%$). The T-HMM significantly performs better than the HMM (+12$\%$ gain in R and F1 for CAR dataset). However, there was not a significant gain in terms of 50Salad dataset either low or high classes. The performance of T-HMM and HMM were close at some instances.  

\textbf{Incremental Update vs. Continual retraining}
The general trend of improvement can be seen in the case of incremental update via a random sampling memory replay as well. The incremental update helped improve the overall recall metric leading to better F1 score. In general, the overall improved F1 score can be found from +5-15$\%$ (for CAR dataset). While a similar trend of not much gain was observed for the 50Salad dataset.

\section{Summary of the Chapter}\label{continual:summary}

In this chapter, we present CSSL workshop and challenge and a latent variable model. In the CSSL workshop and challenge,  approach to the recently defined continual semi-supervised learning problem, in which the labels of the unsupervised data stream are estimated via a time-varying hidden Markov model. \rev{We} formulated the continual semi-supervised learning problem and proposed suitable training and testing protocols for it. To encourage future work in the area we created the first two benchmark datasets, as the foundations of our IJCAI 2021 challenge. We proposed a simple strategy based on batch self-training a baseline model in sessions. The results show that, in both the activity recognition and the crowd counting challenge, the baseline appears in fact to be able to extract information from the unlabelled data stream. Nevertheless, the need for a more sophisticated approach leveraging the dynamics of the data stream is clear.

Next, we proposed a latent variable model approach to the defined continual semi-supervised learning problem, in which the labels of the unsupervised data stream are estimated via a time-varying hidden Markov model. Once the most likely label assignment is estimated, the problem reduces to a continual supervised learning one, and can be tackled by, for instance, random sampling memory replay.

Employing a time-varying hidden Markov model as the latent model allows us to track the evolution of the target domain over time, under realistic assumptions, and therefore achieve better performance. The latent model was evaluated in three different experimental setups on two public datasets adapted to continual learning settings. The proposed solution showed a trend of better performance (a gain of 5-15$\%$ in F1 score) in all the experimental settings.

The model of the label dynamics we proposed makes some weak assumptions on the series of labels: in particular, that they can be described by a Markov chain and that emission probabilities can be described as multivariate Gaussian distributions. Our experiments support the view that these assumptions are fairly realistic -- nevertheless, more general approaches should be explored. 
In the future, we plan to investigate the use of different  logistic models of the dependency of the transition probabilities of the T-HMM from time, and extend the proposed latent variable approach to continual semi-supervised regression.

    \pagestyle{plain}
    \pagestyle{fancy}
    
    \chapter{Conclusions and Future Research Directions}
\label{chapter:conclusion}

In this thesis, we have presented a variety of novel methods for scene understanding divided into three parts. 
Part I \ref{part:part1}: %FAB: there seems to be something wrong with the pointers for the Parts
The ROad event awareness dataset for autonomous driving (Chapter \ref{chapter:road}), focuses on our novel approach to road event detection and the associated 
%dataset with logical requirements 
datasets and baselines. Part II \ref{part:part2} discusses complex activity detection (Chapter \ref{chapter:deformsgraph}, Chapter \ref{chapter:hgraph}) and the spatio-temporal modelling of the scene for holistic scene understanding. Finally, Part III \ref{part:part3} proposes a new paradigm for continual learning (Chapter \ref{chapter:continual}) 
%of updating the models continually.  
allowing us to update models incrementally by leveraging streaming unsupervised data.

To conclude the thesis, this chapter is divided into two sections. First, the methods proposed in this thesis are summarised in Section \ref{con:summary}. Secondly, we illustrated possible research directions for future work in Section \ref{con:future}.

\section{Summary of the Thesis} \label{con:summary}
\subsection{ROad event Awareness Dataset for Autonomous Driving}

We proposed a strategy for situation awareness in autonomous driving based on the notion of road events, and contributed a new ROad event Awareness Dataset for Autonomous Driving (ROAD) as a benchmark for this area of research. The dataset, built on top of videos captured as part of the Oxford RobotCar dataset ~\cite{maddern20171}, has unique features in the field. Its rich annotation follows a multi-label philosophy in which road agents (including the AV), their locations, and the action(s) they perform are all labelled, and road events can be obtained by simply composing labels of the three types. The dataset contains 22 videos with 122K annotated video frames, for a total of 560K detection bounding boxes associated with 1.7M individual labels. 

Baseline tests were conducted on ROAD using a new 3D-RetinaNet architecture, as well as a SlowFast backbone and a YOLOv5 model (for agent detection). Both frame--mAP and video--mAP were evaluated. Our preliminary results 
highlight the challenging nature of ROAD, with the SlowFast baseline achieving a video-mAP on the three main tasks comprised between 20\% and 30\%, at low localisation precision (20\% overlap). YOLOv5, however, was able to achieve significantly better performance. These findings were reinforced by the results of the ROAD @ ICCV 2021 challenge, and support the need for an even broader analysis, while highlighting the significant challenges specific to situation awareness in road scenarios.

%In addition to ROAD, 
In another study we proposed a new learning framework, termed ``learning with requirements", and a new dataset for this task, called ROAD-R (the ROad event Awareness Dataset with logical Requirements), derived from ROAD. We showed that SOTA models most of the time violate the requirements, and how it is possible to exploit the requirements to create models that are compliant with (i.e., strictly satisfy) the requirements while improving their performance. We envision that requirement specification will become a standard step in the development of machine learning models, to guarantee their safety, as it is in any software development process. In this sense, ROAD-R may be followed by many other datasets with formally specified requirements. 

\subsection{Complex Activity Detection}

Moving up in the scene understanding reasoning chain,
%Further down the line toward scene understanding, 
we presented a spatiotemporal complex activity detection framework that leverages both part deformation and a heterogenous graph representation. Our approach is based on three building blocks; action tube detection, part-based deformable 3D RoI pooling for feature extraction, and a GCN module that processes the variable number of detected action tubes to model the overall semantics of a complex activity. In an additional contribution, we temporally annotated two recently released benchmark datasets (ROAD and ESAD) in terms of long-term complex activities. Both datasets come with video-level action tube annotation, making them suitable benchmarks for future work in this area. We thoroughly evaluated our method, showing the effectiveness of our 3D part-based deformable model approach for the detection of complex activities.

This complex activity model works well with datasets endowed with spatiotemporal action annotations. This, however, 
is a serious limitation for the method cannot be applied to
%limited to them can not be generalised to the 
datasets without such ground truth. 
To overcome this limitation, in another study we proposed a novel hybrid graph neural network-based framework that combines both scene graph attention and a temporal graph to model activities of arbitrary duration. Our proposed framework is divided into three main building blocks: agent tube detection and feature extraction; a local scene graph construction with attention; and a temporal graph for recognising the class label and localising each activity instance.
We tested our method on three benchmark datasets, showing the effectiveness of our method in detecting both short-term and long-term activities, thanks to its ability to model their finer-grained structure without the need for extra annotation. 

\subsection{Continual Semi-Supervised Learning}

Lastly, in the last part of this thesis, we formulated an original ``continual semi-supervised learning" problem and proposed suitable training and testing protocols for it. To encourage future work in the area we created the first two benchmark datasets, as the foundations of our CSSL @ IJCAI 2021 challenge. We proposed a simple strategy based on batch self-training a baseline model in sessions. The results show that, in both the activity recognition and the crowd counting challenge, the baseline appears in fact to be able to extract information from the unlabelled data stream. Nevertheless, the need for a more sophisticated approach leveraging the dynamics of the data stream is clear.

To move in this direction we then formulated a latent variable model approach to the continual semi-supervised learning problem, in which the labels of the unsupervised data stream are estimated via a time-varying hidden Markov model. Once the most likely label assignment is estimated, the problem reduces to a continual supervised learning one and can be tackled by, for instance, random sampling memory replay.

Employing a time-varying hidden Markov model as the latent model allows us to track the evolution of the target domain over time, under realistic assumptions, and therefore achieve better performance. The latent model was evaluated in three different experimental setups on two public datasets adapted to continual learning settings. The proposed solution showed a trend of better performance (a gain of 5-15$\%$ in F1 score) in all the experimental settings.

The model of the label dynamics we proposed makes some weak assumptions on the series of labels: in particular, that they can be described by a Markov chain and that emission probabilities can be described as multivariate Gaussian distributions. Our experiments support the view that these assumptions are fairly realistic -- nevertheless, more general approaches should be explored.

\section{Future Research Directions} \label{con:future}

\subsection{Extension of ROAD}

%FAB: you could put some references for the tasks listed below
Our ROAD dataset is extensible to a number of challenging tasks associated with situation awareness in autonomous driving, such as event prediction, trajectory prediction, continual learning, and machine theory of mind.

We are currently actively working to scale up our benchmark by at least one order of magnitude by
%and we pledge to further enrich it in the near future by 
extending ROAD-like annotation to major autonomous driving datasets such as PIE and WAYMO. %FAB: references?

We also intend to propose to organise a follow-up Workshop (possibly a series of events) at the next suitable upcoming major computer vision (e.g. ICCV, CVPR) and robotics (e.g. ICRA, IROS) conferences, in order to start mobilising both communities towards the further development of the benchmark and associated array of tasks and baselines.
An additional workshop on neurosymbolic learning in the real world is being considered as well.

\subsection{Holistic Graph Approach for Complex Activity Detection and Prediction}

In the future we intend to progress further from incremental inference to incremental training, by learning to construct activity graphs in an incremental manner, thus opening the way to applications such as future activity anticipation \cite{liu2022hybrid} and pedestrian intent prediction \cite{cadena2022pedestrian}, while
overcoming the need for a fixed-size snippet representation. A further exciting line of research is the modelling of the uncertainty associated with complex scenes, in either the Bayesian \cite{kendall2017uncertainties} 
or the full epistemic setting \cite{manchingal2022epistemic,osband2021epistemic}.

\subsection{Extension to Spatiotemporal Sentence Grounding}

%FAB: add some references here too?
We also aim to expand our activity modelling to include \emph{spatiotemporal sentence grounding}. More in detail, 
%In the future, 
we plan to integrate our vision models for complex activity detection with language models, in order to localise the specific moment in video based on an input language query. The ultimate goal of our spatiotemporal sentence grounding will be to localise the object tubes in untrimmed videos, where the object may only exist in a very small segment of the video with the help of an input query sentence.

\subsection{Continual Learning Extension to Regression}

Just as in the situations awareness of autonomous driving area,
in the future we intend to organise a follow-up Workshop and Challenge (possibly a series of events) at the next suitable upcoming major machine learning conferences (e.g. IJCAI, ICLR, CVPR) to attract researchers in the field of (semi-supervised or unsupervised) continual learning to develop and improve their methods using our proposed datasets and setup.
Model-wise, 
we plan to investigate the use of different logistic models of the dependency of the transition probabilities of the T-HMM from time, and extend the proposed latent variable approach to continual semi-supervised regression.

%    \begin{appendices}
%    \include{appendices}
%    \end{appendices}
%----MAINMATTER--------------------------------------------------------------
%----BACKMATTER--------------------------------------------------------------

    \backmatter
    \pagestyle{fancy}
    \renewcommand{\bibname}{References}
    \cleardoublepage
    \phantomsection
    \addcontentsline{toc}{chapter}{\bibname}
    \bibliographystyle{ieeetr}
    %\bibliography{../references,../hp_references}
    \bibliography{ref}

    %\printbibliography
%----BACKMATTER--------------------------------------------------------------

\end{document}